\pdfoutput=1
\documentclass[10pt,journal,compsoc]{IEEEtran}
\usepackage{times}
\usepackage[utf8]{inputenc}
\usepackage{amsmath}
\usepackage{amssymb}
\usepackage{graphicx}
\usepackage[font=small]{caption}
\usepackage{tabularx}
\usepackage{booktabs}
\usepackage{multirow}
\usepackage{subfig}
\usepackage{setspace}
\usepackage{color}			
\usepackage{microtype}
\usepackage{setspace}
\usepackage{pifont}
\usepackage{dblfloatfix}

\newcommand{\xmark}{\ding{55}}
\usepackage{tikz}
\usepackage{pstricks}
\usetikzlibrary{spy}
\usepackage{pgfplots}
\pgfplotsset{compat=newest}
\pgfplotsset{plot coordinates/math parser=false}
\newlength \figureheight
\newlength \figurewidth

\renewcommand{\vec}[1]{\boldsymbol{#1}}
\newcommand{\fref}[1]{Fig.~\ref{#1}}
\newcommand{\Fref}[1]{Fig.~\ref{#1}}
\newcommand{\tref}[1]{Table~\ref{#1}}
\newcommand{\Tref}[1]{Table~\ref{#1}}
\newcommand{\sref}[1]{Section~\ref{#1}}
\newcommand{\Sref}[1]{Section~\ref{#1}}
\newcommand{\eg}{e.\,g.~}
\newcommand{\ie}{i.\,e.~}
\newcommand{\wrt}{w.\,r.\,t.~}
\newcommand{\etal}{\textit{et.\,al.}~}

\ifCLASSOPTIONcompsoc
  \usepackage[nocompress]{cite}
\else
  \usepackage{cite}
\fi
\ifCLASSINFOpdf
\else
\fi
\begin{document}
%
\title{Toward Bridging the Simulated-to-Real Gap:\\ Benchmarking Super-Resolution on Real Data}

%
%
%
%

\author{Thomas~Köhler,
	Michel~Bätz,
	Farzad Naderi, 
	André~Kaup,~\IEEEmembership{Fellow,~IEEE},\\
	Andreas~Maier,~\IEEEmembership{Member,~IEEE},
	and~Christian~Riess,~\IEEEmembership{Senior Member,~IEEE}
\IEEEcompsocitemizethanks{
	\IEEEcompsocthanksitem T. Köhler, F. Naderi, and A. Maier are with the Pattern Recognition Lab, Friedrich-Alexander-Universit\"at (FAU) Erlangen-N\"urnberg, Erlangen 91058, Germany.\newline 
	E-mail: \{thomas.koehler, farzad.naderi, andreas.maier\}@fau.de
	\IEEEcompsocthanksitem M. Bätz and A. Kaup are with the Chair of Multimedia Communications and Signal Processing, Friedrich-Alexander-Universit\"at (FAU) Erlangen-N\"urnberg, Erlangen 91058, Germany.\newline 
	E-mail: \{michel.baetz, andre.kaup\}@fau.de
	\IEEEcompsocthanksitem C. Riess is with the IT Security Infrastructures Lab, FAU Erlangen-N\"urnberg, Erlangen 91058, Germany.\newline 
	E-mail:  christian.riess@fau.de
}
}

%
%

\markboth{IEEE TRANSACTIONS ON PATTERN ANALYSIS AND MACHINE INTELLIGENCE}%
{Shell \MakeLowercase{\textit{et al.}}: Bare Demo of IEEEtran.cls for Computer Society Journals}
\IEEEtitleabstractindextext{%
\begin{abstract}
Capturing ground truth data to benchmark super-resolution (SR) is challenging. Therefore, current quantitative studies are mainly evaluated on simulated data artificially sampled from ground truth images. We argue that such evaluations overestimate the actual performance of SR methods compared to their behavior on real images. Toward bridging this \textit{simulated-to-real gap}, we introduce the \textit{Super-Resolution Erlangen} (SupER) database, the first comprehensive laboratory SR database of all-real acquisitions with pixel-wise ground truth. It consists of more than 80k images of 14 scenes combining different facets: CMOS sensor noise, real sampling at four resolution levels, nine scene motion types, two photometric conditions, and lossy video coding at five levels. As such, the database exceeds existing benchmarks by an order of magnitude in quality and quantity. This paper also benchmarks 19 popular single-image and multi-frame algorithms on our data. The benchmark comprises a quantitative study by exploiting ground truth data and qualitative evaluations in a large-scale observer study. We also rigorously investigate agreements between both evaluations from a statistical perspective. One interesting result is that top-performing methods on simulated data may be surpassed by others on real data. Our insights can spur further algorithm development, and the publicy available dataset can foster future evaluations.
\end{abstract}

}

\maketitle

\IEEEdisplaynontitleabstractindextext

%
\IEEEpeerreviewmaketitle

\section{Introduction}
\label{sec:Introduction}

\IEEEPARstart{S}{uper-resolution} (SR) \cite{Milanfar2010} enhances the spatial resolution of digital images without modifying camera hardware. This facilitates low-cost high-resolution (HR) imagery to improve vision tasks, \eg in surveillance \cite{Zhang2010a}, remote sensing \cite{Zhang2012b}, 3D imaging \cite{Schuon2009}, or healthcare \cite{Kohler2014,Kohler2015a}. Single-image SR (SISR) infers HR details from a low-resolution (LR) image using self-similarities \cite{Glasner2009,Huang2015a} or example data via classical regression \cite{Kim2010,Salvador2015,Schulter2015,Timofte2015} or deep learning \cite{Dong2014,Kim2016,Kim2016a,Lai2017,Ledig2017}. Multi-frame SR (MFSR) fuses LR frames with relative motion via interpolation \cite{Batz2016,Takeda2007}, iterative reconstruction \cite{Babacan2011,Bercea2016,Farsiu2004a,Kohler2015c,Liu2014}, or deep learning \cite{Kappeler2016,Liao2015}. Such methods complement related resolution enhancement strategies like frame rate up-conversion (temporal resolution enhancement) \cite{Kim2014} or high dynamic range imaging (radiometric resolution enhancement) \cite{Batz2014}.

SR performance guarantees have also been subject to much research. Examples are seminal works on inherent performance guarantees, like algebraic studies of maximum resolution gains \cite{Baker2002,Lin2004} or statistical validations \cite{Robinson2006}. These works use approximations like linearity and shift invariance of imaging systems, hence they can only roughly predict upper or lower performance bounds. In contrast to that, performance on real data is still widely unexplored, due to the lack of quantitative benchmarks. This is surprising, also in comparison to other computer vision areas like motion analysis \cite{Geiger2012} or deblurring \cite{Lai2016}. Evaluations on real LR data  are performed visually or with no-reference measures due to the absence of ground truth data~\cite{Farsiu2014,Vandewalle2016}. However, this is inappropriate in applications that require evidence that SR has high fidelity \wrt a ground truth (\eg medical imaging) as this property could be anti-correlated to perceptual quality \cite{Blau2018}. Experiments on simulated data with a ground truth to quantify SR quality are by far the most common evaluation strategy. Unfortunately, simulations only partially address practical constraints like physically true image noise, low-light exposures, or photometric variations. This is due to simplifying assumptions on LR image formation, such as bicubic downsampling of HR images \cite{Timofte2016}. 

\begin{figure}[!t]
	\centering
	\subfloat{\includegraphics[width=0.48\linewidth]{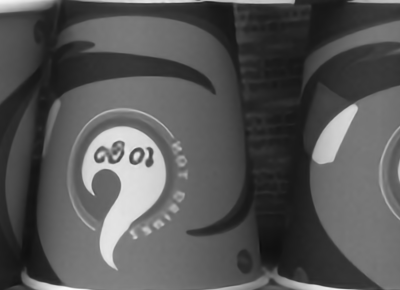}}~
	\subfloat{\includegraphics[width=0.48\linewidth]{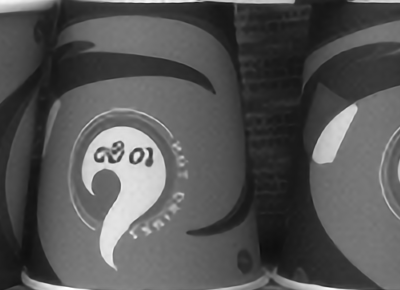}}\\[-1.5ex]
	\setcounter{subfigure}{0}
	\subfloat[VDSR \cite{Kim2016} on simulated data (PSNR: 34.46, IFC: 3.93)][\centering VDSR \cite{Kim2016} on simulated data\par (PSNR: 34.46, IFC: 3.93)]{\includegraphics[width=0.48\linewidth]{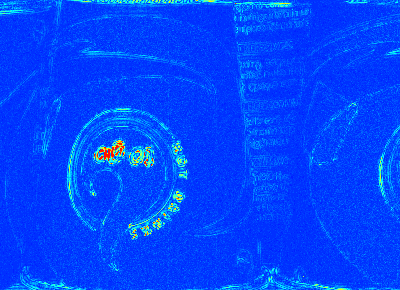}\label{fig:teaser:simulated}}~
	\subfloat[VDSR \cite{Kim2016} on real data (PSNR: 32.46, IFC: 2.63)][\centering VDSR \cite{Kim2016} on real data\par (PSNR: 32.46, IFC: 2.63)]{\includegraphics[width=0.48\linewidth]{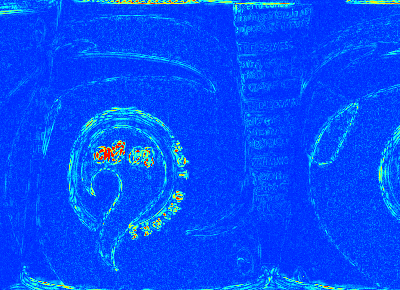}\label{fig:teaser:real}}
	\caption{Can experiments on simulated data predict the behavior of SR on real data? Our study reveals that this is not the case. \protect\subref{fig:teaser:simulated} VDSR \cite{Kim2016} and its color-coded error \wrt the ground truth on simulated low-resolution data. \protect\subref{fig:teaser:real} VDSR on our real acquisitions of the same scene. The simulation considerably overestimates the performance both visually and quantitatively. We benchmark SR algorithms on captured data to overcome shortcomings of simplistic simulations.}
	\label{fig:teaser}
\end{figure}

To demonstrate limitations of simulated data, \fref{fig:teaser} compares the popular VDSR method \cite{Kim2016} on simulated images from bicubic downsampling of the ground truth to real images on the same scene using the hardware-based acquisition proposed here. It shows that simulation is a weak indicator for the performance on real data with degradations like non-Gaussian noise. SR on real data suffers from noise breakthroughs and artifacts at fine structures, \eg text, resulting in considerably lower image quality expressed by PSNR and IFC. In this paper, we show that there is indeed a \textit{simulated-to-real gap} in today's SR benchmarks. More specifically, we reveal that simulation studies often overestimate the actual performance of SR on real images as quantitatively shown in \fref{fig:simulatedToRealDataCorrelation}. Similar conclusions have also been drawn for image denoising \cite{Plotz2017} and motivate the use of real acquisitions to benchmark SR.

\begin{figure*}[!t]
	\centering
		\includegraphics[width=0.985\textwidth]{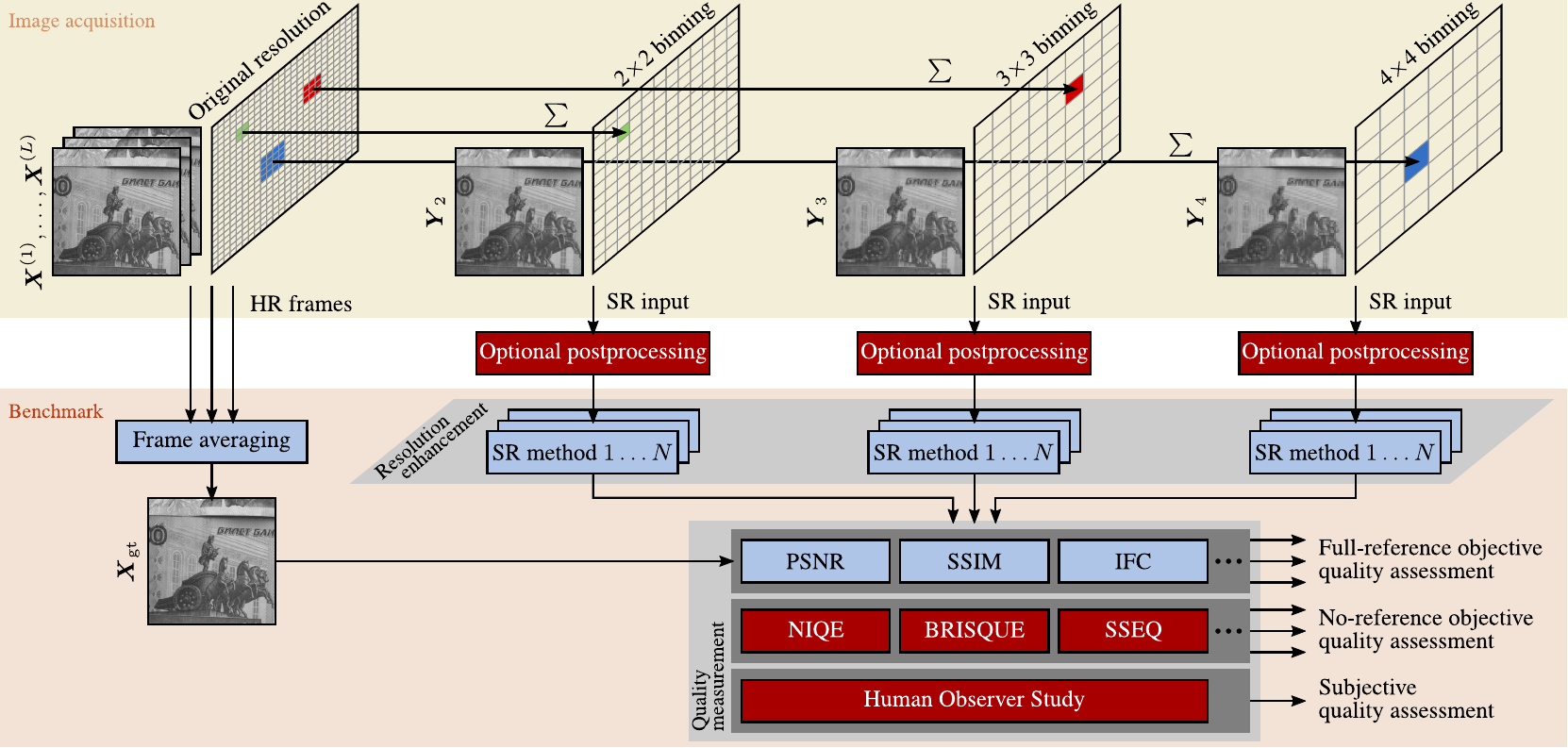}
	\caption{Overview of our data collection and benchmark. In our data collection, we capture multiple frames at the actual pixel resolution to obtain ground truth high-resolution images via frame averaging. We employ hardware binning on the sensor to gain captured low-resolution images and include postprocessing (\eg image/video compression) to collect multiple versions of this low-resolution data (see \sref{sec:SuperResolutionBenchmarkDataset}). In the benchmark, we use 1) full-reference and no-reference measures to quantitatively assess super-resolved data with and without exploiting the ground truth (see Sects.~\ref{sec:BenchmarkSetup} and \ref{sec:QuantitativeStudy}), and 2) observer studies to evaluate image quality according to human perception (see \sref{sec:HumanSubjectStudy}).}
	\label{fig:flowchart}
\end{figure*}

This paper introduces the \textit{Super-Resolution Erlangen \mbox{(SupER)}} database -- a large database of captured LR images at \textit{multiple levels of spatial resolution} and ground truth HR data -- to close the simulated-to-real gap. Data is obtained via \textit{hardware binning}, and covers  difficult conditions like local object motion and photometric variations. It also comes with \textit{postprocessed} images with different levels of H.265/HEVC coding to investigate video compression as shown in \fref{fig:flowchart} (top).
To our knowledge, this is the first comprehensive dataset of \textit{all-real} image sequences to allow quantitative evaluations, as a major step toward an in-the-wild dataset. It comprises more than 80k images of 14 scenes at 4 resolution and 5 compression levels. This size also opens the opportunity to use it for fine-tuning learning-based methods to real data.

We benchmark 19 SR algorithms on the SupER database from two perspectives as shown in \fref{fig:flowchart} (bottom). First, this includes a \textit{quantitative evaluation} using full-reference and no-reference quality assessment. This is by far the most comprehensive SR comparison, which particularly cross-compares SISR and MFSR. Second, we present a \textit{large-scale observer study} to benchmark SR algorithms according to human visual perception. Our experiments reveal some unexpected results on real data as manifestations of the simulated-to-real gap: for instance, several classical methods like shallow regression or sparsity priors for reconstruction-based SR compare very well to much more elaborated deep learning methods. They also show so far unexplored mismatches between quantitative evaluations and human perception: for example, correlations between quantitative and perceptual quality are higher for larger SR factors but deteriorate under real conditions like photometric variations. To maximize the use for the community and to foster benchmarks on real images, we publish all data and source code implementing the evaluation protocols\footnote{\url{https://superresolution.tf.fau.de/}}.

The remainder of this paper is organized as follows. In \sref{sec:RelatedWork}, we review existing SR datasets and evaluation strategies. In \sref{sec:SuperResolutionBenchmarkDataset}, we introduce the proposed benchmark database. In \sref{sec:BenchmarkSetup}, we present the underlying evaluation protocol. In \sref{sec:QuantitativeStudy} and \sref{sec:HumanSubjectStudy}, we evaluate current SR approaches quantitatively and with an observer study. In \sref{sec:Discussion}, we draw conclusions for future algorithm developments. \Sref{sec:Conclusion} concludes this work.

\section{Related Work}
\label{sec:RelatedWork}

In comparison to the great number of algorithmic contributions, there is only few prior work on their comparative experimental evaluations. Most papers follow two evaluation strategies.

\vspace{-0.3em}
\subsection{Benchmarking on Simulated Data}
\label{sec:BenchmarkingOnSimulatedData}

The most common evaluation strategy is the use of simulated data. Yang \etal \cite{Yang2014a} and Timofte \etal \cite{Timofte2016} have evaluated current SISR methods, where LR images are obtained by artificial downsampling of HR ground truth data. In \cite{Liu2014}, Liu and Sun evaluated MFSR on video datasets by artificially sampling HR videos. All of these benchmarks have in common that only simplistic sampling kernels that are known a priori (\eg bicubic \cite{Timofte2016} or Gaussian kernels \cite{Yang2014a,Liu2014}) are simulated but SR in case of more general kernels is unexplored. 

One considerable step forward is the DIVerse 2K resolution (DIV2K) database \cite{Agustsson2017} with LR data generated under bicubic downsampling and more difficult kernels that are hidden to a user. This provides valuable insights to SR performance and the associated NTIRE 2017 challenge \cite{Timofte2017} revealed that overall very deep neural networks \cite{Kim2016,Kim2016a,Ledig2017} and methods evolved thereof \cite{Bae2017,Lim2017,Zhang2018b} are currently the top performing ones. DIV2K covers challenging situations in terms of sampling but the simulations lack MFSR facets, \ie sequences with motion or photometric variations. The LR data is also not the outcome of physical imaging systems with effects like non-Gaussian noise or image/video compression.

In general, the use of simulated data enables comparisons to a ground truth by full-reference quality measures but limits the significance to study SR under realistic constraints. The neglection of physically meaningful sampling kernels, realistic noise models, or environmental conditions manifest crucial limitations and is no realistic depiction of real-world applications. Surprisingly, the impact of realistic image formation models for SR evaluations compared to such simplifying conditions is still widely unexplored. In \sref{sec:ComparisonToExistingDatasets}, we address this question and show the overall weak correlations between benchmarks on simulated and real data termed simulated-to-real gap.

\vspace{-0.3em}
\subsection{Benchmarking on Real Data}
\label{sec:BenchmarkingOnRealData}

Existing real-world image databases \cite{Farsiu2014,Vandewalle2016} lack of ground truth data and are hence designed for perceptual evaluations. This requires no-reference quality measures as for example proposed in \cite{Yeganeh2012,Yuan2012}. In the perception-distortion plane as proposed by Blau and Michaeli \cite{Blau2018}, such methods are used in conjunction with full-reference measures to jointly analyze perceptual quality and fidelity to a ground truth. However, finding appropriate no-reference measures to assess SR on general scenes is difficult. In \sref{sec:CorrelationToQuantitativeStudy}, we show that on real data popular no-reference measures show lower correlations to human visual perception than measures with a reference. Another strategy are large-scale observer studies, as previously conducted for deblurring \cite{Lai2016} or SISR \cite{Yang2014a}. This ensures high agreement to human perception but is cumbersome and the results are difficult to reproduce. Our work aims at constructing a database with ground truths to enable full-reference quality assessment. We further statistically analyze the agreement between quantitative and observer studies.

Other works \cite{Dai2016,Raghavendra2013} also validated SR in specific vision tasks. However, such evaluations have limited informative value for general benchmarks on natural scenes. Our work aims at broadly benchmarking SR algorithms on real captured images.

In a work closely related to ours, Qu \etal \cite{Qu2016} have 
collected a database of real face images with corresponding ground truth data. Their setup utilizes two cameras combined 
with a beam splitter to capture LR and HR images at the same time. However, the required LR/HR
alignment in this \textit{multi-camera} setup is potentially affected by error-prone calibrations 
and image registrations. This makes the use of full-reference 
quality measures for pixel-wise comparisons unreliable. Furthermore, the data of \cite{Qu2016} comprises only single 
images, which excludes MFSR. 
We propose a \textit{single-camera} setup that avoids these limitations and also allows us to acquire more than two resolution levels.

\section{SupER Database}
\label{sec:SuperResolutionBenchmarkDataset}

\begin{figure}[!t]
	\centering
	\includegraphics[width=0.1295\linewidth]{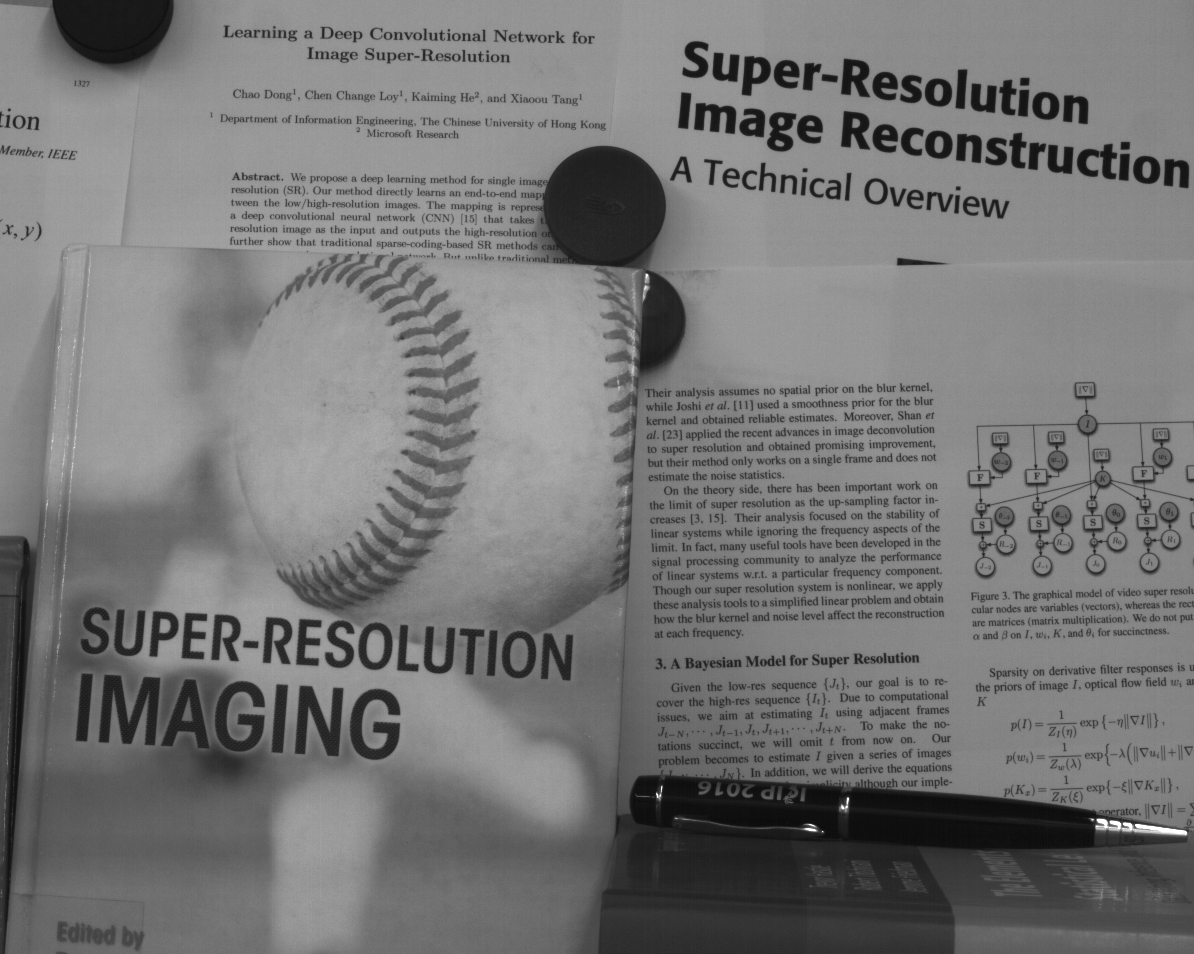}
	\includegraphics[width=0.1295\linewidth]{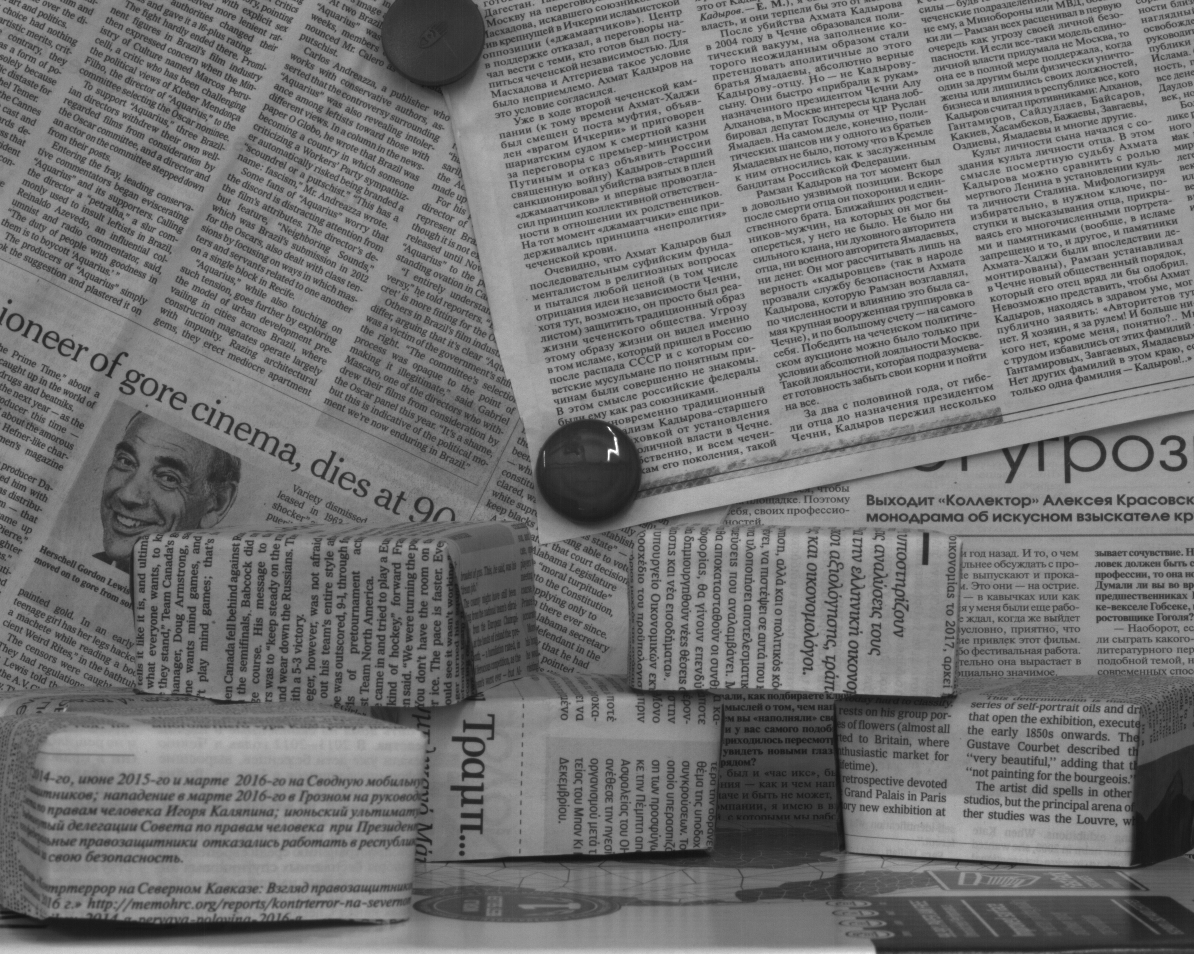}
	\includegraphics[width=0.1295\linewidth]{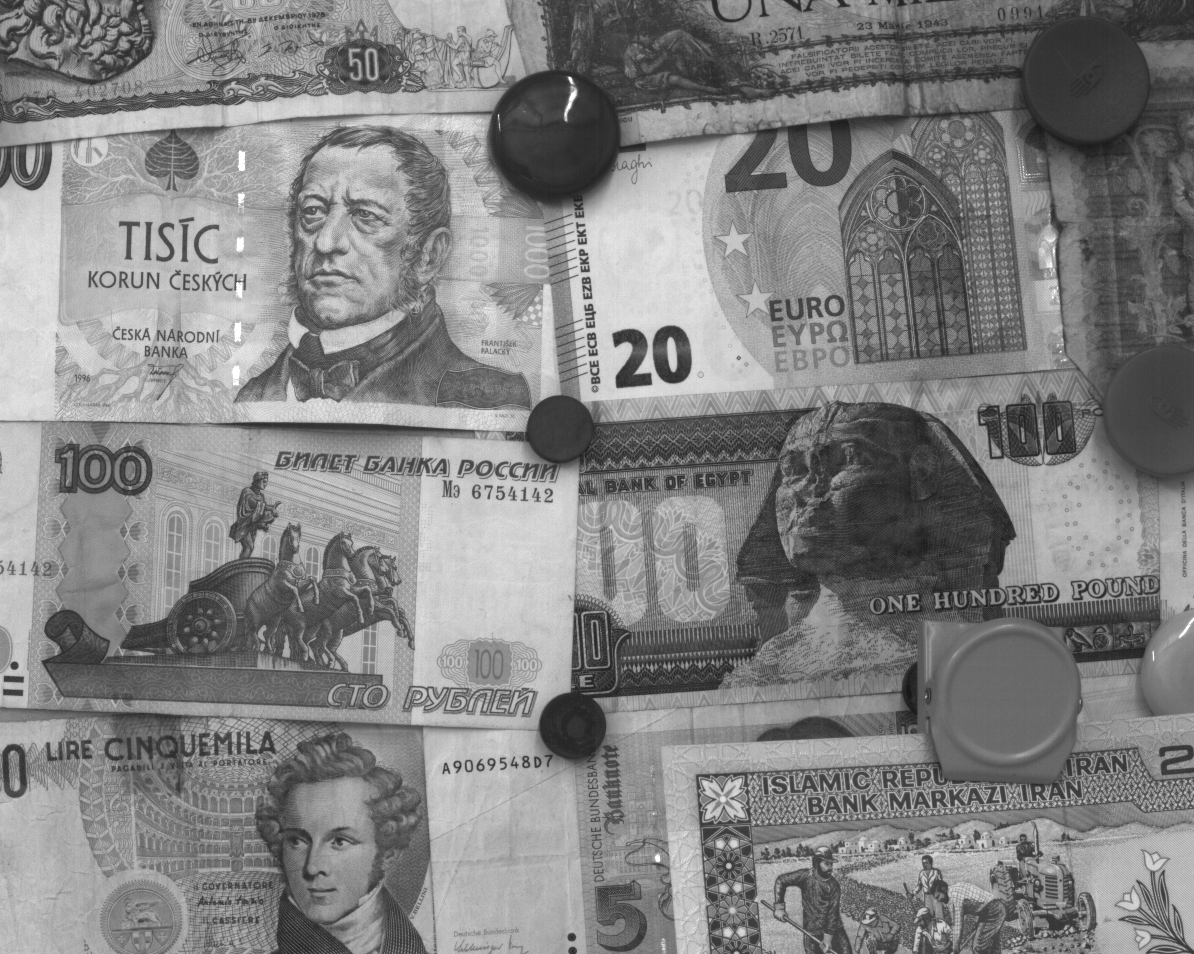}
	\includegraphics[width=0.1295\linewidth]{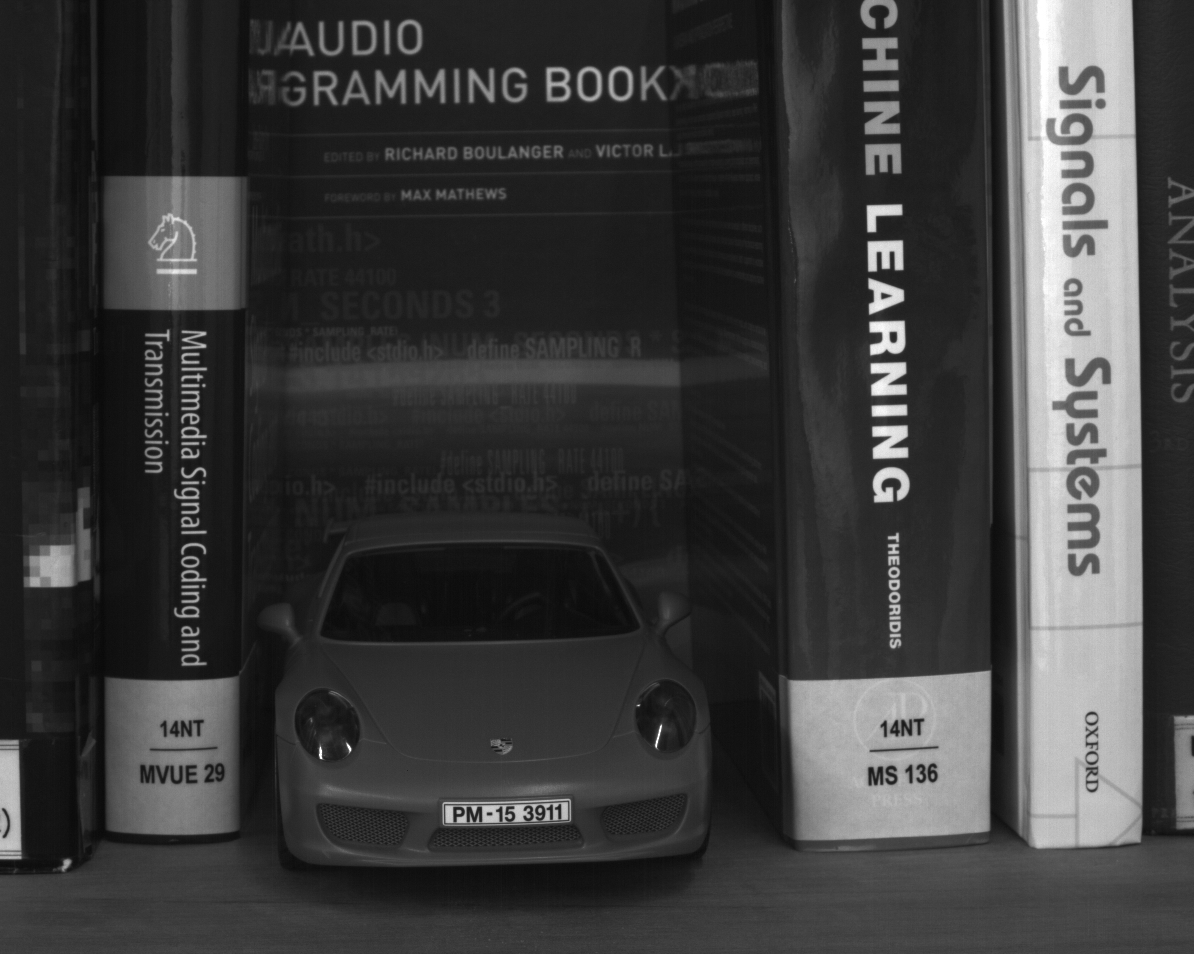}
	\includegraphics[width=0.1295\linewidth]{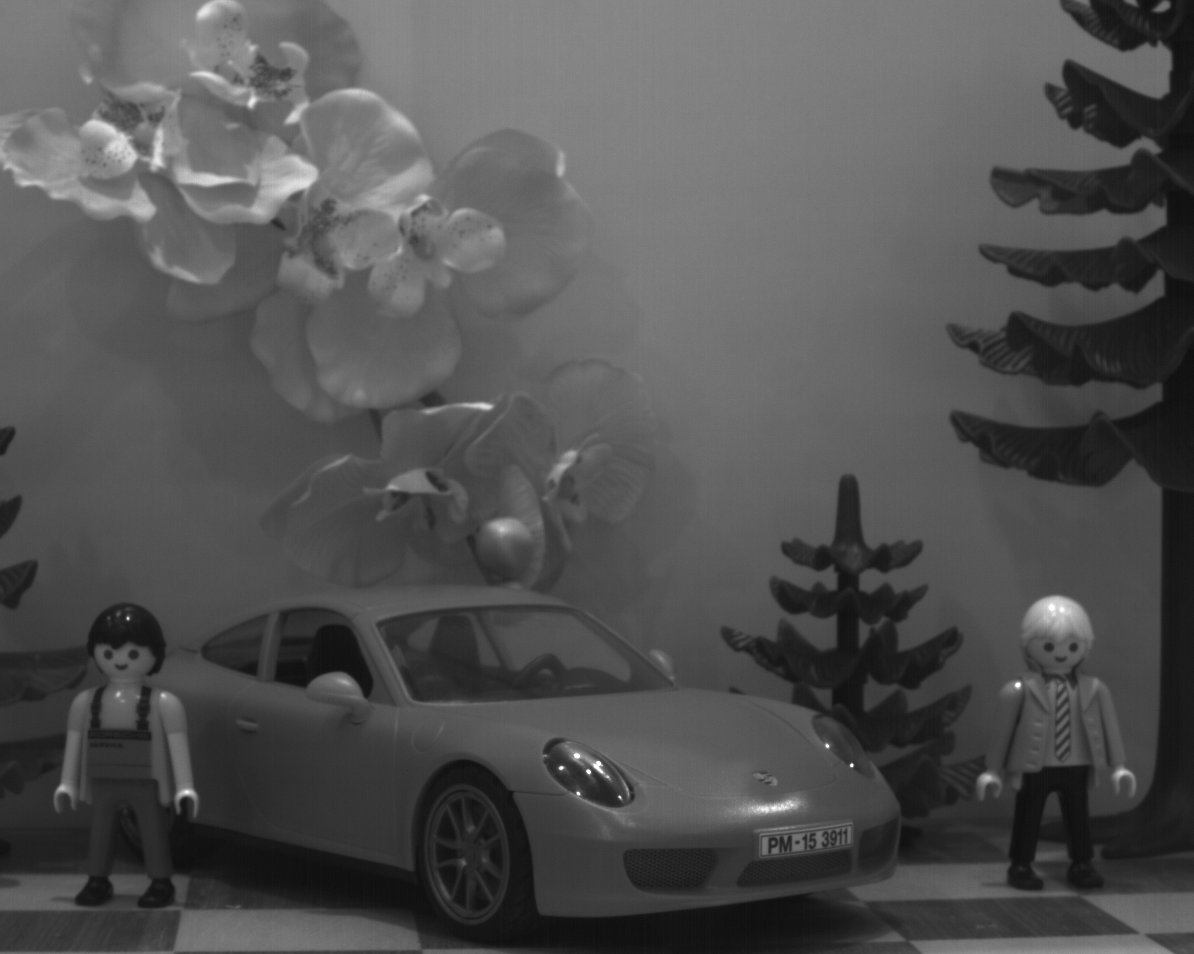} 
	\includegraphics[width=0.1295\linewidth]{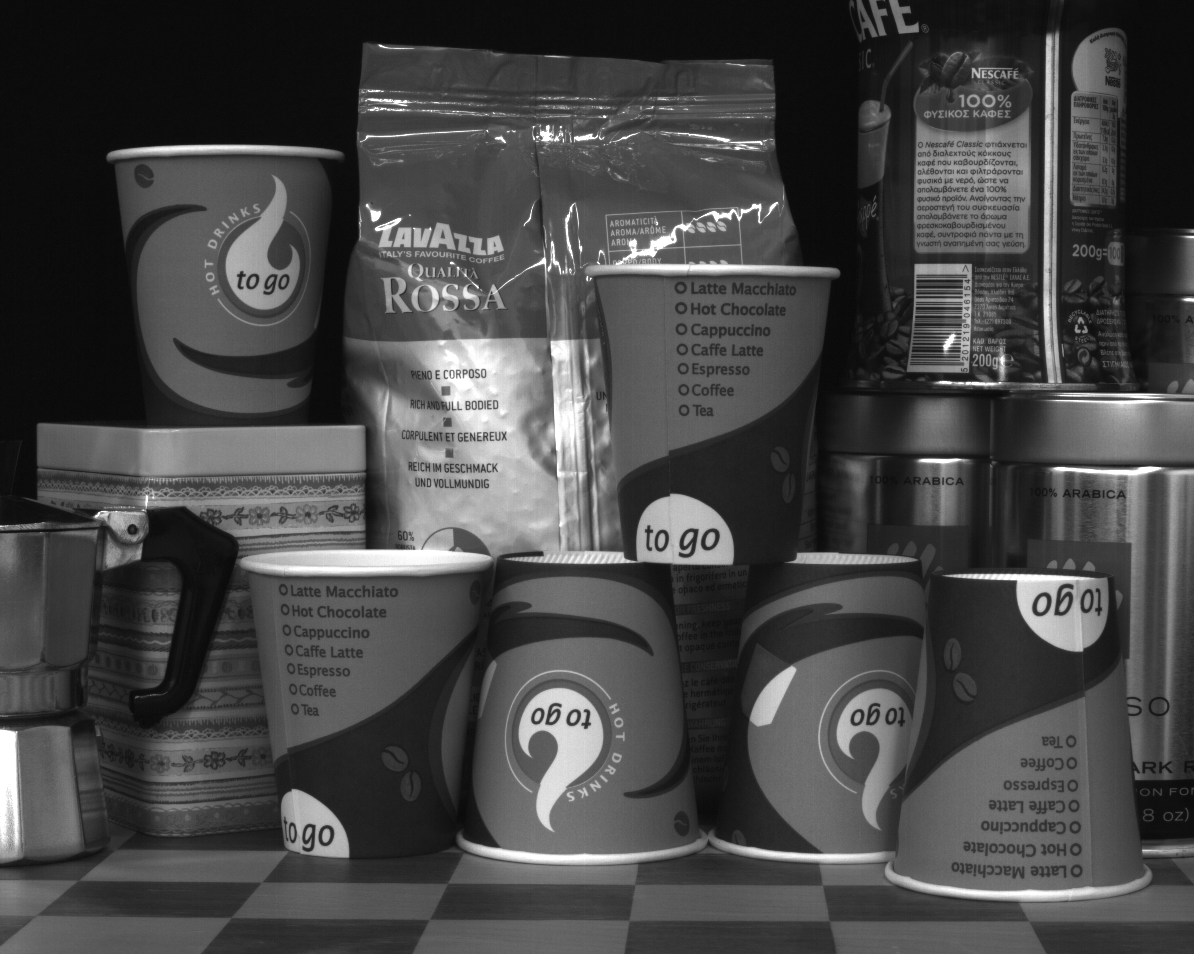}
	\includegraphics[width=0.1295\linewidth]{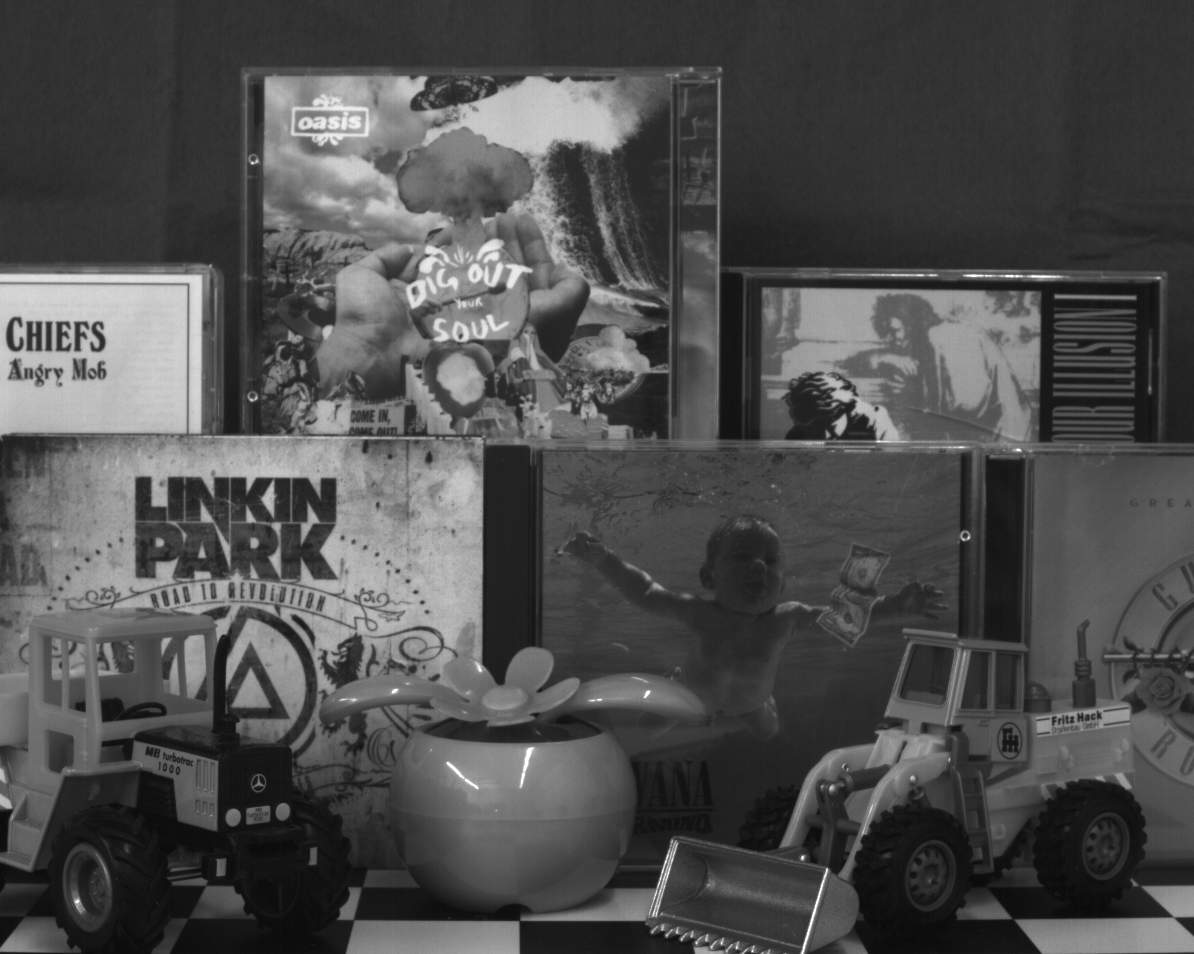}\\[0.7ex]
	\includegraphics[width=0.1295\linewidth]{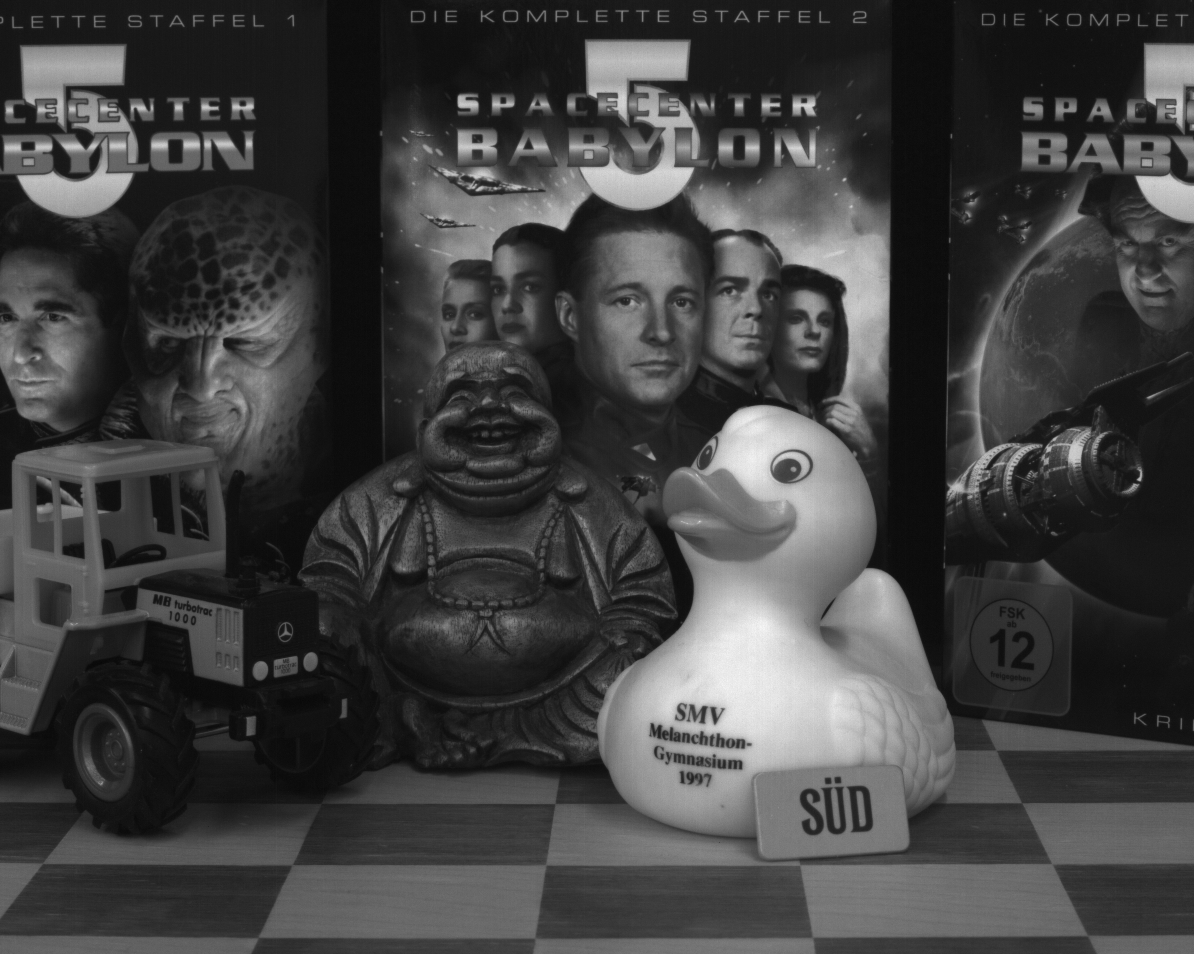}
	\includegraphics[width=0.1295\linewidth]{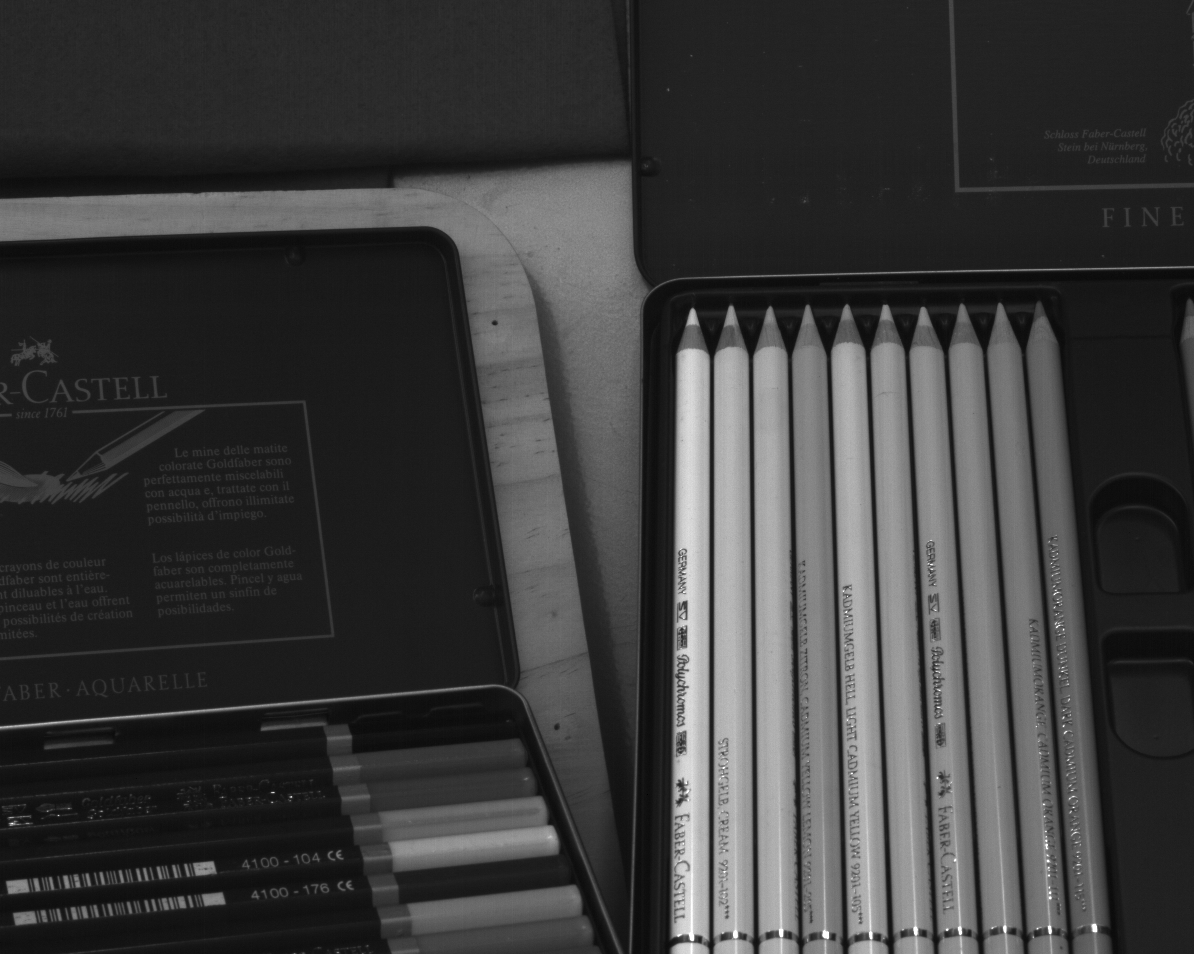}
	\includegraphics[width=0.1295\linewidth]{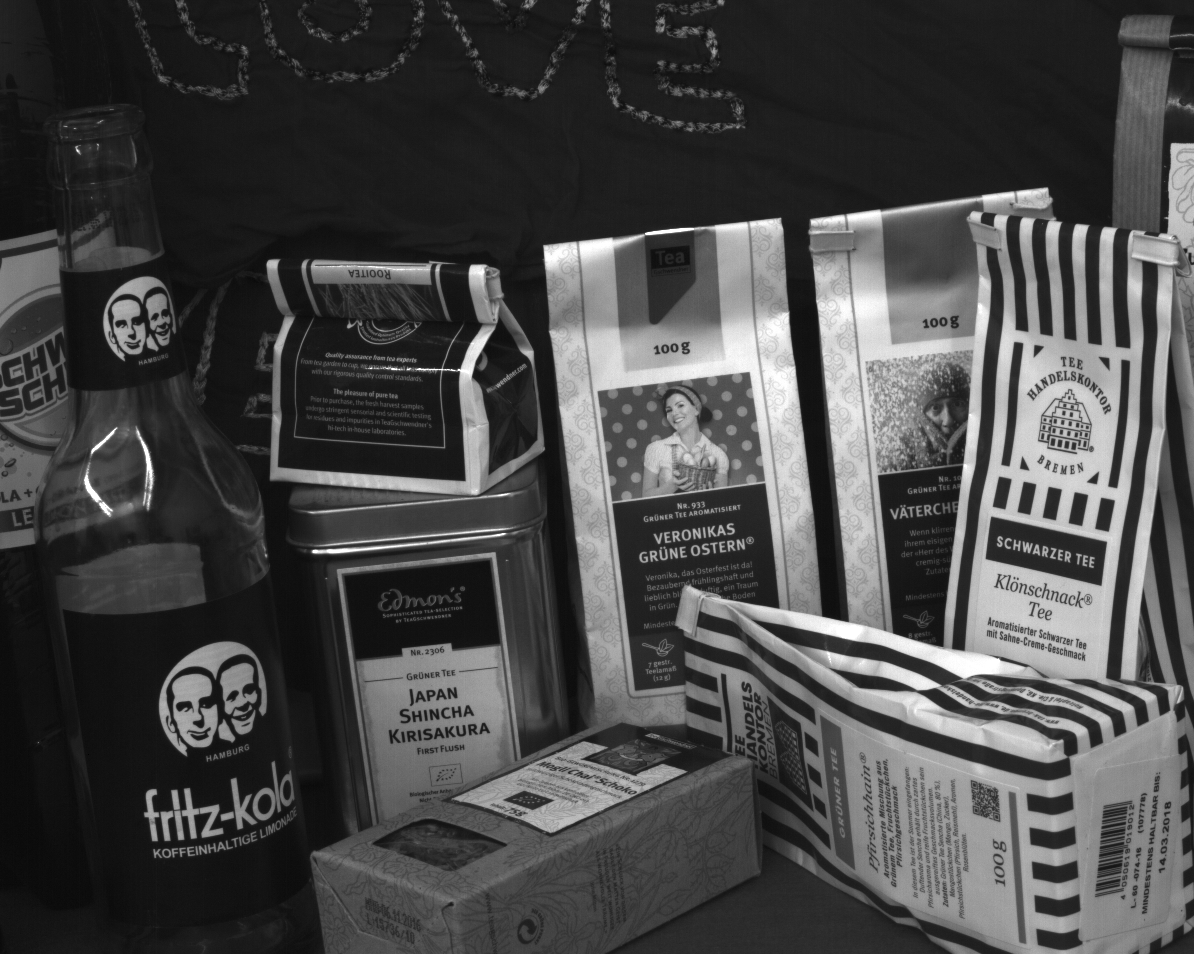}
	\includegraphics[width=0.1295\linewidth]{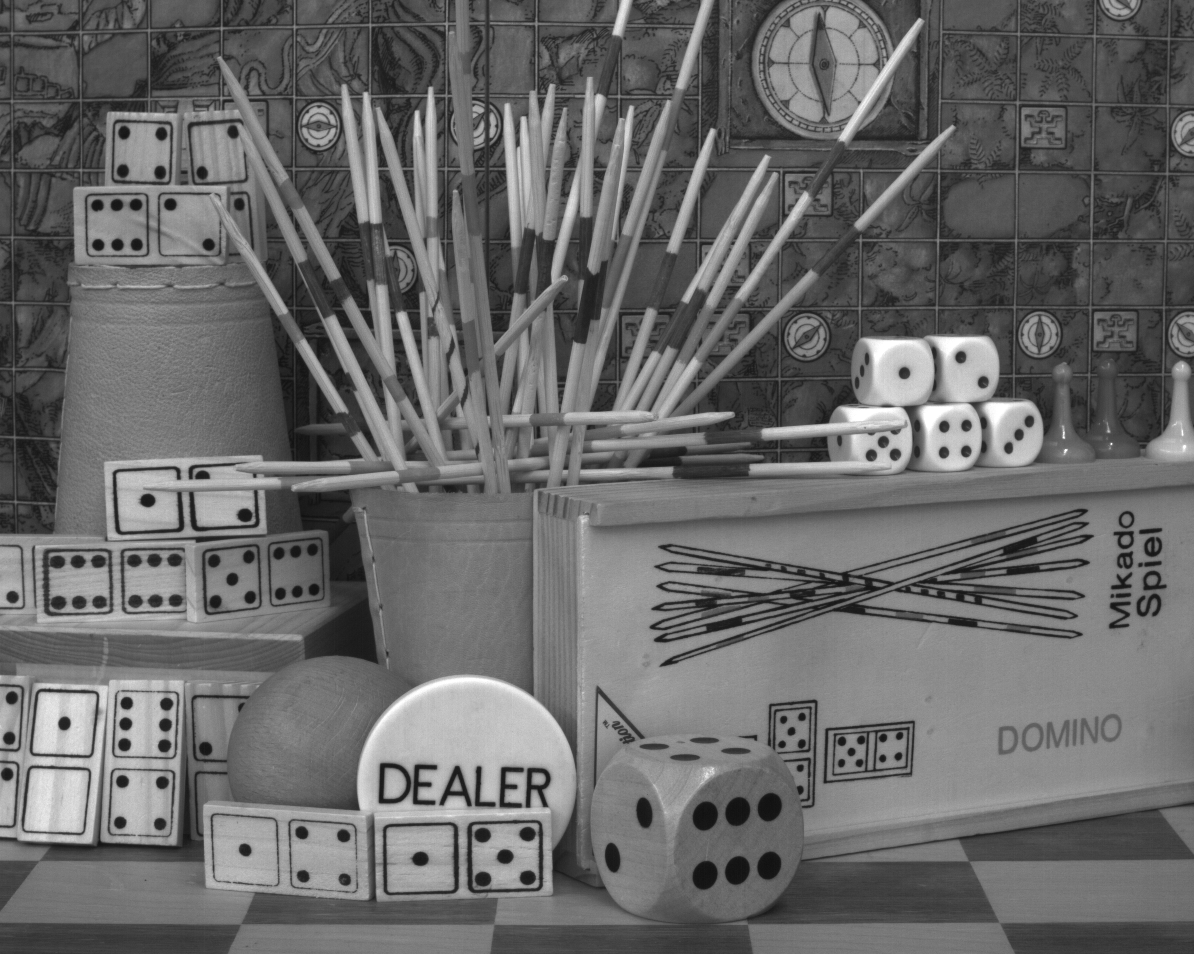}
	\includegraphics[width=0.1295\linewidth]{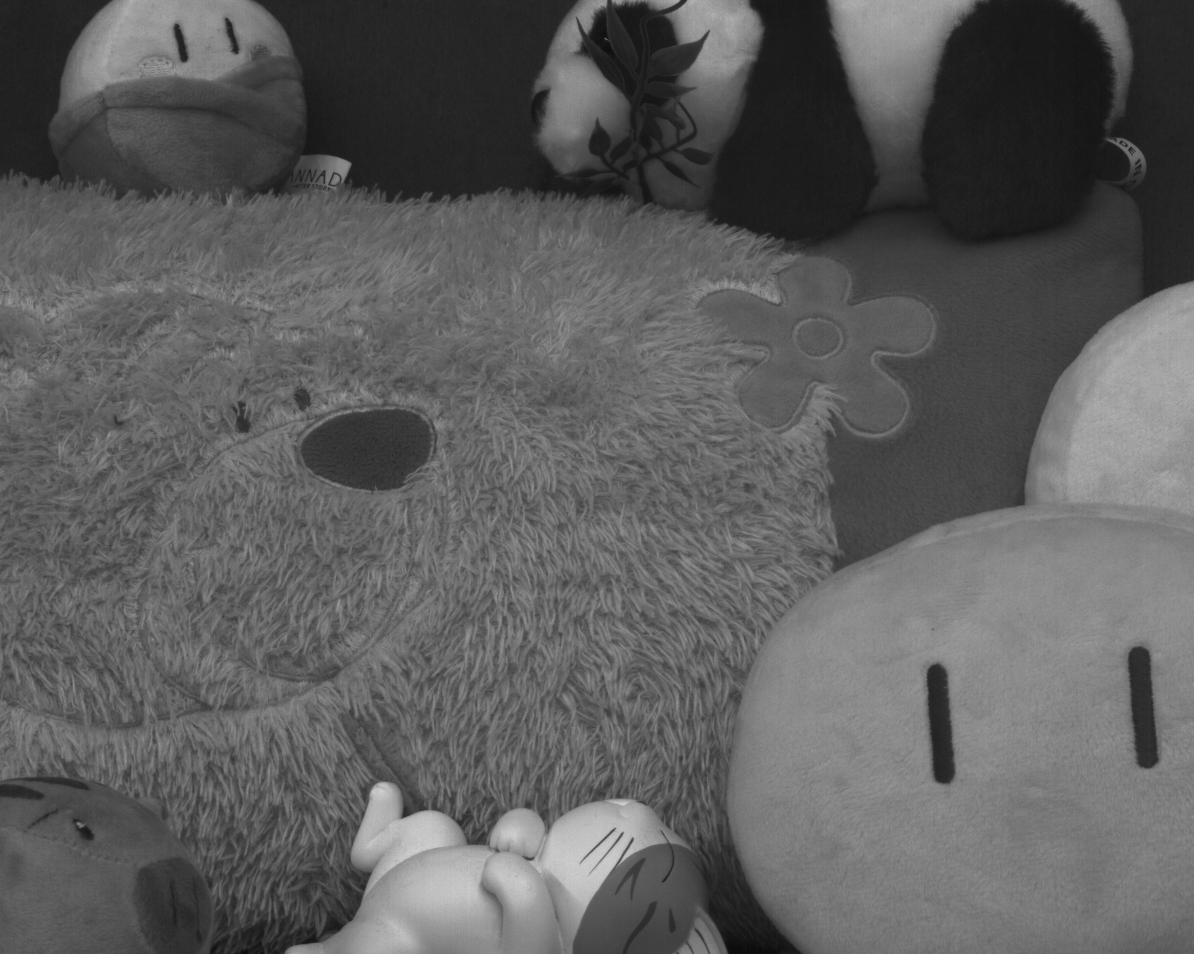}
	\includegraphics[width=0.1295\linewidth]{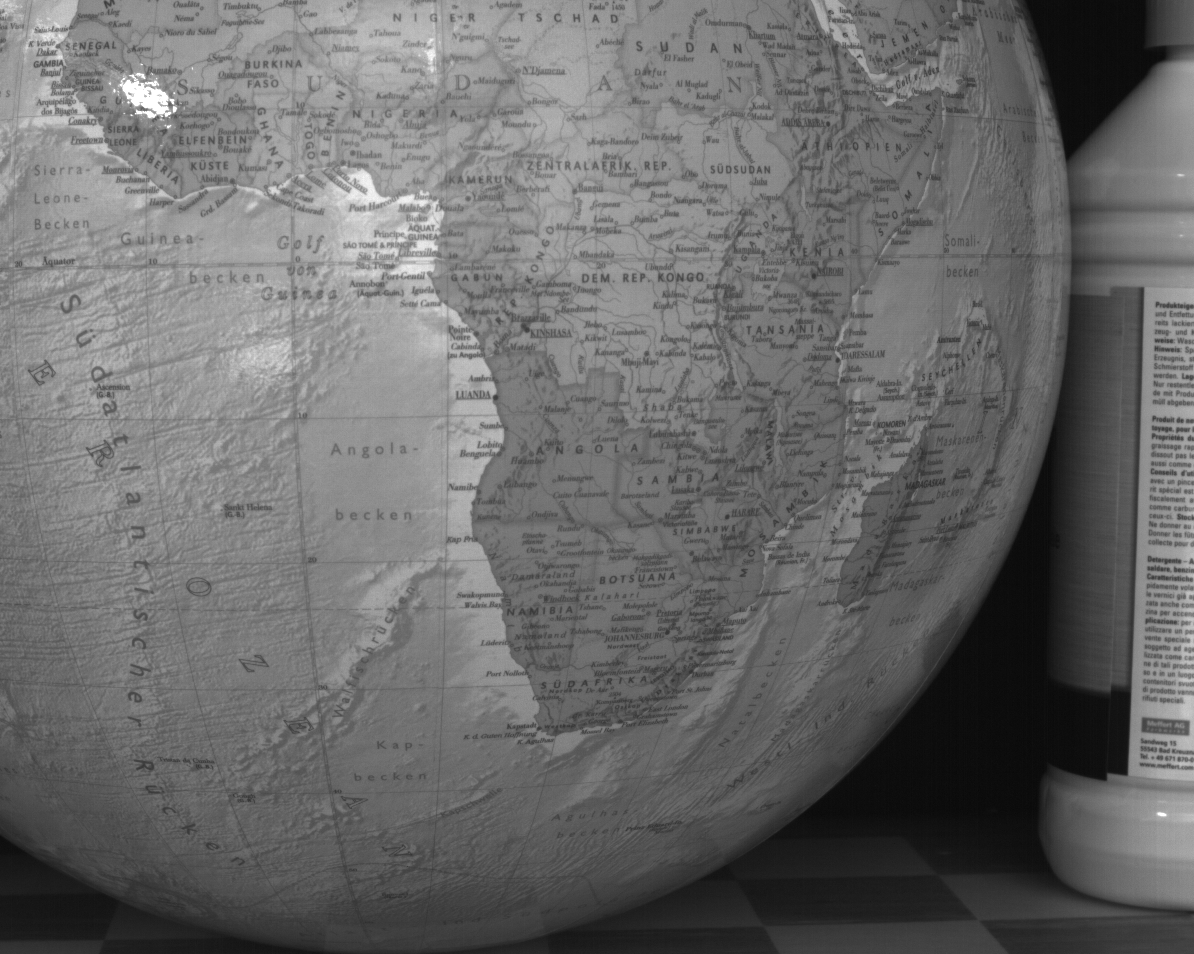}
	\includegraphics[width=0.1295\linewidth]{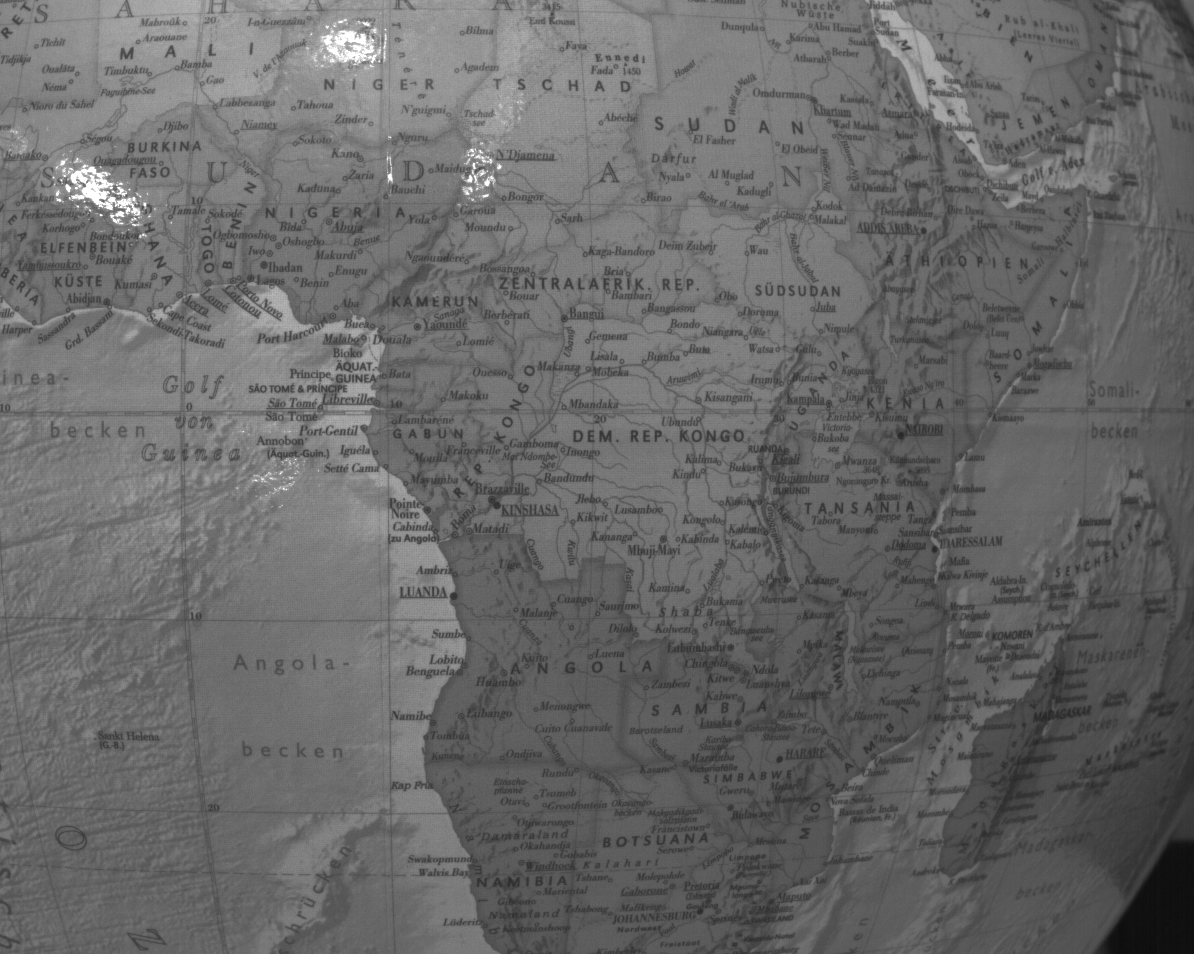}
	\caption{Overview of the scenes covered by our SupER database.}
	\label{fig:databaseOverview}
\end{figure}

We collect sets of LR and HR images at multiple resolutions with a 
single camera by capturing \textit{stop-motion} videos. At each time step of 
a stop-motion video, the underlying scene, environmental conditions, and the camera pose are kept static. For consecutive time steps, the 
scene undergoes changes related to camera and/or object movements and/or 
environmental variations. One time step is 
represented by the $(n+1)$-tuple 
$(\vec{X}_{\mathrm{gt}}, \vec{Y}_{b_1}, \vec{Y}_{b_2} \ldots, \vec{Y}_{b_n})$,
where $\vec{X}_{\mathrm{gt}}$ denotes a ground truth HR image of size 
$N_u \times N_v$ and $\vec{Y}_{b_i}$, $i = 1, \ldots, n$ are LR frames of size 
$N_u/b_i \times N_v/b_i$ at $n$ different hardware binning  factors $b_i$. 
Our database covers 14 lab scenes including text, 
emulated surveillance scenes, and various objects, see \fref{fig:databaseOverview}.

\subsection{Image Formation}
\label{sec:ImageAcquisition}

In order to gain the ground truth $\vec{X}_{\mathrm{gt}}$, we capture $L$ 
frames $\vec{X}^{(l)}$, $l = 1, \ldots, L$ at each time step of a stop-motion 
video using the actual pixel resolution of the camera. These frames are acquired 
under constant illumination and without inter-frame motion. To reduce sensor 
noise, the ground truth is computed by averaging over $L$ ($L = 10$) consecutive 
frames:
\begin{equation}
	\vec{X}_{\mathrm{gt}} = \frac{1}{L} \sum_{l = 1}^L \vec{X}^{(l)}\enspace.
\end{equation} 

To obtain the LR data $\vec{Y}_{b_i}$ associated with 
$\vec{X}_{\mathrm{gt}}$, we use camera hardware binning. This aggregates adjacent pixels on the sensor array as depicted in 
\fref{fig:flowchart} (top). Let $x(\vec{u})$, $\vec{u} \in \mathbb{R}^2$ be 
an irradiance light field \cite{Lin2004}. Hardware binning links 
$x(\vec{u})$ to the image $\vec{Y}_{b}$ according to:
\begin{equation}
	\label{eqn:hardwareBinning}
	\vec{Y}_{b} = \mathcal{Q} \left\{ \mathcal{D}_b \left\{ x(\vec{u}) \right\} 
+ \vec{\epsilon} \right\}\enspace,
\end{equation}
where $\mathcal{D}_b\{ \cdot \}$ denotes sampling according to binning 
factor $b$, $\mathcal{Q}\{ \cdot \}$ denotes quantization and discretization to capture image~intensities, and $\vec{\epsilon}$ is additive noise. The sampling $\mathcal{D}_b\{ \cdot \}$ is described by:
\begin{equation}
	\label{eqn:hardwareBinningSamplingOp}
	\mathcal{D}_b \left\{ x(\vec{u}) \right\} = \left( \vec{H}_{\mathrm{sensor}
, b} \star \vec{H}_{\mathrm{optics}} \star x \right) (\vec{u})\enspace,
\end{equation}
where $\vec{H}_{\mathrm{optics}}$ denotes the optical point spread function
(PSF), $\vec{H}_{\mathrm{sensor}, b}$ models the spatial integration over $b
\times b$ pixels on the sensor array, and $\star$ is the convolution operator
\cite{Lin2004}. As we use a single optical system to capture HR and LR data,
$\vec{H}_{\mathrm{sensor}, b}$ is determined by the binning factor $b$ while
$\vec{H}_{\mathrm{optics}}$ does not depend on the binning. We use $n = 3$ binning factors $b \in \{ 2, 3, 4 \}$ to acquire data at different resolution levels.

Most SR algorithms deal either with grayscale or a single luminance channel, while the chrominance channels are simply interpolated \cite{Geiping2017,Dong2014,Kappeler2016,Timofte2016}. Thus, we limited ourselves to monochromatic acquisitions to compromise between hardware requirements and practical applicability. In order to study full color SR, the setup can be generalized to provide multiple channels, \eg using color filters or a full RGB camera.

\subsection{Image Postprocessing}
\label{sec:ImagePostprocessing}

We capture raw LR and ground truth data in the proposed multi-resolution scheme while camera internal processing is avoided. This enables to explicitly investigate SR under different types of \textit{postprocessing} (or \textit{preprocessing} from the SR perspective). Raw data forms an ideal base for SR while additional (latent) postprocessing steps might deteriorate its performance. This provides a new testbed for quantitative benchmarks.

Some examples of postprocessing steps that have been studied in conjunction with SR are photometric transformations \cite{Capel2003} or demosaicing \cite{Gotoh2004, Farsiu2006}. Lossy image and video compression is a type of postprocessing that is of high practical relevance. We compute an additional compressed version $\vec{Y}_{b,c}$ of the LR image $\vec{Y}_{b}$ according to:
\begin{equation}
	\label{eqn:compression}
	\vec{Y}_{b,c} = \mathcal{C}_c \left\{ \vec{Y}_{b} \right\}\enspace,
\end{equation}
where $\mathcal{C}_c\{ \cdot \}$ denotes compression at compression level $c$. 

We employ H.265/HEVC video coding \cite{Sullivan2012} on our raw data using the \textit{main} profile and \textit{random access} mode (version HM-16.2). Uncoded data and the quantization parameters (QP) $10$, $20$, $30$, and $40$ were chosen so as to cover the full range from a raw data without compression artifacts to a strong compression with heavy artifacts. When combined with the multi-resolution scheme, this approach leads to 15 versions of LR data associated with a given ground truth, \ie three resolutions levels and five compression levels.

\subsection{Sensor and Optics}
\label{sec:SensorAndOptics}

For our data collection, we use a Basler acA2000-50gm CMOS camera \cite{Basler2016} equipped with a $f/1.8$, 16\,mm fixed-focus lens~\cite{Edmundoptics2016}. \Tref{tab:hardwareParameters} summarizes the main parameters of that system. We choose the lens $f$-number scene-dependent as a compromise between a diffraction-limited imaging system and a sufficient depth of field. Using a small $f$-number, the overall resolution is mainly related to sensor properties, namely the used hardware binning. Hence, the pitch between adjacent pixels and their size after hardware binning are main resolution-limiting factors. This leads to aliasing in LR data and is essential to make the use of SR reasonable \cite{Vandewalle2010}\footnote{Modulation transfer function (MTF) measurements to quantify the potential amount of aliasing for certain pixel sizes are reported in \cite{Edmundoptics2016}.}.

The camera-to-scene working distance is chosen between 0.5\,m and 1\,m depending on the scene content and the camera movement type. All scenes comprise planar surfaces (\eg printed text) or 3-D topography with maximum depths of about 0.3\,m. For scenes with multiple objects of different depths, the focus plane is set such that objects close to the image center are in focus. 

\subsection{Motion Types and Environmental Conditions}
\label{sec:MotionAndEnvironmentConditions}

\begin{table}[!t]
	\centering
	\footnotesize
	\caption{Parameters of the Basler acA2000-50gm camera \cite{Basler2016} and the f/1.8 16mm fixed-focus length lens for our data collection \cite{Edmundoptics2016}.}
	\begin{tabular}{cll}
		\toprule
		\multirow{5}{*}{\rotatebox[origin=c]{90}{\textbf{Sensor}}}
		& \textbf{Native resolution [pixels]}								& $2048 \times 1088$ \\
		& \textbf{Pixel size [$\mu$m]}											& $5.5 \times 5.5$ \\
		& \textbf{Sensor size [mm]} 												& $11.26 \times 5.98$ \\
	  & \textbf{Sensor type} 															& CMOS (CMOSIS, CMV2000) \\
		& \textbf{Bit depth} 																& 12 \\
		\midrule
		\multirow{5}{*}{\rotatebox[origin=c]{90}{\textbf{Lens}}}
		& \textbf{Focal length [mm]}												& 16.0 \\
		& \textbf{Field of view [mm/$^\circ]$}							& 60 mm - 31.3$^\circ$ \\
		& \textbf{Aperture [f/\#]}													& f/1.8 - f/16 \\
		& \textbf{Optical resolution [lp/mm]}								& 145 \\
		& \textbf{Working distance [mm]}         						& 100 - $\infty$ \\
		\bottomrule
	\end{tabular}
	\label{tab:hardwareParameters}
\end{table}

\begin{figure}[!t]
	\centering
	\includegraphics[width=0.235\textwidth]{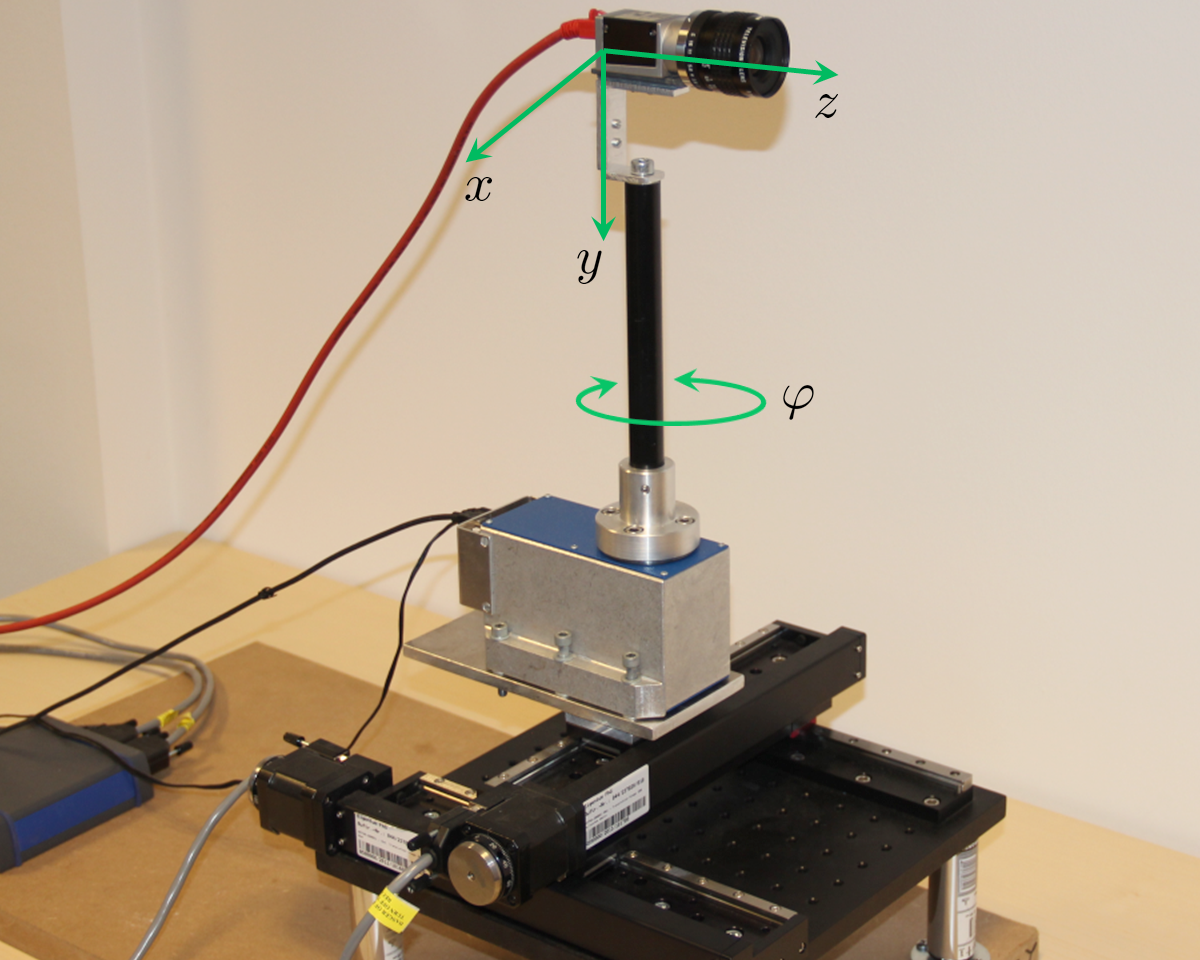}
	\includegraphics[width=0.235\textwidth]{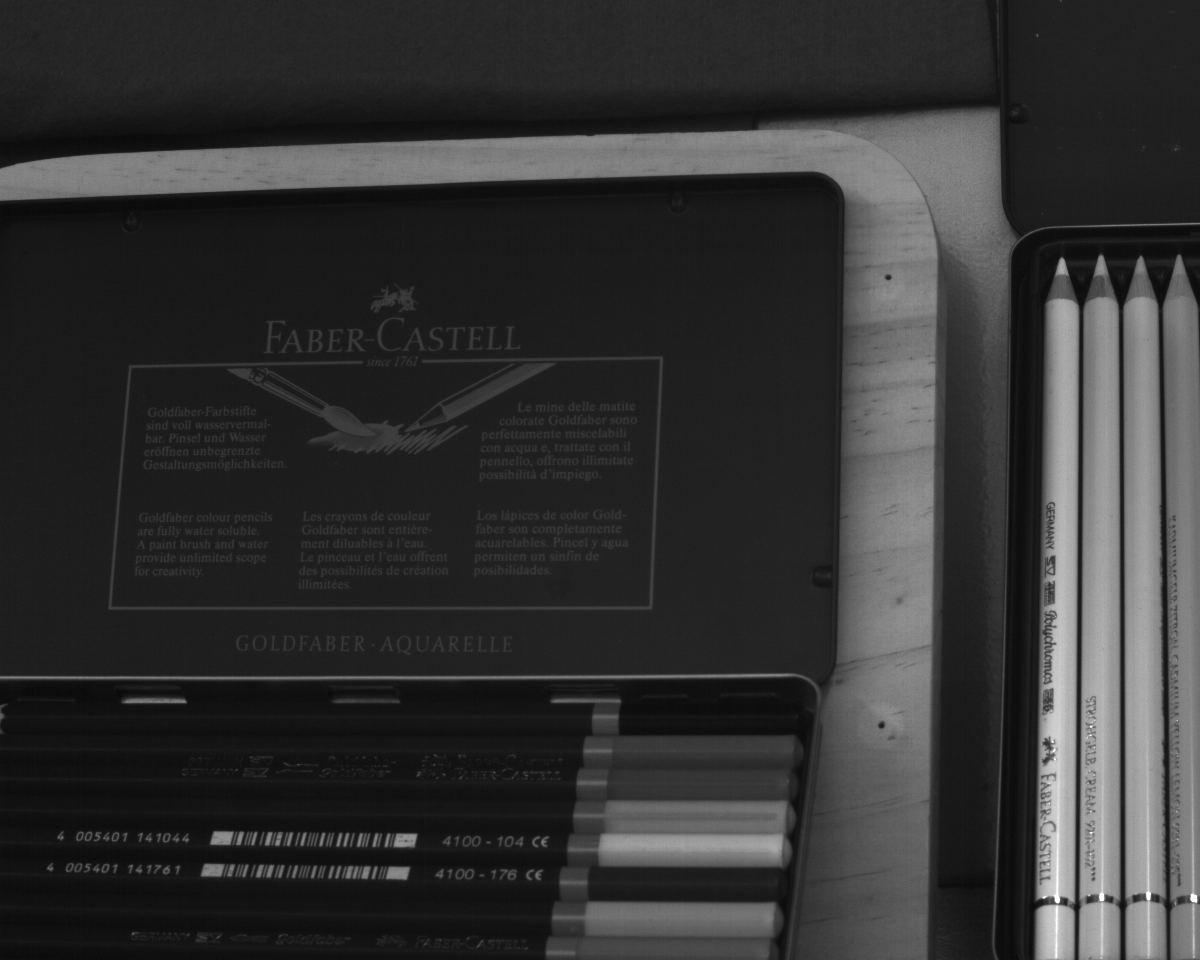}\\[0.7ex]
	\includegraphics[width=0.235\textwidth]{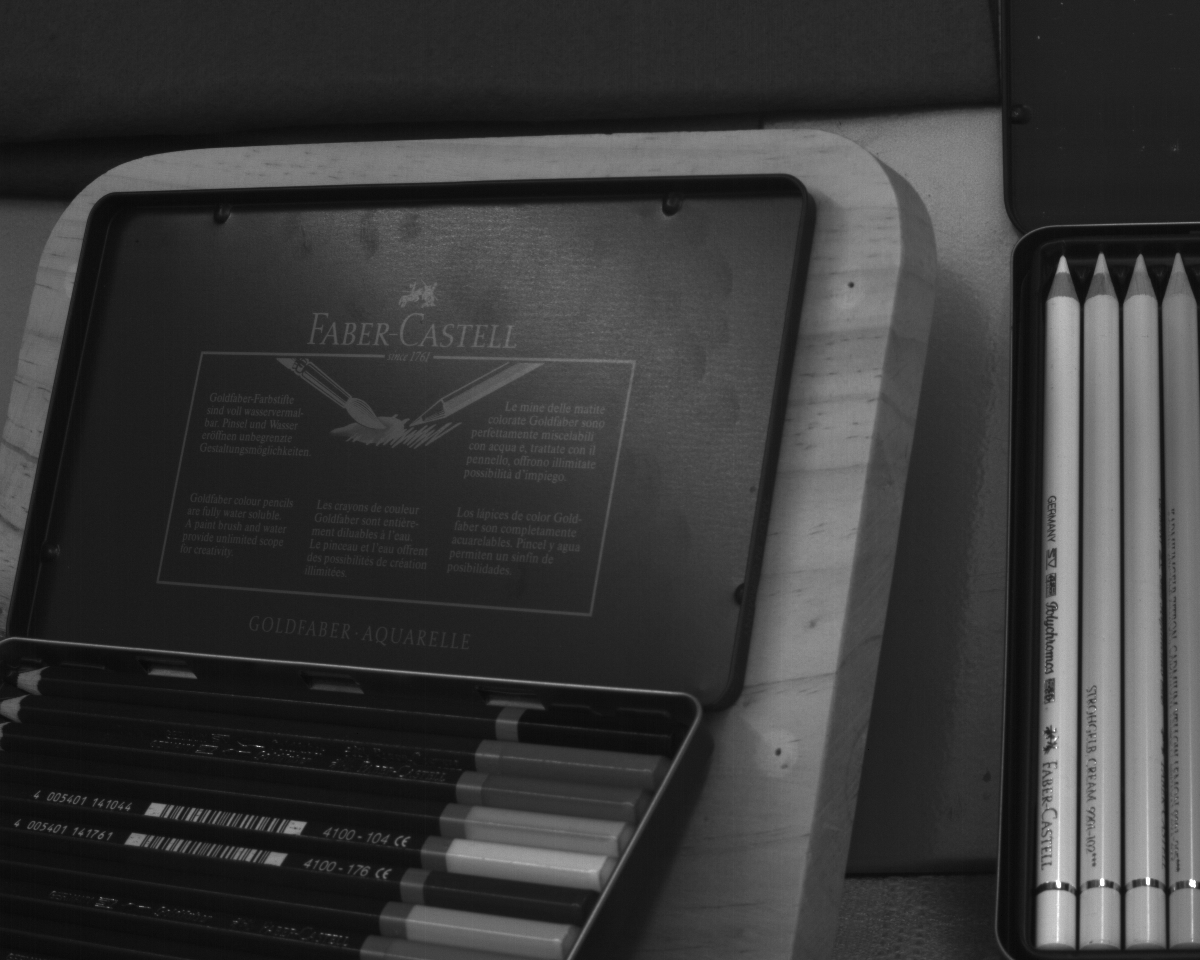}
	\includegraphics[width=0.235\textwidth]{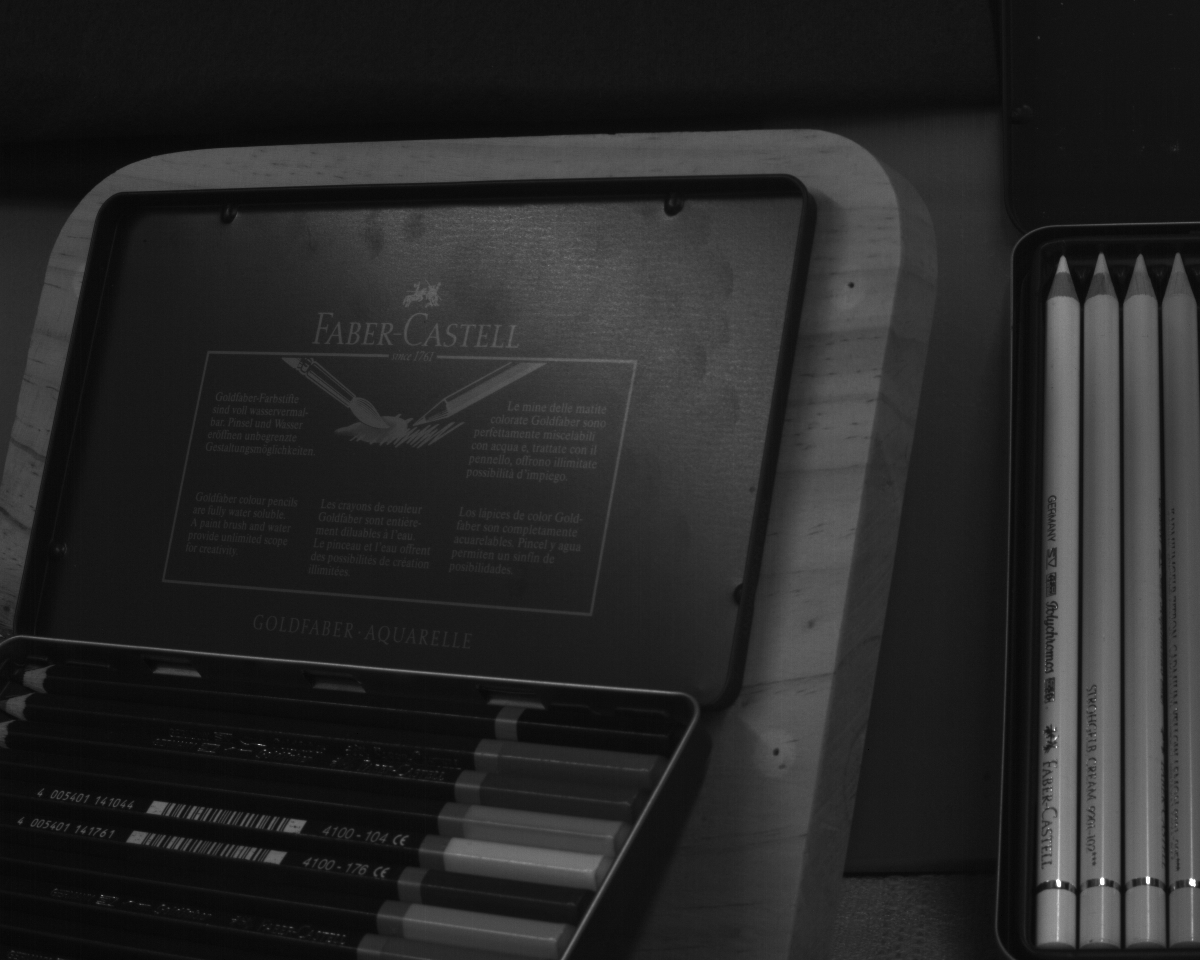}	
	\caption{Proposed hardware setup and an example scene. Our data comprise challenging conditions for SR like local object motion and photometric variations.}
	\label{fig:hardwareSetup}
\end{figure}

The camera is mounted on a position stage as shown in \fref{fig:hardwareSetup} and its pose is controlled by a stepper motor and a height-adjustable table. This enables camera panning in one dimension (in-plane rotation) and translations in three dimensions. We consider four basic motion types using different camera trajectories, see \tref{tab:motionAndPhotometricTypes}. The photometric conditions are controlled by artificial lighting and we consider bright (\textit{daylight}) and low-light illumination (\textit{nightlight}). In conjunction with movements of objects in a scene, this forms five dataset categories with different levels of difficulty.

\newcolumntype{P}[1]{>{\centering\arraybackslash}p{#1}}
\begin{table*}[!bh]
	\footnotesize
	\caption{Comparison of our SupER database to other publicly available benchmark datasets. Unlike existing datasets, we provide captured image sequences at four spatial resolution levels (ground truth HR images plus three levels for LR images) without involving simulation. In addition, our LR data is provided at five compression levels (uncoded plus four quantization levels) using H.265/HEVC video coding. All quantitative properties refer to the original versions of the datasets. We excluded datasets without separate LR data.}
	\centering
	\begin{tabular}{l P{6.6em} P{6.2em} P{6.5em} P{6.2em} P{6.2em} P{6.2em}}
		\toprule
		\textbf{Dataset}		& \textbf{Real/Simulated} & \textbf{\# Res. levels} & \textbf{\# Comp. levels} & \textbf{\# Sequences}	& \textbf{\# LR images}	& \textbf{\# HR images}	\\
		\midrule
		MDSP \cite{Farsiu2014}						& Real				& 1 & 1		& 21						& 915			& \xmark \\
		Vandewalle \cite{Vandewalle2016}	& Real				& 1 & 1		& 3							& 12			& \xmark \\
 		Liu and Sun \cite{Liu2014}				& Simulated		& 2 & 1		& 4							&	171			&	171 \\
		Yang \etal \cite{Yang2014a}				& Simulated		& 4 & 1		& Single images	& 2,061		& 229 \\
		DVI2K \cite{Agustsson2017}				& Simulated		& 4 & 1		& Single images	&	6,000		&	1,000 \\
		Qu \etal \cite{Qu2016}						& Real				& 2 & 1		& Single images & 93			& 93 \\
		\midrule
		\textbf{SupER (ours)}							& Real				& 4 & 5		& 254						& 85,050	& 5,670 \\
		\bottomrule
	\end{tabular}
	\label{tab:databaseOverview}
\end{table*}

\begin{table}[!t]
	\footnotesize
	\caption{Overview of motion types captured in our database.}
	\centering
	\begin{tabular}{p{14em} p{14em}}
		\toprule
		\textbf{Motion type}													& \textbf{Camera trajectory} \\
		\midrule
		Translation $z$											& Linear 								\\
		Translation $x$,$z$									& Sinusoidal 						\\
		Panning															& Circular 							\\
		Translation $x$,$y$,$z$ and panning & Joint sinusoidal and circular \\
		\bottomrule
	\end{tabular}
	\label{tab:motionAndPhotometricTypes}
\end{table}

\textit{Global motion.} 
	This baseline category consists of static scenes with constant daylight 
conditions. All inter-frame motion is global camera motion using the trajectories in \tref{tab:motionAndPhotometricTypes} with uniformly distributed camera positions.
	
\textit{Local motion.}
	This category consists of dynamic scenes captured under daylight conditions 
with a static camera but moving objects, see 
\fref{fig:hardwareSetup}. All inter-frame motion is translational and/or rotational 
object motion. 
 
\textit{Mixed motion.}
	This category combines global and local motion. To this end, 
	each camera trajectory is combined with translational and/or rotational 
	object motion.

\textit{Photometric variation.} 
	This category comprises sequences of $K$ frames, where the first $
K - K_{\mathrm{night}}$ frames are taken from the global, local, and mixed 
motion data and the remaining $K_{\mathrm{night}}$ \textit{outlier} frames are 
obtained under nightlight conditions, see \fref{fig:hardwareSetup}.

\textit{Video compression.} This category further extends the 
aforementioned datasets by five H.265/HEVC compression levels, \ie all LR images are provided as uncompressed and compressed versions.

Overall, our database comprises 56 global, 56 mixed, and 14 local motion image 
sequences with $K = 40$ frames each captured from 14 scenes. The photometric 
variation datasets augment each sequence by $K_{\mathrm{night}} = 5$ 
nightlight images. 

\subsection{Comparison to Existing Datasets}
\label{sec:ComparisonToExistingDatasets}

Our acquisition scheme goes beyond existing real-world databases 
\cite{Farsiu2014,Vandewalle2016} by providing 1) real LR acquisitions, 
2) compressed data using H.265/HEVC coding, and 3) corresponding HR ground 
truth data as summarized in \tref{tab:databaseOverview}. Its size in terms of LR/HR exemplars also exceeds the state-of-the-art by an order of magnitude. This fosters both large-scale benchmarks and training of learning-based methods.

\subsubsection{Hardware Binning for Realistic Image Artifacts}
\label{sec:HardwareBinningVsSoftwareBinning}

In contrast to our image formation in \eqref{eqn:hardwareBinningSamplingOp}, the widely used \textit{software binning} \cite{Liu2014,Yang2014a,Agustsson2017} is based based on the model:
\begin{equation}
	\label{eqn:softwareBinning}
	\vec{Y}_{b} = \mathcal{D}_b \left\{ \vec{X} \right\} + \vec{\eta} \enspace,
\end{equation}
where $\vec{X}$ is a discretization of $x(\vec{u})$ and $\vec{\eta}$ is additive noise. Typically $\vec{X}$ is a reference image from an existing database, \eg LIVE \cite{Sheikh2016}, Set5, Set14, B100, or L20 \cite{Timofte2016}, or from HR videos \cite{Liu2014}. Note that a simulation cannot model the true physics of image formation since it does not have access to the original irradiance $x(\vec{u})$ as used in \eqref{eqn:hardwareBinning}. For instance, simulated data is based on simplified models for $\vec{\eta}$ like ideal quantization noise \cite{Yang2014a} or Gaussian noise \cite{Zhang2018}. The sampling $\mathcal{D}_b\{\cdot\}$ is modeled by bicubic downsampling or other artificial operators \cite{Liu2014,Agustsson2017}. The LR images in our database are degraded by noise and subsampling from actual physical processes.

\begin{figure*}[!t]
	\tiny
	\centering
	\setlength \figurewidth{0.366\textwidth}
	\setlength \figureheight{0.67\figurewidth}
	\raisebox{1.65cm}{\rotatebox{90}{\footnotesize $2\times2$ binning}}\quad
	\subfloat[PSNR ($\rho = -0.12$)]{%
%
\definecolor{mycolor1}{rgb}{0.00000,0.00000,0.87500}%
\definecolor{mycolor2}{rgb}{0.00000,0.12500,1.00000}%
\definecolor{mycolor3}{rgb}{0.00000,0.25000,1.00000}%
\definecolor{mycolor4}{rgb}{0.00000,0.37500,1.00000}%
\definecolor{mycolor5}{rgb}{0.00000,0.50000,1.00000}%
\definecolor{mycolor6}{rgb}{0.00000,0.62500,1.00000}%
\definecolor{mycolor7}{rgb}{0.00000,0.75000,1.00000}%
\definecolor{mycolor8}{rgb}{1.00000,0.87500,0.00000}%
\definecolor{mycolor9}{rgb}{1.00000,0.75000,0.00000}%
\begin{tikzpicture}

\begin{axis}[%
width=0.75\figurewidth,
height=\figureheight,
at={(0\figurewidth,0\figureheight)},
scale only axis,
xmin=34.0,
xmax=39.0,
xtick={34, 35, 36, 37, 38, 39},
xlabel={PSNR on simulated data},
ymin=34.0,
ymax=39.0,
ytick={34, 35, 36, 37, 38, 39},
ylabel={PSNR on real data},
axis background/.style={fill=white},
xmajorgrids,
ymajorgrids,
xlabel near ticks,ylabel near ticks,scaled y ticks=false,yticklabel style={/pgf/number format/fixed, /pgf/number format/precision=2},
]
\addplot [color=mycolor1,only marks,mark=*,mark options={solid,fill=mycolor1},forget plot]
  table[row sep=crcr]{%
36.4727734688047	34.6358766689611\\
};
\node[above left, align=right, text=black]
at (axis cs:36.473,34.636) {EBSR};
\addplot [color=blue,only marks,mark=*,mark options={solid,fill=blue},forget plot]
  table[row sep=crcr]{%
36.5170255291429	34.5187700780102\\
};
\node[below right, align=right, text=black]
at (axis cs:36.517,34.519) {ScSR};
\addplot [color=mycolor2,only marks,mark=*,mark options={solid,fill=mycolor2},forget plot]
  table[row sep=crcr]{%
37.7068484358141	35.5364174357956\\
};
\node[above right, align=right, text=black]
at (axis cs:37.707,35.536) {NBSRF};
\addplot [color=mycolor3,only marks,mark=*,mark options={solid,fill=mycolor3},forget plot]
  table[row sep=crcr]{%
37.5678647558585	35.5443634049649\\
};
\node[above left, align=right, text=black]
at (axis cs:37.568,35.544) {A+};
\addplot [color=mycolor4,only marks,mark=*,mark options={solid,fill=mycolor4},forget plot]
  table[row sep=crcr]{%
37.6325036390408	35.3525032434018\\
};
\node[below left, align=right, text=black]
at (axis cs:37.633,35.353) {SRCNN};
\addplot [color=mycolor5,only marks,mark=*,mark options={solid,fill=mycolor5},forget plot]
  table[row sep=crcr]{%
38.4938245372167	35.6429811041557\\
};
\node[above, align=right, text=black]
at (axis cs:38.494,35.643) {DRCN};
\addplot [color=mycolor6,only marks,mark=*,mark options={solid,fill=mycolor6},forget plot]
  table[row sep=crcr]{%
38.2836345929598	35.3684218413783\\
};
\node[below, align=right, text=black]
at (axis cs:38.284,35.368) {VDSR};
\addplot [color=mycolor7,only marks,mark=*,mark options={solid,fill=mycolor7},forget plot]
  table[row sep=crcr]{%
37.5227747366731	35.0423545999328\\
};
\node[below left, align=right, text=black]
at (axis cs:37.523,35.042) {SESR};
\addplot [color=mycolor8,only marks,mark=*,mark options={solid,fill=mycolor8},forget plot]
  table[row sep=crcr]{%
37.0331073364724	37.1167919129136\\
};
\node[above left, align=right, text=black]
at (axis cs:37.033,37.117) {NUISR};
\addplot [color=mycolor9,only marks,mark=*,mark options={solid,fill=mycolor9},forget plot]
  table[row sep=crcr]{%
36.390579716823	36.3728231014234\\
};
\node[below right, align=right, text=black]
at (axis cs:36.391,36.373) {WNUISR};
\addplot [color=orange!50!mycolor9,only marks,mark=*,mark options={solid,fill=orange!50!mycolor9},forget plot]
  table[row sep=crcr]{%
37.222243343057	36.9854262760396\\
};
\node[below right, align=right, text=black]
at (axis cs:37.222,36.985) {HYSR};
\addplot [color=orange,only marks,mark=*,mark options={solid,fill=orange},forget plot]
  table[row sep=crcr]{%
36.3695599046128	36.424298759886\\
};
\node[above left, align=right, text=black]
at (axis cs:36.37,36.424) {DBRSR};
\addplot [color=red!25!orange,only marks,mark=*,mark options={solid,fill=red!25!orange},forget plot]
  table[row sep=crcr]{%
36.8939817206753	35.8426034255433\\
};
\node[below left, align=right, text=black]
at (axis cs:36.894,35.843) {L1BTV};
\addplot [color=red!50!orange,only marks,mark=*,mark options={solid,fill=red!50!orange},forget plot]
  table[row sep=crcr]{%
37.6703868309325	36.335226176798\\
};
\node[below, align=right, text=black]
at (axis cs:37.67,36.335) {BEPSR};
\addplot [color=red!75!orange,only marks,mark=*,mark options={solid,fill=red!75!orange},forget plot]
  table[row sep=crcr]{%
36.9807569588607	35.6271680802443\\
};
\node[below left, align=right, text=black]
at (axis cs:36.981,35.627) {IRWSR};
\addplot [color=black!25!red,only marks,mark=*,mark options={solid,fill=black!25!red},forget plot]
  table[row sep=crcr]{%
38.1111135968115	36.1771867993378\\
};
\node[below right, align=right, text=black]
at (axis cs:38.111,36.177) {VSRnet};
\addplot [color=black,dashed,line width=1.5pt,forget plot]
  table[row sep=crcr]{%
34.0	34.0\\
39.0	39.0\\
};
\end{axis}
\end{tikzpicture}%
%
}
	\subfloat[SSIM ($\rho = -0.06$)]{%
%
\definecolor{mycolor1}{rgb}{0.00000,0.00000,0.87500}%
\definecolor{mycolor2}{rgb}{0.00000,0.12500,1.00000}%
\definecolor{mycolor3}{rgb}{0.00000,0.25000,1.00000}%
\definecolor{mycolor4}{rgb}{0.00000,0.37500,1.00000}%
\definecolor{mycolor5}{rgb}{0.00000,0.50000,1.00000}%
\definecolor{mycolor6}{rgb}{0.00000,0.62500,1.00000}%
\definecolor{mycolor7}{rgb}{0.00000,0.75000,1.00000}%
\definecolor{mycolor8}{rgb}{1.00000,0.87500,0.00000}%
\definecolor{mycolor9}{rgb}{1.00000,0.75000,0.00000}%
\begin{tikzpicture}

\begin{axis}[%
width=0.75\figurewidth,
height=\figureheight,
at={(0\figurewidth,0\figureheight)},
scale only axis,
xmin=0.91,
xmax=0.96,
xtick={0.91, 0.92, 0.93, 0.94, 0.95, 0.96},
xlabel={SSIM on simulated data},
ymin=0.91,
ymax=0.96,
ylabel={SSIM on real data},
ytick={0.91, 0.92, 0.93, 0.94, 0.95, 0.96},
axis background/.style={fill=white},
xmajorgrids,
ymajorgrids,
xlabel near ticks,ylabel near ticks,scaled y ticks=false,yticklabel style={/pgf/number format/fixed, /pgf/number format/precision=2},
]
\addplot [color=mycolor1,only marks,mark=*,mark options={solid,fill=mycolor1},forget plot]
  table[row sep=crcr]{%
0.941627340130887	0.918591902850193\\
};
\node[below left, align=right, text=black]
at (axis cs:0.942,0.919) {EBSR};
\addplot [color=blue,only marks,mark=*,mark options={solid,fill=blue},forget plot]
  table[row sep=crcr]{%
0.944233098290079	0.908681737387885\\
};
\node[below left, align=right, text=black]
at (axis cs:0.944,0.909) {ScSR};
\addplot [color=mycolor2,only marks,mark=*,mark options={solid,fill=mycolor2},forget plot]
  table[row sep=crcr]{%
0.950794962917594	0.919191430144886\\
};
\node[below right, align=right, text=black]
at (axis cs:0.951,0.919) {NBSRF};
\addplot [color=mycolor3,only marks,mark=*,mark options={solid,fill=mycolor3},forget plot]
  table[row sep=crcr]{%
0.95006844473368	0.920620626806798\\
};
\node[below left, align=right, text=black]
at (axis cs:0.95,0.921) {A+};
\addplot [color=mycolor4,only marks,mark=*,mark options={solid,fill=mycolor4},forget plot]
  table[row sep=crcr]{%
0.949913708220112	0.919097653517722\\
};
\node[below left, align=right, text=black]
at (axis cs:0.95,0.919) {SRCNN};
\addplot [color=mycolor5,only marks,mark=*,mark options={solid,fill=mycolor5},forget plot]
  table[row sep=crcr]{%
0.954561553614212	0.920501665718026\\
};
\node[above, align=right, text=black]
at (axis cs:0.955,0.921) {DRCN};
\addplot [color=mycolor6,only marks,mark=*,mark options={solid,fill=mycolor6},forget plot]
  table[row sep=crcr]{%
0.9542662150658	0.918947653345517\\
};
\node[right, align=right, text=black]
at (axis cs:0.954,0.919) {VDSR};
\addplot [color=mycolor7,only marks,mark=*,mark options={solid,fill=mycolor7},forget plot]
  table[row sep=crcr]{%
0.948236067191986	0.913243057989351\\
};
\node[below left, align=right, text=black]
at (axis cs:0.948,0.913) {SESR};
\addplot [color=mycolor8,only marks,mark=*,mark options={solid,fill=mycolor8},forget plot]
  table[row sep=crcr]{%
0.947252138088177	0.94263449959613\\
};
\node[below left, align=right, text=black]
at (axis cs:0.947,0.943) {NUISR};
\addplot [color=mycolor9,only marks,mark=*,mark options={solid,fill=mycolor9},forget plot]
  table[row sep=crcr]{%
0.943494817648312	0.939700012400012\\
};
\node[below right, align=right, text=black]
at (axis cs:0.943,0.94) {WNUISR};
\addplot [color=orange!50!mycolor9,only marks,mark=*,mark options={solid,fill=orange!50!mycolor9},forget plot]
  table[row sep=crcr]{%
0.947956526035946	0.942746821711145\\
};
\node[right, align=right, text=black]
at (axis cs:0.948,0.943) {HYSR};
\addplot [color=orange,only marks,mark=*,mark options={solid,fill=orange},forget plot]
  table[row sep=crcr]{%
0.9426298550287	0.9406257437563\\
};
\node[below left, align=right, text=black]
at (axis cs:0.943,0.941) {DBRSR};
\addplot [color=red!25!orange,only marks,mark=*,mark options={solid,fill=red!25!orange},forget plot]
  table[row sep=crcr]{%
0.948438905267208	0.922443736946341\\
};
\node[above left, align=right, text=black]
at (axis cs:0.948,0.922) {L1BTV};
\addplot [color=red!50!orange,only marks,mark=*,mark options={solid,fill=red!50!orange},forget plot]
  table[row sep=crcr]{%
0.955632018969766	0.937317042631316\\
};
\node[above, align=right, text=black]
at (axis cs:0.956,0.937) {BEPSR};
\addplot [color=red!75!orange,only marks,mark=*,mark options={solid,fill=red!75!orange},forget plot]
  table[row sep=crcr]{%
0.952990477410079	0.936934881976324\\
};
\node[below left, align=right, text=black]
at (axis cs:0.953,0.937) {IRWSR};
\addplot [color=black!25!red,only marks,mark=*,mark options={solid,fill=black!25!red},forget plot]
  table[row sep=crcr]{%
0.953370317345021	0.933211862860209\\
};
\node[below left, align=right, text=black]
at (axis cs:0.953,0.933) {VSRnet};
\addplot [color=black,dashed,line width=1.5pt,forget plot]
  table[row sep=crcr]{%
0.899594920014006	0.899594920014006\\
0.965188339159464	0.965188339159464\\
};
\end{axis}
\end{tikzpicture}%
%
}
	\subfloat[IFC ($\rho = -0.50$)]{%
%
\definecolor{mycolor1}{rgb}{0.00000,0.00000,0.87500}%
\definecolor{mycolor2}{rgb}{0.00000,0.12500,1.00000}%
\definecolor{mycolor3}{rgb}{0.00000,0.25000,1.00000}%
\definecolor{mycolor4}{rgb}{0.00000,0.37500,1.00000}%
\definecolor{mycolor5}{rgb}{0.00000,0.50000,1.00000}%
\definecolor{mycolor6}{rgb}{0.00000,0.62500,1.00000}%
\definecolor{mycolor7}{rgb}{0.00000,0.75000,1.00000}%
\definecolor{mycolor8}{rgb}{1.00000,0.87500,0.00000}%
\definecolor{mycolor9}{rgb}{1.00000,0.75000,0.00000}%
\begin{tikzpicture}

\begin{axis}[%
width=0.75\figurewidth,
height=\figureheight,
at={(0\figurewidth,0\figureheight)},
scale only axis,
xmin=3.5,
xmax=11.0,
xtick={3.5, 5, 6.5, 8, 9.5, 11},
xlabel={IFC on simulated data},
ymin=3.5,
ymax=11.0,
ytick={3.5, 5, 6.5, 8, 9.5, 11},
ylabel={IFC on real data},
axis background/.style={fill=white},
xmajorgrids,
ymajorgrids,
xlabel near ticks,ylabel near ticks,scaled y ticks=false,yticklabel style={/pgf/number format/fixed, /pgf/number format/precision=2},
]
\addplot [color=mycolor1,only marks,mark=*,mark options={solid,fill=mycolor1},forget plot]
  table[row sep=crcr]{%
6.93545018392183	3.99210747756917\\
};
\node[above, align=right, text=black]
at (axis cs:6.935,3.992) {EBSR};
\addplot [color=blue,only marks,mark=*,mark options={solid,fill=blue},forget plot]
  table[row sep=crcr]{%
6.47813185858394	3.97722305817664\\
};
\node[left, align=right, text=black]
at (axis cs:6.478,3.977) {ScSR};
\addplot [color=mycolor2,only marks,mark=*,mark options={solid,fill=mycolor2},forget plot]
  table[row sep=crcr]{%
9.37337838318253	4.19860799728587\\
};
\node[below right, align=right, text=black]
at (axis cs:9.373,4.199) {NBSRF};
\addplot [color=mycolor3,only marks,mark=*,mark options={solid,fill=mycolor3},forget plot]
  table[row sep=crcr]{%
9.28015661690287	4.20634905788887\\
};
\node[below, align=right, text=black]
at (axis cs:9.28,4.206) {A+};
\addplot [color=mycolor4,only marks,mark=*,mark options={solid,fill=mycolor4},forget plot]
  table[row sep=crcr]{%
8.6112950139362	4.13912070930993\\
};
\node[below left, align=right, text=black]
at (axis cs:8.611,4.139) {SRCNN};
\addplot [color=mycolor5,only marks,mark=*,mark options={solid,fill=mycolor5},forget plot]
  table[row sep=crcr]{%
9.48570930088927	4.2340768968823\\
};
\node[above right, align=right, text=black]
at (axis cs:9.486,4.234) {DRCN};
\addplot [color=mycolor6,only marks,mark=*,mark options={solid,fill=mycolor6},forget plot]
  table[row sep=crcr]{%
8.97594120769305	4.18555932186824\\
};
\node[above left, align=right, text=black]
at (axis cs:8.976,4.186) {VDSR};
\addplot [color=mycolor7,only marks,mark=*,mark options={solid,fill=mycolor7},forget plot]
  table[row sep=crcr]{%
7.61158639861055	4.09378952811317\\
};
\node[above, align=right, text=black]
at (axis cs:7.612,4.094) {SESR};
\addplot [color=mycolor8,only marks,mark=*,mark options={solid,fill=mycolor8},forget plot]
  table[row sep=crcr]{%
6.0311346875589	5.17571396704505\\
};
\node[above left, align=right, text=black]
at (axis cs:6.031,5.176) {NUISR};
\addplot [color=mycolor9,only marks,mark=*,mark options={solid,fill=mycolor9},forget plot]
  table[row sep=crcr]{%
5.43983514465372	4.73643216247299\\
};
\node[above left, align=right, text=black]
at (axis cs:5.44,4.736) {WNUISR};
\addplot [color=orange!50!mycolor9,only marks,mark=*,mark options={solid,fill=orange!50!mycolor9},forget plot]
  table[row sep=crcr]{%
6.69484710770084	5.07400698924607\\
};
\node[below right, align=right, text=black]
at (axis cs:6.695,5.074) {HYSR};
\addplot [color=orange,only marks,mark=*,mark options={solid,fill=orange},forget plot]
  table[row sep=crcr]{%
5.18126597947771	4.72452444424985\\
};
\node[below left, align=right, text=black]
at (axis cs:5.181,4.725) {DBRSR};
\addplot [color=red!25!orange,only marks,mark=*,mark options={solid,fill=red!25!orange},forget plot]
  table[row sep=crcr]{%
5.94540634064281	4.72726813963275\\
};
\node[below right, align=right, text=black]
at (axis cs:5.945,4.727) {L1BTV};
\addplot [color=red!50!orange,only marks,mark=*,mark options={solid,fill=red!50!orange},forget plot]
  table[row sep=crcr]{%
6.37035832727326	5.02388803471728\\
};
\node[above right, align=right, text=black]
at (axis cs:6.37,5.024) {BEPSR};
\addplot [color=red!75!orange,only marks,mark=*,mark options={solid,fill=red!75!orange},forget plot]
  table[row sep=crcr]{%
6.38652506050469	5.15008229609687\\
};
\node[below left, align=right, text=black]
at (axis cs:6.387,5.15) {IRWSR};
\addplot [color=black!25!red,only marks,mark=*,mark options={solid,fill=black!25!red},forget plot]
  table[row sep=crcr]{%
8.69582074293579	4.49976111453502\\
};
\node[above right, align=right, text=black]
at (axis cs:8.696,4.5) {VSRnet};
\addplot [color=black,dashed,line width=1.5pt,forget plot]
  table[row sep=crcr]{%
3.0	3.0\\
11.0	11.0\\
};
\end{axis}
\end{tikzpicture}%
%
}\\[-0.6ex]
	\raisebox{1.65cm}{\rotatebox{90}{\footnotesize $4\times4$ binning}}\quad
	\subfloat[PSNR ($\rho = 0.24$)]{%
%
\definecolor{mycolor1}{rgb}{0.00000,0.00000,0.87500}%
\definecolor{mycolor2}{rgb}{0.00000,0.12500,1.00000}%
\definecolor{mycolor3}{rgb}{0.00000,0.25000,1.00000}%
\definecolor{mycolor4}{rgb}{0.00000,0.37500,1.00000}%
\definecolor{mycolor5}{rgb}{0.00000,0.50000,1.00000}%
\definecolor{mycolor6}{rgb}{0.00000,0.62500,1.00000}%
\definecolor{mycolor7}{rgb}{0.00000,0.75000,1.00000}%
\definecolor{mycolor8}{rgb}{1.00000,0.87500,0.00000}%
\definecolor{mycolor9}{rgb}{1.00000,0.75000,0.00000}%
\begin{tikzpicture}

\begin{axis}[%
width=0.75\figurewidth,
height=\figureheight,
at={(0\figurewidth,0\figureheight)},
scale only axis,
xmin=30.0,
xmax=33.0,
xlabel={PSNR on simulated data},
xtick={30.0, 31, 32, 33},
ymin=30.0,
ymax=33.0,
ylabel={PSNR on real data},
ytick={30.0, 31, 32, 33},
axis background/.style={fill=white},
xmajorgrids,
ymajorgrids,
xlabel near ticks,ylabel near ticks,scaled y ticks=false,yticklabel style={/pgf/number format/fixed, /pgf/number format/precision=2},
]
\addplot [color=mycolor1,only marks,mark=*,mark options={solid,fill=mycolor1},forget plot]
  table[row sep=crcr]{%
31.6234807451061	31.1615866512899\\
};
\node[below left, align=right, text=black]
at (axis cs:31.623,31.162) {EBSR};
\addplot [color=blue,only marks,mark=*,mark options={solid,fill=blue},forget plot]
  table[row sep=crcr]{%
31.0517011520133	30.3858623051139\\
};
\node[below left, align=right, text=black]
at (axis cs:31.052,30.386) {ScSR};
\addplot [color=mycolor2,only marks,mark=*,mark options={solid,fill=mycolor2},forget plot]
  table[row sep=crcr]{%
32.0117982921062	31.3149042676999\\
};
\node[below left, align=right, text=black]
at (axis cs:32.012,31.315) {NBSRF};
\addplot [color=mycolor3,only marks,mark=*,mark options={solid,fill=mycolor3},forget plot]
  table[row sep=crcr]{%
31.7352969404309	31.1337451019311\\
};
\node[below left, align=right, text=black]
at (axis cs:31.735,31.134) {A+};
\addplot [color=mycolor4,only marks,mark=*,mark options={solid,fill=mycolor4},forget plot]
  table[row sep=crcr]{%
31.8766986097007	31.0946685387534\\
};
\node[below right, align=right, text=black]
at (axis cs:31.877,31.095) {SRCNN};
\addplot [color=mycolor5,only marks,mark=*,mark options={solid,fill=mycolor5},forget plot]
  table[row sep=crcr]{%
32.5260463685807	31.3042712075052\\
};
\node[below right, align=right, text=black]
at (axis cs:32.526,31.304) {DRCN};
\addplot [color=mycolor6,only marks,mark=*,mark options={solid,fill=mycolor6},forget plot]
  table[row sep=crcr]{%
32.3549266353463	31.1161930564665\\
};
\node[below right, align=right, text=black]
at (axis cs:32.355,31.116) {VDSR};
\addplot [color=mycolor7,only marks,mark=*,mark options={solid,fill=mycolor7},forget plot]
  table[row sep=crcr]{%
31.7171594431076	30.6089935071775\\
};
\node[below left, align=right, text=black]
at (axis cs:31.717,30.609) {SESR};
\addplot [color=mycolor8,only marks,mark=*,mark options={solid,fill=mycolor8},forget plot]
  table[row sep=crcr]{%
30.5791769810906	31.0777273135743\\
};
\node[below left, align=right, text=black]
at (axis cs:30.579,31.078) {NUISR};
\addplot [color=mycolor9,only marks,mark=*,mark options={solid,fill=mycolor9},forget plot]
  table[row sep=crcr]{%
31.5838025314169	32.1678698160175\\
};
\node[below left, align=right, text=black]
at (axis cs:31.584,32.168) {WNUISR};
\addplot [color=orange!50!mycolor9,only marks,mark=*,mark options={solid,fill=orange!50!mycolor9},forget plot]
  table[row sep=crcr]{%
31.235382421016	31.7071457225608\\
};
\node[below left, align=right, text=black]
at (axis cs:31.235,31.707) {HYSR};
\addplot [color=orange,only marks,mark=*,mark options={solid,fill=orange},forget plot]
  table[row sep=crcr]{%
31.640395164972	32.2588180272785\\
};
\node[below right, align=right, text=black]
at (axis cs:31.64,32.259) {DBRSR};
\addplot [color=red!25!orange,only marks,mark=*,mark options={solid,fill=red!25!orange},forget plot]
  table[row sep=crcr]{%
32.1322490499189	32.097158720787\\
};
\node[below right, align=right, text=black]
at (axis cs:32.132,32.097) {L1BTV};
\addplot [color=red!50!orange,only marks,mark=*,mark options={solid,fill=red!50!orange},forget plot]
  table[row sep=crcr]{%
32.2680779968444	32.3203526436632\\
};
\node[above left, align=right, text=black]
at (axis cs:32.268,32.32) {BEPSR};
\addplot [color=red!75!orange,only marks,mark=*,mark options={solid,fill=red!75!orange},forget plot]
  table[row sep=crcr]{%
31.8484712806603	31.472815516692\\
};
\node[below left, align=right, text=black]
at (axis cs:31.848,31.473) {IRWSR};
\addplot [color=black!25!red,only marks,mark=*,mark options={solid,fill=black!25!red},forget plot]
  table[row sep=crcr]{%
32.1888381930584	31.5596474053808\\
};
\node[below right, align=right, text=black]
at (axis cs:32.189,31.56) {VSRnet};
\addplot [color=black,dashed,line width=1.5pt,forget plot]
  table[row sep=crcr]{%
30.0	30.0\\
33.0	33.0\\
};
\end{axis}
\end{tikzpicture}%
%
}
	\subfloat[SSIM ($\rho = 0.64$)]{%
%
\definecolor{mycolor1}{rgb}{0.00000,0.00000,0.87500}%
\definecolor{mycolor2}{rgb}{0.00000,0.12500,1.00000}%
\definecolor{mycolor3}{rgb}{0.00000,0.25000,1.00000}%
\definecolor{mycolor4}{rgb}{0.00000,0.37500,1.00000}%
\definecolor{mycolor5}{rgb}{0.00000,0.50000,1.00000}%
\definecolor{mycolor6}{rgb}{0.00000,0.62500,1.00000}%
\definecolor{mycolor7}{rgb}{0.00000,0.75000,1.00000}%
\definecolor{mycolor8}{rgb}{1.00000,0.87500,0.00000}%
\definecolor{mycolor9}{rgb}{1.00000,0.75000,0.00000}%
\begin{tikzpicture}

\begin{axis}[%
width=0.75\figurewidth,
height=\figureheight,
at={(0\figurewidth,0\figureheight)},
scale only axis,
xmin=0.82,
xmax=0.90,
xlabel={SSIM on simulated data},
ymin=0.82,
ymax=0.90,
ylabel={SSIM on real data},
axis background/.style={fill=white},
xmajorgrids,
ymajorgrids,
xlabel near ticks,ylabel near ticks,scaled y ticks=false,yticklabel style={/pgf/number format/fixed, /pgf/number format/precision=2},
]
\addplot [color=mycolor1,only marks,mark=*,mark options={solid,fill=mycolor1},forget plot]
  table[row sep=crcr]{%
0.859690334264647	0.856416206384163\\
};
\node[below left, align=right, text=black]
at (axis cs:0.86,0.856) {EBSR};
\addplot [color=blue,only marks,mark=*,mark options={solid,fill=blue},forget plot]
  table[row sep=crcr]{%
0.852648399062092	0.841508072577499\\
};
\node[below left, align=right, text=black]
at (axis cs:0.853,0.842) {ScSR};
\addplot [color=mycolor2,only marks,mark=*,mark options={solid,fill=mycolor2},forget plot]
  table[row sep=crcr]{%
0.867092506945947	0.858349776485079\\
};
\node[above, align=right, text=black]
at (axis cs:0.867,0.858) {NBSRF};
\addplot [color=mycolor3,only marks,mark=*,mark options={solid,fill=mycolor3},forget plot]
  table[row sep=crcr]{%
0.861022203490453	0.854623290579988\\
};
\node[below right, align=right, text=black]
at (axis cs:0.861,0.855) {A+};
\addplot [color=mycolor4,only marks,mark=*,mark options={solid,fill=mycolor4},forget plot]
  table[row sep=crcr]{%
0.864709284158895	0.857105426494556\\
};
\node[below right, align=right, text=black]
at (axis cs:0.865,0.857) {SRCNN};
\addplot [color=mycolor5,only marks,mark=*,mark options={solid,fill=mycolor5},forget plot]
  table[row sep=crcr]{%
0.875741370852102	0.863244530667138\\
};
\node[above right, align=right, text=black]
at (axis cs:0.876,0.863) {DRCN};
\addplot [color=mycolor6,only marks,mark=*,mark options={solid,fill=mycolor6},forget plot]
  table[row sep=crcr]{%
0.875637775466003	0.862479628372608\\
};
\node[below right, align=right, text=black]
at (axis cs:0.876,0.862) {VDSR};
\addplot [color=mycolor7,only marks,mark=*,mark options={solid,fill=mycolor7},forget plot]
  table[row sep=crcr]{%
0.863338263863315	0.848288883910185\\
};
\node[below left, align=right, text=black]
at (axis cs:0.863,0.848) {SESR};
\addplot [color=mycolor8,only marks,mark=*,mark options={solid,fill=mycolor8},forget plot]
  table[row sep=crcr]{%
0.831635521136856	0.847855654715386\\
};
\node[below left, align=right, text=black]
at (axis cs:0.832,0.848) {NUISR};
\addplot [color=mycolor9,only marks,mark=*,mark options={solid,fill=mycolor9},forget plot]
  table[row sep=crcr]{%
0.855895915953595	0.872496156354921\\
};
\node[below left, align=right, text=black]
at (axis cs:0.856,0.872) {WNUISR};
\addplot [color=orange!50!mycolor9,only marks,mark=*,mark options={solid,fill=orange!50!mycolor9},forget plot]
  table[row sep=crcr]{%
0.847425718490648	0.862394534754597\\
};
\node[below left, align=right, text=black]
at (axis cs:0.847,0.862) {HYSR};
\addplot [color=orange,only marks,mark=*,mark options={solid,fill=orange},forget plot]
  table[row sep=crcr]{%
0.858183339715091	0.875724598658509\\
};
\node[above left, align=right, text=black]
at (axis cs:0.858,0.876) {DBRSR};
\addplot [color=red!25!orange,only marks,mark=*,mark options={solid,fill=red!25!orange},forget plot]
  table[row sep=crcr]{%
0.881486730762282	0.881095833427546\\
};
\node[below right, align=right, text=black]
at (axis cs:0.881,0.881) {L1BTV};
\addplot [color=red!50!orange,only marks,mark=*,mark options={solid,fill=red!50!orange},forget plot]
  table[row sep=crcr]{%
0.883156499247047	0.89570990926857\\
};
\node[below left, align=right, text=black]
at (axis cs:0.883,0.896) {BEPSR};
\addplot [color=red!75!orange,only marks,mark=*,mark options={solid,fill=red!75!orange},forget plot]
  table[row sep=crcr]{%
0.880589417788299	0.883965535737172\\
};
\node[above left, align=right, text=black]
at (axis cs:0.881,0.884) {IRWSR};
\addplot [color=black!25!red,only marks,mark=*,mark options={solid,fill=black!25!red},forget plot]
  table[row sep=crcr]{%
0.872056942880644	0.86924017172172\\
};
\node[above right, align=right, text=black]
at (axis cs:0.872,0.869) {VSRnet};
\addplot [color=black,dashed,line width=1.5pt,forget plot]
  table[row sep=crcr]{%
0.82	0.82\\
0.90	0.90\\
};
\end{axis}
\end{tikzpicture}%
%
}
	\subfloat[IFC ($\rho = -0.19$)]{%
%
\definecolor{mycolor1}{rgb}{0.00000,0.00000,0.87500}%
\definecolor{mycolor2}{rgb}{0.00000,0.12500,1.00000}%
\definecolor{mycolor3}{rgb}{0.00000,0.25000,1.00000}%
\definecolor{mycolor4}{rgb}{0.00000,0.37500,1.00000}%
\definecolor{mycolor5}{rgb}{0.00000,0.50000,1.00000}%
\definecolor{mycolor6}{rgb}{0.00000,0.62500,1.00000}%
\definecolor{mycolor7}{rgb}{0.00000,0.75000,1.00000}%
\definecolor{mycolor8}{rgb}{1.00000,0.87500,0.00000}%
\definecolor{mycolor9}{rgb}{1.00000,0.75000,0.00000}%
\begin{tikzpicture}

\begin{axis}[%
width=0.75\figurewidth,
height=\figureheight,
at={(0\figurewidth,0\figureheight)},
scale only axis,
xmin=2.0,
xmax=4.0,
xtick={2.0, 2.5, 3.0, 3.5, 4.0},
xlabel={IFC on simulated data},
ymin=2.0,
ymax=4.0,
ytick={2.0, 2.5, 3.0, 3.5, 4.0},
ylabel={IFC on real data},
axis background/.style={fill=white},
xmajorgrids,
ymajorgrids,
xlabel near ticks,ylabel near ticks,scaled y ticks=false,yticklabel style={/pgf/number format/fixed, /pgf/number format/precision=2},
]
\addplot [color=mycolor1,only marks,mark=*,mark options={solid,fill=mycolor1},forget plot]
  table[row sep=crcr]{%
2.76178585668044	2.19922682681529\\
};
\node[below left, align=right, text=black]
at (axis cs:2.762,2.199) {EBSR};
\addplot [color=blue,only marks,mark=*,mark options={solid,fill=blue},forget plot]
  table[row sep=crcr]{%
2.54451405853695	2.05965841379034\\
};
\node[above left, align=right, text=black]
at (axis cs:2.545,2.06) {ScSR};
\addplot [color=mycolor2,only marks,mark=*,mark options={solid,fill=mycolor2},forget plot]
  table[row sep=crcr]{%
3.37623253508522	2.26102573901571\\
};
\node[below right, align=right, text=black]
at (axis cs:3.376,2.261) {NBSRF};
\addplot [color=mycolor3,only marks,mark=*,mark options={solid,fill=mycolor3},forget plot]
  table[row sep=crcr]{%
3.30842694138642	2.2500310907896\\
};
\node[left, align=right, text=black]
at (axis cs:3.308,2.25) {A+};
\addplot [color=mycolor4,only marks,mark=*,mark options={solid,fill=mycolor4},forget plot]
  table[row sep=crcr]{%
3.05929814073833	2.20324794602667\\
};
\node[above left, align=right, text=black]
at (axis cs:3.059,2.203) {SRCNN};
\addplot [color=mycolor5,only marks,mark=*,mark options={solid,fill=mycolor5},forget plot]
  table[row sep=crcr]{%
3.41216674524437	2.33510907521803\\
};
\node[above left, align=right, text=black]
at (axis cs:3.412,2.335) {DRCN};
\addplot [color=mycolor6,only marks,mark=*,mark options={solid,fill=mycolor6},forget plot]
  table[row sep=crcr]{%
3.42578792932659	2.30261324178185\\
};
\node[right, align=right, text=black]
at (axis cs:3.426,2.303) {VDSR};
\addplot [color=mycolor7,only marks,mark=*,mark options={solid,fill=mycolor7},forget plot]
  table[row sep=crcr]{%
2.94172193744536	2.12521957637553\\
};
\node[below left, align=right, text=black]
at (axis cs:2.942,2.125) {SESR};
\addplot [color=mycolor8,only marks,mark=*,mark options={solid,fill=mycolor8},forget plot]
  table[row sep=crcr]{%
2.41453331025546	2.66551962731853\\
};
\node[below left, align=right, text=black]
at (axis cs:2.415,2.666) {NUISR};
\addplot [color=mycolor9,only marks,mark=*,mark options={solid,fill=mycolor9},forget plot]
  table[row sep=crcr]{%
2.50650079214384	2.76875535986401\\
};
\node[above right, align=right, text=black]
at (axis cs:2.507,2.769) {WNUISR};
\addplot [color=orange!50!mycolor9,only marks,mark=*,mark options={solid,fill=orange!50!mycolor9},forget plot]
  table[row sep=crcr]{%
2.56362964356328	2.73412807957454\\
};
\node[below right, align=right, text=black]
at (axis cs:2.564,2.734) {HYSR};
\addplot [color=orange,only marks,mark=*,mark options={solid,fill=orange},forget plot]
  table[row sep=crcr]{%
2.47643512656103	2.78051114024322\\
};
\node[above left, align=right, text=black]
at (axis cs:2.476,2.781) {DBRSR};
\addplot [color=red!25!orange,only marks,mark=*,mark options={solid,fill=red!25!orange},forget plot]
  table[row sep=crcr]{%
2.91522212505677	2.90569611769856\\
};
\node[below right, align=right, text=black]
at (axis cs:2.915,2.906) {L1BTV};
\addplot [color=red!50!orange,only marks,mark=*,mark options={solid,fill=red!50!orange},forget plot]
  table[row sep=crcr]{%
2.989217787924	3.16801311003749\\
};
\node[above left, align=right, text=black]
at (axis cs:2.989,3.168) {BEPSR};
\addplot [color=red!75!orange,only marks,mark=*,mark options={solid,fill=red!75!orange},forget plot]
  table[row sep=crcr]{%
2.95468418127529	3.05031142290774\\
};
\node[above left, align=right, text=black]
at (axis cs:2.955,3.05) {IRWSR};
\addplot [color=black!25!red,only marks,mark=*,mark options={solid,fill=black!25!red},forget plot]
  table[row sep=crcr]{%
3.11433345018079	2.45676479127335\\
};
\node[below left, align=right, text=black]
at (axis cs:3.114,2.457) {VSRnet};
\addplot [color=black,dashed,line width=1.5pt,forget plot]
  table[row sep=crcr]{%
2.0	2.0\\
4.0	4.0\\
};
\end{axis}
\end{tikzpicture}%
%
}
	\caption{Correlation between the benchmarks of SR algorithms on simulated data and our captured LR data in terms of different full-reference quality measures for $2\times2$ (top row) and $4\times4$ binning (bottom row). The individual algorithms are categorized either as SISR (shown with blue color map) or MFSR (shown with red color map). Algorithms located below the line of equal image quality perform worse on real data compared to simulated data. For each quality measure, the corresponding Spearman rank correlation $\rho$ is shown.}
	\label{fig:simulatedToRealDataCorrelation}
\end{figure*}
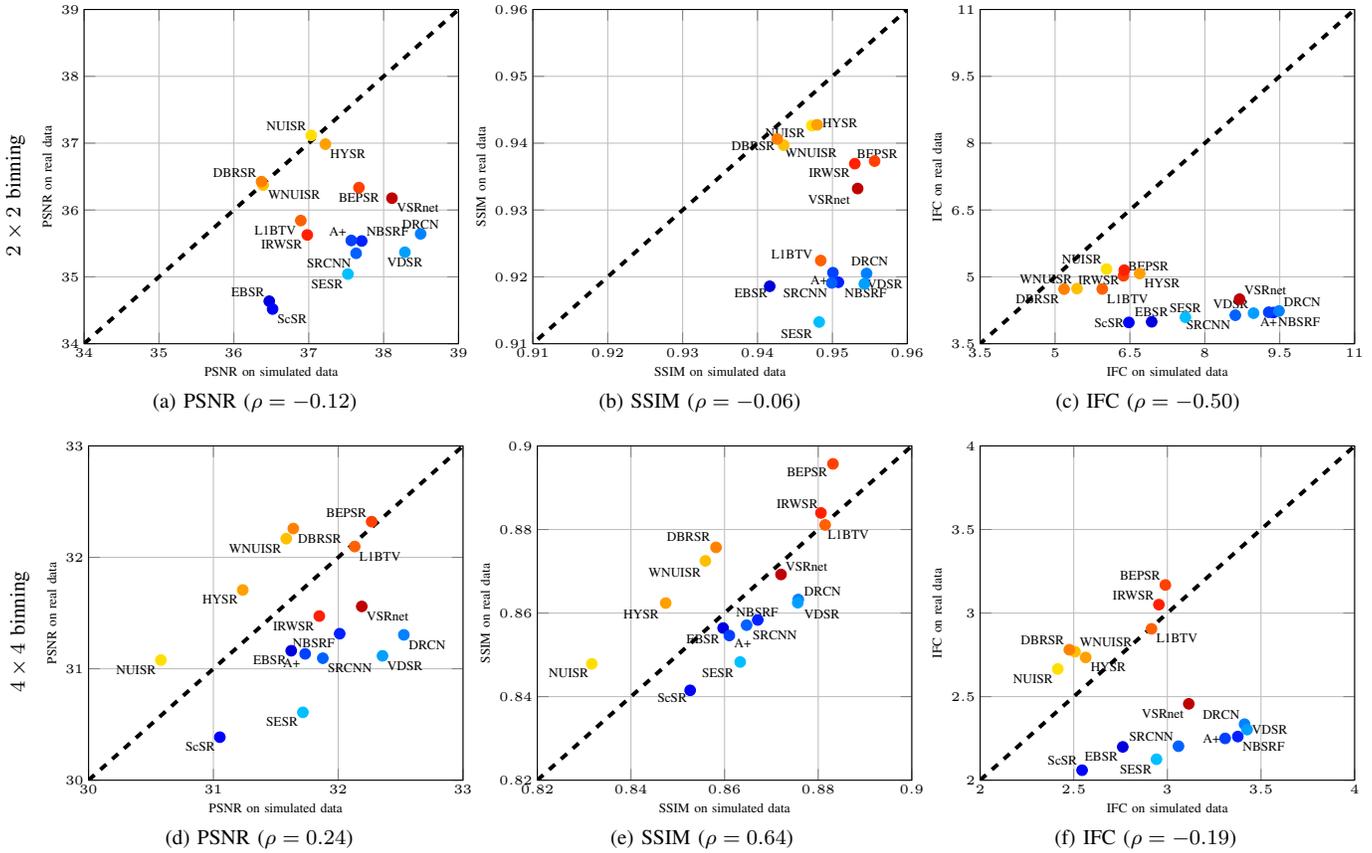

We validate the importance of real acquisitions by analyzing correlations between SR benchmarks on simulated and real data. Following prior work \cite{Agustsson2017}, we simulate LR images from our ground truth data using bicubic downsampling as a counterpart to LR images that were captured by hardware binning. \Fref{fig:simulatedToRealDataCorrelation} shows the performance of various algorithms of our benchmark (see \sref{sec:BenchmarkSetup}) for different binning factors on simulated versus real data. The performance is depicted \wrt different full-reference quality measures (PSNR, SSIM, and IFC) averaged over 14 scenes with global motion. We can observe that the performance of most algorithms is higher on simulated data but drops on real data as depicted in \fref{fig:teaser}. For instance, two recent deep networks (VDSR \cite{Kim2016}, DRCN \cite{Kim2016a}) outperform most classical algorithms on simulated data. Interestingly, these deep networks are outperformed by several other approaches on real data. Overall, there are low Spearman rank correlations of $\rho < 0.6$ between the quality measures on simulated and real data. There are even anti-correlations in case of small magnifications corresponding to small binning factors ($2\times2$). This demonstrates that benchmarks on simulated data are weak indicators for an algorithm ranking on real data, especially for small magnifications. We argue that this simulated-to-real gap appears because simulations follows a relatively simple model, while real data can be affected by more realistic artifacts like non-Gaussian noise or oversaturated pixels, among others. Such observations have also been reported in related areas like blind deblurring \cite{Lai2016} or denoising \cite{Plotz2017}, which underlines the importance of real data for thorough SR evaluations.

Some prior works also use the same models for simulations and SR. This can be seen as \textit{inverse crime} \cite{Wirgin2004} and limits the significance of experiments. For instance, many deep learning methods are trained and validated for LR/HR exemplars related to each other by bicubic downsampling. This explains their lower performance on real data that follows a more complicated image formation. It also adds another merit of using real LR data: beyond the aspect of evaluation, our datasets can also be divided into training, validation, and test subsets to complement the fine-tuning of learning-based methods.

\subsubsection{Single-Camera Setup for Registered Ground Truth}
\label{sec:SingleCameraVsMultiCameraSetup}

The used single-camera setup overcomes several limitations of related multi-camera setups as proposed in \cite{Qu2016}. First and foremost, it guarantees by design a perfect alignment between LR and ground truth data. This allows pixel-wise comparisons among super-resolved images and the ground truth by full-reference quality measures. In contrast to data captured with multi-camera setups, our ground truth is not the outcome of a potentially error-prone registration procedure. 

The proposed single-camera setup further extends the scope of our database in comparison to \cite{Qu2016}: one important benefit is the existence of multiple resolution levels. Additionally, we collect image sequences instead of single images. This makes the data usable for both, SISR and MFSR.

\subsection{Alternative Camera Setups}

In principle, there is a variety of possible setup designs to acquire LR and HR data. These approaches typically represent trade-offs of setup complexity against the level of realism. 

One extension toward in-the-wild data could be to collect LR/HR exemplars from far/near distances with the same hardware binning factor instead of using different factors at the same distance. Thus, an acquisition from a near distance can represent ground truth data and acquisitions from far distances serve as LR images. A related approach is to use optical zoom to gain near and far images instead of moving the camera. Similar to a multi-camera setup, one practical challenge of such setups is to achieve the required accuracy for a native pixel-wise registration of the images and the handling of potential occlusion or lens distortion effects.

Another possibility is to aim for a higher level of realism of simulations by carefully estimating and applying a mapping between HR and LR images. In context of the first setup, one could estimate a PSF to simulate far images from near ones. This considerably reduces the effort during acquisition, but again relies on the quality of the simulation. 

Our proposed setup based on hardware binning compromises between these two example variants, in providing real acquisitions but still leaves the acquisition complexity manageable.

\section{Benchmark Setup}
\label{sec:BenchmarkSetup}

\subsection{Evaluation Protocol}
\label{sec:EvaluationProtocol}

We perform SR on $K = 2M + 1$ consecutive LR frames $\vec{Y}^{(-M)}, \ldots, 
\vec{Y}^{(0)}, \ldots, \vec{Y}^{(M)}$ in a sliding window scheme. $\vec{Y}^{(0)}$ is referred to as the 
reference frame. For SISR, $\vec{Y}^{(0)}$ serves as input to determine the 
corresponding HR image $\vec{X}_{\mathrm{sr}}$. In case of MFSR, $\vec{Y}^{(-M
)}, \ldots, \vec{Y}^{(M)}$ is exploited to obtain $\vec{X}_{\mathrm{sr}}$ 
using variational optical flow \cite{Liu2009} to estimate subpixel motion 
towards $\vec{Y}^{(0)}$. For MFSR with customized motion compensation 
\cite{Liu2014,Ma2015,Kappeler2016}, we employ the optical flow estimation 
used in the original versions of the algorithms. 

We study the magnification factors 2, 3, and 4 
to super-resolve LR images at the respective binning factors to the 
resolution of the ground truth. The number of input frames for MFSR is chosen 
according to the desired magnification and therefore we use 5, 11, and 17 frames 
for the factors 2, 3, and 4, respectively. Our benchmark comprises ten sliding windows per dataset. We crop central regions of size $1200 \times 960$ pixels with sufficient details in the ground truth of each dataset to increase the significance of evaluating SR quality.

\subsection{Quantitative Quality Measures}

We use ten full-reference and no-reference measures $Q(\vec{X}_{\mathrm{sr}})$ to quantitatively assess image quality. The full-reference methods comprise the PSNR, SSIM \cite{Wang2004}, MS-SSIM \cite{Wang2003}, and IFC \cite{Sheikh2005}. These measures focus on different hand-crafted features, namely intensities (PSNR), structural information (SSIM, MS-SSIM), and natural scene statistics (IFC) to assess the fidelity of super-resolved data $\vec{X}_{\mathrm{sr}}$ \wrt the ground truth $\vec{X}_{\mathrm{gt}}$. Additionally we employ the recently proposed learned perceptual image patch similarity (LPIPS) metric \cite{Zhang2018c} using SqueezeNet with linear calibration (version 0.1). Different to hand-crafted features, LPIPS exploits similarities of deep features learned from perceptual judgments. We assume that $\vec{X}_{\mathrm{sr}}$ is aligned with $\vec{X}_{\mathrm{gt}}$. A three-pixel boundary is cropped to provide space for compensating alignment differences by some algorithms. 

We additionally use five popular no-reference measures, namely S3 \cite{Vu2012}, BRISQUE \cite{Mittal2012}, SSEQ \cite{Liu2014a}, NIQE \cite{Mittal2013}, and SRM \cite{Ma2017}, which is a recent measure trained from SISR examples. Higher S3 and SRM measures express higher perceptual quality of the assessed SR image. For BRISQUE, SSEQ and NIQE, we used the negated scores such that higher measures express higher quality.

Note that scene content can considerably influence the absolute values of 
these measures \cite{Yang2014a}. To reduce dependencies from scene content and 
to analyze the improvement of SR over the input data, we also evaluate 
normalized quality measures:
\begin{equation}
	\tilde{Q}(\vec{X}_{\mathrm{sr}}) = ( Q(\vec{X}_{\mathrm{sr}}) - Q(\tilde{\vec{Y}}) ) / Q(\tilde{\vec{Y}})\enspace,
\end{equation}
where $\tilde{\vec{Y}}$ denotes the nearest-neighbor interpolation of the
reference frame $\vec{Y}^{(0)}$ on the target HR grid.

\subsection{Perceptual Pairwise Comparisons}

A large-scale observer study is set up to assess image quality according to human visual perception. The study adopts \textit{forced-choice pairwise comparisons} \cite{Mantiuk2012}, where two images obtained by different 
SR algorithms from the same data are presented side-by-side. An observer is 
requested to choose from each pair the image with higher quality. The pairs are randomly sampled without replacement out of $n_{
\text{data}} \binom{n_{\text{sr}}}{2}$ pairs with $n_{\text{sr}}$ algorithms and $n_{\text{data}}$ datasets. Within a pair, both images are exposed to identical conditions, i.\,e. identical binning, motion, lighting, and compression. Thus the only differences in the presented images are due to the applied SR algorithms. Observers are guided interactively through sessions of $n_{\text{pairs}}$ pairs by a dynamic webpage. 
Among the $n_{\text{pairs}}$ pairs, $n_{\text{sanity}}$ pairs are 
randomly mixed in as sanity checks to identify careless observers. Sanity checks comprise ground truth images 
and results with severe artifacts (aliasing, noise, or motion artifacts). We 
discard a session if the sanity check is failed more than once.

We denote by $\vec{M} \in \mathbb{Z}^{n_{\text{sr}} \times n_{\text{sr}}}$ 
the winning matrix for a session, where 
$M_{ij}$, $i \neq j$ denotes the number of times that the $i$-th method is 
preferred over the $j$-th method. To globally rank algorithms from pairwise votes, we use $\vec{M}$ to fit a Bradley-Terry (B-T) model \cite{Lai2016}. This describes the probability 
$P(i \succ j)$ that the $i$-th method is ranked over the $j$-th method:
\begin{equation}
	\label{eqn:btModelProb}
	P(i \succ j) = \frac{e^{\delta_i}}{e^{\delta_i} + e^{\delta_j}}\enspace,
\end{equation}
where $\delta_i$ and $\delta_j$ are quality scores associated with the $i$-th 
and the $j$-th method, respectively. Then, based on \eqref{eqn:btModelProb}, 
the negative log-likelihood for the B-T scores 
$\vec{\delta} \in \mathbb{R}^{n_{\text{sr}}}$ is given by:
\begin{equation}
	\label{eqn:btModelLogLikelihood}
	\mathcal{L}(\vec{\delta}) 
	= - \log \left( \prod_{i = 1}^{n_{\text{sr}}} \prod_{j = 1, j \neq i}^{n_{\text{sr}}} P(i \succ j)^{M_{ij}} \right)\enspace.
\end{equation}
The B-T scores $\vec{\delta}$ are obtained by 
minimizing \eqref{eqn:btModelLogLikelihood} using expectation-maximization 
\cite{Lai2016}.

To analyze the agreement among the votes from $n_{\text{observer}}$ 
sessions, we employ the Kendall coefficient of agreement \cite{Kendall1940}:
\begin{equation}
	\label{eqn:kendallCoeff}
	u 
	= \frac{2W}{\binom{n_{\text{sr}}}{2} \binom{n_{\text{observer}}}{2}} - 1,
	\quad
	W = \sum_{i = 1}^{n_{\text{sr}}} \sum_{j = 1, j \neq i}^{n_{\text{sr}}} \binom{M_{ij}}{2}
	\enspace.
\end{equation}
This describes inter-observer variances and perfect agreement leads to $u = 1$, while uniformly random votes lead to $u = -\frac{1}{n_{\text{sr}}}$.

\subsection{Evaluated Algorithms}

We investigate 18 classical and state-of-the-art SR
methods and bicubic interpolation as categorized in \tref{tab:algorithmOverview}.

For MFSR, we study interpolation, reconstruction, and deep 
learning methods. Interpolation MFSR comprises non-uniform 
interpolation (NUISR) \cite{Park2003}, NUISR with outlier weighting 
(WNUISR) \cite{Batz2016}, and denoising-based refinement (DBRSR) \cite{Batz2016b}. 
The reconstruction methods are non-blind $L_1$ norm minimization 
with bilateral total variation prior (L1BTV) \cite{Farsiu2004a}, bilateral edge preserving prior (BEPSR), and iteratively re-weighted 
minimization (IRWSR) \cite{Kohler2015c}. We also evaluate blind 
Bayesian video SR (BVSR) \cite{Liu2014} and SR with motion blur handling (SRB)
 \cite{Ma2015}. We further use the video SR network (VSRnet) of \cite{Kappeler2016} and the hybrid approach (HYSR) of \cite{Batz2015}.

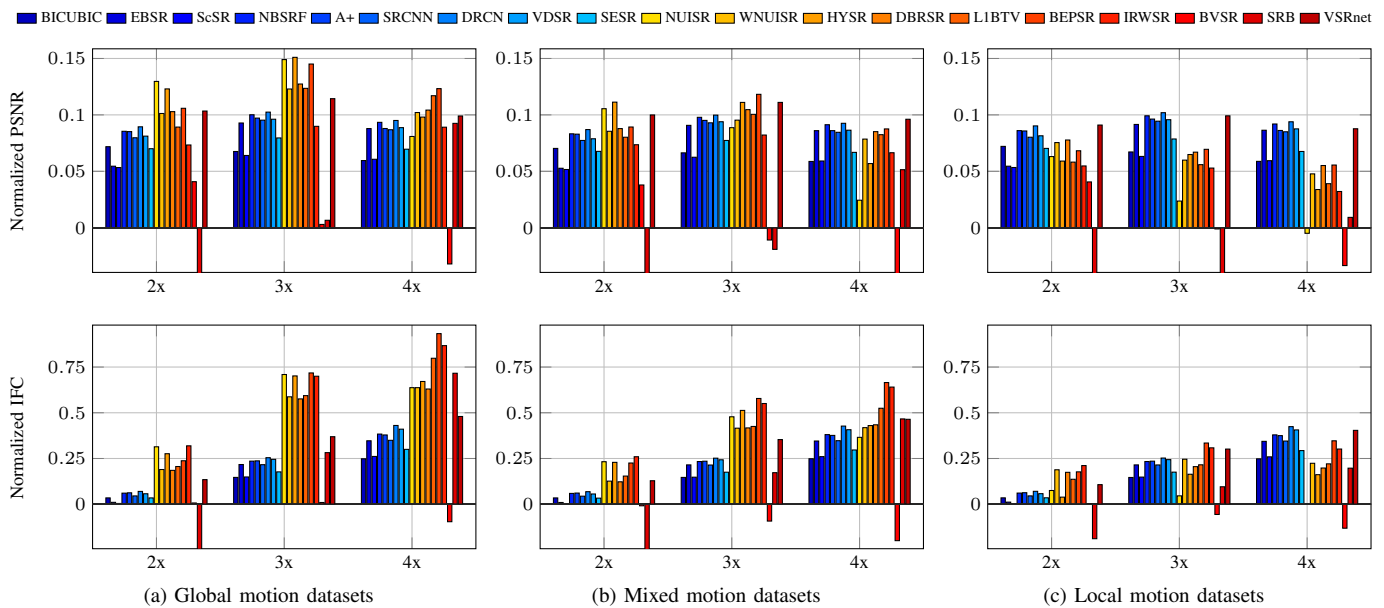
\begin{figure*}[!b]
	\centering
	\subfloat{%
%
\definecolor{mycolor1}{rgb}{0.00000,0.00000,0.87500}%
\definecolor{mycolor2}{rgb}{0.00000,0.12500,1.00000}%
\definecolor{mycolor3}{rgb}{0.00000,0.25000,1.00000}%
\definecolor{mycolor4}{rgb}{0.00000,0.37500,1.00000}%
\definecolor{mycolor5}{rgb}{0.00000,0.50000,1.00000}%
\definecolor{mycolor6}{rgb}{0.00000,0.62500,1.00000}%
\definecolor{mycolor7}{rgb}{0.00000,0.75000,1.00000}%
\definecolor{mycolor8}{rgb}{1.00000,0.87500,0.00000}%
\definecolor{mycolor9}{rgb}{1.00000,0.75000,0.00000}%
\definecolor{mycolor10}{rgb}{0.87500,0.00000,0.00000}%
\begin{tikzpicture}
\scriptsize
\begin{axis}[%
    const plot,
		hide axis,
    xmin=10,
    xmax=50,
    ymin=0,
    ymax=0,
    legend style={
			legend image code/.code={2mm
				\draw [#1] (0cm,-0.065cm) rectangle (0.25cm,0.065cm);
			},
			anchor=north,
			inner ysep=0pt,
			draw=none, 
			legend cell align=center,
			legend columns=-1, 
			nodes={scale=0.815, transform shape}
		}
	]
	\addlegendimage{black, mark=none, fill=black!25!blue}
	\addlegendentry{BICUBIC}
	\addlegendimage{black, mark=none, fill=mycolor1}
	\addlegendentry{EBSR}
	\addlegendimage{black, mark=none, fill=blue}
	\addlegendentry{ScSR}
	\addlegendimage{black, mark=none, fill=mycolor2}
	\addlegendentry{NBSRF}
	\addlegendimage{black, mark=none, fill=mycolor3}
	\addlegendentry{A+}
	\addlegendimage{black, mark=none, fill=mycolor4}
	\addlegendentry{SRCNN}
	\addlegendimage{black, mark=none, fill=mycolor5}
	\addlegendentry{DRCN}
	\addlegendimage{black, mark=none, fill=mycolor6}
	\addlegendentry{VDSR}
	\addlegendimage{black, mark=none, fill=mycolor7}
	\addlegendentry{SESR}
	\addlegendimage{black, mark=none, fill=mycolor8}
	\addlegendentry{NUISR}
	\addlegendimage{black, mark=none, fill=mycolor9}
	\addlegendentry{WNUISR}
	\addlegendimage{black, mark=none, fill=orange!50!mycolor9}
	\addlegendentry{HYSR}
	\addlegendimage{black, mark=none, fill=orange}
	\addlegendentry{DBRSR}
	\addlegendimage{black, mark=none, fill=red!25!orange}
	\addlegendentry{L1BTV}
	\addlegendimage{black, mark=none, fill=red!50!orange}
	\addlegendentry{BEPSR}
	\addlegendimage{black, mark=none, fill=red!75!orange}
	\addlegendentry{IRWSR}
	\addlegendimage{black, mark=none, fill=red}
	\addlegendentry{BVSR}
	\addlegendimage{black, mark=none, fill=mycolor10}
	\addlegendentry{SRB}
	\addlegendimage{black, mark=none, fill=black!25!red}
	\addlegendentry{VSRnet}
\end{axis}

\end{tikzpicture}%
%
}\\[-0.75ex]
	\scriptsize
	\setlength \figurewidth{0.281\textwidth}
	\setlength \figureheight{1.40\figurewidth}
	\raisebox{0.9cm}{\rotatebox{90}{\scriptsize Normalized PSNR}}\quad
	\subfloat{%
%
\definecolor{mycolor1}{rgb}{0.00000,0.00000,0.87500}%
\definecolor{mycolor2}{rgb}{0.00000,0.12500,1.00000}%
\definecolor{mycolor3}{rgb}{0.00000,0.25000,1.00000}%
\definecolor{mycolor4}{rgb}{0.00000,0.37500,1.00000}%
\definecolor{mycolor5}{rgb}{0.00000,0.50000,1.00000}%
\definecolor{mycolor6}{rgb}{0.00000,0.62500,1.00000}%
\definecolor{mycolor7}{rgb}{0.00000,0.75000,1.00000}%
\definecolor{mycolor8}{rgb}{1.00000,0.87500,0.00000}%
\definecolor{mycolor9}{rgb}{1.00000,0.75000,0.00000}%
\definecolor{mycolor10}{rgb}{0.87500,0.00000,0.00000}%
\begin{tikzpicture}

\begin{axis}[%
width=\figurewidth,
height=0.417\figureheight,
at={(0\figurewidth,0\figureheight)},
scale only axis,
xmin=0.5,
xmax=3.5,
xtick={1,2,3,4,5,6,7,8,9,10,11,12,13,14,15,16,17,18,19,20,21,22,23,24,25,26,27,28,29,30,31,32,33,34,35,36,37,38,39,40,41,42,43,44,45,46,47,48,49,50,51,52,53,54,55,56,57},
xticklabels={{2x},{3x},{4x}},
ymin=-0.0396430705941855,
ymax=0.158572282376742,
axis background/.style={fill=white},
xmajorgrids,
ymajorgrids,
xlabel near ticks,ylabel near ticks,scaled y ticks=false,yticklabel style={/pgf/number format/fixed, /pgf/number format/precision=2},
]
\addplot[ybar,bar width=0.034,bar shift=-0.379,fill=black!25!blue,draw=black,area legend] plot table[row sep=crcr] {%
1	0.0718067656445448\\
2	0.0675544756732017\\
3	0.0595125917961722\\
};
\addplot[forget plot,color=white!15!black] table[row sep=crcr] {%
0.5	0\\
3.5	0\\
};
\addplot[ybar,bar width=0.034,bar shift=-0.337,fill=mycolor1,draw=black,area legend] plot table[row sep=crcr] {%
1	0.0545138345717285\\
2	0.0928193546847919\\
3	0.0878613646866779\\
};
\addplot[forget plot,color=white!15!black] table[row sep=crcr] {%
0.5	0\\
3.5	0\\
};
\addplot[ybar,bar width=0.034,bar shift=-0.295,fill=blue,draw=black,area legend] plot table[row sep=crcr] {%
1	0.0533347247923587\\
2	0.0640388264611097\\
3	0.0606605988302488\\
};
\addplot[forget plot,color=white!15!black] table[row sep=crcr] {%
0.5	0\\
3.5	0\\
};
\addplot[ybar,bar width=0.034,bar shift=-0.253,fill=mycolor2,draw=black,area legend] plot table[row sep=crcr] {%
1	0.0854583672569841\\
2	0.100115620361536\\
3	0.0933962846840464\\
};
\addplot[forget plot,color=white!15!black] table[row sep=crcr] {%
0.5	0\\
3.5	0\\
};
\addplot[ybar,bar width=0.034,bar shift=-0.211,fill=mycolor3,draw=black,area legend] plot table[row sep=crcr] {%
1	0.0851695490073981\\
2	0.0971645233467304\\
3	0.0879364965964269\\
};
\addplot[forget plot,color=white!15!black] table[row sep=crcr] {%
0.5	0\\
3.5	0\\
};
\addplot[ybar,bar width=0.034,bar shift=-0.168,fill=mycolor4,draw=black,area legend] plot table[row sep=crcr] {%
1	0.0797202832573437\\
2	0.0953318617021718\\
3	0.0868453104468814\\
};
\addplot[forget plot,color=white!15!black] table[row sep=crcr] {%
0.5	0\\
3.5	0\\
};
\addplot[ybar,bar width=0.034,bar shift=-0.126,fill=mycolor5,draw=black,area legend] plot table[row sep=crcr] {%
1	0.0894307631166125\\
2	0.102459503177351\\
3	0.0950519006639159\\
};
\addplot[forget plot,color=white!15!black] table[row sep=crcr] {%
0.5	0\\
3.5	0\\
};
\addplot[ybar,bar width=0.034,bar shift=-0.084,fill=mycolor6,draw=black,area legend] plot table[row sep=crcr] {%
1	0.0811678116456141\\
2	0.0961560748360261\\
3	0.0886899540055777\\
};
\addplot[forget plot,color=white!15!black] table[row sep=crcr] {%
0.5	0\\
3.5	0\\
};
\addplot[ybar,bar width=0.034,bar shift=-0.042,fill=mycolor7,draw=black,area legend] plot table[row sep=crcr] {%
1	0.0700219514030107\\
2	0.079572394765626\\
3	0.0695800175530972\\
};
\addplot[forget plot,color=white!15!black] table[row sep=crcr] {%
0.5	0\\
3.5	0\\
};
\addplot[ybar,bar width=0.034,bar shift=-0,fill=mycolor8,draw=black,area legend] plot table[row sep=crcr] {%
1	0.129674449565356\\
2	0.149023160450096\\
3	0.0808653595631601\\
};
\addplot[forget plot,color=white!15!black] table[row sep=crcr] {%
0.5	0\\
3.5	0\\
};
\addplot[ybar,bar width=0.034,bar shift=0.042,fill=mycolor9,draw=black,area legend] plot table[row sep=crcr] {%
1	0.101246830345493\\
2	0.122950565994942\\
3	0.102097444269872\\
};
\addplot[forget plot,color=white!15!black] table[row sep=crcr] {%
0.5	0\\
3.5	0\\
};
\addplot[ybar,bar width=0.034,bar shift=0.084,fill=orange!50!mycolor9,draw=black,area legend] plot table[row sep=crcr] {%
1	0.123029076323291\\
2	0.151021221311183\\
3	0.0980534813232744\\
};
\addplot[forget plot,color=white!15!black] table[row sep=crcr] {%
0.5	0\\
3.5	0\\
};
\addplot[ybar,bar width=0.034,bar shift=0.126,fill=orange,draw=black,area legend] plot table[row sep=crcr] {%
1	0.102878251942983\\
2	0.127356168504388\\
3	0.104255720584305\\
};
\addplot[forget plot,color=white!15!black] table[row sep=crcr] {%
0.5	0\\
3.5	0\\
};
\addplot[ybar,bar width=0.034,bar shift=0.168,fill=red!25!orange,draw=black,area legend] plot table[row sep=crcr] {%
1	0.0892033885059617\\
2	0.123527186155759\\
3	0.11699732832305\\
};
\addplot[forget plot,color=white!15!black] table[row sep=crcr] {%
0.5	0\\
3.5	0\\
};
\addplot[ybar,bar width=0.034,bar shift=0.211,fill=red!50!orange,draw=black,area legend] plot table[row sep=crcr] {%
1	0.105843940217868\\
2	0.145084527998751\\
3	0.12327165305627\\
};
\addplot[forget plot,color=white!15!black] table[row sep=crcr] {%
0.5	0\\
3.5	0\\
};
\addplot[ybar,bar width=0.034,bar shift=0.253,fill=red!75!orange,draw=black,area legend] plot table[row sep=crcr] {%
1	0.073339310930307\\
2	0.0898571284135134\\
3	0.0890929807150915\\
};
\addplot[forget plot,color=white!15!black] table[row sep=crcr] {%
0.5	0\\
3.5	0\\
};
\addplot[ybar,bar width=0.034,bar shift=0.295,fill=red,draw=black,area legend] plot table[row sep=crcr] {%
1	0.0407886260724873\\
2	0.00290451673543416\\
3	-0.0321387107750672\\
};
\addplot[forget plot,color=white!15!black] table[row sep=crcr] {%
0.5	0\\
3.5	0\\
};
\addplot[ybar,bar width=0.034,bar shift=0.337,fill=mycolor10,draw=black,area legend] plot table[row sep=crcr] {%
1	-0.173414388822268\\
2	0.0065835543755435\\
3	0.0924953042922807\\
};
\addplot[forget plot,color=white!15!black] table[row sep=crcr] {%
0.5	0\\
3.5	0\\
};
\addplot[ybar,bar width=0.034,bar shift=0.379,fill=black!25!red,draw=black,area legend] plot table[row sep=crcr] {%
1	0.103418009902451\\
2	0.114377144959567\\
3	0.0990341008434151\\
};
\addplot[forget plot,color=white!15!black] table[row sep=crcr] {%
0.5	0\\
3.5	0\\
};
\end{axis}
\end{tikzpicture}%
%
}~
	\subfloat{%
%
\definecolor{mycolor1}{rgb}{0.00000,0.00000,0.87500}%
\definecolor{mycolor2}{rgb}{0.00000,0.12500,1.00000}%
\definecolor{mycolor3}{rgb}{0.00000,0.25000,1.00000}%
\definecolor{mycolor4}{rgb}{0.00000,0.37500,1.00000}%
\definecolor{mycolor5}{rgb}{0.00000,0.50000,1.00000}%
\definecolor{mycolor6}{rgb}{0.00000,0.62500,1.00000}%
\definecolor{mycolor7}{rgb}{0.00000,0.75000,1.00000}%
\definecolor{mycolor8}{rgb}{1.00000,0.87500,0.00000}%
\definecolor{mycolor9}{rgb}{1.00000,0.75000,0.00000}%
\definecolor{mycolor10}{rgb}{0.87500,0.00000,0.00000}%
\begin{tikzpicture}

\begin{axis}[%
width=\figurewidth,
height=0.417\figureheight,
at={(0\figurewidth,0\figureheight)},
scale only axis,
xmin=0.5,
xmax=3.5,
xtick={1,2,3,4,5,6,7,8,9,10,11,12,13,14,15,16,17,18,19,20,21,22,23,24,25,26,27,28,29,30,31,32,33,34,35,36,37,38,39,40,41,42,43,44,45,46,47,48,49,50,51,52,53,54,55,56,57},
xticklabels={{2x},{3x},{4x}},
ymin=-0.0396430705941855,
ymax=0.158572282376742,
axis background/.style={fill=white},
xmajorgrids,
ymajorgrids,
xlabel near ticks,ylabel near ticks,scaled y ticks=false,yticklabel style={/pgf/number format/fixed, /pgf/number format/precision=2},
]
\addplot[ybar,bar width=0.034,bar shift=-0.379,fill=black!25!blue,draw=black,area legend] plot table[row sep=crcr] {%
1	0.070284365369231\\
2	0.0663497858994235\\
3	0.0587409795890587\\
};
\addplot[forget plot,color=white!15!black] table[row sep=crcr] {%
0.5	0\\
3.5	0\\
};
\addplot[ybar,bar width=0.034,bar shift=-0.337,fill=mycolor1,draw=black,area legend] plot table[row sep=crcr] {%
1	0.0526956407551614\\
2	0.0907269672673494\\
3	0.0860202022769042\\
};
\addplot[forget plot,color=white!15!black] table[row sep=crcr] {%
0.5	0\\
3.5	0\\
};
\addplot[ybar,bar width=0.034,bar shift=-0.295,fill=blue,draw=black,area legend] plot table[row sep=crcr] {%
1	0.0515939514991259\\
2	0.0624659741874914\\
3	0.059134403739555\\
};
\addplot[forget plot,color=white!15!black] table[row sep=crcr] {%
0.5	0\\
3.5	0\\
};
\addplot[ybar,bar width=0.034,bar shift=-0.253,fill=mycolor2,draw=black,area legend] plot table[row sep=crcr] {%
1	0.083221616380681\\
2	0.0978452595765837\\
3	0.0912838297481567\\
};
\addplot[forget plot,color=white!15!black] table[row sep=crcr] {%
0.5	0\\
3.5	0\\
};
\addplot[ybar,bar width=0.034,bar shift=-0.211,fill=mycolor3,draw=black,area legend] plot table[row sep=crcr] {%
1	0.0829269431619043\\
2	0.0951208612553684\\
3	0.0860342433543562\\
};
\addplot[forget plot,color=white!15!black] table[row sep=crcr] {%
0.5	0\\
3.5	0\\
};
\addplot[ybar,bar width=0.034,bar shift=-0.168,fill=mycolor4,draw=black,area legend] plot table[row sep=crcr] {%
1	0.0773522698465775\\
2	0.0929664971099697\\
3	0.0845666345320816\\
};
\addplot[forget plot,color=white!15!black] table[row sep=crcr] {%
0.5	0\\
3.5	0\\
};
\addplot[ybar,bar width=0.034,bar shift=-0.126,fill=mycolor5,draw=black,area legend] plot table[row sep=crcr] {%
1	0.0869828311425487\\
2	0.0997929657440815\\
3	0.0925405426152866\\
};
\addplot[forget plot,color=white!15!black] table[row sep=crcr] {%
0.5	0\\
3.5	0\\
};
\addplot[ybar,bar width=0.034,bar shift=-0.084,fill=mycolor6,draw=black,area legend] plot table[row sep=crcr] {%
1	0.078829866769553\\
2	0.0939299762466022\\
3	0.0864268342591471\\
};
\addplot[forget plot,color=white!15!black] table[row sep=crcr] {%
0.5	0\\
3.5	0\\
};
\addplot[ybar,bar width=0.034,bar shift=-0.042,fill=mycolor7,draw=black,area legend] plot table[row sep=crcr] {%
1	0.0676706341699478\\
2	0.0773412877295669\\
3	0.0667835996521201\\
};
\addplot[forget plot,color=white!15!black] table[row sep=crcr] {%
0.5	0\\
3.5	0\\
};
\addplot[ybar,bar width=0.034,bar shift=-0,fill=mycolor8,draw=black,area legend] plot table[row sep=crcr] {%
1	0.105482275670991\\
2	0.0886578676237481\\
3	0.024367066695719\\
};
\addplot[forget plot,color=white!15!black] table[row sep=crcr] {%
0.5	0\\
3.5	0\\
};
\addplot[ybar,bar width=0.034,bar shift=0.042,fill=mycolor9,draw=black,area legend] plot table[row sep=crcr] {%
1	0.0854714887214545\\
2	0.0953417764366599\\
3	0.0785479402646431\\
};
\addplot[forget plot,color=white!15!black] table[row sep=crcr] {%
0.5	0\\
3.5	0\\
};
\addplot[ybar,bar width=0.034,bar shift=0.084,fill=orange!50!mycolor9,draw=black,area legend] plot table[row sep=crcr] {%
1	0.111300919025584\\
2	0.111058897819255\\
3	0.0568432986677339\\
};
\addplot[forget plot,color=white!15!black] table[row sep=crcr] {%
0.5	0\\
3.5	0\\
};
\addplot[ybar,bar width=0.034,bar shift=0.126,fill=orange,draw=black,area legend] plot table[row sep=crcr] {%
1	0.0878532980472672\\
2	0.104655249116435\\
3	0.0851377581042994\\
};
\addplot[forget plot,color=white!15!black] table[row sep=crcr] {%
0.5	0\\
3.5	0\\
};
\addplot[ybar,bar width=0.034,bar shift=0.168,fill=red!25!orange,draw=black,area legend] plot table[row sep=crcr] {%
1	0.0801906728039708\\
2	0.100492170002975\\
3	0.0824593861394637\\
};
\addplot[forget plot,color=white!15!black] table[row sep=crcr] {%
0.5	0\\
3.5	0\\
};
\addplot[ybar,bar width=0.034,bar shift=0.211,fill=red!50!orange,draw=black,area legend] plot table[row sep=crcr] {%
1	0.0892681069362418\\
2	0.118228756467012\\
3	0.0875613815782573\\
};
\addplot[forget plot,color=white!15!black] table[row sep=crcr] {%
0.5	0\\
3.5	0\\
};
\addplot[ybar,bar width=0.034,bar shift=0.253,fill=red!75!orange,draw=black,area legend] plot table[row sep=crcr] {%
1	0.0735496161008224\\
2	0.0821482164247768\\
3	0.0664573301631479\\
};
\addplot[forget plot,color=white!15!black] table[row sep=crcr] {%
0.5	0\\
3.5	0\\
};
\addplot[ybar,bar width=0.034,bar shift=0.295,fill=red,draw=black,area legend] plot table[row sep=crcr] {%
1	0.0378771730385575\\
2	-0.0109237569323275\\
3	-0.0419407337833587\\
};
\addplot[forget plot,color=white!15!black] table[row sep=crcr] {%
0.5	0\\
3.5	0\\
};
\addplot[ybar,bar width=0.034,bar shift=0.337,fill=mycolor10,draw=black,area legend] plot table[row sep=crcr] {%
1	-0.17685789674282\\
2	-0.0192020983807493\\
3	0.051427606124294\\
};
\addplot[forget plot,color=white!15!black] table[row sep=crcr] {%
0.5	0\\
3.5	0\\
};
\addplot[ybar,bar width=0.034,bar shift=0.379,fill=black!25!red,draw=black,area legend] plot table[row sep=crcr] {%
1	0.0999406381353746\\
2	0.111076930604056\\
3	0.0961358271327967\\
};
\addplot[forget plot,color=white!15!black] table[row sep=crcr] {%
0.5	0\\
3.5	0\\
};
\end{axis}
\end{tikzpicture}%
%
}~
	\subfloat{%
%
\definecolor{mycolor1}{rgb}{0.00000,0.00000,0.87500}%
\definecolor{mycolor2}{rgb}{0.00000,0.12500,1.00000}%
\definecolor{mycolor3}{rgb}{0.00000,0.25000,1.00000}%
\definecolor{mycolor4}{rgb}{0.00000,0.37500,1.00000}%
\definecolor{mycolor5}{rgb}{0.00000,0.50000,1.00000}%
\definecolor{mycolor6}{rgb}{0.00000,0.62500,1.00000}%
\definecolor{mycolor7}{rgb}{0.00000,0.75000,1.00000}%
\definecolor{mycolor8}{rgb}{1.00000,0.87500,0.00000}%
\definecolor{mycolor9}{rgb}{1.00000,0.75000,0.00000}%
\definecolor{mycolor10}{rgb}{0.87500,0.00000,0.00000}%
\begin{tikzpicture}

\begin{axis}[%
width=\figurewidth,
height=0.417\figureheight,
at={(0\figurewidth,0\figureheight)},
scale only axis,
xmin=0.5,
xmax=3.5,
xtick={1,2,3,4,5,6,7,8,9,10,11,12,13,14,15,16,17,18,19,20,21,22,23,24,25,26,27,28,29,30,31,32,33,34,35,36,37,38,39,40,41,42,43,44,45,46,47,48,49,50,51,52,53,54,55,56,57},
xticklabels={{2x},{3x},{4x}},
ymin=-0.0396430705941855,
ymax=0.158572282376742,
axis background/.style={fill=white},
xmajorgrids,
ymajorgrids,
xlabel near ticks,ylabel near ticks,scaled y ticks=false,yticklabel style={/pgf/number format/fixed, /pgf/number format/precision=2},
]
\addplot[ybar,bar width=0.034,bar shift=-0.379,fill=black!25!blue,draw=black,area legend] plot table[row sep=crcr] {%
1	0.0720882268790726\\
2	0.0671409423467453\\
3	0.0588114988379233\\
};
\addplot[forget plot,color=white!15!black] table[row sep=crcr] {%
0.5	0\\
3.5	0\\
};
\addplot[ybar,bar width=0.034,bar shift=-0.337,fill=mycolor1,draw=black,area legend] plot table[row sep=crcr] {%
1	0.0545239743562981\\
2	0.0915365857619137\\
3	0.086402367313312\\
};
\addplot[forget plot,color=white!15!black] table[row sep=crcr] {%
0.5	0\\
3.5	0\\
};
\addplot[ybar,bar width=0.034,bar shift=-0.295,fill=blue,draw=black,area legend] plot table[row sep=crcr] {%
1	0.0533752953544629\\
2	0.0631918867223099\\
3	0.0594019243235682\\
};
\addplot[forget plot,color=white!15!black] table[row sep=crcr] {%
0.5	0\\
3.5	0\\
};
\addplot[ybar,bar width=0.034,bar shift=-0.253,fill=mycolor2,draw=black,area legend] plot table[row sep=crcr] {%
1	0.0860511183572481\\
2	0.0992017642179022\\
3	0.091932946053154\\
};
\addplot[forget plot,color=white!15!black] table[row sep=crcr] {%
0.5	0\\
3.5	0\\
};
\addplot[ybar,bar width=0.034,bar shift=-0.211,fill=mycolor3,draw=black,area legend] plot table[row sep=crcr] {%
1	0.0857487580495451\\
2	0.09628878046589\\
3	0.0861967155427517\\
};
\addplot[forget plot,color=white!15!black] table[row sep=crcr] {%
0.5	0\\
3.5	0\\
};
\addplot[ybar,bar width=0.034,bar shift=-0.168,fill=mycolor4,draw=black,area legend] plot table[row sep=crcr] {%
1	0.0802086505958559\\
2	0.0944270724172775\\
3	0.0850266906982501\\
};
\addplot[forget plot,color=white!15!black] table[row sep=crcr] {%
0.5	0\\
3.5	0\\
};
\addplot[ybar,bar width=0.034,bar shift=-0.126,fill=mycolor5,draw=black,area legend] plot table[row sep=crcr] {%
1	0.0902373859233698\\
2	0.101943719506296\\
3	0.0938925850069973\\
};
\addplot[forget plot,color=white!15!black] table[row sep=crcr] {%
0.5	0\\
3.5	0\\
};
\addplot[ybar,bar width=0.034,bar shift=-0.084,fill=mycolor6,draw=black,area legend] plot table[row sep=crcr] {%
1	0.0814759946644935\\
2	0.0957281795884601\\
3	0.0876025351385149\\
};
\addplot[forget plot,color=white!15!black] table[row sep=crcr] {%
0.5	0\\
3.5	0\\
};
\addplot[ybar,bar width=0.034,bar shift=-0.042,fill=mycolor7,draw=black,area legend] plot table[row sep=crcr] {%
1	0.0702930292287381\\
2	0.0786457421296113\\
3	0.0675746540501674\\
};
\addplot[forget plot,color=white!15!black] table[row sep=crcr] {%
0.5	0\\
3.5	0\\
};
\addplot[ybar,bar width=0.034,bar shift=-0,fill=mycolor8,draw=black,area legend] plot table[row sep=crcr] {%
1	0.0630425842256763\\
2	0.0236178155406836\\
3	-0.00497199448508565\\
};
\addplot[forget plot,color=white!15!black] table[row sep=crcr] {%
0.5	0\\
3.5	0\\
};
\addplot[ybar,bar width=0.034,bar shift=0.042,fill=mycolor9,draw=black,area legend] plot table[row sep=crcr] {%
1	0.0754456507375864\\
2	0.0599052958913066\\
3	0.047709019872419\\
};
\addplot[forget plot,color=white!15!black] table[row sep=crcr] {%
0.5	0\\
3.5	0\\
};
\addplot[ybar,bar width=0.034,bar shift=0.084,fill=orange!50!mycolor9,draw=black,area legend] plot table[row sep=crcr] {%
1	0.059057898267216\\
2	0.0648697484734611\\
3	0.0338455031169931\\
};
\addplot[forget plot,color=white!15!black] table[row sep=crcr] {%
0.5	0\\
3.5	0\\
};
\addplot[ybar,bar width=0.034,bar shift=0.126,fill=orange,draw=black,area legend] plot table[row sep=crcr] {%
1	0.0777382925779194\\
2	0.0669563896660624\\
3	0.0551152833062923\\
};
\addplot[forget plot,color=white!15!black] table[row sep=crcr] {%
0.5	0\\
3.5	0\\
};
\addplot[ybar,bar width=0.034,bar shift=0.168,fill=red!25!orange,draw=black,area legend] plot table[row sep=crcr] {%
1	0.0581448635166289\\
2	0.0558856780929937\\
3	0.0390199751027088\\
};
\addplot[forget plot,color=white!15!black] table[row sep=crcr] {%
0.5	0\\
3.5	0\\
};
\addplot[ybar,bar width=0.034,bar shift=0.211,fill=red!50!orange,draw=black,area legend] plot table[row sep=crcr] {%
1	0.0682050973823168\\
2	0.0694827523932628\\
3	0.0555544484660593\\
};
\addplot[forget plot,color=white!15!black] table[row sep=crcr] {%
0.5	0\\
3.5	0\\
};
\addplot[ybar,bar width=0.034,bar shift=0.253,fill=red!75!orange,draw=black,area legend] plot table[row sep=crcr] {%
1	0.0547442844295805\\
2	0.0528650464063534\\
3	0.0321009935868159\\
};
\addplot[forget plot,color=white!15!black] table[row sep=crcr] {%
0.5	0\\
3.5	0\\
};
\addplot[ybar,bar width=0.034,bar shift=0.295,fill=red,draw=black,area legend] plot table[row sep=crcr] {%
1	0.0405774783033952\\
2	-0.00111568595422315\\
3	-0.0335737619395627\\
};
\addplot[forget plot,color=white!15!black] table[row sep=crcr] {%
0.5	0\\
3.5	0\\
};
\addplot[ybar,bar width=0.034,bar shift=0.337,fill=mycolor10,draw=black,area legend] plot table[row sep=crcr] {%
1	-0.176779964325623\\
2	-0.0411598232024296\\
3	0.00915922824133331\\
};
\addplot[forget plot,color=white!15!black] table[row sep=crcr] {%
0.5	0\\
3.5	0\\
};
\addplot[ybar,bar width=0.034,bar shift=0.379,fill=black!25!red,draw=black,area legend] plot table[row sep=crcr] {%
1	0.0910015577113705\\
2	0.0991650117721198\\
3	0.0877070482001649\\
};
\addplot[forget plot,color=white!15!black] table[row sep=crcr] {%
0.5	0\\
3.5	0\\
};
\end{axis}
\end{tikzpicture}%
%
}\\
	\setcounter{subfigure}{0}
	\raisebox{1.0cm}{\rotatebox{90}{\scriptsize Normalized IFC}}\quad
	\subfloat[Global motion datasets]{%
%
\definecolor{mycolor1}{rgb}{0.00000,0.00000,0.87500}%
\definecolor{mycolor2}{rgb}{0.00000,0.12500,1.00000}%
\definecolor{mycolor3}{rgb}{0.00000,0.25000,1.00000}%
\definecolor{mycolor4}{rgb}{0.00000,0.37500,1.00000}%
\definecolor{mycolor5}{rgb}{0.00000,0.50000,1.00000}%
\definecolor{mycolor6}{rgb}{0.00000,0.62500,1.00000}%
\definecolor{mycolor7}{rgb}{0.00000,0.75000,1.00000}%
\definecolor{mycolor8}{rgb}{1.00000,0.87500,0.00000}%
\definecolor{mycolor9}{rgb}{1.00000,0.75000,0.00000}%
\definecolor{mycolor10}{rgb}{0.87500,0.00000,0.00000}%
\begin{tikzpicture}

\begin{axis}[%
width=\figurewidth,
height=0.417\figureheight,
at={(0\figurewidth,0\figureheight)},
scale only axis,
xmin=0.5,
xmax=3.5,
xtick={1,2,3,4,5,6,7,8,9,10,11,12,13,14,15,16,17,18,19,20,21,22,23,24,25,26,27,28,29,30,31,32,33,34,35,36,37,38,39,40,41,42,43,44,45,46,47,48,49,50,51,52,53,54,55,56,57},
xticklabels={{2x},{3x},{4x}},
ymin=-0.243268842916461,
ymax=0.979385609077267,
ytick={0, 0.25, 0.5, 0.75},axis background/.style={fill=white},
xmajorgrids,
ymajorgrids,
xlabel near ticks,ylabel near ticks,scaled y ticks=false,yticklabel style={/pgf/number format/fixed, /pgf/number format/precision=2},
]
\addplot[ybar,bar width=0.034,bar shift=-0.379,fill=black!25!blue,draw=black,area legend] plot table[row sep=crcr] {%
1	0.0346105943708696\\
2	0.146671785937909\\
3	0.248405594246889\\
};
\addplot[forget plot,color=white!15!black] table[row sep=crcr] {%
0.5	0\\
3.5	0\\
};
\addplot[ybar,bar width=0.034,bar shift=-0.337,fill=mycolor1,draw=black,area legend] plot table[row sep=crcr] {%
1	0.010334298684155\\
2	0.217051173914317\\
3	0.346855272983661\\
};
\addplot[forget plot,color=white!15!black] table[row sep=crcr] {%
0.5	0\\
3.5	0\\
};
\addplot[ybar,bar width=0.034,bar shift=-0.295,fill=blue,draw=black,area legend] plot table[row sep=crcr] {%
1	0.00411754592158344\\
2	0.149066408833909\\
3	0.261296271815287\\
};
\addplot[forget plot,color=white!15!black] table[row sep=crcr] {%
0.5	0\\
3.5	0\\
};
\addplot[ybar,bar width=0.034,bar shift=-0.253,fill=mycolor2,draw=black,area legend] plot table[row sep=crcr] {%
1	0.0605605128974474\\
2	0.235632522529961\\
3	0.383527168282461\\
};
\addplot[forget plot,color=white!15!black] table[row sep=crcr] {%
0.5	0\\
3.5	0\\
};
\addplot[ybar,bar width=0.034,bar shift=-0.211,fill=mycolor3,draw=black,area legend] plot table[row sep=crcr] {%
1	0.0624099847451456\\
2	0.237064172342369\\
3	0.378712238131791\\
};
\addplot[forget plot,color=white!15!black] table[row sep=crcr] {%
0.5	0\\
3.5	0\\
};
\addplot[ybar,bar width=0.034,bar shift=-0.168,fill=mycolor4,draw=black,area legend] plot table[row sep=crcr] {%
1	0.0454244804546661\\
2	0.21621063937027\\
3	0.349472809971264\\
};
\addplot[forget plot,color=white!15!black] table[row sep=crcr] {%
0.5	0\\
3.5	0\\
};
\addplot[ybar,bar width=0.034,bar shift=-0.126,fill=mycolor5,draw=black,area legend] plot table[row sep=crcr] {%
1	0.0700499360539537\\
2	0.254375232522845\\
3	0.430649464215933\\
};
\addplot[forget plot,color=white!15!black] table[row sep=crcr] {%
0.5	0\\
3.5	0\\
};
\addplot[ybar,bar width=0.034,bar shift=-0.084,fill=mycolor6,draw=black,area legend] plot table[row sep=crcr] {%
1	0.0572180205939611\\
2	0.245372186700142\\
3	0.41035668990152\\
};
\addplot[forget plot,color=white!15!black] table[row sep=crcr] {%
0.5	0\\
3.5	0\\
};
\addplot[ybar,bar width=0.034,bar shift=-0.042,fill=mycolor7,draw=black,area legend] plot table[row sep=crcr] {%
1	0.0345059847266191\\
2	0.176871436380782\\
3	0.299936882411215\\
};
\addplot[forget plot,color=white!15!black] table[row sep=crcr] {%
0.5	0\\
3.5	0\\
};
\addplot[ybar,bar width=0.034,bar shift=-0,fill=mycolor8,draw=black,area legend] plot table[row sep=crcr] {%
1	0.313995038480457\\
2	0.709191912209464\\
3	0.637243879755838\\
};
\addplot[forget plot,color=white!15!black] table[row sep=crcr] {%
0.5	0\\
3.5	0\\
};
\addplot[ybar,bar width=0.034,bar shift=0.042,fill=mycolor9,draw=black,area legend] plot table[row sep=crcr] {%
1	0.189723781501552\\
2	0.587282052863463\\
3	0.637312947677521\\
};
\addplot[forget plot,color=white!15!black] table[row sep=crcr] {%
0.5	0\\
3.5	0\\
};
\addplot[ybar,bar width=0.034,bar shift=0.084,fill=orange!50!mycolor9,draw=black,area legend] plot table[row sep=crcr] {%
1	0.276148338766905\\
2	0.70151936105688\\
3	0.671119425355009\\
};
\addplot[forget plot,color=white!15!black] table[row sep=crcr] {%
0.5	0\\
3.5	0\\
};
\addplot[ybar,bar width=0.034,bar shift=0.126,fill=orange,draw=black,area legend] plot table[row sep=crcr] {%
1	0.185137509476007\\
2	0.575871022872472\\
3	0.629949531878794\\
};
\addplot[forget plot,color=white!15!black] table[row sep=crcr] {%
0.5	0\\
3.5	0\\
};
\addplot[ybar,bar width=0.034,bar shift=0.168,fill=red!25!orange,draw=black,area legend] plot table[row sep=crcr] {%
1	0.206158721688332\\
2	0.593743518566645\\
3	0.798067401945281\\
};
\addplot[forget plot,color=white!15!black] table[row sep=crcr] {%
0.5	0\\
3.5	0\\
};
\addplot[ybar,bar width=0.034,bar shift=0.211,fill=red!50!orange,draw=black,area legend] plot table[row sep=crcr] {%
1	0.237652565425821\\
2	0.717789059529604\\
3	0.932748199121207\\
};
\addplot[forget plot,color=white!15!black] table[row sep=crcr] {%
0.5	0\\
3.5	0\\
};
\addplot[ybar,bar width=0.034,bar shift=0.253,fill=red!75!orange,draw=black,area legend] plot table[row sep=crcr] {%
1	0.319307187833851\\
2	0.699854358835014\\
3	0.867071588544234\\
};
\addplot[forget plot,color=white!15!black] table[row sep=crcr] {%
0.5	0\\
3.5	0\\
};
\addplot[ybar,bar width=0.034,bar shift=0.295,fill=red,draw=black,area legend] plot table[row sep=crcr] {%
1	0.00684740485681463\\
2	0.00949718418997463\\
3	-0.0956962530622943\\
};
\addplot[forget plot,color=white!15!black] table[row sep=crcr] {%
0.5	0\\
3.5	0\\
};
\addplot[ybar,bar width=0.034,bar shift=0.337,fill=mycolor10,draw=black,area legend] plot table[row sep=crcr] {%
1	-0.245726103956022\\
2	0.282040648293659\\
3	0.716143724600444\\
};
\addplot[forget plot,color=white!15!black] table[row sep=crcr] {%
0.5	0\\
3.5	0\\
};
\addplot[ybar,bar width=0.034,bar shift=0.379,fill=black!25!red,draw=black,area legend] plot table[row sep=crcr] {%
1	0.134400086625632\\
2	0.369171946668426\\
3	0.479774983009539\\
};
\addplot[forget plot,color=white!15!black] table[row sep=crcr] {%
0.5	0\\
3.5	0\\
};
\end{axis}
\end{tikzpicture}%
%
\label{fig:srBenchmarkMotionTypes:global}}~
	\subfloat[Mixed motion datasets]{%
%
\definecolor{mycolor1}{rgb}{0.00000,0.00000,0.87500}%
\definecolor{mycolor2}{rgb}{0.00000,0.12500,1.00000}%
\definecolor{mycolor3}{rgb}{0.00000,0.25000,1.00000}%
\definecolor{mycolor4}{rgb}{0.00000,0.37500,1.00000}%
\definecolor{mycolor5}{rgb}{0.00000,0.50000,1.00000}%
\definecolor{mycolor6}{rgb}{0.00000,0.62500,1.00000}%
\definecolor{mycolor7}{rgb}{0.00000,0.75000,1.00000}%
\definecolor{mycolor8}{rgb}{1.00000,0.87500,0.00000}%
\definecolor{mycolor9}{rgb}{1.00000,0.75000,0.00000}%
\definecolor{mycolor10}{rgb}{0.87500,0.00000,0.00000}%
\begin{tikzpicture}

\begin{axis}[%
width=\figurewidth,
height=0.417\figureheight,
at={(0\figurewidth,0\figureheight)},
scale only axis,
xmin=0.5,
xmax=3.5,
xtick={1,2,3,4,5,6,7,8,9,10,11,12,13,14,15,16,17,18,19,20,21,22,23,24,25,26,27,28,29,30,31,32,33,34,35,36,37,38,39,40,41,42,43,44,45,46,47,48,49,50,51,52,53,54,55,56,57},
xticklabels={{2x},{3x},{4x}},
ymin=-0.243268842916461,
ymax=0.979385609077267,
ytick={0, 0.25, 0.5, 0.75},axis background/.style={fill=white},
xmajorgrids,
ymajorgrids,
xlabel near ticks,ylabel near ticks,scaled y ticks=false,yticklabel style={/pgf/number format/fixed, /pgf/number format/precision=2},
]
\addplot[ybar,bar width=0.034,bar shift=-0.379,fill=black!25!blue,draw=black,area legend] plot table[row sep=crcr] {%
1	0.0344397876210232\\
2	0.14657677336705\\
3	0.248646677002269\\
};
\addplot[forget plot,color=white!15!black] table[row sep=crcr] {%
0.5	0\\
3.5	0\\
};
\addplot[ybar,bar width=0.034,bar shift=-0.337,fill=mycolor1,draw=black,area legend] plot table[row sep=crcr] {%
1	0.00952794103693039\\
2	0.215079599165365\\
3	0.345614996768088\\
};
\addplot[forget plot,color=white!15!black] table[row sep=crcr] {%
0.5	0\\
3.5	0\\
};
\addplot[ybar,bar width=0.034,bar shift=-0.295,fill=blue,draw=black,area legend] plot table[row sep=crcr] {%
1	0.00309997336355322\\
2	0.14787568566474\\
3	0.26015788336533\\
};
\addplot[forget plot,color=white!15!black] table[row sep=crcr] {%
0.5	0\\
3.5	0\\
};
\addplot[ybar,bar width=0.034,bar shift=-0.253,fill=mycolor2,draw=black,area legend] plot table[row sep=crcr] {%
1	0.0591392084189326\\
2	0.233367409170956\\
3	0.380527014906312\\
};
\addplot[forget plot,color=white!15!black] table[row sep=crcr] {%
0.5	0\\
3.5	0\\
};
\addplot[ybar,bar width=0.034,bar shift=-0.211,fill=mycolor3,draw=black,area legend] plot table[row sep=crcr] {%
1	0.0610728594976171\\
2	0.235173703572348\\
3	0.376278687780214\\
};
\addplot[forget plot,color=white!15!black] table[row sep=crcr] {%
0.5	0\\
3.5	0\\
};
\addplot[ybar,bar width=0.034,bar shift=-0.168,fill=mycolor4,draw=black,area legend] plot table[row sep=crcr] {%
1	0.0439994358648686\\
2	0.214322947508238\\
3	0.347380134938269\\
};
\addplot[forget plot,color=white!15!black] table[row sep=crcr] {%
0.5	0\\
3.5	0\\
};
\addplot[ybar,bar width=0.034,bar shift=-0.126,fill=mycolor5,draw=black,area legend] plot table[row sep=crcr] {%
1	0.0685159244693324\\
2	0.251706539970467\\
3	0.427188051502294\\
};
\addplot[forget plot,color=white!15!black] table[row sep=crcr] {%
0.5	0\\
3.5	0\\
};
\addplot[ybar,bar width=0.034,bar shift=-0.084,fill=mycolor6,draw=black,area legend] plot table[row sep=crcr] {%
1	0.0559761545455902\\
2	0.242913594104961\\
3	0.40731710589702\\
};
\addplot[forget plot,color=white!15!black] table[row sep=crcr] {%
0.5	0\\
3.5	0\\
};
\addplot[ybar,bar width=0.034,bar shift=-0.042,fill=mycolor7,draw=black,area legend] plot table[row sep=crcr] {%
1	0.0332005523003169\\
2	0.175059969252329\\
3	0.296608696124324\\
};
\addplot[forget plot,color=white!15!black] table[row sep=crcr] {%
0.5	0\\
3.5	0\\
};
\addplot[ybar,bar width=0.034,bar shift=-0,fill=mycolor8,draw=black,area legend] plot table[row sep=crcr] {%
1	0.232429870987585\\
2	0.478053827981364\\
3	0.365629638147883\\
};
\addplot[forget plot,color=white!15!black] table[row sep=crcr] {%
0.5	0\\
3.5	0\\
};
\addplot[ybar,bar width=0.034,bar shift=0.042,fill=mycolor9,draw=black,area legend] plot table[row sep=crcr] {%
1	0.12630075090213\\
2	0.415784235566078\\
3	0.41861917202234\\
};
\addplot[forget plot,color=white!15!black] table[row sep=crcr] {%
0.5	0\\
3.5	0\\
};
\addplot[ybar,bar width=0.034,bar shift=0.084,fill=orange!50!mycolor9,draw=black,area legend] plot table[row sep=crcr] {%
1	0.22890147937609\\
2	0.512916498968336\\
3	0.430005206528112\\
};
\addplot[forget plot,color=white!15!black] table[row sep=crcr] {%
0.5	0\\
3.5	0\\
};
\addplot[ybar,bar width=0.034,bar shift=0.126,fill=orange,draw=black,area legend] plot table[row sep=crcr] {%
1	0.122588775612834\\
2	0.417213236141841\\
3	0.434242206772952\\
};
\addplot[forget plot,color=white!15!black] table[row sep=crcr] {%
0.5	0\\
3.5	0\\
};
\addplot[ybar,bar width=0.034,bar shift=0.168,fill=red!25!orange,draw=black,area legend] plot table[row sep=crcr] {%
1	0.153901809204779\\
2	0.425233907495483\\
3	0.524744908527659\\
};
\addplot[forget plot,color=white!15!black] table[row sep=crcr] {%
0.5	0\\
3.5	0\\
};
\addplot[ybar,bar width=0.034,bar shift=0.211,fill=red!50!orange,draw=black,area legend] plot table[row sep=crcr] {%
1	0.225335850488748\\
2	0.578229307622062\\
3	0.665230499323814\\
};
\addplot[forget plot,color=white!15!black] table[row sep=crcr] {%
0.5	0\\
3.5	0\\
};
\addplot[ybar,bar width=0.034,bar shift=0.253,fill=red!75!orange,draw=black,area legend] plot table[row sep=crcr] {%
1	0.259613242652425\\
2	0.55061331234387\\
3	0.640566311348252\\
};
\addplot[forget plot,color=white!15!black] table[row sep=crcr] {%
0.5	0\\
3.5	0\\
};
\addplot[ybar,bar width=0.034,bar shift=0.295,fill=red,draw=black,area legend] plot table[row sep=crcr] {%
1	-0.00863930549473035\\
2	-0.0922312289028135\\
3	-0.199305818361219\\
};
\addplot[forget plot,color=white!15!black] table[row sep=crcr] {%
0.5	0\\
3.5	0\\
};
\addplot[ybar,bar width=0.034,bar shift=0.337,fill=mycolor10,draw=black,area legend] plot table[row sep=crcr] {%
1	-0.265901243487266\\
2	0.172443472655706\\
3	0.466428217540643\\
};
\addplot[forget plot,color=white!15!black] table[row sep=crcr] {%
0.5	0\\
3.5	0\\
};
\addplot[ybar,bar width=0.034,bar shift=0.379,fill=black!25!red,draw=black,area legend] plot table[row sep=crcr] {%
1	0.128400536376423\\
2	0.35337394146496\\
3	0.464325658992906\\
};
\addplot[forget plot,color=white!15!black] table[row sep=crcr] {%
0.5	0\\
3.5	0\\
};
\end{axis}
\end{tikzpicture}%
%
\label{fig:srBenchmarkMotionTypes:mixed}}~
	\subfloat[Local motion datasets]{%
%
\definecolor{mycolor1}{rgb}{0.00000,0.00000,0.87500}%
\definecolor{mycolor2}{rgb}{0.00000,0.12500,1.00000}%
\definecolor{mycolor3}{rgb}{0.00000,0.25000,1.00000}%
\definecolor{mycolor4}{rgb}{0.00000,0.37500,1.00000}%
\definecolor{mycolor5}{rgb}{0.00000,0.50000,1.00000}%
\definecolor{mycolor6}{rgb}{0.00000,0.62500,1.00000}%
\definecolor{mycolor7}{rgb}{0.00000,0.75000,1.00000}%
\definecolor{mycolor8}{rgb}{1.00000,0.87500,0.00000}%
\definecolor{mycolor9}{rgb}{1.00000,0.75000,0.00000}%
\definecolor{mycolor10}{rgb}{0.87500,0.00000,0.00000}%
\begin{tikzpicture}

\begin{axis}[%
width=\figurewidth,
height=0.417\figureheight,
at={(0\figurewidth,0\figureheight)},
scale only axis,
xmin=0.5,
xmax=3.5,
xtick={1,2,3,4,5,6,7,8,9,10,11,12,13,14,15,16,17,18,19,20,21,22,23,24,25,26,27,28,29,30,31,32,33,34,35,36,37,38,39,40,41,42,43,44,45,46,47,48,49,50,51,52,53,54,55,56,57},
xticklabels={{2x},{3x},{4x}},
ymin=-0.243268842916461,
ymax=0.979385609077267,
ytick={0, 0.25, 0.5, 0.75},axis background/.style={fill=white},
xmajorgrids,
ymajorgrids,
xlabel near ticks,ylabel near ticks,scaled y ticks=false,yticklabel style={/pgf/number format/fixed, /pgf/number format/precision=2},
]
\addplot[ybar,bar width=0.034,bar shift=-0.379,fill=black!25!blue,draw=black,area legend] plot table[row sep=crcr] {%
1	0.0348020615162899\\
2	0.14630596727952\\
3	0.248122064022826\\
};
\addplot[forget plot,color=white!15!black] table[row sep=crcr] {%
0.5	0\\
3.5	0\\
};
\addplot[ybar,bar width=0.034,bar shift=-0.337,fill=mycolor1,draw=black,area legend] plot table[row sep=crcr] {%
1	0.0114359256678006\\
2	0.214977734747985\\
3	0.343797657845924\\
};
\addplot[forget plot,color=white!15!black] table[row sep=crcr] {%
0.5	0\\
3.5	0\\
};
\addplot[ybar,bar width=0.034,bar shift=-0.295,fill=blue,draw=black,area legend] plot table[row sep=crcr] {%
1	0.00354464642622407\\
2	0.148378401792146\\
3	0.259086074058898\\
};
\addplot[forget plot,color=white!15!black] table[row sep=crcr] {%
0.5	0\\
3.5	0\\
};
\addplot[ybar,bar width=0.034,bar shift=-0.253,fill=mycolor2,draw=black,area legend] plot table[row sep=crcr] {%
1	0.0609739110423952\\
2	0.233471358539718\\
3	0.379170510001626\\
};
\addplot[forget plot,color=white!15!black] table[row sep=crcr] {%
0.5	0\\
3.5	0\\
};
\addplot[ybar,bar width=0.034,bar shift=-0.211,fill=mycolor3,draw=black,area legend] plot table[row sep=crcr] {%
1	0.0629026868757296\\
2	0.235382854476057\\
3	0.374983996676879\\
};
\addplot[forget plot,color=white!15!black] table[row sep=crcr] {%
0.5	0\\
3.5	0\\
};
\addplot[ybar,bar width=0.034,bar shift=-0.168,fill=mycolor4,draw=black,area legend] plot table[row sep=crcr] {%
1	0.0455892821121758\\
2	0.214780767419041\\
3	0.34540873678648\\
};
\addplot[forget plot,color=white!15!black] table[row sep=crcr] {%
0.5	0\\
3.5	0\\
};
\addplot[ybar,bar width=0.034,bar shift=-0.126,fill=mycolor5,draw=black,area legend] plot table[row sep=crcr] {%
1	0.070521300981638\\
2	0.252369178198677\\
3	0.424168609624727\\
};
\addplot[forget plot,color=white!15!black] table[row sep=crcr] {%
0.5	0\\
3.5	0\\
};
\addplot[ybar,bar width=0.034,bar shift=-0.084,fill=mycolor6,draw=black,area legend] plot table[row sep=crcr] {%
1	0.0577142154072153\\
2	0.243745829631051\\
3	0.406285235776815\\
};
\addplot[forget plot,color=white!15!black] table[row sep=crcr] {%
0.5	0\\
3.5	0\\
};
\addplot[ybar,bar width=0.034,bar shift=-0.042,fill=mycolor7,draw=black,area legend] plot table[row sep=crcr] {%
1	0.0351467570209687\\
2	0.175450069758422\\
3	0.293173228853547\\
};
\addplot[forget plot,color=white!15!black] table[row sep=crcr] {%
0.5	0\\
3.5	0\\
};
\addplot[ybar,bar width=0.034,bar shift=-0,fill=mycolor8,draw=black,area legend] plot table[row sep=crcr] {%
1	0.0749899878788156\\
2	0.04576387725868\\
3	-0.00123631563979869\\
};
\addplot[forget plot,color=white!15!black] table[row sep=crcr] {%
0.5	0\\
3.5	0\\
};
\addplot[ybar,bar width=0.034,bar shift=0.042,fill=mycolor9,draw=black,area legend] plot table[row sep=crcr] {%
1	0.188347505794424\\
2	0.246037106939179\\
3	0.223807563371349\\
};
\addplot[forget plot,color=white!15!black] table[row sep=crcr] {%
0.5	0\\
3.5	0\\
};
\addplot[ybar,bar width=0.034,bar shift=0.084,fill=orange!50!mycolor9,draw=black,area legend] plot table[row sep=crcr] {%
1	0.038718926223364\\
2	0.16390815314813\\
3	0.161530138292714\\
};
\addplot[forget plot,color=white!15!black] table[row sep=crcr] {%
0.5	0\\
3.5	0\\
};
\addplot[ybar,bar width=0.034,bar shift=0.126,fill=orange,draw=black,area legend] plot table[row sep=crcr] {%
1	0.174403296435295\\
2	0.205433279241348\\
3	0.197933955410359\\
};
\addplot[forget plot,color=white!15!black] table[row sep=crcr] {%
0.5	0\\
3.5	0\\
};
\addplot[ybar,bar width=0.034,bar shift=0.168,fill=red!25!orange,draw=black,area legend] plot table[row sep=crcr] {%
1	0.136477414926454\\
2	0.215529308639099\\
3	0.22071080308343\\
};
\addplot[forget plot,color=white!15!black] table[row sep=crcr] {%
0.5	0\\
3.5	0\\
};
\addplot[ybar,bar width=0.034,bar shift=0.211,fill=red!50!orange,draw=black,area legend] plot table[row sep=crcr] {%
1	0.176922728266741\\
2	0.334577232872867\\
3	0.346746291041673\\
};
\addplot[forget plot,color=white!15!black] table[row sep=crcr] {%
0.5	0\\
3.5	0\\
};
\addplot[ybar,bar width=0.034,bar shift=0.253,fill=red!75!orange,draw=black,area legend] plot table[row sep=crcr] {%
1	0.21126677866115\\
2	0.308303898376491\\
3	0.301755282581958\\
};
\addplot[forget plot,color=white!15!black] table[row sep=crcr] {%
0.5	0\\
3.5	0\\
};
\addplot[ybar,bar width=0.034,bar shift=0.295,fill=red,draw=black,area legend] plot table[row sep=crcr] {%
1	0.00267116636047432\\
2	-0.056122919306289\\
3	-0.131205051783149\\
};
\addplot[forget plot,color=white!15!black] table[row sep=crcr] {%
0.5	0\\
3.5	0\\
};
\addplot[ybar,bar width=0.034,bar shift=0.337,fill=mycolor10,draw=black,area legend] plot table[row sep=crcr] {%
1	-0.188603703161181\\
2	0.095430319546927\\
3	0.197245525797141\\
};
\addplot[forget plot,color=white!15!black] table[row sep=crcr] {%
0.5	0\\
3.5	0\\
};
\addplot[ybar,bar width=0.034,bar shift=0.379,fill=black!25!red,draw=black,area legend] plot table[row sep=crcr] {%
1	0.107001342728802\\
2	0.301551934350373\\
3	0.403836730827122\\
};
\addplot[forget plot,color=white!15!black] table[row sep=crcr] {%
0.5	0\\
3.5	0\\
};
\end{axis}
\end{tikzpicture}%
%
\label{fig:srBenchmarkMotionTypes:local}}\\
	\caption{Benchmark of the SR algorithms for global \protect\subref{fig:srBenchmarkMotionTypes:global}, mixed \protect\subref{fig:srBenchmarkMotionTypes:mixed}, and local motion \protect\subref{fig:srBenchmarkMotionTypes:local} using the mean normalized PSNR and IFC for $2\times$, $3\times$ and $4\times$ magnification. The algorithms are categorized either as SISR (shown with blue color map) or MFSR (shown with red color map).}
	\label{fig:srBenchmarkMotionTypes}
\end{figure*}

For SISR, we study dictionary and deep learning 
methods. The dictionary methods are example-based ridge regression (EBSR) 
\cite{Kim2010}, sparse coding (ScSR) \cite{Yang2010}, Naive 
Bayes SR forests (NBSRF) \cite{Salvador2015}, and adjusted anchored 
neighborhood regression (A+) \cite{Timofte2015}. The deep learning methods 
comprise CNNs (SRCNN) \cite{Dong2014}, very deep networks
(VDSR) \cite{Kim2016}, and deeply-recursive networks (DRCN) 
\cite{Kim2016a}. As an internal method, we evaluate transformed self-exemplars (SESR) 
\cite{Huang2015a}.

\begin{table}[!t]
	\scriptsize
	\caption{Categorization of the SR algorithms in our benchmark.}
	\centering
	\begin{tabular}{lll}
		\toprule
		\textbf{Category}			& \textbf{Single-image (SISR)}										& \textbf{Multi-frame (MFSR)}				\\
		\midrule
		Self-exemplars				& SESR \cite{Huang2015a}														&															\\
		\midrule
		Deep learning					& SRCNN \cite{Dong2014},VDSR \cite{Kim2016}					&	VSRnet \cite{Kappeler2016}	\\
		architectures					& DRCN \cite{Kim2016a}															&															\\
		\midrule
		Shallow								& ScSR \cite{Yang2010}, EBSR \cite{Kim2010}					&	HYSR \cite{Batz2015}				\\
		architectures					& NBSRF \cite{Salvador2015}, A+ \cite{Timofte2015}	&			\\
		
		\midrule		
		Interpolation					&	BICUBIC																						&	NUISR	\cite{Park2003}, WNUISR \cite{Batz2016}			\\
													& 																									& DBRSR \cite{Batz2016b}			\\
		\midrule	
		Non-blind							&																										&	L1BTV \cite{Farsiu2004a}, BEPSR \cite{Zeng2013}		\\
		reconstruction				& 																									&	IRWSR \cite{Kohler2015c}		\\
		\midrule
		Blind	reconstruction	&																										& BVSR \cite{Liu2014}, SRB \cite{Ma2015}					\\
		\bottomrule
	\end{tabular}
	\label{tab:algorithmOverview}
\end{table}

\begin{figure*}[!t]
	\flushleft
	\subfloat[LR input]{
	\begin{minipage}{0.158\textwidth} 
		\includegraphics[trim={32.7cm 14.5cm 1cm 12.5cm},clip,width=1.00\textwidth]
		{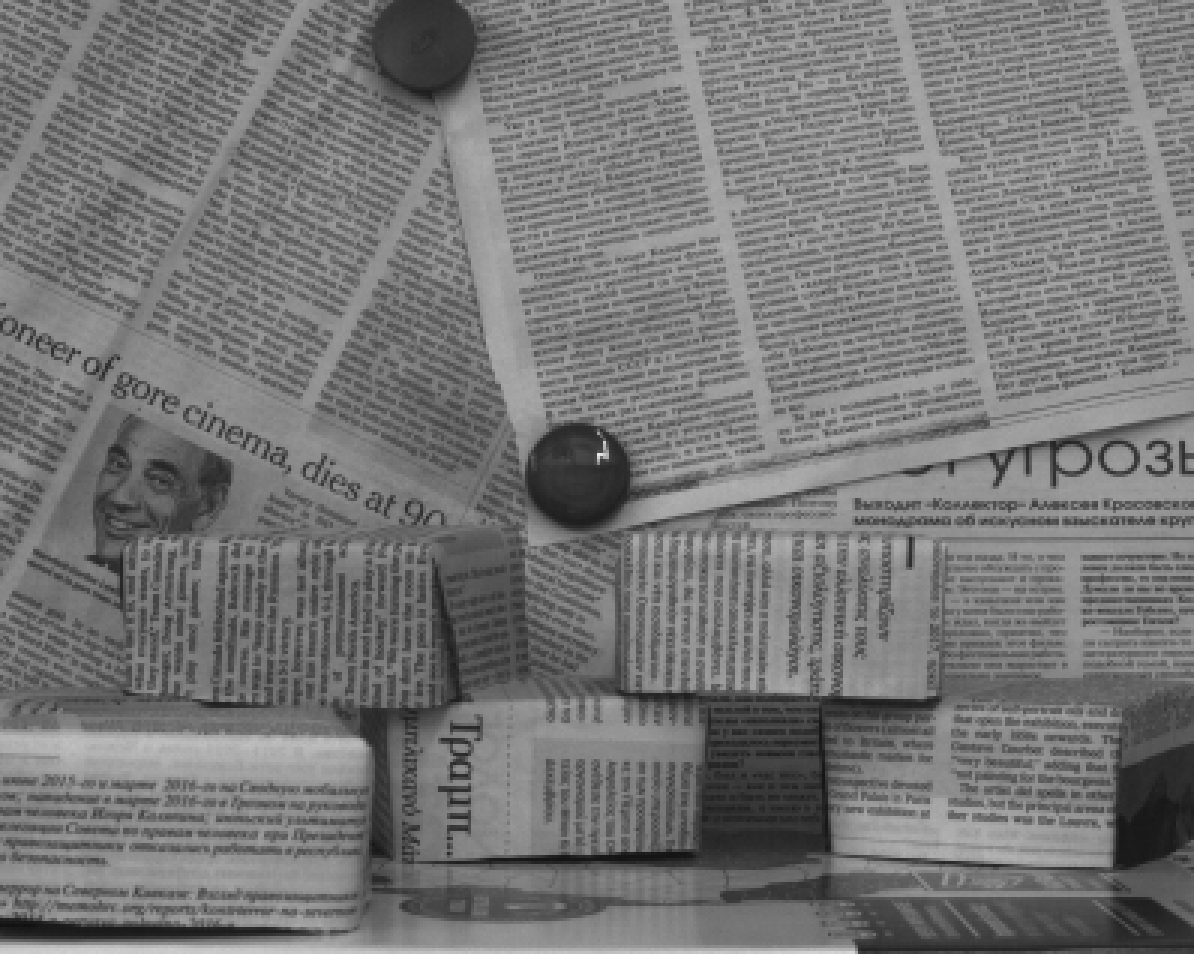}\\[0.4ex]
		\includegraphics[trim={32.7cm 14.5cm 1cm 12.5cm},clip,width=1.00\textwidth]
		{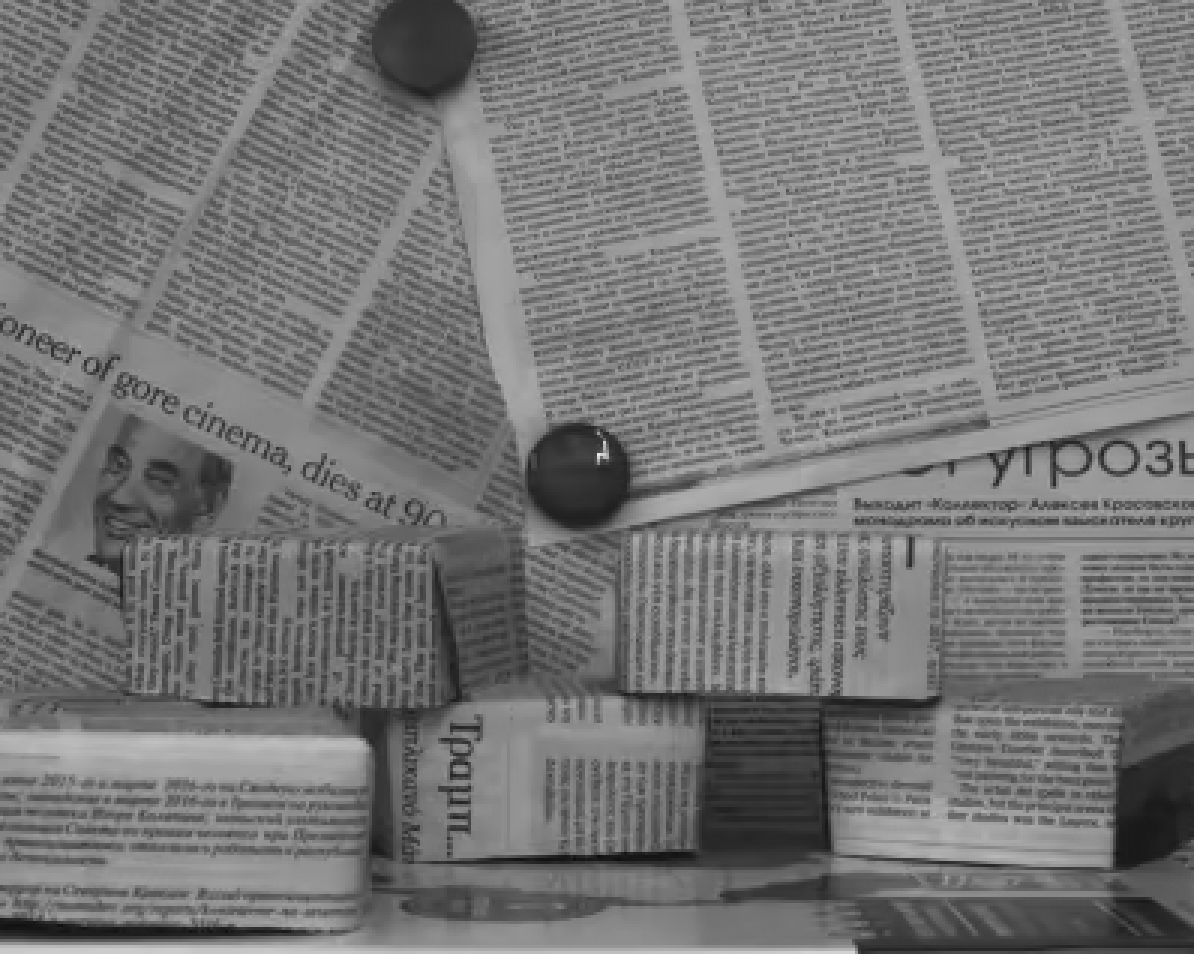}
	\end{minipage}}
	\subfloat[A+ \cite{Timofte2015}]{
	\begin{minipage}{0.158\textwidth} 
		\includegraphics[trim={32.7cm 14.5cm 1cm 12.5cm},clip,width=1.00\textwidth]
		{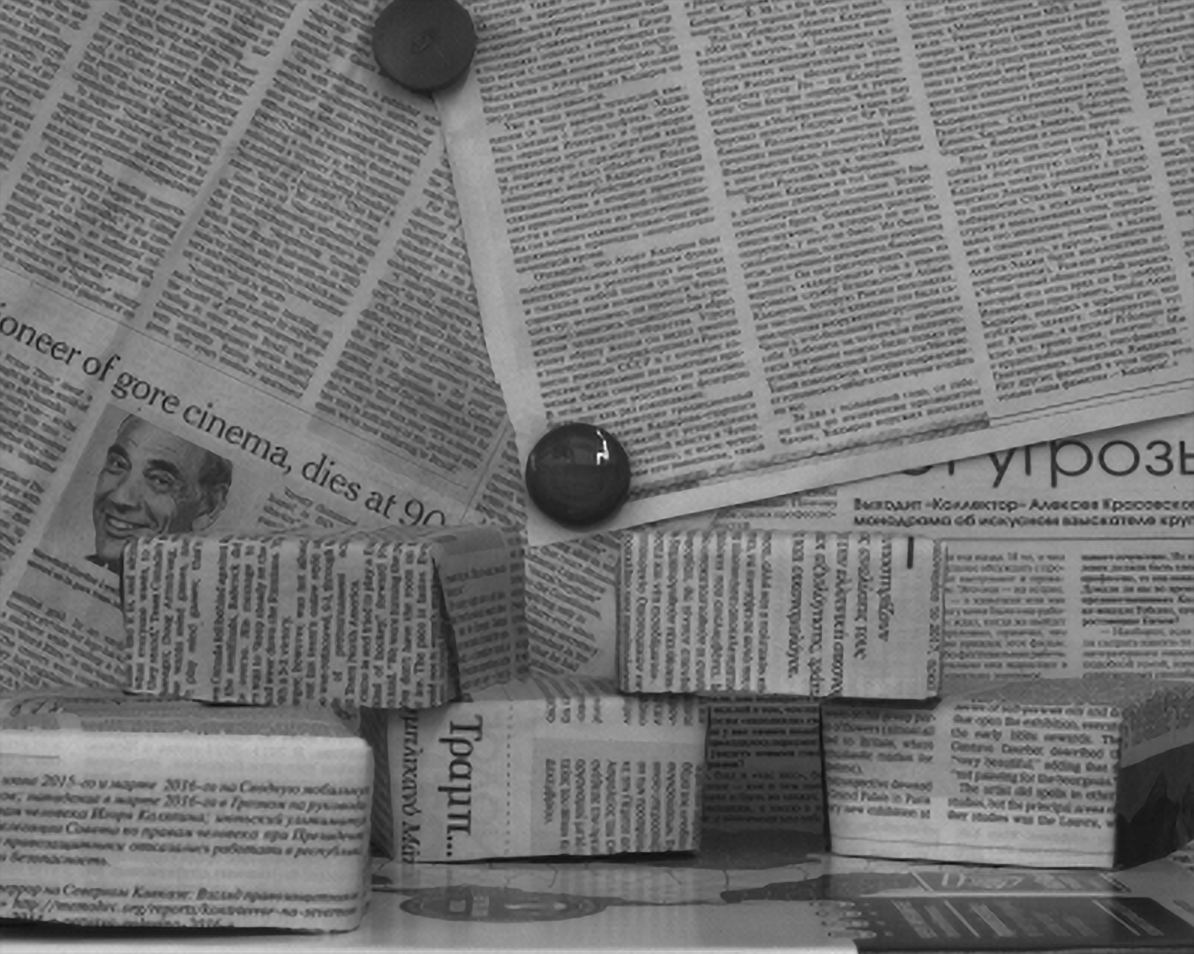}\\[0.4ex]
		\includegraphics[trim={32.7cm 14.5cm 1cm 12.5cm},clip,width=1.00\textwidth]
		{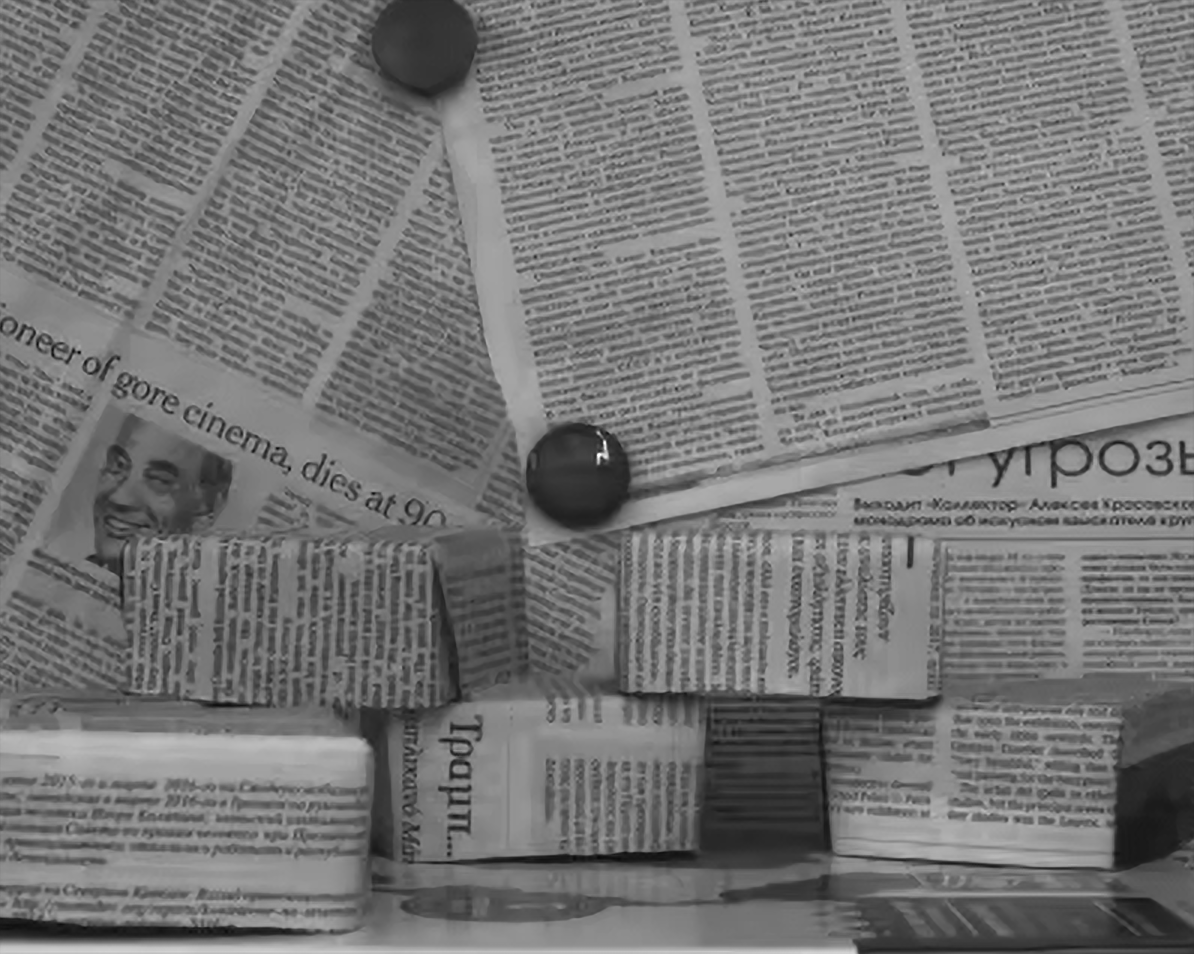}
	\end{minipage}}
	\subfloat[VDSR \cite{Kim2016}]{
	\begin{minipage}{0.158\textwidth} 
		\includegraphics[trim={32.7cm 14.5cm 1cm 12.5cm},clip,width=1.00\textwidth]
		{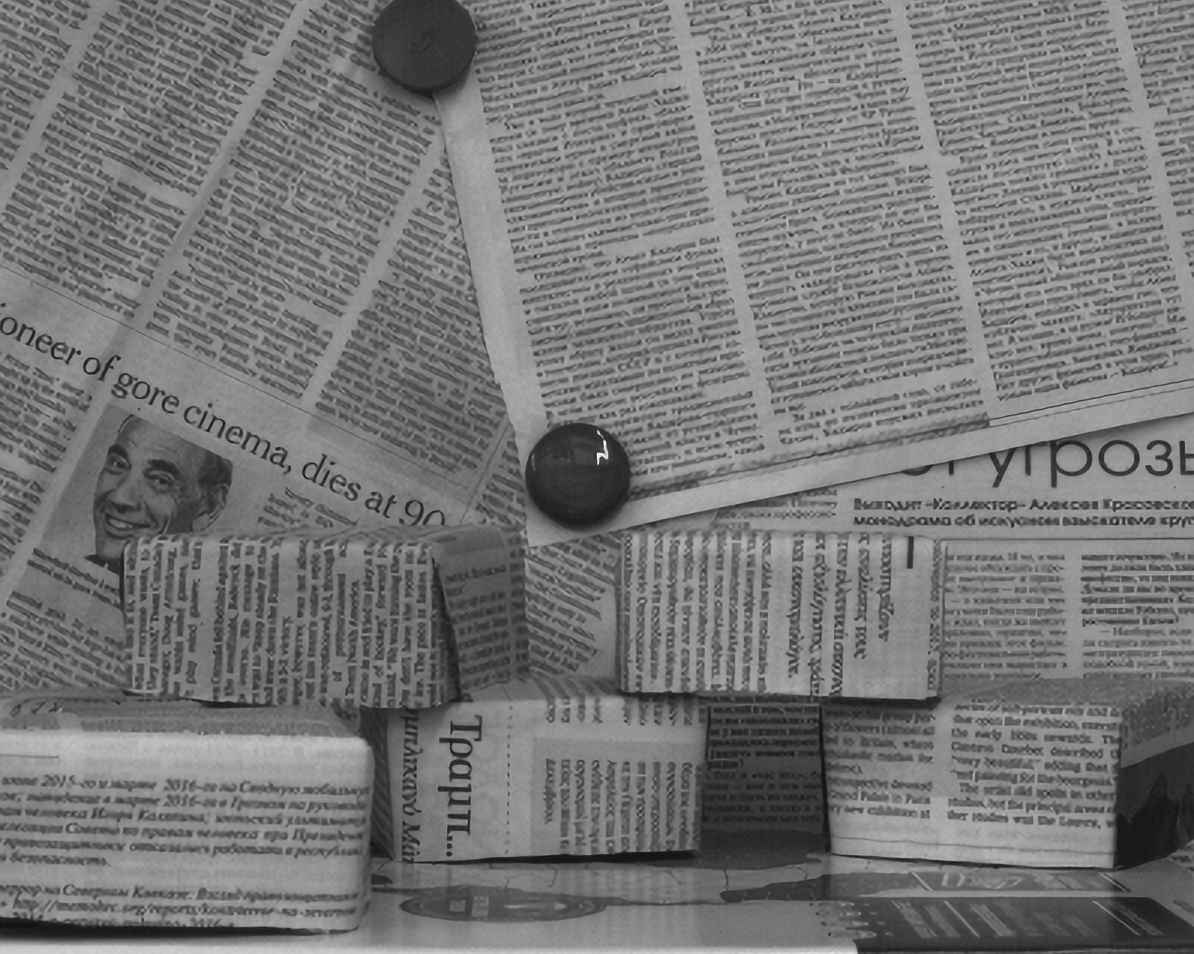}\\[0.4ex]
		\includegraphics[trim={32.7cm 14.5cm 1cm 12.5cm},clip,width=1.00\textwidth]
		{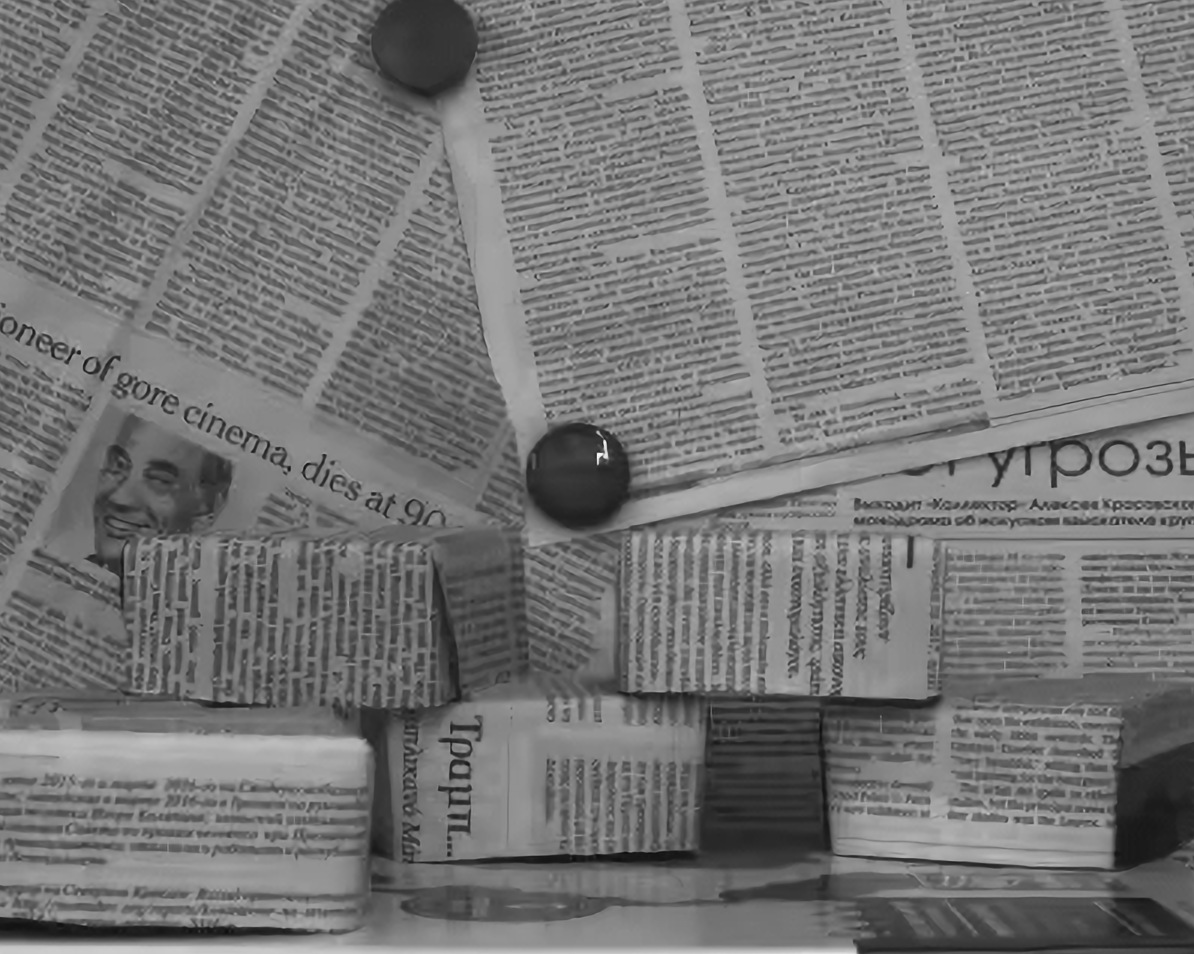}
	\end{minipage}}
	\subfloat[WNUISR \cite{Batz2016}]{
	\begin{minipage}{0.158\textwidth} 
		\includegraphics[trim={32.7cm 14.5cm 1cm 12.5cm},clip,width=1.00\textwidth]
		{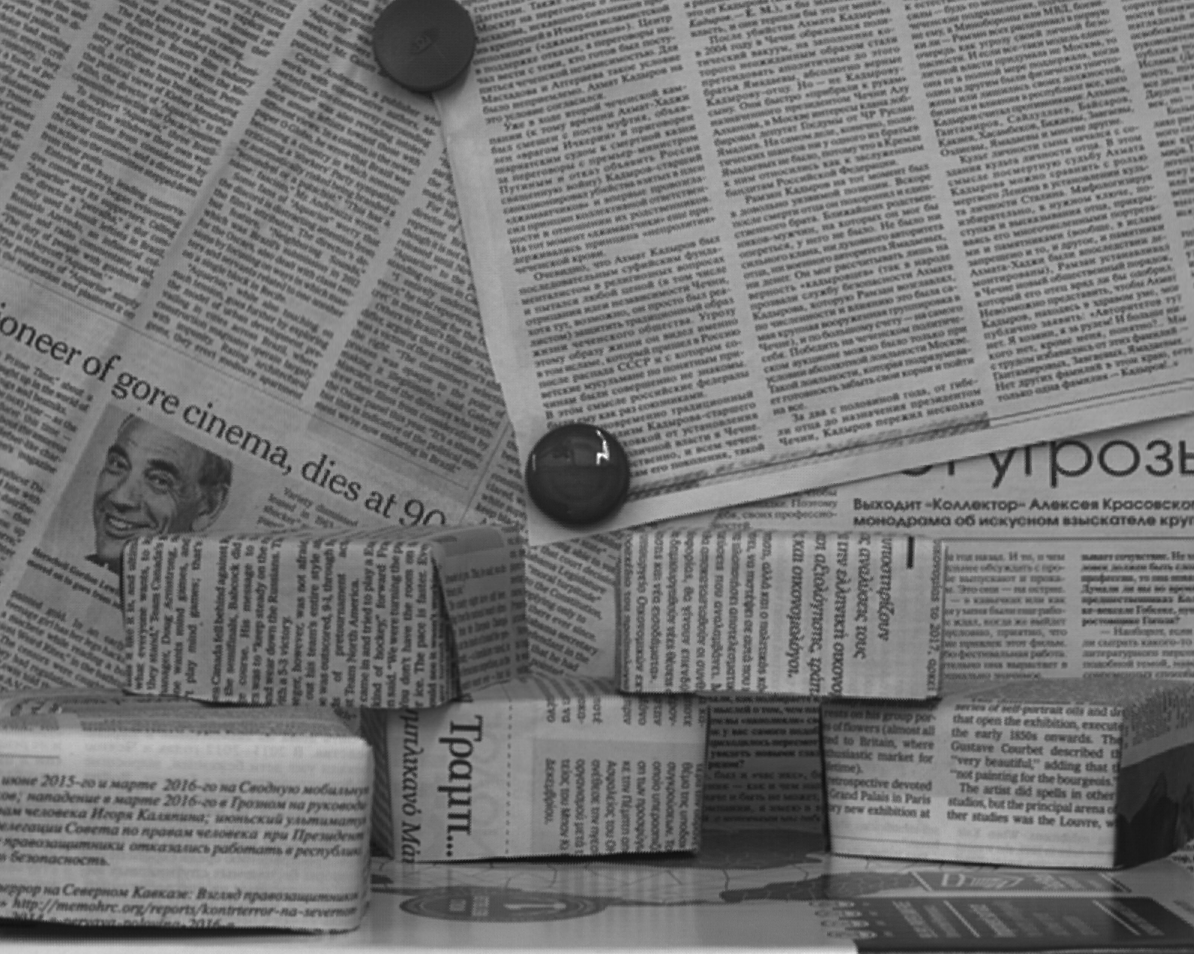}\\[0.4ex]
		\includegraphics[trim={32.7cm 14.5cm 1cm 12.5cm},clip,width=1.00\textwidth]
		{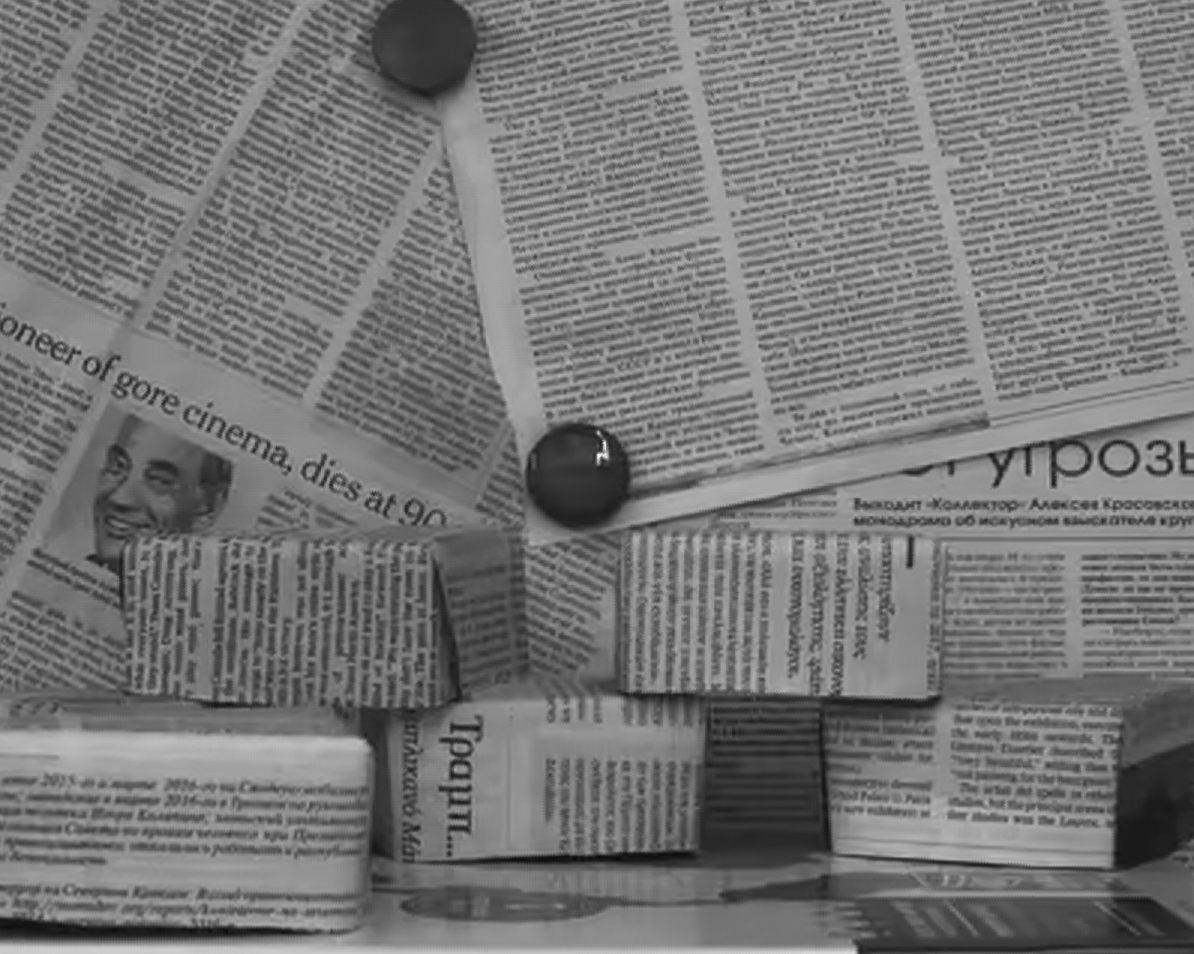}
	\end{minipage}}
	\subfloat[IRWSR \cite{Kohler2015c}]{
	\begin{minipage}{0.158\textwidth} 
		\includegraphics[trim={32.7cm 14.5cm 1cm 12.5cm},clip,width=1.00\textwidth]
		{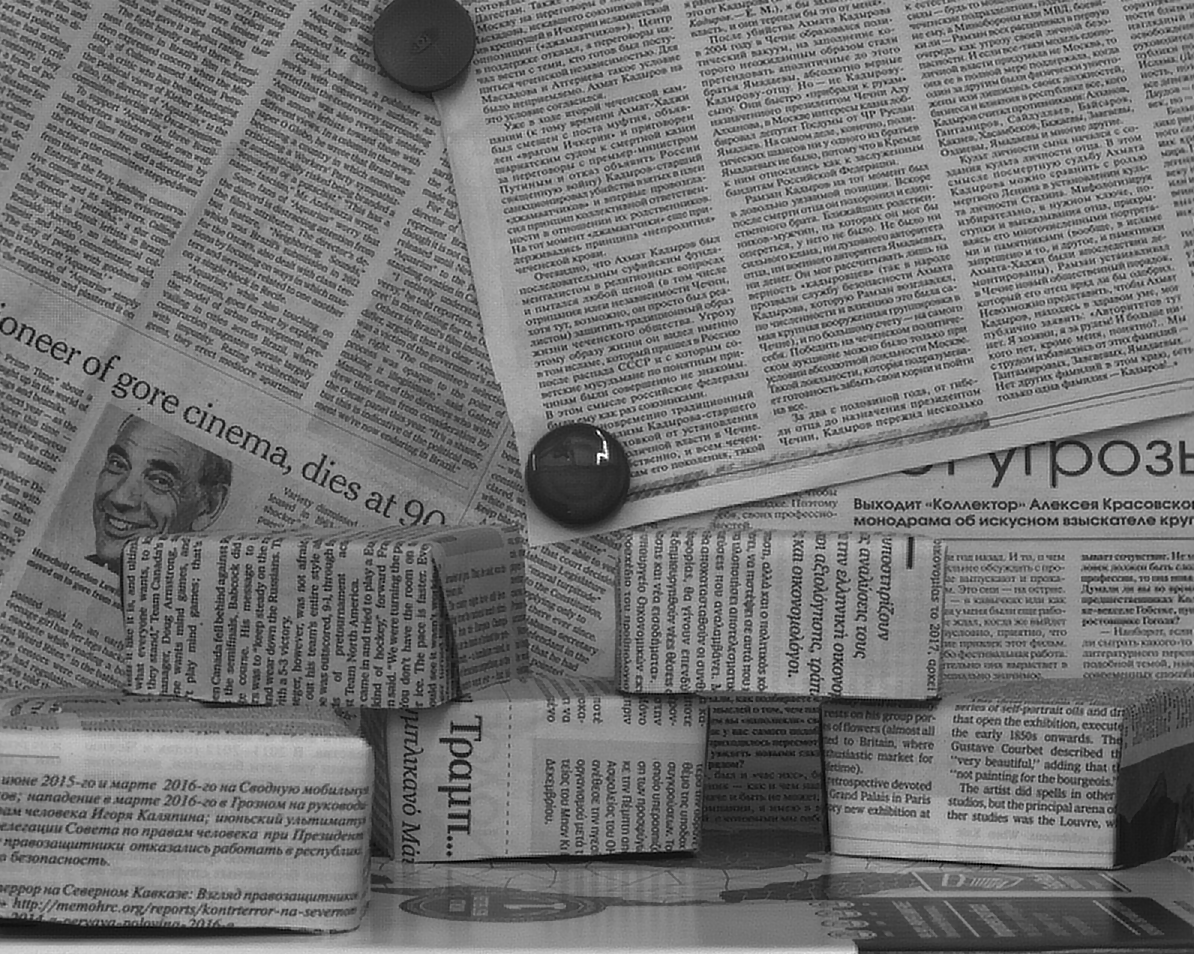}\\[0.4ex]
		\includegraphics[trim={32.7cm 14.5cm 1cm 12.5cm},clip,width=1.00\textwidth]
		{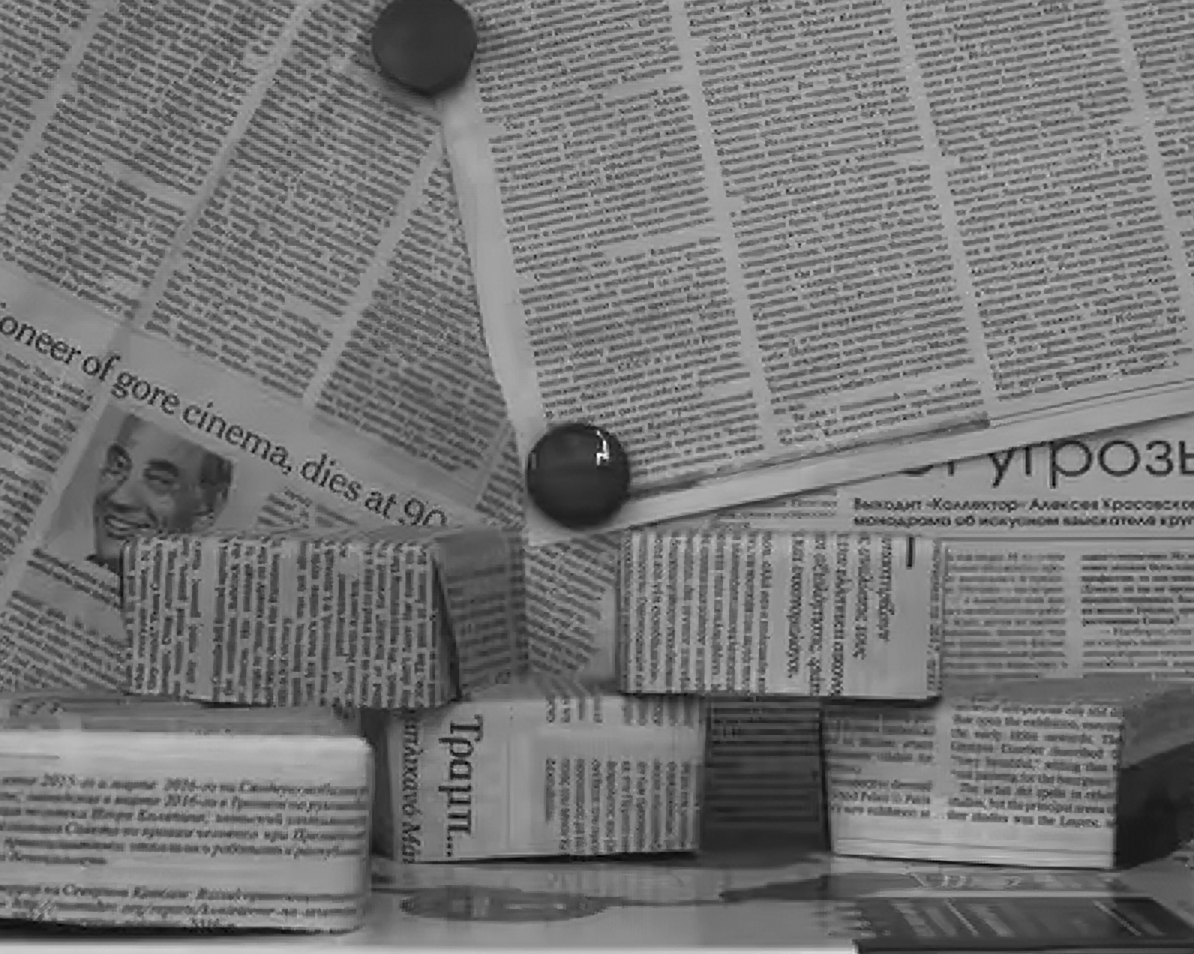}
	\end{minipage}}
	~\vline
	\subfloat[Ground truth]{
	\begin{minipage}{0.158\textwidth} 
		\begin{tikzpicture}
			\draw (0, 0) node[inner sep=0] {\includegraphics[width=1.00\textwidth]
			{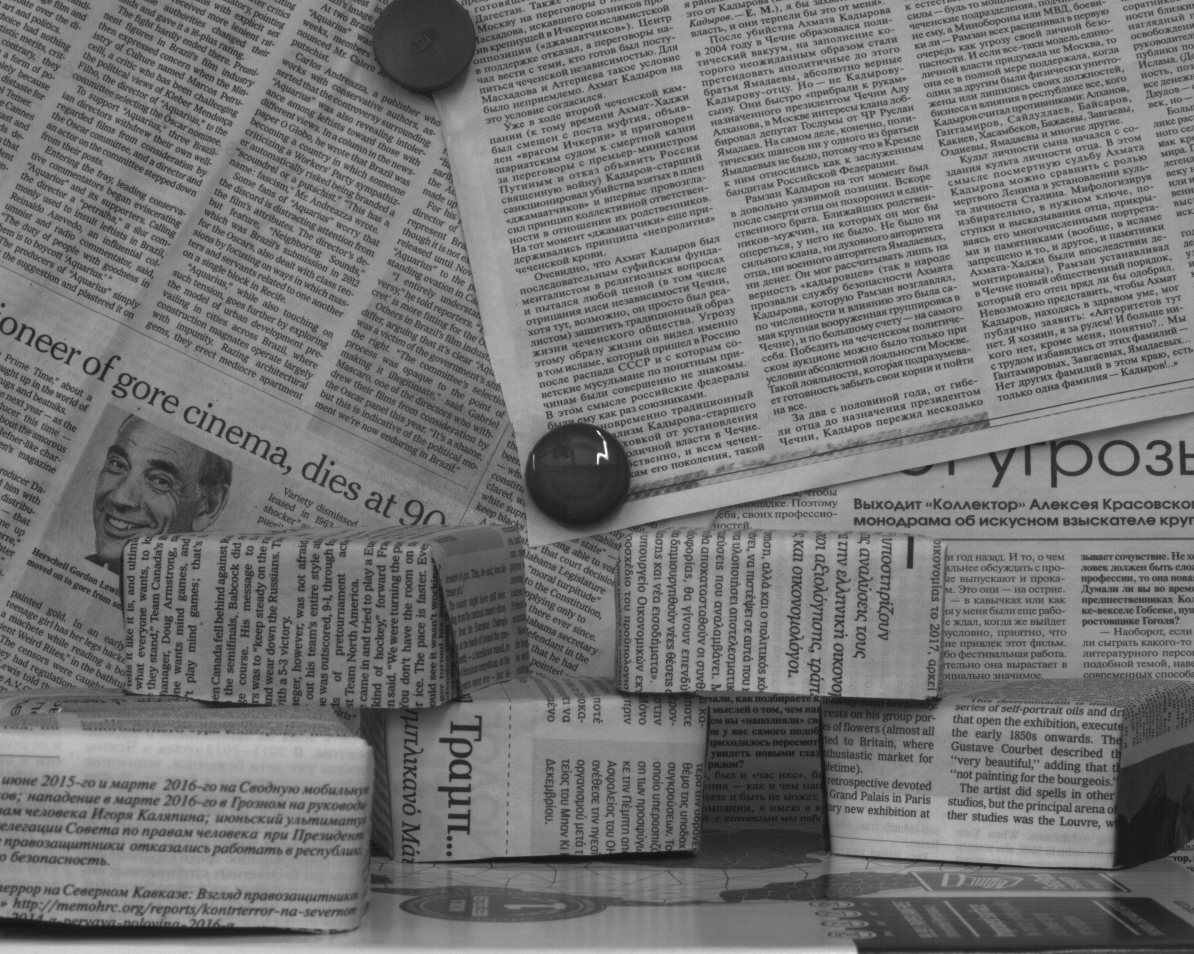}};
			\draw [draw=red, line width=0.3mm] (1.355cm, 0.29cm) rectangle (0.8cm,-0.17cm);
		\end{tikzpicture}\\[0.3ex]
		\includegraphics[trim={32.7cm 14.5cm 1cm 12.5cm},clip,width=1.00\textwidth]
		{./images/newspapers_global_pan_translation_xyz_inlier_bin3_sr0_f11_win01.png}
	\end{minipage}}
	\hfill
	\caption{SR methods under global motion on the \textit{newspapers} dataset ($3 \times$ magnification). Top: SR on raw data. MFSR (\eg, WNUISR and IRWSR) outperforms SISR algorithms (\eg, A+ and VDSR) w.r.t. the recovery of fine structures like text and reconstruction algorithms with sparsity priors (\eg, IRWSR) enhanced the recovery of HR details compared to interpolation-based methods (\eg, WNUISR). Bottom: SR under H.265/HEVC coding (quantization QP30). All methods are affected by compression artifacts and become indistinguishable.}
	\label{fig:globalMotionExample}
\end{figure*}

\begin{figure*}[!t]
	\scriptsize 
	\centering
	\subfloat{%
%
\definecolor{mycolor1}{rgb}{0.00000,0.00000,0.87500}%
\definecolor{mycolor2}{rgb}{0.00000,0.12500,1.00000}%
\definecolor{mycolor3}{rgb}{0.00000,0.25000,1.00000}%
\definecolor{mycolor4}{rgb}{0.00000,0.37500,1.00000}%
\definecolor{mycolor5}{rgb}{0.00000,0.50000,1.00000}%
\definecolor{mycolor6}{rgb}{0.00000,0.62500,1.00000}%
\definecolor{mycolor7}{rgb}{0.00000,0.75000,1.00000}%
\definecolor{mycolor8}{rgb}{1.00000,0.87500,0.00000}%
\definecolor{mycolor9}{rgb}{1.00000,0.75000,0.00000}%
\definecolor{mycolor10}{rgb}{0.87500,0.00000,0.00000}%
\begin{tikzpicture}
\scriptsize
\begin{axis}[%
    const plot,
		hide axis,
    xmin=10,
    xmax=50,
    ymin=0,
    ymax=0,
    legend style={
			legend image code/.code={2mm
				\draw [#1] (0cm,-0.065cm) rectangle (0.25cm,0.065cm);
			},
			anchor=north,
			inner ysep=0pt,
			draw=none, 
			legend cell align=center,
			legend columns=-1, 
			nodes={scale=0.815, transform shape}
		}
	]
	\addlegendimage{black, mark=none, fill=black!25!blue}
	\addlegendentry{BICUBIC}
	\addlegendimage{black, mark=none, fill=mycolor1}
	\addlegendentry{EBSR}
	\addlegendimage{black, mark=none, fill=blue}
	\addlegendentry{ScSR}
	\addlegendimage{black, mark=none, fill=mycolor2}
	\addlegendentry{NBSRF}
	\addlegendimage{black, mark=none, fill=mycolor3}
	\addlegendentry{A+}
	\addlegendimage{black, mark=none, fill=mycolor4}
	\addlegendentry{SRCNN}
	\addlegendimage{black, mark=none, fill=mycolor5}
	\addlegendentry{DRCN}
	\addlegendimage{black, mark=none, fill=mycolor6}
	\addlegendentry{VDSR}
	\addlegendimage{black, mark=none, fill=mycolor7}
	\addlegendentry{SESR}
	\addlegendimage{black, mark=none, fill=mycolor8}
	\addlegendentry{NUISR}
	\addlegendimage{black, mark=none, fill=mycolor9}
	\addlegendentry{WNUISR}
	\addlegendimage{black, mark=none, fill=orange!50!mycolor9}
	\addlegendentry{HYSR}
	\addlegendimage{black, mark=none, fill=orange}
	\addlegendentry{DBRSR}
	\addlegendimage{black, mark=none, fill=red!25!orange}
	\addlegendentry{L1BTV}
	\addlegendimage{black, mark=none, fill=red!50!orange}
	\addlegendentry{BEPSR}
	\addlegendimage{black, mark=none, fill=red!75!orange}
	\addlegendentry{IRWSR}
	\addlegendimage{black, mark=none, fill=red}
	\addlegendentry{BVSR}
	\addlegendimage{black, mark=none, fill=mycolor10}
	\addlegendentry{SRB}
	\addlegendimage{black, mark=none, fill=black!25!red}
	\addlegendentry{VSRnet}
\end{axis}

\end{tikzpicture}%
%
}\\[-1.2ex]
	\setlength \figurewidth{0.435\textwidth}
	\setlength \figureheight{0.9\figurewidth}
	\subfloat{%
%
\definecolor{mycolor1}{rgb}{0.00000,0.00000,0.87500}%
\definecolor{mycolor2}{rgb}{0.00000,0.12500,1.00000}%
\definecolor{mycolor3}{rgb}{0.00000,0.25000,1.00000}%
\definecolor{mycolor4}{rgb}{0.00000,0.37500,1.00000}%
\definecolor{mycolor5}{rgb}{0.00000,0.50000,1.00000}%
\definecolor{mycolor6}{rgb}{0.00000,0.62500,1.00000}%
\definecolor{mycolor7}{rgb}{0.00000,0.75000,1.00000}%
\definecolor{mycolor8}{rgb}{1.00000,0.87500,0.00000}%
\definecolor{mycolor9}{rgb}{1.00000,0.75000,0.00000}%
\definecolor{mycolor10}{rgb}{0.87500,0.00000,0.00000}%
\begin{tikzpicture}

\begin{axis}[%
width=\figurewidth,
height=0.417\figureheight,
at={(0\figurewidth,0\figureheight)},
scale only axis,
xmin=0.5,
xmax=5.5,
xtick={1,2,3,4,5,6,7,8,9,10,11,12,13,14,15,16,17,18,19,20,21,22,23,24,25,26,27,28,29,30,31,32,33,34,35,36,37,38,39,40,41,42,43,44,45,46,47,48,49,50,51,52,53,54,55,56,57,58,59,60,61,62,63,64,65,66,67,68,69,70,71,72,73,74,75,76,77,78,79,80,81,82,83,84,85,86,87,88,89,90,91,92,93,94,95},
xticklabels={{Uncoded},{QP10},{QP20},{QP30},{QP40}},
ymin=-0.035019877298227,
ymax=0.16283506152712,
ylabel={Normalized PSNR},
axis background/.style={fill=white},
xmajorgrids,
ymajorgrids,
xlabel near ticks,ylabel near ticks,scaled y ticks=false,yticklabel style={/pgf/number format/fixed, /pgf/number format/precision=2},
]
\addplot[ybar,bar width=0.034,bar shift=-0.379,fill=black!25!blue,draw=black,area legend] plot table[row sep=crcr] {%
1	0.0688977897971624\\
2	0.068621755464783\\
3	0.0647290831190001\\
4	0.0505814293632146\\
5	0.0251391866537291\\
};
\addplot[forget plot,color=white!15!black] table[row sep=crcr] {%
0.5	0\\
5.5	0\\
};
\addplot[ybar,bar width=0.034,bar shift=-0.337,fill=mycolor1,draw=black,area legend] plot table[row sep=crcr] {%
1	0.0950951868249174\\
2	0.0945678016583107\\
3	0.0908612637035567\\
4	0.0690328198449928\\
5	0.0301811449869154\\
};
\addplot[forget plot,color=white!15!black] table[row sep=crcr] {%
0.5	0\\
5.5	0\\
};
\addplot[ybar,bar width=0.034,bar shift=-0.295,fill=blue,draw=black,area legend] plot table[row sep=crcr] {%
1	0.064965618316259\\
2	0.06463503595293\\
3	0.0673724739782594\\
4	0.055050387778437\\
5	0.0251074766295973\\
};
\addplot[forget plot,color=white!15!black] table[row sep=crcr] {%
0.5	0\\
5.5	0\\
};
\addplot[ybar,bar width=0.034,bar shift=-0.253,fill=mycolor2,draw=black,area legend] plot table[row sep=crcr] {%
1	0.101435846391175\\
2	0.100814999683849\\
3	0.0981164282811699\\
4	0.0730393301582545\\
5	0.0302529210879248\\
};
\addplot[forget plot,color=white!15!black] table[row sep=crcr] {%
0.5	0\\
5.5	0\\
};
\addplot[ybar,bar width=0.034,bar shift=-0.211,fill=mycolor3,draw=black,area legend] plot table[row sep=crcr] {%
1	0.0981907952271786\\
2	0.0976531390815574\\
3	0.0948288417975863\\
4	0.0709004400985789\\
5	0.0299102028643868\\
};
\addplot[forget plot,color=white!15!black] table[row sep=crcr] {%
0.5	0\\
5.5	0\\
};
\addplot[ybar,bar width=0.034,bar shift=-0.168,fill=mycolor4,draw=black,area legend] plot table[row sep=crcr] {%
1	0.0957515596483877\\
2	0.0952441695286421\\
3	0.0930785959634183\\
4	0.0701559024725587\\
5	0.0289830196127223\\
};
\addplot[forget plot,color=white!15!black] table[row sep=crcr] {%
0.5	0\\
5.5	0\\
};
\addplot[ybar,bar width=0.034,bar shift=-0.126,fill=mycolor5,draw=black,area legend] plot table[row sep=crcr] {%
1	0.102241010845575\\
2	0.101826869631998\\
3	0.0986866725769136\\
4	0.0701930146497588\\
5	0.0270381624477082\\
};
\addplot[forget plot,color=white!15!black] table[row sep=crcr] {%
0.5	0\\
5.5	0\\
};
\addplot[ybar,bar width=0.034,bar shift=-0.084,fill=mycolor6,draw=black,area legend] plot table[row sep=crcr] {%
1	0.0953592446200996\\
2	0.0947743970323944\\
3	0.0919245036116485\\
4	0.0650483893195888\\
5	0.0245899027401814\\
};
\addplot[forget plot,color=white!15!black] table[row sep=crcr] {%
0.5	0\\
5.5	0\\
};
\addplot[ybar,bar width=0.034,bar shift=-0.042,fill=mycolor7,draw=black,area legend] plot table[row sep=crcr] {%
1	0.0804783785864329\\
2	0.0805784841605046\\
3	0.080562251666323\\
4	0.0631386922554249\\
5	0.0277220861444433\\
};
\addplot[forget plot,color=white!15!black] table[row sep=crcr] {%
0.5	0\\
5.5	0\\
};
\addplot[ybar,bar width=0.034,bar shift=-0,fill=mycolor8,draw=black,area legend] plot table[row sep=crcr] {%
1	0.15309257134379\\
2	0.149693873325847\\
3	0.122799313016274\\
4	0.0704267387035089\\
5	0.0290768508457694\\
};
\addplot[forget plot,color=white!15!black] table[row sep=crcr] {%
0.5	0\\
5.5	0\\
};
\addplot[ybar,bar width=0.034,bar shift=0.042,fill=mycolor9,draw=black,area legend] plot table[row sep=crcr] {%
1	0.135281564150254\\
2	0.132431230700902\\
3	0.110292607417784\\
4	0.0651863836559508\\
5	0.0274239210547701\\
};
\addplot[forget plot,color=white!15!black] table[row sep=crcr] {%
0.5	0\\
5.5	0\\
};
\addplot[ybar,bar width=0.034,bar shift=0.084,fill=orange!50!mycolor9,draw=black,area legend] plot table[row sep=crcr] {%
1	0.155081010978209\\
2	0.151269817514056\\
3	0.125689927175014\\
4	0.074460537461985\\
5	0.0308285557130555\\
};
\addplot[forget plot,color=white!15!black] table[row sep=crcr] {%
0.5	0\\
5.5	0\\
};
\addplot[ybar,bar width=0.034,bar shift=0.126,fill=orange,draw=black,area legend] plot table[row sep=crcr] {%
1	0.139907913182175\\
2	0.13678981427726\\
3	0.113438079348243\\
4	0.0668431510847081\\
5	0.0281784218815737\\
};
\addplot[forget plot,color=white!15!black] table[row sep=crcr] {%
0.5	0\\
5.5	0\\
};
\addplot[ybar,bar width=0.034,bar shift=0.168,fill=red!25!orange,draw=black,area legend] plot table[row sep=crcr] {%
1	0.12767382683248\\
2	0.127073543409151\\
3	0.111801030497384\\
4	0.0645135327123174\\
5	0.027043351160388\\
};
\addplot[forget plot,color=white!15!black] table[row sep=crcr] {%
0.5	0\\
5.5	0\\
};
\addplot[ybar,bar width=0.034,bar shift=0.211,fill=red!50!orange,draw=black,area legend] plot table[row sep=crcr] {%
1	0.14714650549545\\
2	0.146138930234385\\
3	0.12338754201876\\
4	0.0661126733005126\\
5	0.0261574938209656\\
};
\addplot[forget plot,color=white!15!black] table[row sep=crcr] {%
0.5	0\\
5.5	0\\
};
\addplot[ybar,bar width=0.034,bar shift=0.253,fill=red!75!orange,draw=black,area legend] plot table[row sep=crcr] {%
1	0.0988010774895904\\
2	0.0959046687666738\\
3	0.102922758616113\\
4	0.0554201631021667\\
5	0.0231419964914075\\
};
\addplot[forget plot,color=white!15!black] table[row sep=crcr] {%
0.5	0\\
5.5	0\\
};
\addplot[ybar,bar width=0.034,bar shift=0.295,fill=red,draw=black,area legend] plot table[row sep=crcr] {%
1	-0.00605916698241661\\
2	-0.00776421005142588\\
3	-0.0321684443108196\\
4	-0.0252672420274612\\
5	-0.0125999177742886\\
};
\addplot[forget plot,color=white!15!black] table[row sep=crcr] {%
0.5	0\\
5.5	0\\
};
\addplot[ybar,bar width=0.034,bar shift=0.337,fill=mycolor10,draw=black,area legend] plot table[row sep=crcr] {%
1	0.00957212626732039\\
2	0.0882177925426464\\
3	0.081011289827219\\
4	0.0525058760669765\\
5	0.0205693501396694\\
};
\addplot[forget plot,color=white!15!black] table[row sep=crcr] {%
0.5	0\\
5.5	0\\
};
\addplot[ybar,bar width=0.034,bar shift=0.379,fill=black!25!red,draw=black,area legend] plot table[row sep=crcr] {%
1	0.113763868011629\\
2	0.112076402287431\\
3	0.104297459027569\\
4	0.0728606915943377\\
5	0.0291652761948736\\
};
\addplot[forget plot,color=white!15!black] table[row sep=crcr] {%
0.5	0\\
5.5	0\\
};
\end{axis}
\end{tikzpicture}%
%
}\quad
	\subfloat{%
%
\definecolor{mycolor1}{rgb}{0.00000,0.00000,0.87500}%
\definecolor{mycolor2}{rgb}{0.00000,0.12500,1.00000}%
\definecolor{mycolor3}{rgb}{0.00000,0.25000,1.00000}%
\definecolor{mycolor4}{rgb}{0.00000,0.37500,1.00000}%
\definecolor{mycolor5}{rgb}{0.00000,0.50000,1.00000}%
\definecolor{mycolor6}{rgb}{0.00000,0.62500,1.00000}%
\definecolor{mycolor7}{rgb}{0.00000,0.75000,1.00000}%
\definecolor{mycolor8}{rgb}{1.00000,0.87500,0.00000}%
\definecolor{mycolor9}{rgb}{1.00000,0.75000,0.00000}%
\definecolor{mycolor10}{rgb}{0.87500,0.00000,0.00000}%
\begin{tikzpicture}

\begin{axis}[%
width=\figurewidth,
height=0.417\figureheight,
at={(0\figurewidth,0\figureheight)},
scale only axis,
xmin=0.5,
xmax=5.5,
xtick={1,2,3,4,5,6,7,8,9,10,11,12,13,14,15,16,17,18,19,20,21,22,23,24,25,26,27,28,29,30,31,32,33,34,35,36,37,38,39,40,41,42,43,44,45,46,47,48,49,50,51,52,53,54,55,56,57,58,59,60,61,62,63,64,65,66,67,68,69,70,71,72,73,74,75,76,77,78,79,80,81,82,83,84,85,86,87,88,89,90,91,92,93,94,95},
xticklabels={{Uncoded},{QP10},{QP20},{QP30},{QP40}},
ymin=-0.206484191790332,
ymax=0.80,
ytick={0, 0.25, 0.5, 0.75},ylabel={Normalized IFC},
axis background/.style={fill=white},
xmajorgrids,
ymajorgrids,
xlabel near ticks,ylabel near ticks,scaled y ticks=false,yticklabel style={/pgf/number format/fixed, /pgf/number format/precision=2},
]
\addplot[ybar,bar width=0.034,bar shift=-0.379,fill=black!25!blue,draw=black,area legend] plot table[row sep=crcr] {%
1	0.148173845995369\\
2	0.151528156810247\\
3	0.164164794515769\\
4	0.169614987664301\\
5	0.16619792270334\\
};
\addplot[forget plot,color=white!15!black] table[row sep=crcr] {%
0.5	0\\
5.5	0\\
};
\addplot[ybar,bar width=0.034,bar shift=-0.337,fill=mycolor1,draw=black,area legend] plot table[row sep=crcr] {%
1	0.219332210019973\\
2	0.224627358725964\\
3	0.246275985943316\\
4	0.226961252383757\\
5	0.181809584199027\\
};
\addplot[forget plot,color=white!15!black] table[row sep=crcr] {%
0.5	0\\
5.5	0\\
};
\addplot[ybar,bar width=0.034,bar shift=-0.295,fill=blue,draw=black,area legend] plot table[row sep=crcr] {%
1	0.149231353321125\\
2	0.155037199501981\\
3	0.190872912205852\\
4	0.17893776914489\\
5	0.135294428300503\\
};
\addplot[forget plot,color=white!15!black] table[row sep=crcr] {%
0.5	0\\
5.5	0\\
};
\addplot[ybar,bar width=0.034,bar shift=-0.253,fill=mycolor2,draw=black,area legend] plot table[row sep=crcr] {%
1	0.236792006179539\\
2	0.242516067314353\\
3	0.270952479818739\\
4	0.236660038763597\\
5	0.169578403913959\\
};
\addplot[forget plot,color=white!15!black] table[row sep=crcr] {%
0.5	0\\
5.5	0\\
};
\addplot[ybar,bar width=0.034,bar shift=-0.211,fill=mycolor3,draw=black,area legend] plot table[row sep=crcr] {%
1	0.237346270119672\\
2	0.243239018155325\\
3	0.268414800712262\\
4	0.235079585195609\\
5	0.171147465520023\\
};
\addplot[forget plot,color=white!15!black] table[row sep=crcr] {%
0.5	0\\
5.5	0\\
};
\addplot[ybar,bar width=0.034,bar shift=-0.168,fill=mycolor4,draw=black,area legend] plot table[row sep=crcr] {%
1	0.216263343623866\\
2	0.222161555894454\\
3	0.252386262270459\\
4	0.228288698177313\\
5	0.166410720046937\\
};
\addplot[forget plot,color=white!15!black] table[row sep=crcr] {%
0.5	0\\
5.5	0\\
};
\addplot[ybar,bar width=0.034,bar shift=-0.126,fill=mycolor5,draw=black,area legend] plot table[row sep=crcr] {%
1	0.253984216258468\\
2	0.260430022878933\\
3	0.289581893081499\\
4	0.237865435701268\\
5	0.16260681780421\\
};
\addplot[forget plot,color=white!15!black] table[row sep=crcr] {%
0.5	0\\
5.5	0\\
};
\addplot[ybar,bar width=0.034,bar shift=-0.084,fill=mycolor6,draw=black,area legend] plot table[row sep=crcr] {%
1	0.244877714863301\\
2	0.250898438438133\\
3	0.277979606615\\
4	0.225382885134155\\
5	0.150023401027994\\
};
\addplot[forget plot,color=white!15!black] table[row sep=crcr] {%
0.5	0\\
5.5	0\\
};
\addplot[ybar,bar width=0.034,bar shift=-0.042,fill=mycolor7,draw=black,area legend] plot table[row sep=crcr] {%
1	0.17710915438815\\
2	0.184095355206511\\
3	0.221791284705141\\
4	0.202662166557041\\
5	0.153941418897477\\
};
\addplot[forget plot,color=white!15!black] table[row sep=crcr] {%
0.5	0\\
5.5	0\\
};
\addplot[ybar,bar width=0.034,bar shift=-0,fill=mycolor8,draw=black,area legend] plot table[row sep=crcr] {%
1	0.689751249424748\\
2	0.635807545064766\\
3	0.423061696341859\\
4	0.221311516908517\\
5	0.15219072906096\\
};
\addplot[forget plot,color=white!15!black] table[row sep=crcr] {%
0.5	0\\
5.5	0\\
};
\addplot[ybar,bar width=0.034,bar shift=0.042,fill=mycolor9,draw=black,area legend] plot table[row sep=crcr] {%
1	0.583504450296663\\
2	0.537966220960295\\
3	0.366138363643686\\
4	0.197062082971678\\
5	0.136541384375839\\
};
\addplot[forget plot,color=white!15!black] table[row sep=crcr] {%
0.5	0\\
5.5	0\\
};
\addplot[ybar,bar width=0.034,bar shift=0.084,fill=orange!50!mycolor9,draw=black,area legend] plot table[row sep=crcr] {%
1	0.686629467507338\\
2	0.624933835096273\\
3	0.425243539971137\\
4	0.238812530939047\\
5	0.169590996539069\\
};
\addplot[forget plot,color=white!15!black] table[row sep=crcr] {%
0.5	0\\
5.5	0\\
};
\addplot[ybar,bar width=0.034,bar shift=0.126,fill=orange,draw=black,area legend] plot table[row sep=crcr] {%
1	0.58307009807509\\
2	0.541350452119997\\
3	0.375251497539681\\
4	0.206165300622883\\
5	0.146870115247453\\
};
\addplot[forget plot,color=white!15!black] table[row sep=crcr] {%
0.5	0\\
5.5	0\\
};
\addplot[ybar,bar width=0.034,bar shift=0.168,fill=red!25!orange,draw=black,area legend] plot table[row sep=crcr] {%
1	0.568662996978401\\
2	0.535184511388021\\
3	0.368285923268515\\
4	0.17919674360818\\
5	0.12916962432113\\
};
\addplot[forget plot,color=white!15!black] table[row sep=crcr] {%
0.5	0\\
5.5	0\\
};
\addplot[ybar,bar width=0.034,bar shift=0.211,fill=red!50!orange,draw=black,area legend] plot table[row sep=crcr] {%
1	0.707200155060589\\
2	0.648498275492632\\
3	0.433361737109824\\
4	0.202964172171971\\
5	0.131299522627744\\
};
\addplot[forget plot,color=white!15!black] table[row sep=crcr] {%
0.5	0\\
5.5	0\\
};
\addplot[ybar,bar width=0.034,bar shift=0.253,fill=red!75!orange,draw=black,area legend] plot table[row sep=crcr] {%
1	0.686209124104792\\
2	0.606603726068993\\
3	0.42752482206729\\
4	0.17683799761368\\
5	0.110980875655669\\
};
\addplot[forget plot,color=white!15!black] table[row sep=crcr] {%
0.5	0\\
5.5	0\\
};
\addplot[ybar,bar width=0.034,bar shift=0.295,fill=red,draw=black,area legend] plot table[row sep=crcr] {%
1	-0.037194829934054\\
2	-0.086186004959773\\
3	-0.205066918234221\\
4	-0.11466761520673\\
5	0.0244751508120396\\
};
\addplot[forget plot,color=white!15!black] table[row sep=crcr] {%
0.5	0\\
5.5	0\\
};
\addplot[ybar,bar width=0.034,bar shift=0.337,fill=mycolor10,draw=black,area legend] plot table[row sep=crcr] {%
1	0.265565354223431\\
2	0.273995776653354\\
3	0.243648895124791\\
4	0.187192719429534\\
5	0.164885330255514\\
};
\addplot[forget plot,color=white!15!black] table[row sep=crcr] {%
0.5	0\\
5.5	0\\
};
\addplot[ybar,bar width=0.034,bar shift=0.379,fill=black!25!red,draw=black,area legend] plot table[row sep=crcr] {%
1	0.354536605613016\\
2	0.339043603490803\\
3	0.316721002911014\\
4	0.247378085673451\\
5	0.17623076464922\\
};
\addplot[forget plot,color=white!15!black] table[row sep=crcr] {%
0.5	0\\
5.5	0\\
};
\end{axis}
\end{tikzpicture}%
%
}
	\caption{Robustness analysis of SR \wrt video compression. The $x$-axis denotes the compression level in terms of H.265/HEVC quantization. The $y$-axis depicts the normalized PSNR and IFC averaged over 14 scenes with global motion and $3\times$ magnification. The individual algorithms are categorized either as SISR (shown with blue color map) or MFSR (shown with red color map).}
	\label{fig:qualityMeasuresCompression}
\end{figure*}
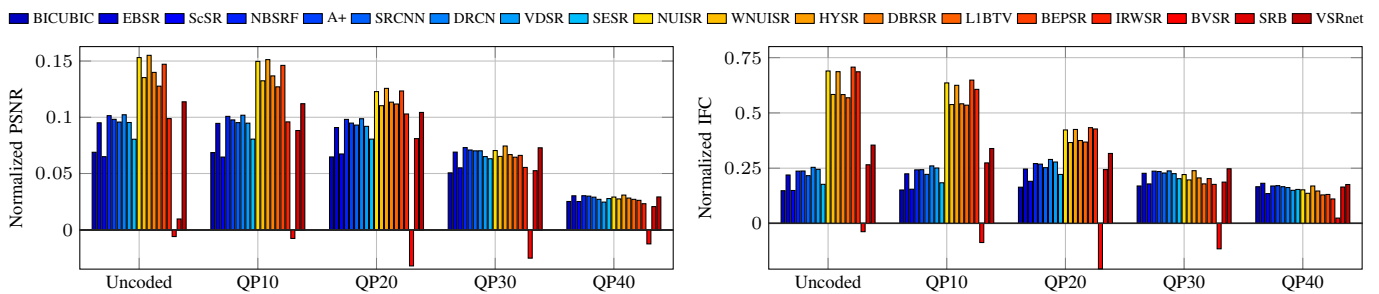

We used published reference implementations if available. For L1BTV and BEPSR, we used the publicly available MATLAB SR 
toolbox \cite{Kohler2015c}. For NUISR, we adopted the method in \cite{Batz2015}. 
BVSR is based on source code provided by the authors 
of \cite{Kappeler2016}. All learning-based methods use their original pretrained 
models wherever possible. NBSRF and ScSR were 
retrained for $3\times$ and $4\times$ magnification on the original training 
data as pretrained models were unavailable. For VSRnet, we used the pretrained network for $K = 5$ frames for all magnifications. We 
selected free parameters following the guidelines in the source codes. For methods that require knowledge on the 
camera PSF, an isotropic Gaussian kernel of size $\lceil 6 \sigma_{\text{PSF}} \rceil \times \lceil 6 
\sigma_{\text{PSF}} \rceil$ pixels was used, where $\sigma_{\text{PSF}} = b \sigma_0$, $b$ is the binning factor, 
and $\sigma_0 = 0.4$ is the standard deviation on the LR grid.

\section{Quantitative Study}
\label{sec:QuantitativeStudy}

In this section, we quantitatively evaluate SR on our database. This benchmark aims at evaluating the fidelity of SR images \wrt the ground truth data using normalized full-reference quality measures. Results for other measures including no-reference assessment for a perceptual benchmark are provided on our project website. We analyze different motion and environmental conditions as well as video compression.

\subsection{Super-Resolution on Static Scenes}

\Fref{fig:srBenchmarkMotionTypes:global} shows a comparison of the SR methods with different magnification factors on the global motion datasets using the normalized PSNR and IFC. Overall, we found that relative performances of algorithms depend on the utilized quality measure. Except for large magnifications,
interpolation-based MFSR (NUISR, HYSR) performed best in terms of PSNR.
In case of IFC, reconstruction-based methods (BEPSR, IRWSR) achieved better
scores. This can be explained by two properties. 1) PSNR
weighs deviations to the ground truth in homogeneous and textured regions
uniformly. We observed that PSNR tends to prefer slightly oversmoothed
images, which is consistent with evaluations of full-reference quality
assessment \cite{Sheikh2006}. As interpolation-based SR tends to introduce
blur, these methods are ranked higher by
PSNR. 2) IFC puts the emphasis on high frequencies
\cite{Yang2014a}. Reconstruction-based SR use statistical priors on
natural images, \eg sparsity~\cite{Kohler2015c, Farsiu2004a}, which leads to a
better recovery of high frequencies and thus a higher IFC score.

Interestingly, blind SR (SRB, BVSR) did not perform better than computationally
more efficient non-blind methods. SRB was prone to ringing artifacts while BVSR was affected by oversmoothing due to inaccurate
PSF estimation, which partly led to negative normalized measures.
\Fref{fig:globalMotionExample} (top) shows a comparison among different
methodologies. Here, the sparsity priors contributed to the recovery of
the printed text.

The benchmark also reveals that under pure global motion MFSR outperforms SISR. 
This is because MFSR exploits complementary
information across multiple images to recover HR details, while SISR "hallucinates" these details. 

In SISR, it is worth noting that the use of external data outperformed the self-exemplar approach (SESR). However, even some of the sophisticated deep nets (VDSR, DRCN) performed only comparable to classical shallow architectures. This is quite surprising as results on simulated data \cite{Kim2016, Kim2016a} indicate that deep learning surpasses such classical approaches. It is worth noting that similar benefits over sophisticated techniques have also been observed for computationally cheap learning-based filters like RAISR \cite{Romano2016} running on cell phones with limited resources. In MFSR, interpolation-based algorithms were suitable for small magnification ($2\times$) while reconstruction and deep learning approaches (VSRnet) performed better for larger factors ($3\times$ and $4\times$). We explain this behavior by the use of statistical priors or the use of LR/HR exemplars, which guide the recovery of fine structures.

\subsection{Super-Resolution under Video Compression}

\begin{figure*}[!t]
	\flushleft
	\subfloat[A+ \cite{Timofte2015}]{
	\begin{minipage}{0.158\textwidth} 
		\includegraphics[trim={27.7cm 2.0cm 2cm 21.75cm},clip,width=1.00\textwidth]
		{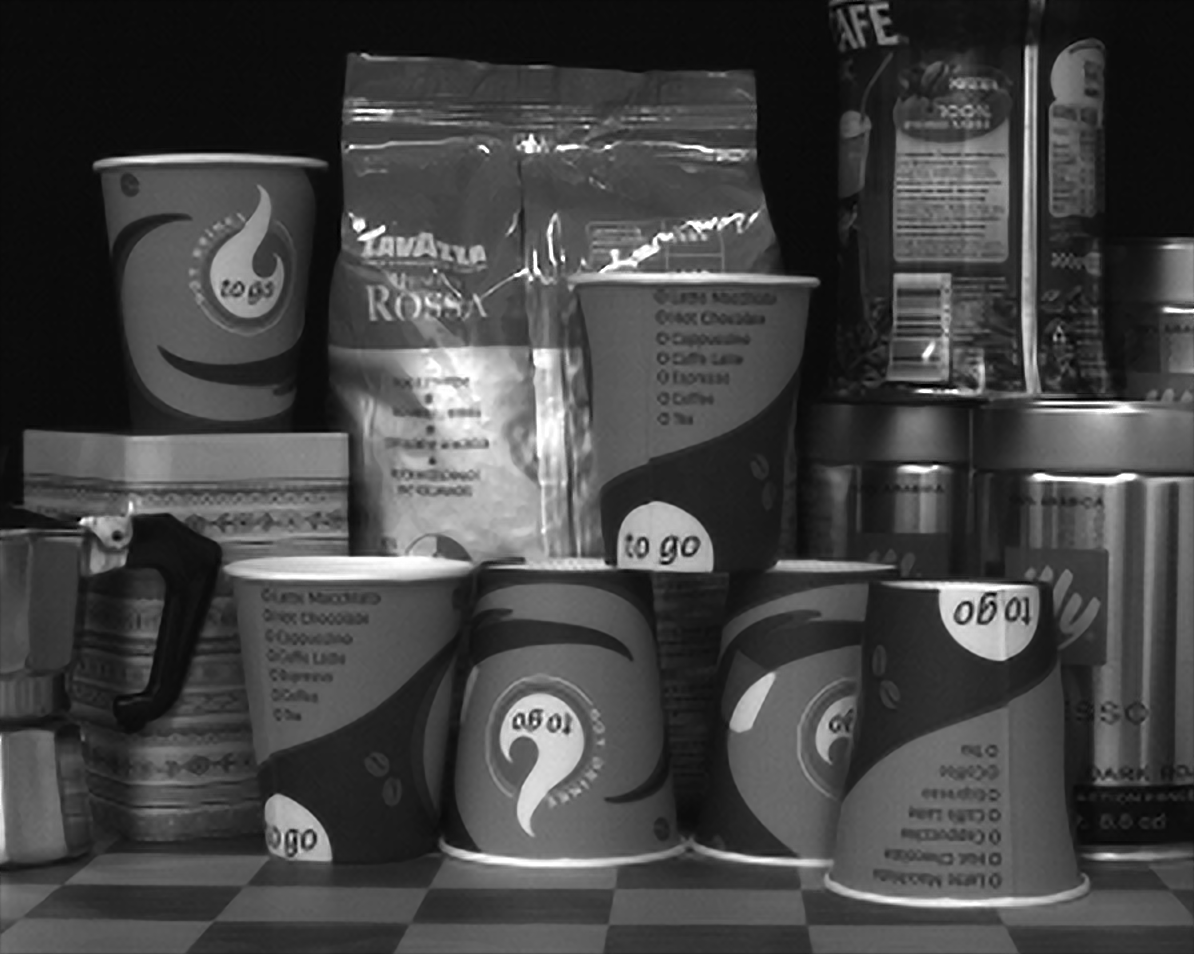}\\[0.4ex]
		\includegraphics[trim={20.7cm 4.0cm 13cm 23.0cm},clip,width=1.00\textwidth]
		{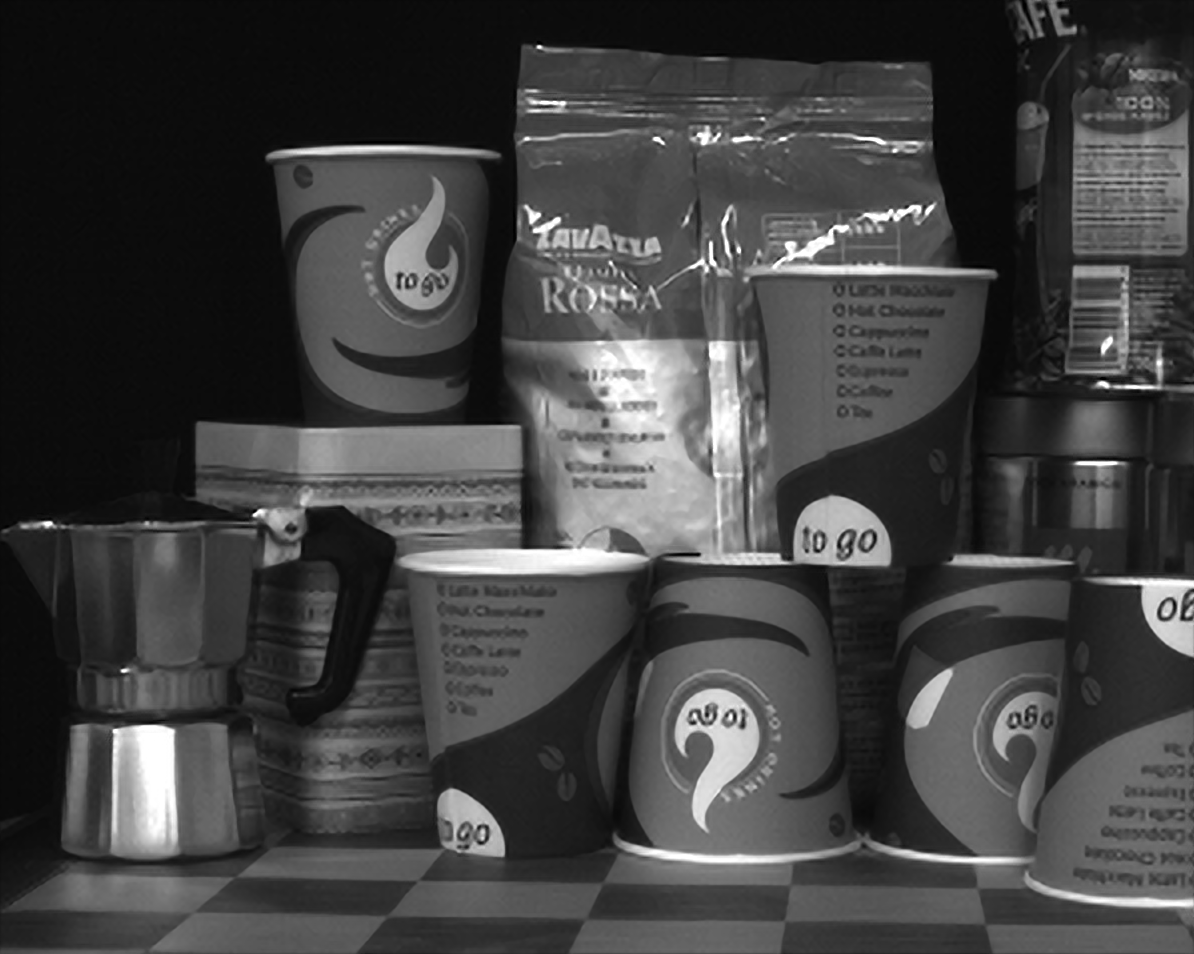}
	\end{minipage}}
	\subfloat[NUISR \cite{Batz2016}]{
	\begin{minipage}{0.158\textwidth} 
		\includegraphics[trim={27.7cm 2.0cm 2cm 21.75cm},clip,width=1.00\textwidth]
		{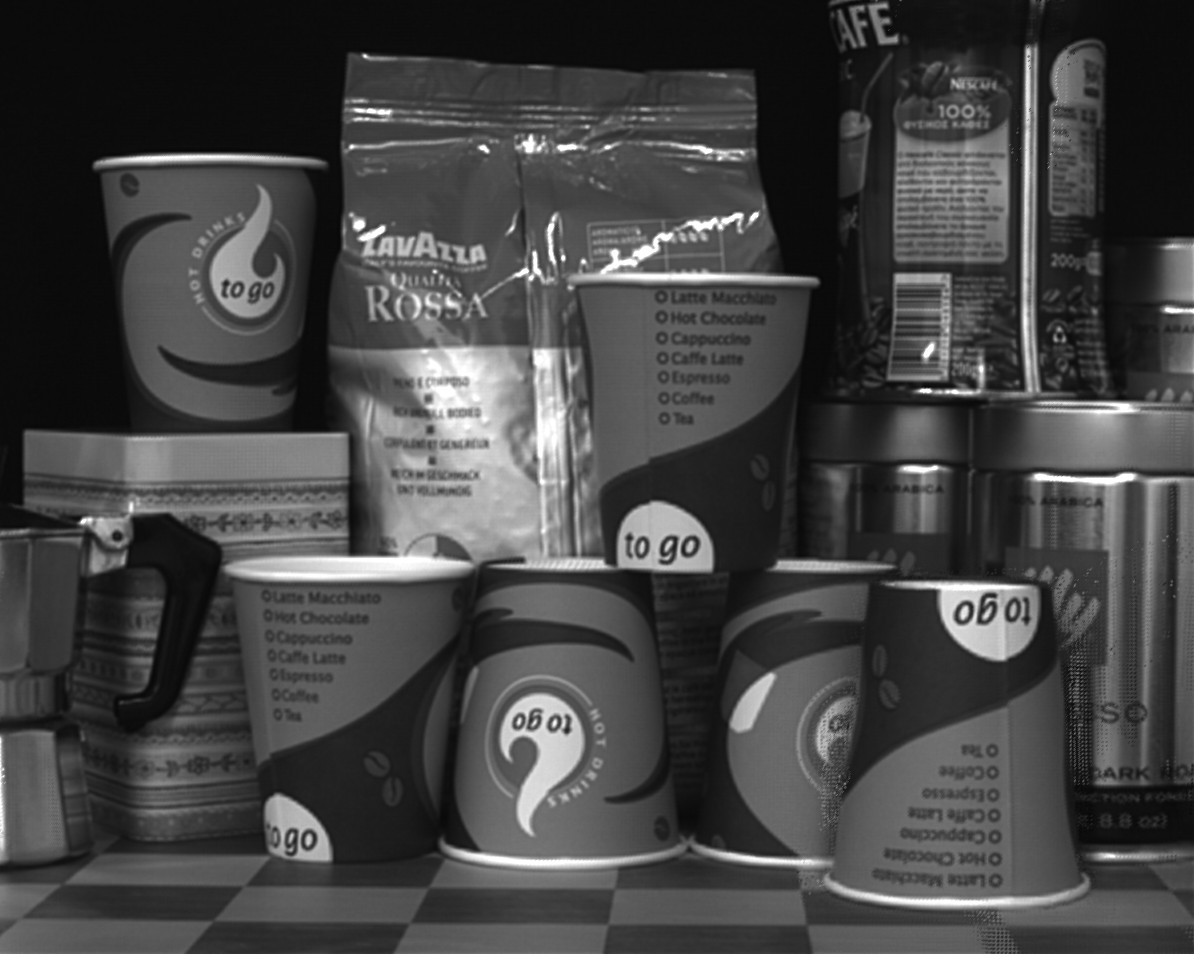}\\[0.4ex]
		\includegraphics[trim={20.7cm 4.0cm 13cm 23.0cm},clip,width=1.00\textwidth]
		{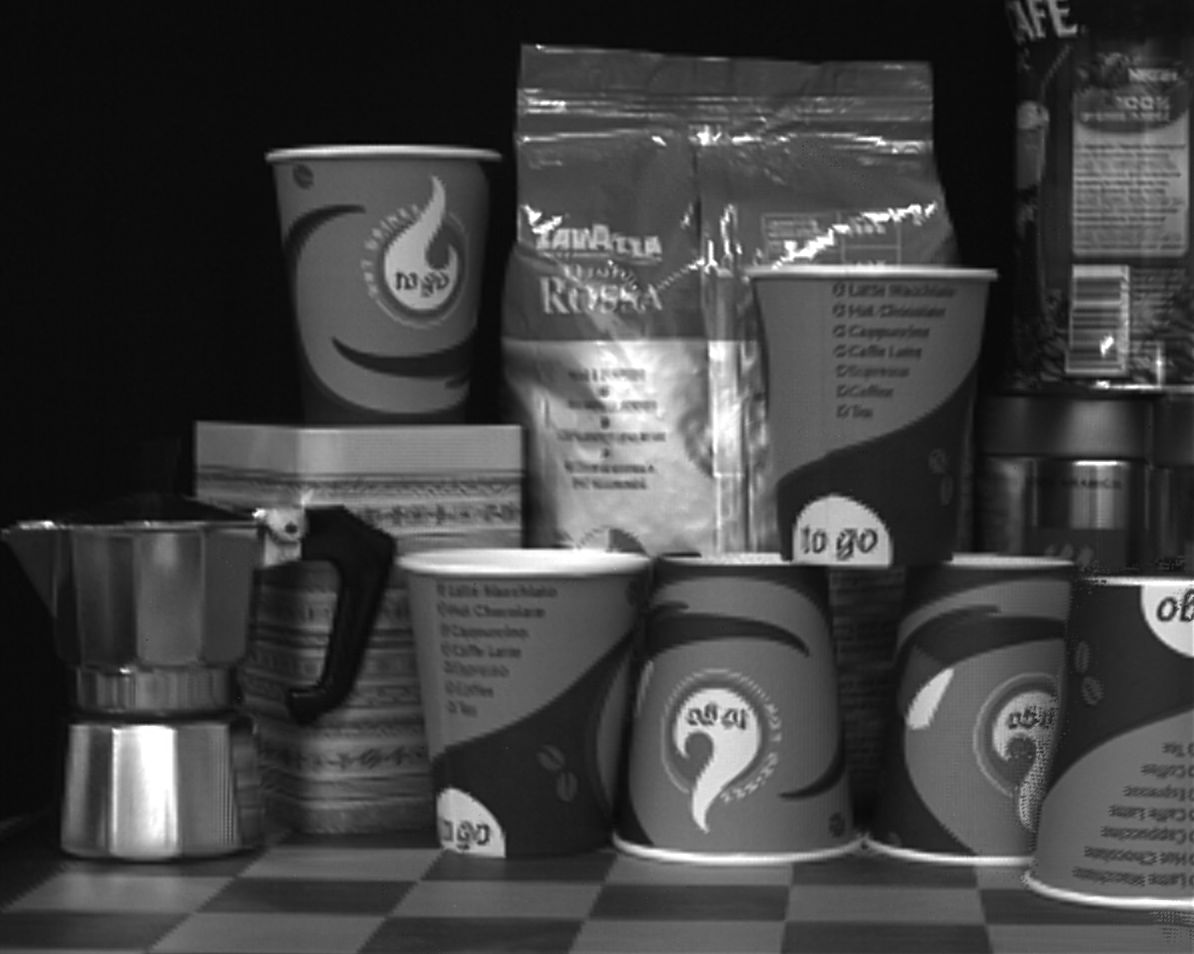}
	\end{minipage}}
	\subfloat[IRWSR \cite{Kohler2015c}]{
	\begin{minipage}{0.158\textwidth} 
		\includegraphics[trim={27.7cm 2.0cm 2cm 21.75cm},clip,width=1.00\textwidth]
		{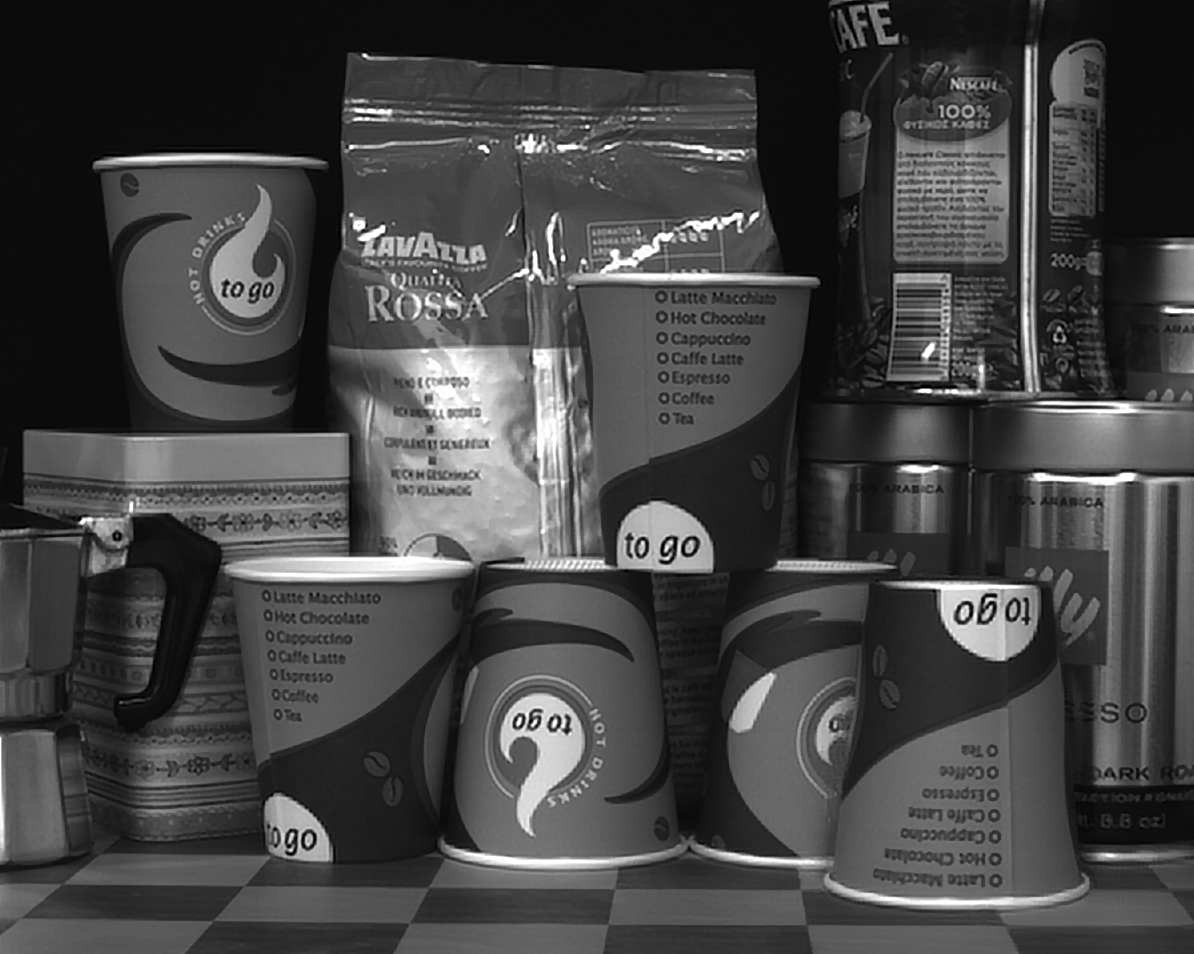}\\[0.4ex]
		\includegraphics[trim={20.7cm 4.0cm 13cm 23.0cm},clip,width=1.00\textwidth]
		{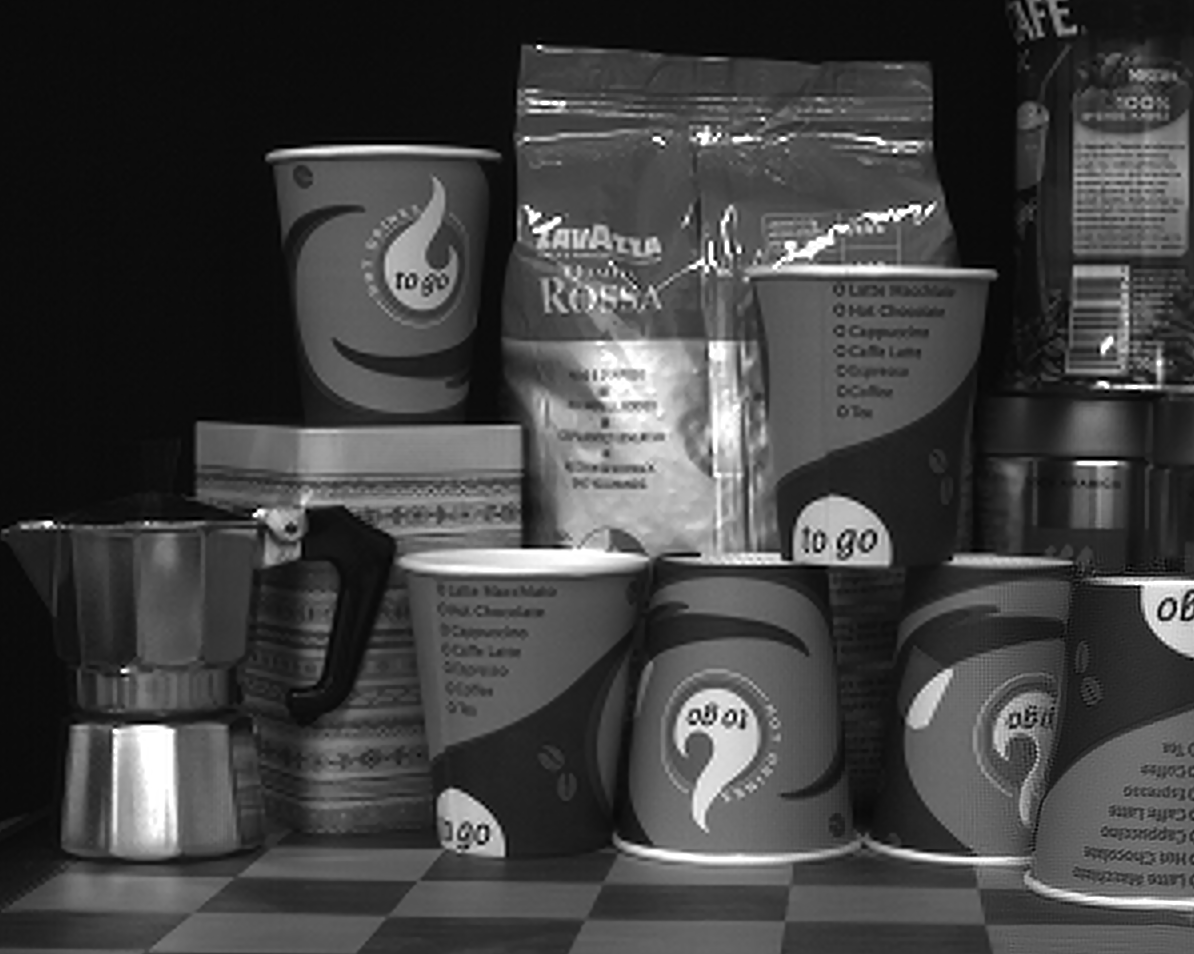}
	\end{minipage}}
	\subfloat[VSRnet \cite{Kappeler2016}]{
	\begin{minipage}{0.158\textwidth} 
		\includegraphics[trim={27.7cm 2.0cm 2cm 21.75cm},clip,width=1.00\textwidth]
		{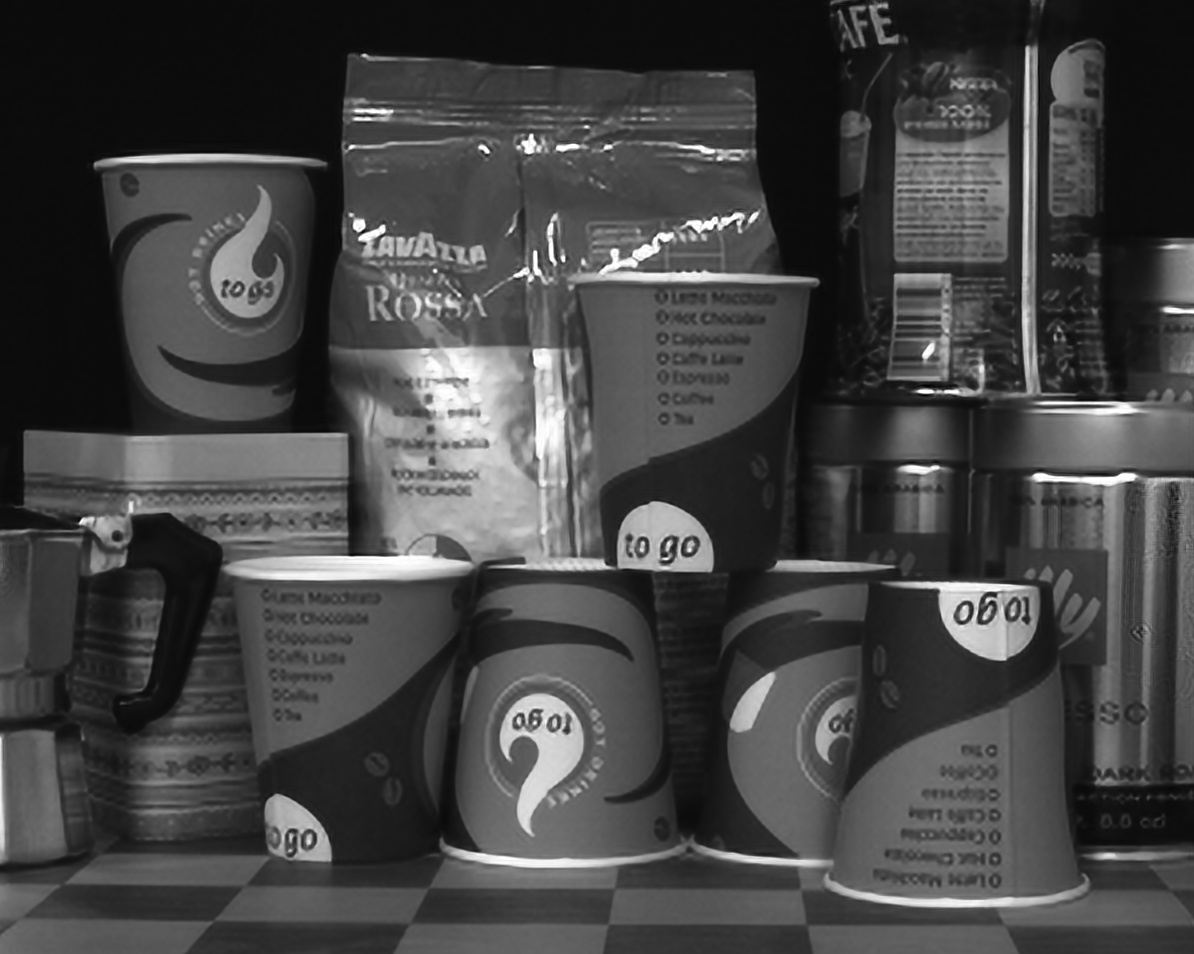}\\[0.4ex]
		\includegraphics[trim={20.7cm 4.0cm 13cm 23.0cm},clip,width=1.00\textwidth]
		{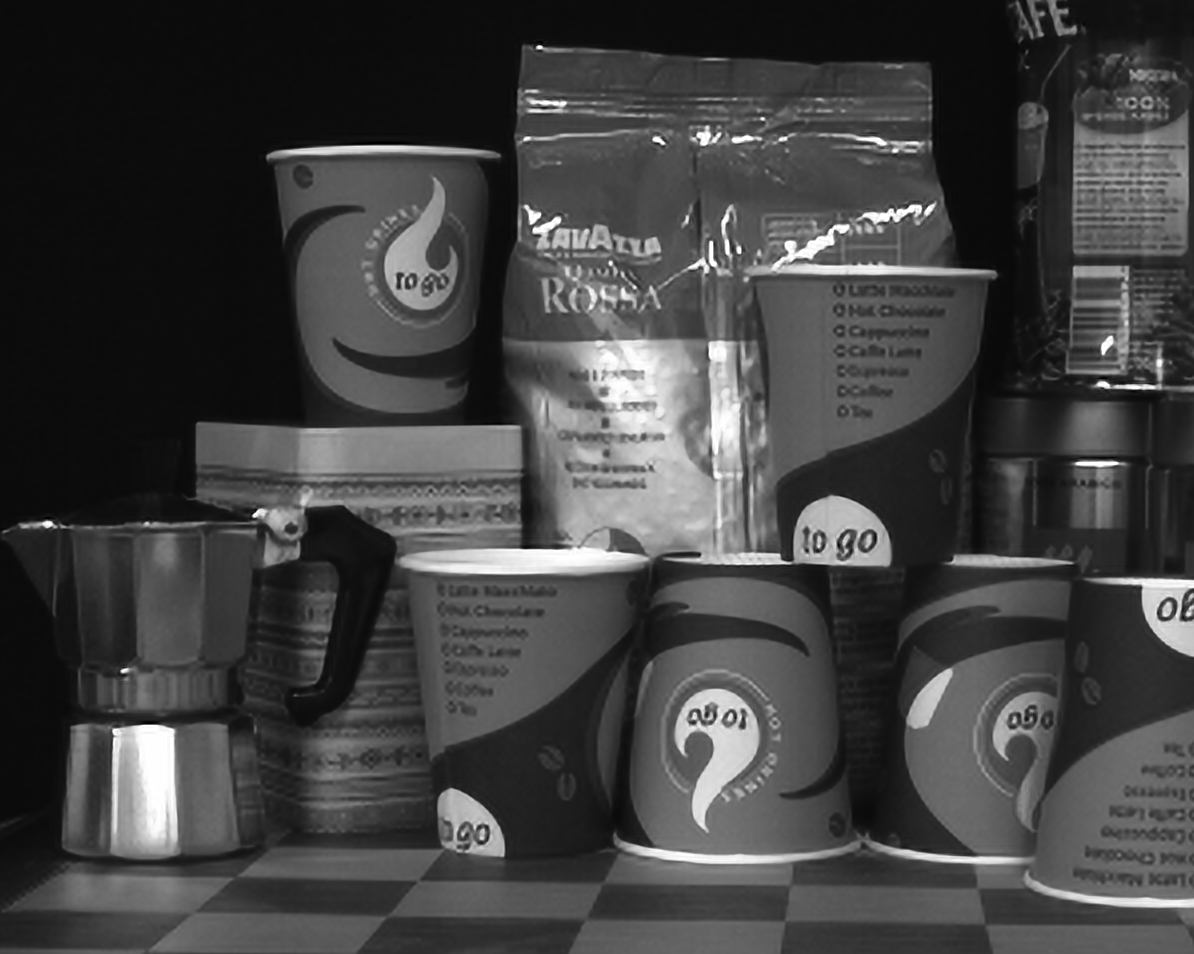}
	\end{minipage}}
	~\vline
	\subfloat[Ground truth]{
	\begin{minipage}{0.158\textwidth} 
		\includegraphics[trim={27.7cm 2.0cm 2cm 21.75cm},clip,width=1.00\textwidth]
		{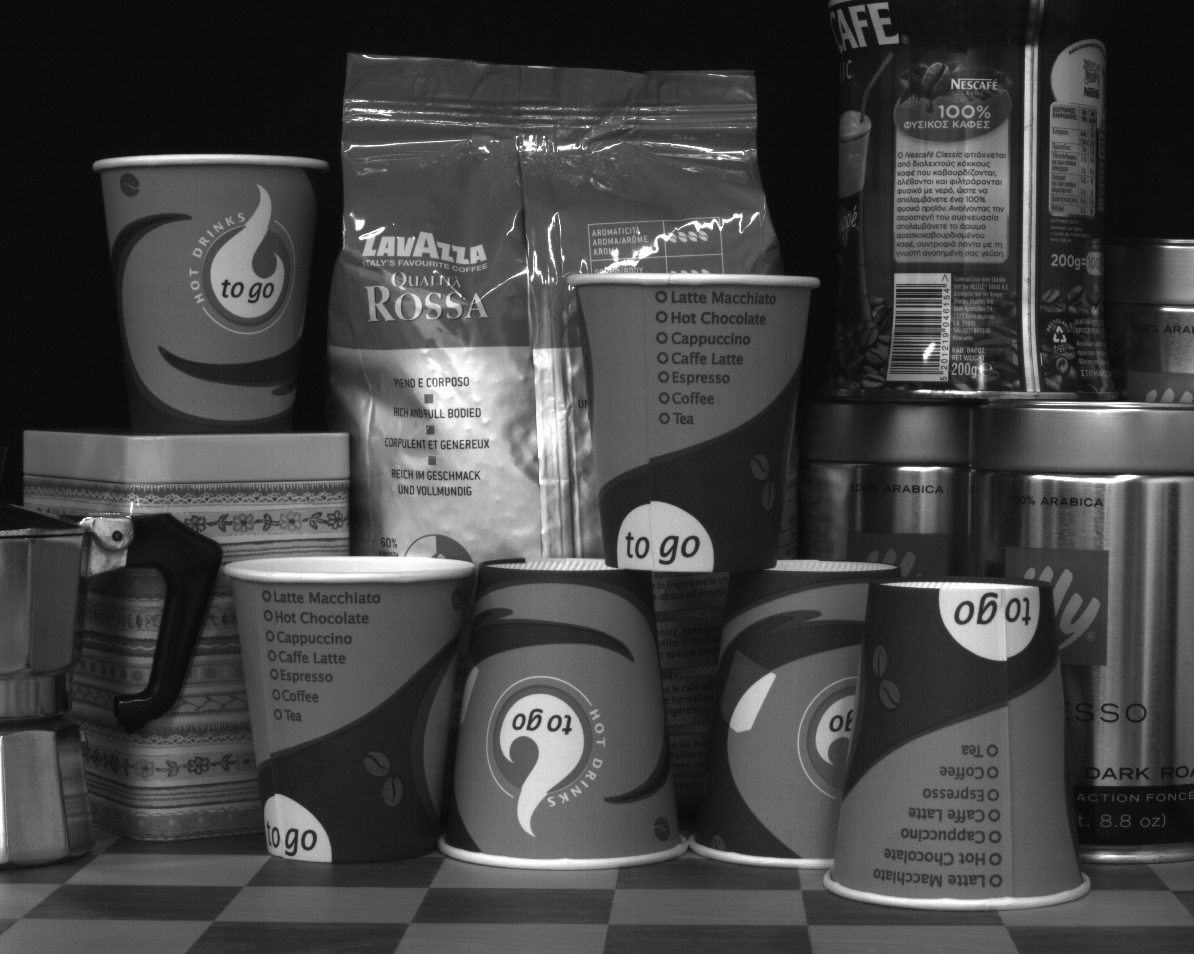}\\[0.4ex]
		\includegraphics[trim={20.7cm 4.0cm 13cm 23.0cm},clip,width=1.00\textwidth]
		{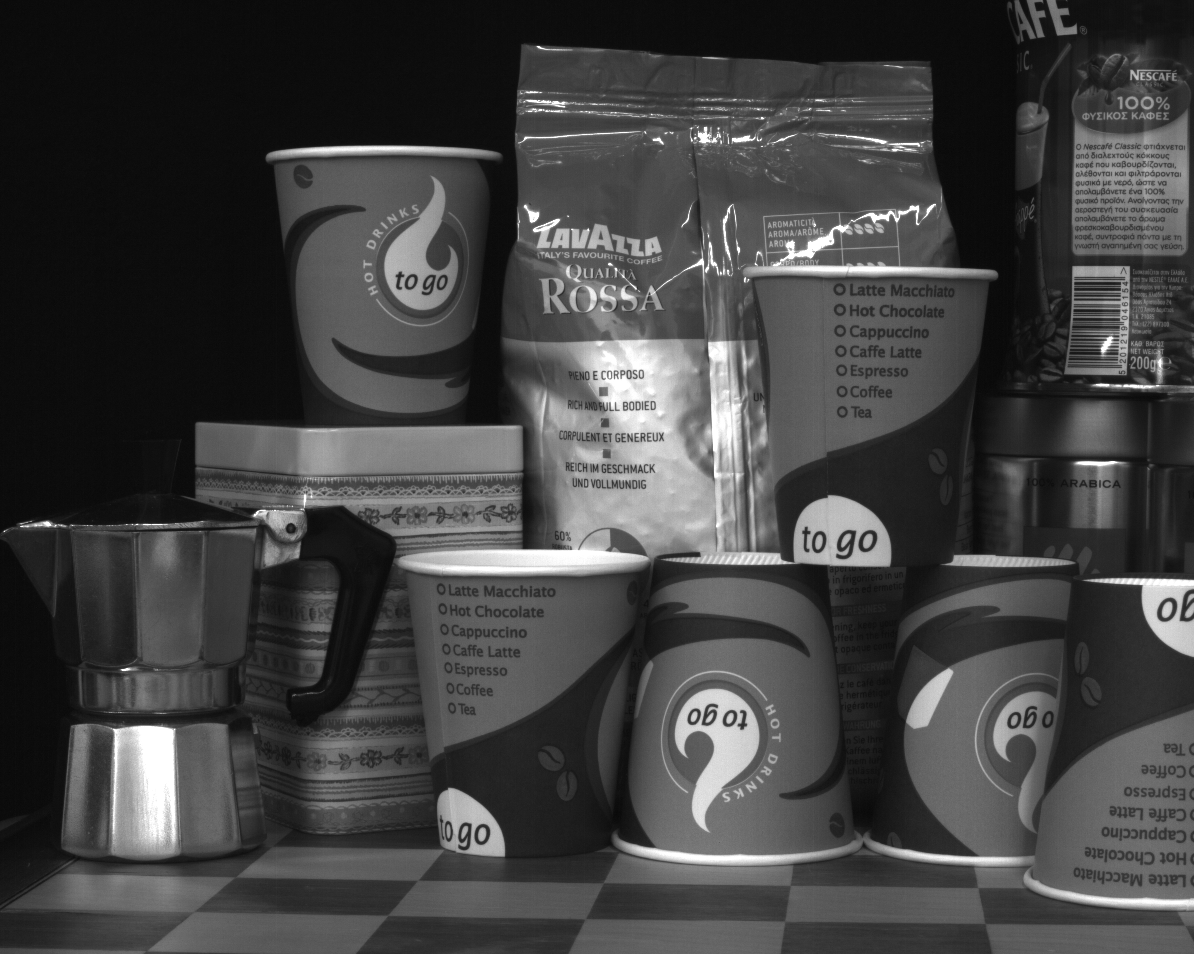}
	\end{minipage}}
	\subfloat{
	\begin{minipage}{0.158\textwidth} 
		\begin{tikzpicture}
			\draw (0, 0) node[inner sep=0] {\includegraphics[width=1.00\textwidth]
			{./images/coffee_local_pan_translation_xyz_inlier_bin4_sr0_f17_win01.png}};
			\draw [draw=red, line width=0.3mm] (1.3cm, -0.275cm) rectangle (0.45cm,-1.05cm);
			\draw [draw=green, line width=0.3mm] (-0.4cm, -0.45cm) rectangle (0.05cm,-0.85cm);
		\end{tikzpicture}
	\end{minipage}}
	\hfill
	\caption{SR on the dynamic \textit{coffee} dataset ($3 \times$ magnification). Top: Mixed motion due movements of a coffee cup. In contrast to SISR (\eg A+ \cite{Timofte2015}), simple interpolation-based MFSR (\eg NUISR \cite{Park2003}) is prone to erroneous optical flow caused by occlusions near object boundaries, while robust reconstruction (\eg IRWSR \cite{Kohler2015c}) and deep learning methods (\eg VSRnet \cite{Kappeler2016}) partly compensate for uncertainties. Bottom: Local object motion without camera motion. Except VSRnet \cite{Kappeler2016} that exploits training data, MFSR cannot effectively enhance the resolution.}
	\label{fig:mixedAndLocalMotionExample}
\end{figure*}

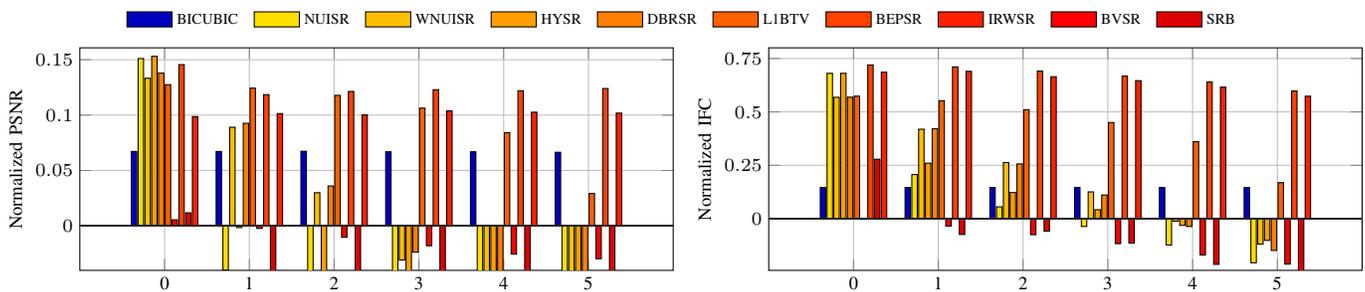
\begin{figure*}[!t]
	\scriptsize 
	\centering
	\subfloat{%
%
\definecolor{mycolor1}{rgb}{0.00000,0.00000,0.87500}%
\definecolor{mycolor2}{rgb}{0.00000,0.12500,1.00000}%
\definecolor{mycolor3}{rgb}{0.00000,0.25000,1.00000}%
\definecolor{mycolor4}{rgb}{0.00000,0.37500,1.00000}%
\definecolor{mycolor5}{rgb}{0.00000,0.50000,1.00000}%
\definecolor{mycolor6}{rgb}{0.00000,0.62500,1.00000}%
\definecolor{mycolor7}{rgb}{0.00000,0.75000,1.00000}%
\definecolor{mycolor8}{rgb}{1.00000,0.87500,0.00000}%
\definecolor{mycolor9}{rgb}{1.00000,0.75000,0.00000}%
\definecolor{mycolor10}{rgb}{0.87500,0.00000,0.00000}%
\begin{tikzpicture}
\scriptsize
\begin{axis}[%
    const plot,
		hide axis,
    xmin=10,
    xmax=50,
    ymin=0,
    ymax=0,
    legend style={
			legend image code/.code={2mm
				\draw [#1] (0cm,-0.065cm) rectangle (0.25cm,0.065cm);
			},
			anchor=north,
			inner ysep=0pt,
			draw=none, 
			legend cell align=center,
			legend columns=-1,
			/tikz/every even column/.append style={column sep=0.15cm},
			nodes={scale=0.815, transform shape}
		}
	]
	\addlegendimage{black, mark=none, fill=black!25!blue, area legend}
	\addlegendentry{BICUBIC}
	\addlegendimage{black, mark=none, fill=mycolor8, area legend}
	\addlegendentry{NUISR}
	\addlegendimage{black, mark=none, fill=mycolor9, area legend}
	\addlegendentry{WNUISR}
	\addlegendimage{black, mark=none, fill=orange!50!mycolor9, area legend}
	\addlegendentry{HYSR}
	\addlegendimage{black, mark=none, fill=orange, area legend}
	\addlegendentry{DBRSR}
	\addlegendimage{black, mark=none, fill=red!25!orange, area legend}
	\addlegendentry{L1BTV}
	\addlegendimage{black, mark=none, fill=red!50!orange, area legend}
	\addlegendentry{BEPSR}
	\addlegendimage{black, mark=none, fill=red!75!orange, area legend}
	\addlegendentry{IRWSR}
	\addlegendimage{black, mark=none, fill=red, area legend}
	\addlegendentry{BVSR}
	\addlegendimage{black, mark=none, fill=mycolor10, area legend}
	\addlegendentry{SRB}
\end{axis}

\end{tikzpicture}%
%
}\\[-1.2ex]
	\setlength \figurewidth{0.435\textwidth}
	\setlength \figureheight{0.9\figurewidth}
	\setcounter{subfigure}{0}
	\subfloat{%
%
\definecolor{mycolor1}{rgb}{1.00000,0.87500,0.00000}%
\definecolor{mycolor2}{rgb}{1.00000,0.75000,0.00000}%
\definecolor{mycolor3}{rgb}{0.87500,0.00000,0.00000}%
\begin{tikzpicture}

\begin{axis}[%
width=\figurewidth,
height=0.417\figureheight,
at={(0\figurewidth,0\figureheight)},
scale only axis,
xmin=0,
xmax=7,
xtick={1,2,3,4,5,6},
xticklabels={{0},{1},{2},{3},{4},{5}},
ymin=-0.0401897338690963,
ymax=0.160758935476385,
ylabel={Normalized PSNR},
axis background/.style={fill=white},
xmajorgrids,
ymajorgrids,
xlabel near ticks,ylabel near ticks,scaled y ticks=false,yticklabel style={/pgf/number format/fixed, /pgf/number format/precision=2},
]
\addplot[ybar,bar width=0.064,bar shift=-0.36,fill=black!25!blue,draw=black,area legend] plot table[row sep=crcr] {%
1	0.0671719865369982\\
2	0.0671356052995594\\
3	0.0673532166347459\\
4	0.0668855753488249\\
5	0.0668880576103005\\
6	0.0664453704072383\\
};
\addplot[forget plot,color=white!15!black] table[row sep=crcr] {%
0	0\\
7	0\\
};
\addplot[ybar,bar width=0.064,bar shift=-0.28,fill=mycolor1,draw=black,area legend] plot table[row sep=crcr] {%
1	0.15109423690482\\
2	-0.0399393606841239\\
3	-0.106475078129164\\
4	-0.1553874400994\\
5	-0.200448287827324\\
6	-0.24121615447559\\
};
\addplot[forget plot,color=white!15!black] table[row sep=crcr] {%
0	0\\
7	0\\
};
\addplot[ybar,bar width=0.064,bar shift=-0.2,fill=mycolor2,draw=black,area legend] plot table[row sep=crcr] {%
1	0.133286855287587\\
2	0.0890353419751549\\
3	0.0299460014939972\\
4	-0.0307925011041798\\
5	-0.0868237111109396\\
6	-0.128565852109316\\
};
\addplot[forget plot,color=white!15!black] table[row sep=crcr] {%
0	0\\
7	0\\
};
\addplot[ybar,bar width=0.064,bar shift=-0.12,fill=orange!50!mycolor2,draw=black,area legend] plot table[row sep=crcr] {%
1	0.153103748072748\\
2	-0.00151587582282763\\
3	-0.0591950554327069\\
4	-0.101822175643498\\
5	-0.141674405984811\\
6	-0.176923616623052\\
};
\addplot[forget plot,color=white!15!black] table[row sep=crcr] {%
0	0\\
7	0\\
};
\addplot[ybar,bar width=0.064,bar shift=-0.04,fill=orange,draw=black,area legend] plot table[row sep=crcr] {%
1	0.13796747002924\\
2	0.0926225939789486\\
3	0.0359876968633083\\
4	-0.0238411427186078\\
5	-0.078180205697942\\
6	-0.118500927625138\\
};
\addplot[forget plot,color=white!15!black] table[row sep=crcr] {%
0	0\\
7	0\\
};
\addplot[ybar,bar width=0.064,bar shift=0.04,fill=red!25!orange,draw=black,area legend] plot table[row sep=crcr] {%
1	0.127516679822096\\
2	0.124363776062916\\
3	0.117851954657321\\
4	0.106313429312328\\
5	0.084105831314354\\
6	0.0292610752107406\\
};
\addplot[forget plot,color=white!15!black] table[row sep=crcr] {%
0	0\\
7	0\\
};
\addplot[ybar,bar width=0.064,bar shift=0.12,fill=red,draw=black,area legend] plot table[row sep=crcr] {%
1	0.00545644711400756\\
2	-0.00231549149838662\\
3	-0.0103498515798158\\
4	-0.0179649346001822\\
5	-0.0255189001856621\\
6	-0.0296822304481539\\
};
\addplot[forget plot,color=white!15!black] table[row sep=crcr] {%
0	0\\
7	0\\
};
\addplot[ybar,bar width=0.064,bar shift=0.2,fill=red!50!orange,draw=black,area legend] plot table[row sep=crcr] {%
1	0.145574307098239\\
2	0.118506853019622\\
3	0.1213826581331\\
4	0.122959831274124\\
5	0.122029482110345\\
6	0.124009051912793\\
};
\addplot[forget plot,color=white!15!black] table[row sep=crcr] {%
0	0\\
7	0\\
};
\addplot[ybar,bar width=0.064,bar shift=0.28,fill=mycolor3,draw=black,area legend] plot table[row sep=crcr] {%
1	0.0117839952054566\\
2	-0.183910932312258\\
3	-0.173359735009983\\
4	-0.201058363011994\\
5	-0.275877800267849\\
6	-0.312568229850199\\
};
\addplot[forget plot,color=white!15!black] table[row sep=crcr] {%
0	0\\
7	0\\
};
\addplot[ybar,bar width=0.064,bar shift=0.36,fill=red!75!orange,draw=black,area legend] plot table[row sep=crcr] {%
1	0.0986688969619948\\
2	0.101405464445265\\
3	0.10019055230977\\
4	0.103914738175101\\
5	0.102633663300348\\
6	0.101820365976337\\
};
\addplot[forget plot,color=white!15!black] table[row sep=crcr] {%
0	0\\
7	0\\
};
\end{axis}
\end{tikzpicture}%
%
}\quad
	\subfloat{%
%
\definecolor{mycolor1}{rgb}{1.00000,0.87500,0.00000}%
\definecolor{mycolor2}{rgb}{1.00000,0.75000,0.00000}%
\definecolor{mycolor3}{rgb}{0.87500,0.00000,0.00000}%
\begin{tikzpicture}

\begin{axis}[%
width=\figurewidth,
height=0.417\figureheight,
at={(0\figurewidth,0\figureheight)},
scale only axis,
xmin=0,
xmax=7,
xtick={1,2,3,4,5,6},
xticklabels={{0},{1},{2},{3},{4},{5}},
ymin=-0.241814814673824,
ymax=0.80,
ytick={0, 0.25, 0.5, 0.75},
ylabel={Normalized IFC},
axis background/.style={fill=white},
xmajorgrids,
ymajorgrids,
xlabel near ticks,ylabel near ticks,scaled y ticks=false,yticklabel style={/pgf/number format/fixed, /pgf/number format/precision=2},
]
\addplot[ybar,bar width=0.064,bar shift=-0.36,fill=black!25!blue,draw=black,area legend] plot table[row sep=crcr] {%
1	0.14622298788427\\
2	0.14620842545483\\
3	0.146257207446152\\
4	0.146286197215796\\
5	0.146195358776088\\
6	0.145839615627436\\
};
\addplot[forget plot,color=white!15!black] table[row sep=crcr] {%
0	0\\
7	0\\
};
\addplot[ybar,bar width=0.064,bar shift=-0.28,fill=mycolor1,draw=black,area legend] plot table[row sep=crcr] {%
1	0.680611504720186\\
2	0.206665663927956\\
3	0.0557481560181877\\
4	-0.0363415638692197\\
5	-0.123100486957692\\
6	-0.206281328807995\\
};
\addplot[forget plot,color=white!15!black] table[row sep=crcr] {%
0	0\\
7	0\\
};
\addplot[ybar,bar width=0.064,bar shift=-0.2,fill=mycolor2,draw=black,area legend] plot table[row sep=crcr] {%
1	0.568273405581111\\
2	0.419070223683841\\
3	0.262904062548552\\
4	0.125549049849534\\
5	-0.0117394365191089\\
6	-0.119216816686424\\
};
\addplot[forget plot,color=white!15!black] table[row sep=crcr] {%
0	0\\
7	0\\
};
\addplot[ybar,bar width=0.064,bar shift=-0.12,fill=orange!50!mycolor2,draw=black,area legend] plot table[row sep=crcr] {%
1	0.680715476429955\\
2	0.259963673764144\\
3	0.122732368061377\\
4	0.0421564597591487\\
5	-0.0312946435335568\\
6	-0.101604116343361\\
};
\addplot[forget plot,color=white!15!black] table[row sep=crcr] {%
0	0\\
7	0\\
};
\addplot[ybar,bar width=0.064,bar shift=-0.04,fill=orange,draw=black,area legend] plot table[row sep=crcr] {%
1	0.568887525698531\\
2	0.421381047965986\\
3	0.256451329049686\\
4	0.110993354082642\\
5	-0.0365279199373902\\
6	-0.148503082713908\\
};
\addplot[forget plot,color=white!15!black] table[row sep=crcr] {%
0	0\\
7	0\\
};
\addplot[ybar,bar width=0.064,bar shift=0.04,fill=red!25!orange,draw=black,area legend] plot table[row sep=crcr] {%
1	0.5738241308402\\
2	0.552582576996231\\
3	0.510559896876152\\
4	0.450289000076663\\
5	0.360761298527637\\
6	0.168929251929564\\
};
\addplot[forget plot,color=white!15!black] table[row sep=crcr] {%
0	0\\
7	0\\
};
\addplot[ybar,bar width=0.064,bar shift=0.12,fill=red,draw=black,area legend] plot table[row sep=crcr] {%
1	-0.00307492394466845\\
2	-0.0347552770557738\\
3	-0.0747725116406037\\
4	-0.11673135114406\\
5	-0.170059836499523\\
6	-0.212469887869631\\
};
\addplot[forget plot,color=white!15!black] table[row sep=crcr] {%
0	0\\
7	0\\
};
\addplot[ybar,bar width=0.064,bar shift=0.2,fill=red!50!orange,draw=black,area legend] plot table[row sep=crcr] {%
1	0.719861889344916\\
2	0.710956993506497\\
3	0.691707908319795\\
4	0.667939326732896\\
5	0.640479585441581\\
6	0.59805982806603\\
};
\addplot[forget plot,color=white!15!black] table[row sep=crcr] {%
0	0\\
7	0\\
};
\addplot[ybar,bar width=0.064,bar shift=0.28,fill=mycolor3,draw=black,area legend] plot table[row sep=crcr] {%
1	0.278462140912413\\
2	-0.0738318439004537\\
3	-0.058340887262815\\
4	-0.114652946025602\\
5	-0.213804184577406\\
6	-0.283006266039472\\
};
\addplot[forget plot,color=white!15!black] table[row sep=crcr] {%
0	0\\
7	0\\
};
\addplot[ybar,bar width=0.064,bar shift=0.36,fill=red!75!orange,draw=black,area legend] plot table[row sep=crcr] {%
1	0.685964680010486\\
2	0.689861566932844\\
3	0.664752904146581\\
4	0.64620006837341\\
5	0.616023351092224\\
6	0.573484658303994\\
};
\addplot[forget plot,color=white!15!black] table[row sep=crcr] {%
0	0\\
7	0\\
};
\end{axis}
\end{tikzpicture}%
%
}
	\caption{Robustness analysis of MFSR \wrt photometric variations. The $x$-axis denote the number of photometric outliers within a set of $K = 11$ frames. The $y$-axis depicts the normalized PSNR and IFC averaged over 14 scenes with global motion and $3\times$ magnification.}
	\label{fig:photometricOutlierDatasets}
\end{figure*}

We investigated the influence of video compression using H.265/HEVC coding. \Fref{fig:qualityMeasuresCompression} benchmarks the different SR methods at five compression levels for $3\times$ magnification. As expected, video compression considerably affects the overall performance of SR. In particular, we found that at large compression levels the algorithms become indistinguishable as shown in \fref{fig:globalMotionExample} (bottom). This is related to the deficiency of current SR methodologies to handle video compression. It is interesting to note that most of the current deep learning (SRCNN, VDSR, DRCN) or dictionary learning techniques (ScSR, EBSR, NBSRF, A+) hold the potential to consider video compression in their underlying models. However, this aspect is often ignored within the image formation and training of these methods, which explains their behavior on real data affected by video compression.    

Interestingly, video compression has more influence to MFSR. We can explain this observation by antialiasing that is implicitly performed by H.265/HEVC coding. Since MFSR exploits aliased signal components to recover HR details, the performance of these algorithms is inherently limited.

\subsection{Super-Resolution on Dynamic Scenes}

\Fref{fig:srBenchmarkMotionTypes:mixed} benchmarks SR under mixed motion. Here, the performance of most MFSR
algorithms considerably deteriorated compared to static scenes while SISR
was unaffected. Motion artifacts were more
significant for more input frames at larger magnifications. That is
because optical flow estimation becomes more difficult for large displacements
related to local motion over longer input sequences.
We found that algorithms
building on simple interpolation (NUISR, HYSR) were most sensitive.
Interpolation-based SR with proper outlier weighting (WNUISR) or refinement
(DBRSR) as well as reconstruction-based SR with outlier-insensitive models
showed higher robustness. Interestingly, VSRnet was only slightly affected by
local motion. We explain this observation by the neural network architecture
that was trained for a fixed number of frames and the underlying adaptive motion
compensation. \Fref{fig:mixedAndLocalMotionExample} (top) depicts
some representative methods, where local motion is related to translational movements of a cup. The insufficient optical flow estimation leads to motion artifacts.

\Fref{fig:srBenchmarkMotionTypes:local} depicts our benchmark 
under local motion. The absence of global motion inherently 
affected MFSR algorithms as complementary information across LR frames does not exist. 
Thus, they effectively perform multi-frame deblurring/denoising 
but cannot overcome undersampling. In our benchmark, SISR partly 
outperformed MFSR. Among the MFSR algorithms, VSRnet performed best. This can 
be explained by the external training data used for VSRnet. Without global motion, it drops back to SISR and better 
recovers HR details than other MFSR methods, as shown in 
\fref{fig:mixedAndLocalMotionExample} (bottom).

\subsection{Super-Resolution under Photometric Variations}

We also studied SR under photometric variations over the input frames. This 
situation appears if input frames are collected over a longer period of time 
with environmental changes, \eg in remote sensing, and is crucial for MFSR. 
An exact handling requires a photometric registration \cite{Capel2003}, which is omitted by most 
state-of-the-art algorithms.

\begin{figure}[!b]
	\flushleft
	\subfloat[NUISR \cite{Park2003}]{
	\begin{minipage}{0.156\textwidth} 
		\includegraphics[trim={16.7cm 2.0cm 17cm 25.0cm},clip,width=1.00\textwidth]
		{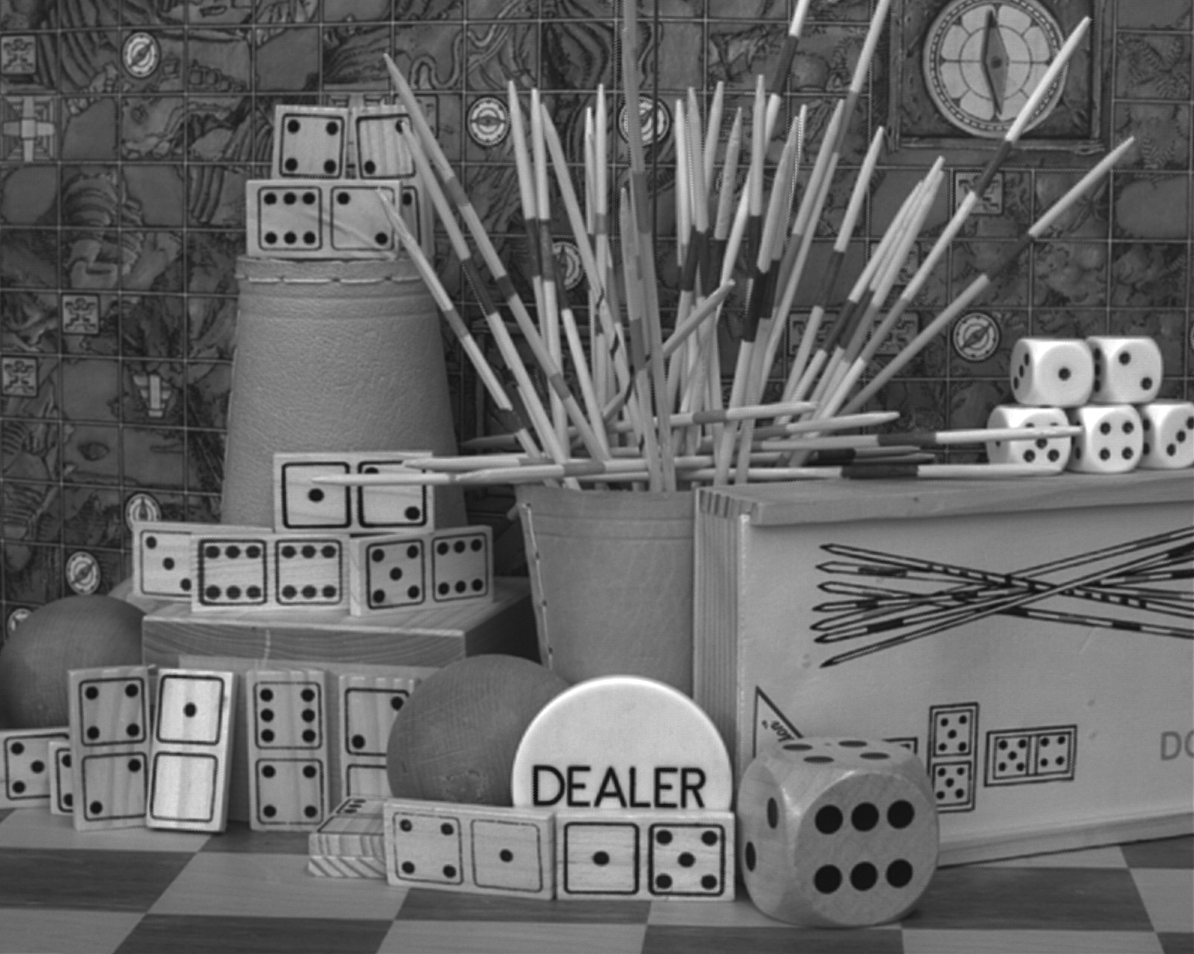}\\[0.4ex]
		\includegraphics[trim={17.3cm 2.35cm 16.4cm 24.65cm},clip,width=1.00\textwidth]
		{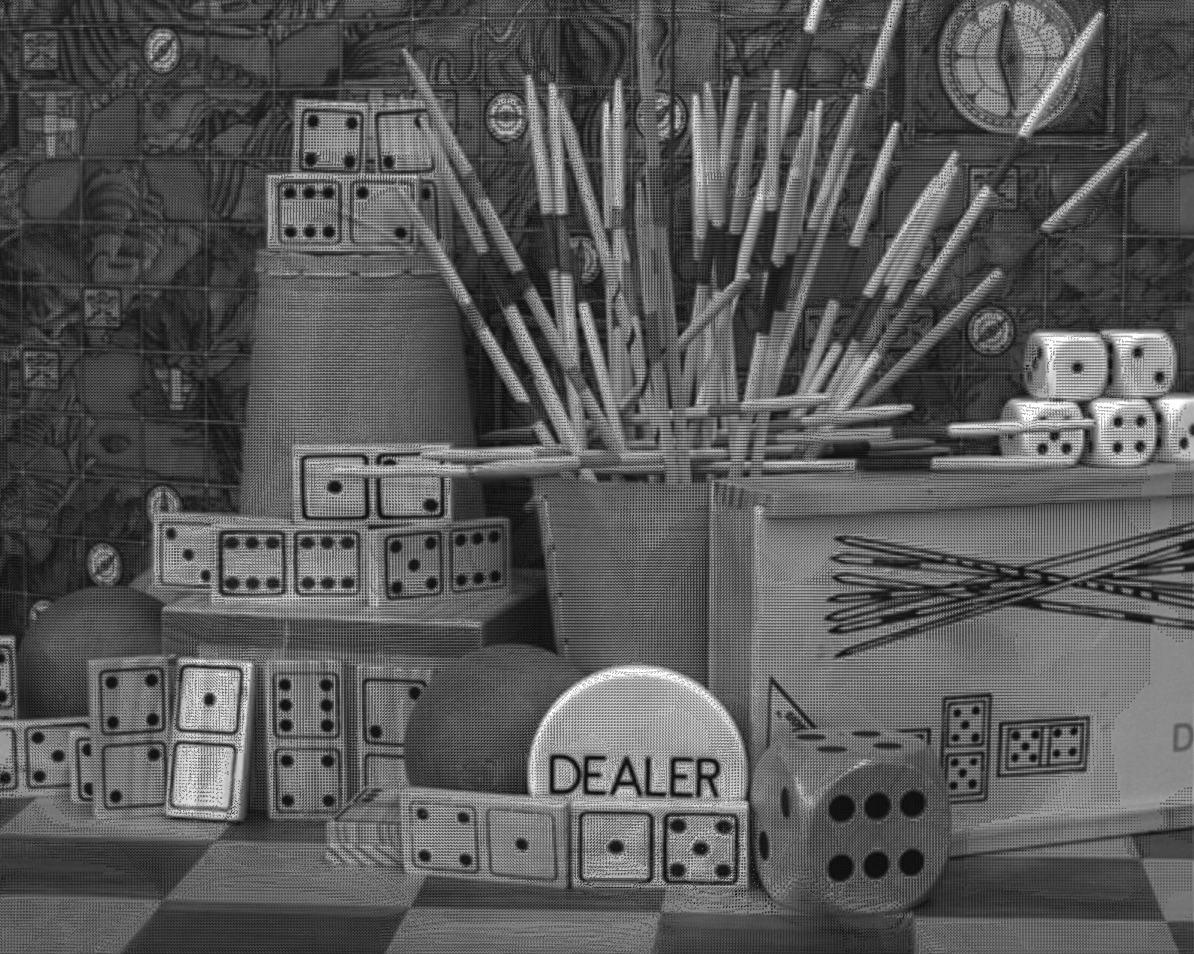}
	\end{minipage}}
	\subfloat[BEPSR \cite{Zeng2013}]{
	\begin{minipage}{0.156\textwidth} 
		\includegraphics[trim={16.7cm 2.0cm 17cm 25.0cm},clip,width=1.00\textwidth]
		{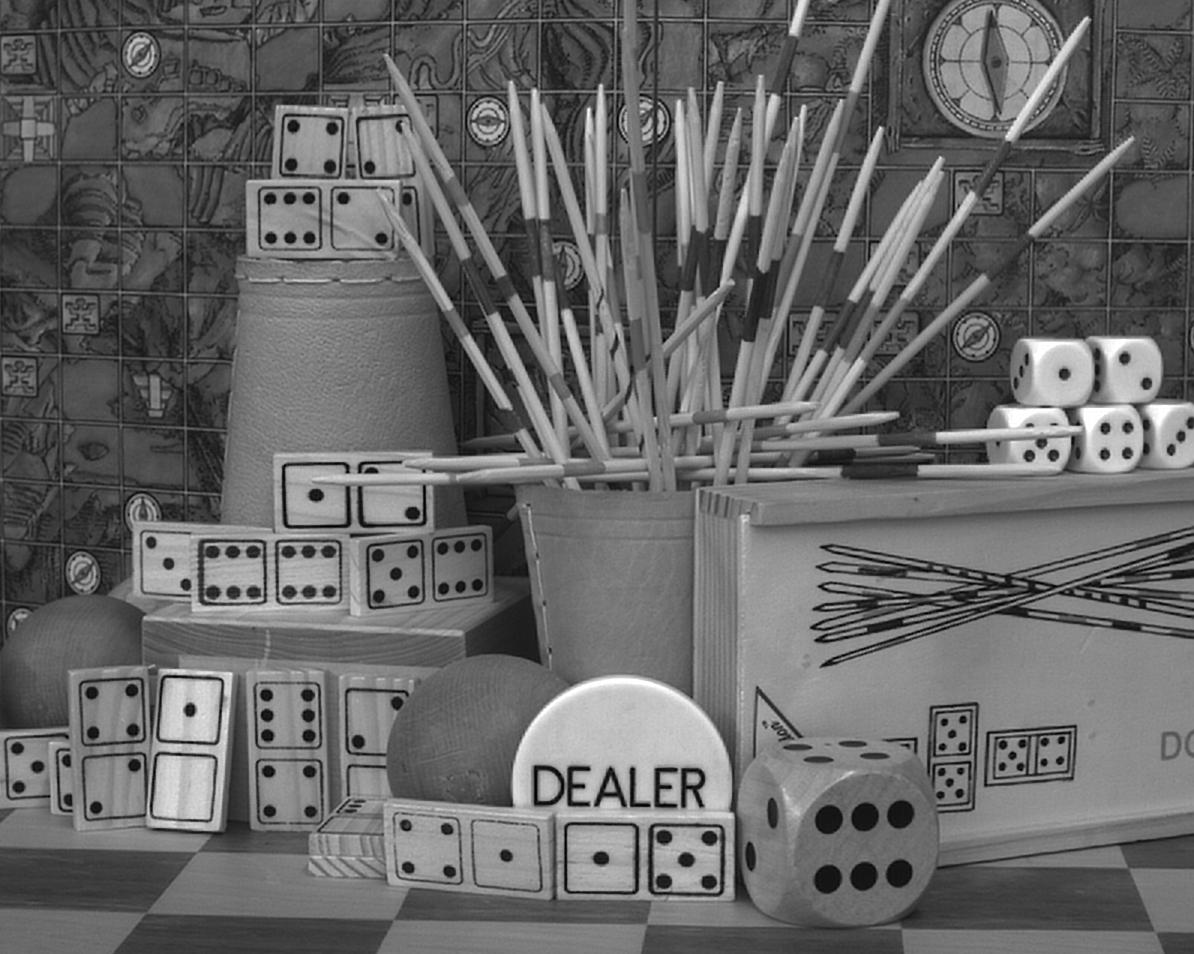}\\[0.4ex]
		\includegraphics[trim={17.3cm 2.35cm 16.4cm 24.65cm},clip,width=1.00\textwidth]
		{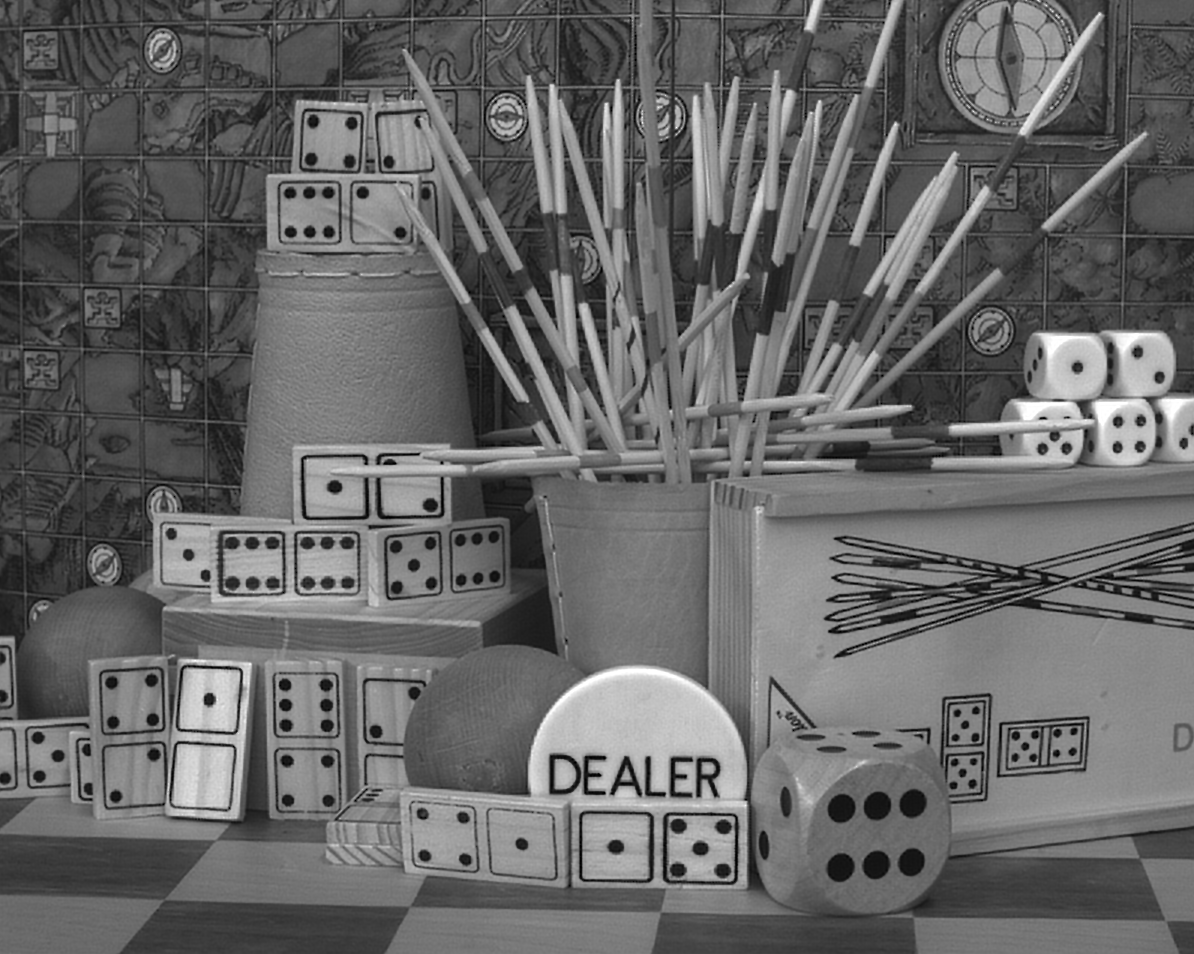}
	\end{minipage}}
	~\vline
	\subfloat[Ground truth]{
	\begin{minipage}{0.156\textwidth} 
		\begin{tikzpicture}
			\draw (0, 0) node[inner sep=0] {\includegraphics[width=1.00\textwidth]
			{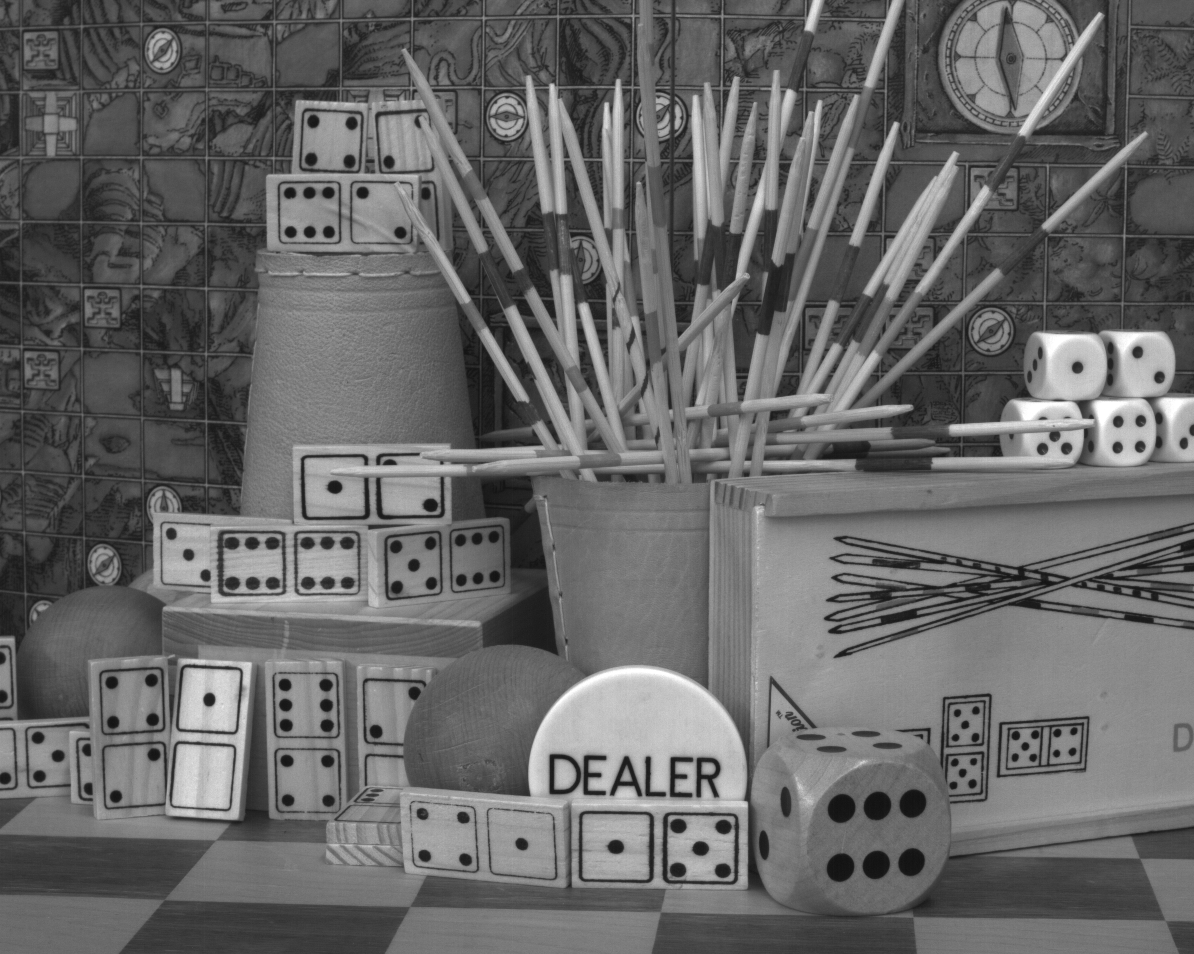}};
			\draw [draw=red, line width=0.3mm] (-0.25cm, -0.5cm) rectangle (0.35cm,-1.0cm);
		\end{tikzpicture}\\[0.3ex]
		\includegraphics[trim={17.3cm 2.35cm 16.4cm 24.65cm},clip,width=1.00\textwidth]
		{./images/games_global_pan_translation_xyz_outlier_photometric_bin3_sr0_f11_win06.png}
	\end{minipage}}
	\hfill
	\caption{MFSR without (top) and with photometric variations across multiple LR frames (bottom) on the \textit{games} dataset ($3\times$ magnification).}
	\label{fig:photometricVariationsExample}
\end{figure}

\begin{figure*}[!t]
	\scriptsize 
	\centering
	\setlength \figurewidth{0.185\textwidth}
	\setlength \figureheight{1.3\figurewidth}
	\subfloat[Global motion datasets]{%
%
\definecolor{mycolor1}{rgb}{0.87500,0.00000,0.00000}%
\definecolor{mycolor2}{rgb}{0.00000,0.12500,1.00000}%
\definecolor{mycolor3}{rgb}{0.00000,0.00000,0.62500}%
\definecolor{mycolor4}{rgb}{0.00000,0.75000,1.00000}%
\definecolor{mycolor5}{rgb}{1.00000,0.75000,0.00000}%
\definecolor{mycolor6}{rgb}{0.00000,0.00000,0.87500}%
\definecolor{mycolor7}{rgb}{1.00000,0.87500,0.00000}%
\begin{tikzpicture}

\begin{axis}[%
width=0.951\figurewidth,
height=\figureheight,
at={(0\figurewidth,0\figureheight)},
scale only axis,
xmin=-4,
xmax=2,
xlabel={Mean B-T score $\bar{\delta}$},
ymin=0,
ymax=21,
ytick={1,2,3,4,5,6,7,8,9,10,11,12,13,14,15,16,17,18,19,20},
yticklabels={{BVSR},{NN},{BICUBIC},{EBSR},{A+},{NUISR},{HYSR},{NBSRF},{SESR},{SRCNN},{VSRnet},{DRCN},{DBRSR},{WNUISR},{ScSR},{VDSR},{L1BTV},{SRB},{IRWSR},{BEPSR}},
axis background/.style={fill=white},
xmajorgrids,
ymajorgrids,
xlabel near ticks,ylabel near ticks,scaled y ticks=false,yticklabel style={/pgf/number format/fixed, /pgf/number format/precision=2},
]
\addplot[xbar,bar width=0.8,fill=mycolor1,draw=black,area legend] plot table[row sep=crcr] {%
-3.21068191600352	1\\
};
\addplot[forget plot,color=white!15!black] table[row sep=crcr] {%
0	0\\
0	21\\
};
\addplot[xbar,bar width=0.8,fill=blue,draw=black,area legend] plot table[row sep=crcr] {%
-2.52170392981464	2\\
};
\addplot[forget plot,color=white!15!black] table[row sep=crcr] {%
0	0\\
0	21\\
};
\addplot[xbar,bar width=0.8,fill=mycolor2,draw=black,area legend] plot table[row sep=crcr] {%
-2.06211785383346	3\\
};
\addplot[forget plot,color=white!15!black] table[row sep=crcr] {%
0	0\\
0	21\\
};
\addplot[xbar,bar width=0.8,fill=mycolor3,draw=black,area legend] plot table[row sep=crcr] {%
-0.49437845427705	4\\
};
\addplot[forget plot,color=white!15!black] table[row sep=crcr] {%
0	0\\
0	21\\
};
\addplot[xbar,bar width=0.8,fill=mycolor4,draw=black,area legend] plot table[row sep=crcr] {%
-0.163040158426084	5\\
};
\addplot[forget plot,color=white!15!black] table[row sep=crcr] {%
0	0\\
0	21\\
};
\addplot[xbar,bar width=0.8,fill=mycolor5,draw=black,area legend] plot table[row sep=crcr] {%
-0.139042968391532	6\\
};
\addplot[forget plot,color=white!15!black] table[row sep=crcr] {%
0	0\\
0	21\\
};
\addplot[xbar,bar width=0.8,fill=orange,draw=black,area legend] plot table[row sep=crcr] {%
-0.131805897636123	7\\
};
\addplot[forget plot,color=white!15!black] table[row sep=crcr] {%
0	0\\
0	21\\
};
\addplot[xbar,bar width=0.8,fill=mycolor6,draw=black,area legend] plot table[row sep=crcr] {%
-0.1174022485736	8\\
};
\addplot[forget plot,color=white!15!black] table[row sep=crcr] {%
0	0\\
0	21\\
};
\addplot[xbar,bar width=0.8,fill=blue!50!mycolor4,draw=black,area legend] plot table[row sep=crcr] {%
-0.0348370715967708	9\\
};
\addplot[forget plot,color=white!15!black] table[row sep=crcr] {%
0	0\\
0	21\\
};
\addplot[xbar,bar width=0.8,fill=mycolor2!80!mycolor4,draw=black,area legend] plot table[row sep=crcr] {%
0.0257052859266844	10\\
};
\addplot[forget plot,color=white!15!black] table[row sep=crcr] {%
0	0\\
0	21\\
};
\addplot[xbar,bar width=0.8,fill=mycolor7,draw=black,area legend] plot table[row sep=crcr] {%
0.123093422667461	11\\
};
\addplot[forget plot,color=white!15!black] table[row sep=crcr] {%
0	0\\
0	21\\
};
\addplot[xbar,bar width=0.8,fill=mycolor2!40!mycolor4,draw=black,area legend] plot table[row sep=crcr] {%
0.263252404369464	12\\
};
\addplot[forget plot,color=white!15!black] table[row sep=crcr] {%
0	0\\
0	21\\
};
\addplot[xbar,bar width=0.8,fill=red!25!orange,draw=black,area legend] plot table[row sep=crcr] {%
0.327040299738351	13\\
};
\addplot[forget plot,color=white!15!black] table[row sep=crcr] {%
0	0\\
0	21\\
};
\addplot[xbar,bar width=0.8,fill=orange!50!mycolor5,draw=black,area legend] plot table[row sep=crcr] {%
0.329393995658466	14\\
};
\addplot[forget plot,color=white!15!black] table[row sep=crcr] {%
0	0\\
0	21\\
};
\addplot[xbar,bar width=0.8,fill=black!25!blue,draw=black,area legend] plot table[row sep=crcr] {%
0.333485976702655	15\\
};
\addplot[forget plot,color=white!15!black] table[row sep=crcr] {%
0	0\\
0	21\\
};
\addplot[xbar,bar width=0.8,fill=mycolor2!20!mycolor4,draw=black,area legend] plot table[row sep=crcr] {%
0.360630048053914	16\\
};
\addplot[forget plot,color=white!15!black] table[row sep=crcr] {%
0	0\\
0	21\\
};
\addplot[xbar,bar width=0.8,fill=red!75!orange,draw=black,area legend] plot table[row sep=crcr] {%
1.6022229250763	17\\
};
\addplot[forget plot,color=white!15!black] table[row sep=crcr] {%
0	0\\
0	21\\
};
\addplot[xbar,bar width=0.8,fill=red!50!orange,draw=black,area legend] plot table[row sep=crcr] {%
1.80913513433377	18\\
};
\addplot[forget plot,color=white!15!black] table[row sep=crcr] {%
0	0\\
0	21\\
};
\addplot[xbar,bar width=0.8,fill=red,draw=black,area legend] plot table[row sep=crcr] {%
1.84953232832462	19\\
};
\addplot[forget plot,color=white!15!black] table[row sep=crcr] {%
0	0\\
0	21\\
};
\addplot[xbar,bar width=0.8,fill=black!25!red,draw=black,area legend] plot table[row sep=crcr] {%
1.8515186777011	20\\
};
\addplot[forget plot,color=white!15!black] table[row sep=crcr] {%
0	0\\
0	21\\
};
\end{axis}
\end{tikzpicture}%
%
}~
	\subfloat[Mixed motion datasets]{%
%
\definecolor{mycolor1}{rgb}{0.87500,0.00000,0.00000}%
\definecolor{mycolor2}{rgb}{0.00000,0.12500,1.00000}%
\definecolor{mycolor3}{rgb}{1.00000,0.75000,0.00000}%
\definecolor{mycolor4}{rgb}{0.00000,0.00000,0.62500}%
\definecolor{mycolor5}{rgb}{0.00000,0.75000,1.00000}%
\definecolor{mycolor6}{rgb}{0.00000,0.00000,0.87500}%
\definecolor{mycolor7}{rgb}{1.00000,0.87500,0.00000}%
\begin{tikzpicture}

\begin{axis}[%
width=0.951\figurewidth,
height=\figureheight,
at={(0\figurewidth,0\figureheight)},
scale only axis,
xmin=-3,
xmax=1.5,
xlabel={Mean B-T score $\bar{\delta}$},
ymin=0,
ymax=21,
ytick={1,2,3,4,5,6,7,8,9,10,11,12,13,14,15,16,17,18,19,20},
yticklabels={{BVSR},{NN},{BICUBIC},{NUISR},{HYSR},{EBSR},{WNUISR},{DBRSR},{A+},{NBSRF},{SESR},{SRCNN},{VSRnet},{VDSR},{DRCN},{ScSR},{SRB},{L1BTV},{IRWSR},{BEPSR}},
axis background/.style={fill=white},
xmajorgrids,
ymajorgrids,
xlabel near ticks,ylabel near ticks,scaled y ticks=false,yticklabel style={/pgf/number format/fixed, /pgf/number format/precision=2},
]
\addplot[xbar,bar width=0.8,fill=mycolor1,draw=black,area legend] plot table[row sep=crcr] {%
-2.82457608765266	1\\
};
\addplot[forget plot,color=white!15!black] table[row sep=crcr] {%
0	0\\
0	21\\
};
\addplot[xbar,bar width=0.8,fill=blue,draw=black,area legend] plot table[row sep=crcr] {%
-2.03426712130284	2\\
};
\addplot[forget plot,color=white!15!black] table[row sep=crcr] {%
0	0\\
0	21\\
};
\addplot[xbar,bar width=0.8,fill=mycolor2,draw=black,area legend] plot table[row sep=crcr] {%
-1.65044056427776	3\\
};
\addplot[forget plot,color=white!15!black] table[row sep=crcr] {%
0	0\\
0	21\\
};
\addplot[xbar,bar width=0.8,fill=mycolor3,draw=black,area legend] plot table[row sep=crcr] {%
-0.842214322912392	4\\
};
\addplot[forget plot,color=white!15!black] table[row sep=crcr] {%
0	0\\
0	21\\
};
\addplot[xbar,bar width=0.8,fill=orange,draw=black,area legend] plot table[row sep=crcr] {%
-0.462956666625039	5\\
};
\addplot[forget plot,color=white!15!black] table[row sep=crcr] {%
0	0\\
0	21\\
};
\addplot[xbar,bar width=0.8,fill=mycolor4,draw=black,area legend] plot table[row sep=crcr] {%
-0.126444045764507	6\\
};
\addplot[forget plot,color=white!15!black] table[row sep=crcr] {%
0	0\\
0	21\\
};
\addplot[xbar,bar width=0.8,fill=orange!50!mycolor3,draw=black,area legend] plot table[row sep=crcr] {%
-0.017368372595249	7\\
};
\addplot[forget plot,color=white!15!black] table[row sep=crcr] {%
0	0\\
0	21\\
};
\addplot[xbar,bar width=0.8,fill=red!25!orange,draw=black,area legend] plot table[row sep=crcr] {%
0.0336855346082461	8\\
};
\addplot[forget plot,color=white!15!black] table[row sep=crcr] {%
0	0\\
0	21\\
};
\addplot[xbar,bar width=0.8,fill=mycolor5,draw=black,area legend] plot table[row sep=crcr] {%
0.150936572438186	9\\
};
\addplot[forget plot,color=white!15!black] table[row sep=crcr] {%
0	0\\
0	21\\
};
\addplot[xbar,bar width=0.8,fill=mycolor6,draw=black,area legend] plot table[row sep=crcr] {%
0.213104380918641	10\\
};
\addplot[forget plot,color=white!15!black] table[row sep=crcr] {%
0	0\\
0	21\\
};
\addplot[xbar,bar width=0.8,fill=blue!50!mycolor5,draw=black,area legend] plot table[row sep=crcr] {%
0.258923826709304	11\\
};
\addplot[forget plot,color=white!15!black] table[row sep=crcr] {%
0	0\\
0	21\\
};
\addplot[xbar,bar width=0.8,fill=mycolor2!80!mycolor5,draw=black,area legend] plot table[row sep=crcr] {%
0.336769475019614	12\\
};
\addplot[forget plot,color=white!15!black] table[row sep=crcr] {%
0	0\\
0	21\\
};
\addplot[xbar,bar width=0.8,fill=mycolor7,draw=black,area legend] plot table[row sep=crcr] {%
0.393294425822709	13\\
};
\addplot[forget plot,color=white!15!black] table[row sep=crcr] {%
0	0\\
0	21\\
};
\addplot[xbar,bar width=0.8,fill=mycolor2!20!mycolor5,draw=black,area legend] plot table[row sep=crcr] {%
0.62143592004631	14\\
};
\addplot[forget plot,color=white!15!black] table[row sep=crcr] {%
0	0\\
0	21\\
};
\addplot[xbar,bar width=0.8,fill=mycolor2!40!mycolor5,draw=black,area legend] plot table[row sep=crcr] {%
0.634873392881432	15\\
};
\addplot[forget plot,color=white!15!black] table[row sep=crcr] {%
0	0\\
0	21\\
};
\addplot[xbar,bar width=0.8,fill=black!25!blue,draw=black,area legend] plot table[row sep=crcr] {%
0.684527249697011	16\\
};
\addplot[forget plot,color=white!15!black] table[row sep=crcr] {%
0	0\\
0	21\\
};
\addplot[xbar,bar width=0.8,fill=red!50!orange,draw=black,area legend] plot table[row sep=crcr] {%
1.03378369441082	17\\
};
\addplot[forget plot,color=white!15!black] table[row sep=crcr] {%
0	0\\
0	21\\
};
\addplot[xbar,bar width=0.8,fill=red!75!orange,draw=black,area legend] plot table[row sep=crcr] {%
1.1232429149644	18\\
};
\addplot[forget plot,color=white!15!black] table[row sep=crcr] {%
0	0\\
0	21\\
};
\addplot[xbar,bar width=0.8,fill=red,draw=black,area legend] plot table[row sep=crcr] {%
1.18163650629129	19\\
};
\addplot[forget plot,color=white!15!black] table[row sep=crcr] {%
0	0\\
0	21\\
};
\addplot[xbar,bar width=0.8,fill=black!25!red,draw=black,area legend] plot table[row sep=crcr] {%
1.29205328732248	20\\
};
\addplot[forget plot,color=white!15!black] table[row sep=crcr] {%
0	0\\
0	21\\
};
\end{axis}
\end{tikzpicture}%
%
}~
	\subfloat[Local motion datasets]{%
%
\definecolor{mycolor1}{rgb}{0.87500,0.00000,0.00000}%
\definecolor{mycolor2}{rgb}{1.00000,0.75000,0.00000}%
\definecolor{mycolor3}{rgb}{0.00000,0.12500,1.00000}%
\definecolor{mycolor4}{rgb}{0.00000,0.00000,0.62500}%
\definecolor{mycolor5}{rgb}{0.00000,0.75000,1.00000}%
\definecolor{mycolor6}{rgb}{0.00000,0.00000,0.87500}%
\definecolor{mycolor7}{rgb}{1.00000,0.87500,0.00000}%
\begin{tikzpicture}

\begin{axis}[%
width=0.951\figurewidth,
height=\figureheight,
at={(0\figurewidth,0\figureheight)},
scale only axis,
xmin=-2.5,
xmax=1.5,
xlabel={Mean B-T score $\bar{\delta}$},
ymin=0,
ymax=21,
ytick={1,2,3,4,5,6,7,8,9,10,11,12,13,14,15,16,17,18,19,20},
yticklabels={{BVSR},{NN},{NUISR},{BICUBIC},{DBRSR},{WNUISR},{HYSR},{IRWSR},{L1BTV},{EBSR},{SRB},{A+},{NBSRF},{BEPSR},{SRCNN},{VSRnet},{SESR},{ScSR},{VDSR},{DRCN}},
axis background/.style={fill=white},
xmajorgrids,
ymajorgrids,
xlabel near ticks,ylabel near ticks,scaled y ticks=false,yticklabel style={/pgf/number format/fixed, /pgf/number format/precision=2},
]
\addplot[xbar,bar width=0.8,fill=mycolor1,draw=black,area legend] plot table[row sep=crcr] {%
-2.45761832641972	1\\
};
\addplot[forget plot,color=white!15!black] table[row sep=crcr] {%
0	0\\
0	21\\
};
\addplot[xbar,bar width=0.8,fill=blue,draw=black,area legend] plot table[row sep=crcr] {%
-1.91132920764394	2\\
};
\addplot[forget plot,color=white!15!black] table[row sep=crcr] {%
0	0\\
0	21\\
};
\addplot[xbar,bar width=0.8,fill=mycolor2,draw=black,area legend] plot table[row sep=crcr] {%
-1.78807142917529	3\\
};
\addplot[forget plot,color=white!15!black] table[row sep=crcr] {%
0	0\\
0	21\\
};
\addplot[xbar,bar width=0.8,fill=mycolor3,draw=black,area legend] plot table[row sep=crcr] {%
-1.33656805259574	4\\
};
\addplot[forget plot,color=white!15!black] table[row sep=crcr] {%
0	0\\
0	21\\
};
\addplot[xbar,bar width=0.8,fill=red!25!orange,draw=black,area legend] plot table[row sep=crcr] {%
-0.72122529486502	5\\
};
\addplot[forget plot,color=white!15!black] table[row sep=crcr] {%
0	0\\
0	21\\
};
\addplot[xbar,bar width=0.8,fill=orange!50!mycolor2,draw=black,area legend] plot table[row sep=crcr] {%
-0.71786704332083	6\\
};
\addplot[forget plot,color=white!15!black] table[row sep=crcr] {%
0	0\\
0	21\\
};
\addplot[xbar,bar width=0.8,fill=orange,draw=black,area legend] plot table[row sep=crcr] {%
-0.470294776644207	7\\
};
\addplot[forget plot,color=white!15!black] table[row sep=crcr] {%
0	0\\
0	21\\
};
\addplot[xbar,bar width=0.8,fill=red,draw=black,area legend] plot table[row sep=crcr] {%
0.14059808249499	8\\
};
\addplot[forget plot,color=white!15!black] table[row sep=crcr] {%
0	0\\
0	21\\
};
\addplot[xbar,bar width=0.8,fill=red!75!orange,draw=black,area legend] plot table[row sep=crcr] {%
0.332932612504089	9\\
};
\addplot[forget plot,color=white!15!black] table[row sep=crcr] {%
0	0\\
0	21\\
};
\addplot[xbar,bar width=0.8,fill=mycolor4,draw=black,area legend] plot table[row sep=crcr] {%
0.392873360005392	10\\
};
\addplot[forget plot,color=white!15!black] table[row sep=crcr] {%
0	0\\
0	21\\
};
\addplot[xbar,bar width=0.8,fill=red!50!orange,draw=black,area legend] plot table[row sep=crcr] {%
0.555854774211428	11\\
};
\addplot[forget plot,color=white!15!black] table[row sep=crcr] {%
0	0\\
0	21\\
};
\addplot[xbar,bar width=0.8,fill=mycolor5,draw=black,area legend] plot table[row sep=crcr] {%
0.610029312585727	12\\
};
\addplot[forget plot,color=white!15!black] table[row sep=crcr] {%
0	0\\
0	21\\
};
\addplot[xbar,bar width=0.8,fill=mycolor6,draw=black,area legend] plot table[row sep=crcr] {%
0.697568579571008	13\\
};
\addplot[forget plot,color=white!15!black] table[row sep=crcr] {%
0	0\\
0	21\\
};
\addplot[xbar,bar width=0.8,fill=black!25!red,draw=black,area legend] plot table[row sep=crcr] {%
0.767981731743068	14\\
};
\addplot[forget plot,color=white!15!black] table[row sep=crcr] {%
0	0\\
0	21\\
};
\addplot[xbar,bar width=0.8,fill=mycolor3!80!mycolor5,draw=black,area legend] plot table[row sep=crcr] {%
0.823730339432294	15\\
};
\addplot[forget plot,color=white!15!black] table[row sep=crcr] {%
0	0\\
0	21\\
};
\addplot[xbar,bar width=0.8,fill=mycolor7,draw=black,area legend] plot table[row sep=crcr] {%
0.828329299940606	16\\
};
\addplot[forget plot,color=white!15!black] table[row sep=crcr] {%
0	0\\
0	21\\
};
\addplot[xbar,bar width=0.8,fill=blue!50!mycolor5,draw=black,area legend] plot table[row sep=crcr] {%
0.840277910395276	17\\
};
\addplot[forget plot,color=white!15!black] table[row sep=crcr] {%
0	0\\
0	21\\
};
\addplot[xbar,bar width=0.8,fill=black!25!blue,draw=black,area legend] plot table[row sep=crcr] {%
1.09806336412734	18\\
};
\addplot[forget plot,color=white!15!black] table[row sep=crcr] {%
0	0\\
0	21\\
};
\addplot[xbar,bar width=0.8,fill=mycolor3!20!mycolor5,draw=black,area legend] plot table[row sep=crcr] {%
1.1134219279472	19\\
};
\addplot[forget plot,color=white!15!black] table[row sep=crcr] {%
0	0\\
0	21\\
};
\addplot[xbar,bar width=0.8,fill=mycolor3!40!mycolor5,draw=black,area legend] plot table[row sep=crcr] {%
1.20131283570633	20\\
};
\addplot[forget plot,color=white!15!black] table[row sep=crcr] {%
0	0\\
0	21\\
};
\end{axis}
\end{tikzpicture}%
%
}~
	\subfloat[Photometric variation datasets]{%
%
\definecolor{mycolor1}{rgb}{1.00000,0.75000,0.00000}%
\definecolor{mycolor2}{rgb}{0.87500,0.00000,0.00000}%
\definecolor{mycolor3}{rgb}{1.00000,0.87500,0.00000}%
\definecolor{mycolor4}{rgb}{0.00000,0.12500,1.00000}%
\begin{tikzpicture}

\begin{axis}[%
width=0.951\figurewidth,
height=\figureheight,
at={(0\figurewidth,0\figureheight)},
scale only axis,
xmin=-1,
xmax=0.8,
xlabel={Mean B-T score $\bar{\delta}$},
ymin=0,
ymax=13,
ytick={1,2,3,4,5,6,7,8,9,10,11,12},
yticklabels={{NUISR},{BVSR},{HYSR},{NN},{VSRnet},{WNUISR},{DBRSR},{BICUBIC},{SRB},{L1BTV},{BEPSR},{IRWSR}},
axis background/.style={fill=white},
xmajorgrids,
ymajorgrids,
xlabel near ticks,ylabel near ticks,scaled y ticks=false,yticklabel style={/pgf/number format/fixed, /pgf/number format/precision=2},
]
\addplot[xbar,bar width=0.8,fill=mycolor1,draw=black,area legend] plot table[row sep=crcr] {%
-0.942048437797492	1\\
};
\addplot[forget plot,color=white!15!black] table[row sep=crcr] {%
0	0\\
0	13\\
};
\addplot[xbar,bar width=0.8,fill=mycolor2,draw=black,area legend] plot table[row sep=crcr] {%
-0.617820994145243	2\\
};
\addplot[forget plot,color=white!15!black] table[row sep=crcr] {%
0	0\\
0	13\\
};
\addplot[xbar,bar width=0.8,fill=orange,draw=black,area legend] plot table[row sep=crcr] {%
-0.313263689749766	3\\
};
\addplot[forget plot,color=white!15!black] table[row sep=crcr] {%
0	0\\
0	13\\
};
\addplot[xbar,bar width=0.8,fill=blue,draw=black,area legend] plot table[row sep=crcr] {%
-0.211064442287951	4\\
};
\addplot[forget plot,color=white!15!black] table[row sep=crcr] {%
0	0\\
0	13\\
};
\addplot[xbar,bar width=0.8,fill=mycolor3,draw=black,area legend] plot table[row sep=crcr] {%
-0.0923585717263472	5\\
};
\addplot[forget plot,color=white!15!black] table[row sep=crcr] {%
0	0\\
0	13\\
};
\addplot[xbar,bar width=0.8,fill=orange!50!mycolor1,draw=black,area legend] plot table[row sep=crcr] {%
-0.0562942342118862	6\\
};
\addplot[forget plot,color=white!15!black] table[row sep=crcr] {%
0	0\\
0	13\\
};
\addplot[xbar,bar width=0.8,fill=red!25!orange,draw=black,area legend] plot table[row sep=crcr] {%
-0.0505410469980514	7\\
};
\addplot[forget plot,color=white!15!black] table[row sep=crcr] {%
0	0\\
0	13\\
};
\addplot[xbar,bar width=0.8,fill=mycolor4,draw=black,area legend] plot table[row sep=crcr] {%
0.0699010869310685	8\\
};
\addplot[forget plot,color=white!15!black] table[row sep=crcr] {%
0	0\\
0	13\\
};
\addplot[xbar,bar width=0.8,fill=red!50!orange,draw=black,area legend] plot table[row sep=crcr] {%
0.112146522090725	9\\
};
\addplot[forget plot,color=white!15!black] table[row sep=crcr] {%
0	0\\
0	13\\
};
\addplot[xbar,bar width=0.8,fill=red!75!orange,draw=black,area legend] plot table[row sep=crcr] {%
0.479810062245435	10\\
};
\addplot[forget plot,color=white!15!black] table[row sep=crcr] {%
0	0\\
0	13\\
};
\addplot[xbar,bar width=0.8,fill=black!25!red,draw=black,area legend] plot table[row sep=crcr] {%
0.694491547184156	11\\
};
\addplot[forget plot,color=white!15!black] table[row sep=crcr] {%
0	0\\
0	13\\
};
\addplot[xbar,bar width=0.8,fill=red,draw=black,area legend] plot table[row sep=crcr] {%
0.721101867083605	12\\
};
\addplot[forget plot,color=white!15!black] table[row sep=crcr] {%
0	0\\
0	13\\
};
\end{axis}
\end{tikzpicture}%
%
}
	\caption{Ranking of the competing SR algorithms on the different datasets in our observer study. We ranked the algorithms \wrt their mean B-T scores $\bar{\delta}$, where higher, positive scores express better image quality according to human visual perception. The individual algorithms are categorized either as SISR (shown with blue color map) or MFSR (shown with red color map).}
	\label{fig:btScoreRanking}
\end{figure*}
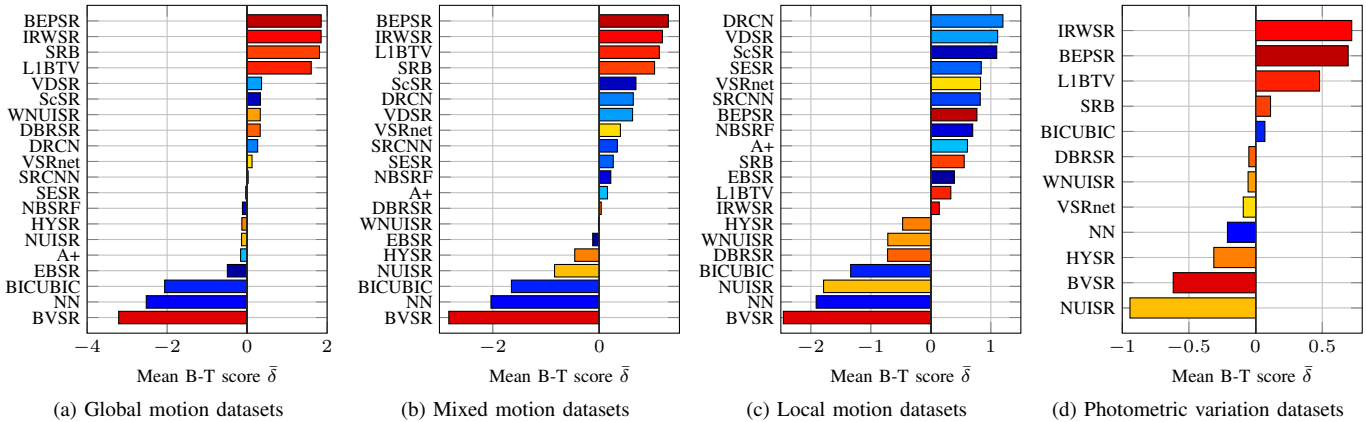

\Fref{fig:photometricOutlierDatasets} evaluates MFSR for an 
increasing number of photometric outlier within $K = 11$ 
consecutive frames with global motion and $3\times$ magnification. We found that even for a few outliers most methods 
performed worse than simple bicubic interpolation as photometric variations are neither considered 
implicitly by generative models nor explicitly by proper correction methods. 
Reconstruction-based algorithms with robust and adaptive models (IRWSR, BEPSR)
 were less sensitive and adaptively handled photometric variations. \Fref{fig:photometricVariationsExample} depicts 
this behavior on the \textit{games} dataset. The 
photometric variations resulted in intensity distortions and noise in 
interpolation-based SR (NUISR) while adaptive reconstruction-based SR (BEPSR) 
was unaffected.

\section{Human Observer Study}
\label{sec:HumanSubjectStudy}

We conducted an observer study for global motion, mixed motion, local motion, and photometric variations. The data comprises uncompressed images, one camera trajectory (translation $x$, $y$, $z$ and pan), and one sliding window per image sequence. Overall, the study was conducted with 3,024 images obtained from 19 algorithms at three resolution levels, which yields 26,712 image pairs. The observers comprise paid participants from Amazon Mechanical Turk as well as volunteers including experts in computer vision. We collected 292,400 votes in 5,848 sessions\footnote{We tolerated repeated participations for the observers, but ensured that each session consists of randomly selected image pairs.}. Each session comprises 50 image pairs, where 8 pairs served as sanity checks. The median time to complete one session was 14.8 minutes. We discarded 16.8\% of the votes due to observers who failed the sanity checks.

\subsection{Ranking of the Super-Resolution Methods}

The algorithm benchmark in our observer study is based on the B-T model. In \fref{fig:btScoreRanking}, we rank the competing SR methods based on their mean B-T scores $\bar{\delta}$ on different datasets. One can observe that the ranking heavily depends on the underlying motion and photometric conditions. We make the following observations that partly agree with our quantitative evaluation:
\begin{itemize}
	\item In case of global motion, reconstruction-based MFSR (BEPSR, IRWSR, SRB, L1BTV) ranked highest.
	\item In case of mixed motion, MFSR is partly outperformed by SISR. We explain this behavior by the sensitivity of MFSR against inaccurate local motion estimates. 
	\item In case of local motion, SISR algorithms are ranked highest. Particularly, recent deep learning (DRCN, VDSR) and dictionary methods (ScSR) performed best. 
	\item In case of photometric variations, robust reconstruction methods (IRWSR, BEPSR) are ranked highest.
\end{itemize}
 
\subsection{Inter-Observer Variance}

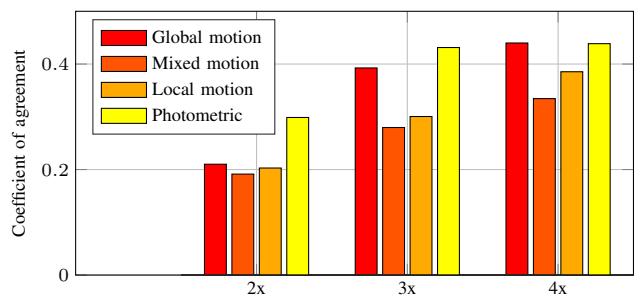
\begin{figure}[!b]
	\scriptsize 
	\centering
	\setlength \figurewidth{0.43\textwidth}
	\setlength \figureheight{0.45\figurewidth}
	%
%
\definecolor{mycolor1}{rgb}{1.00000,0.33333,0.00000}%
\definecolor{mycolor2}{rgb}{1.00000,0.66667,0.00000}%
\definecolor{mycolor3}{rgb}{1.00000,1.00000,0.00000}%
\begin{tikzpicture}

\begin{axis}[%
width=0.951\figurewidth,
height=\figureheight,
at={(0\figurewidth,0\figureheight)},
scale only axis,
xmin=0.8,
xmax=4.5,
xtick={2,3,4},
xticklabels={{2x},{3x},{4x}},
ymin=0,
ymax=0.5,
ylabel={Coefficient of agreement},
axis background/.style={fill=white},
xmajorgrids,
ymajorgrids,
legend style={at={(0.03,0.97)},anchor=north west,legend cell align=left,align=left,draw=white!15!black},
xlabel near ticks,ylabel near ticks,scaled y ticks=false,yticklabel style={/pgf/number format/fixed, /pgf/number format/precision=2},
]
\addplot[ybar,bar width=0.145,bar shift=-0.273,fill=red,draw=black,area legend] plot table[row sep=crcr] {%
2	0.210100684943756\\
3	0.392589019719109\\
4	0.439902716292288\\
};
\addplot[forget plot,color=white!15!black] table[row sep=crcr] {%
1.5	0\\
4.5	0\\
};
\addlegendentry{Global motion};

\addplot[ybar,bar width=0.145,bar shift=-0.091,fill=mycolor1,draw=black,area legend] plot table[row sep=crcr] {%
2	0.191390209688291\\
3	0.27967790712503\\
4	0.334488487132131\\
};
\addplot[forget plot,color=white!15!black] table[row sep=crcr] {%
1.5	0\\
4.5	0\\
};
\addlegendentry{Mixed motion};

\addplot[ybar,bar width=0.145,bar shift=0.091,fill=mycolor2,draw=black,area legend] plot table[row sep=crcr] {%
2	0.202896649723999\\
3	0.300528319760981\\
4	0.385396212243146\\
};
\addplot[forget plot,color=white!15!black] table[row sep=crcr] {%
1.5	0\\
4.5	0\\
};
\addlegendentry{Local motion};

\addplot[ybar,bar width=0.145,bar shift=0.273,fill=mycolor3,draw=black,area legend] plot table[row sep=crcr] {%
2	0.298696449181886\\
3	0.431297744025017\\
4	0.438611388611389\\
};
\addplot[forget plot,color=white!15!black] table[row sep=crcr] {%
1.5	0\\
4.5	0\\
};
\addlegendentry{Photometric};

\end{axis}
\end{tikzpicture}%
%

	\caption{Kendall coefficients of agreement among the observers in our study for different datasets and magnification factors.}
	\label{fig:kendall}
\end{figure}

The inter-observer variance is analyzed by the Kendall coefficient of agreement $u$ for different datasets and magnification factors in \fref{fig:kendall}. We found that the agreement among different observers increases with the magnification factor. This can be explained by a higher perceptual consensus at large factors, where differences among SR algorithms are often clearly visible. For smaller factors, observers are more often controversial about algorithm performance. Moreover, the agreement is higher under global motion compared to local and mixed motion. This is because motion artifacts in MFSR are more likely under local motion. Therefore, some observers subjectively prefer SISR that hallucinates HR details without motion artifacts, while others tolerate slight motion artifacts but prefer the recovery of true HR details as achieved by MFSR. Photometric variations lead to the highest agreement, since here quality differences are clearly visible.

\begin{figure}[!t]
	\scriptsize 
	\centering
	\setlength \figurewidth{0.43\textwidth}
	\setlength \figureheight{0.47\figurewidth}
	%
%
\definecolor{mycolor1}{rgb}{1.00000,0.65000,0.00000}%
\begin{tikzpicture}

\begin{axis}[%
width=0.951\figurewidth,
height=\figureheight,
at={(0\figurewidth,0\figureheight)},
scale only axis,
xmin=500,
xmax=4000,
xlabel={User sessions},
ymin=0.18,
ymax=0.32,
ylabel={Coefficient of agreement},
axis background/.style={fill=white},
xmajorgrids,
ymajorgrids,
legend style={legend columns=3,legend cell align=left,align=left,draw=white!15!black},
xlabel near ticks,ylabel near ticks,scaled y ticks=false,yticklabel style={/pgf/number format/fixed, /pgf/number format/precision=2},
]
\addplot [color=blue,line width=1.5pt]
 plot [error bars/.cd, y dir = both, y explicit]
 table[row sep=crcr, y error plus index=2, y error minus index=3]{%
500	0.289098197589188	0.0145321322507454	0.0145321322507454\\
675	0.293493401242995	0.0102980222657699	0.0102980222657699\\
850	0.286996927604687	0.00914865229267218	0.00914865229267218\\
1025	0.288452844677807	0.00862581228827856	0.00862581228827856\\
1200	0.287664147633234	0.0068928932056448	0.0068928932056448\\
1375	0.284832630689552	0.0065670519519461	0.0065670519519461\\
1550	0.287107518751468	0.00607202361003293	0.00607202361003293\\
1725	0.285864326709854	0.00503707856751491	0.00503707856751491\\
1900	0.284025915737145	0.0052808419322579	0.0052808419322579\\
2075	0.285551837656606	0.0048287952169118	0.0048287952169118\\
2250	0.284837508668308	0.00420573982726692	0.00420573982726692\\
2425	0.283930506170519	0.00391908847950946	0.00391908847950946\\
2600	0.284281534522932	0.00372319730787073	0.00372319730787073\\
2775	0.284277040175848	0.00332100292936668	0.00332100292936668\\
2950	0.28269154625134	0.0030119930138506	0.0030119930138506\\
3125	0.284425180360213	0.0028255739200589	0.0028255739200589\\
3300	0.283745571113602	0.00236351375271325	0.00236351375271325\\
3475	0.282519444527837	0.00212113281535919	0.00212113281535919\\
3650	0.284368726302833	0.00192522925196445	0.00192522925196445\\
3825	0.283411182606701	0.00160462166446891	0.00160462166446891\\
4000	0.282298635413652	0.00119613009529966	0.00119613009529966\\
};
\addlegendentry{$n_f$ = 0};

\addplot [color=red,dashed,line width=1.5pt]
 plot [error bars/.cd, y dir = both, y explicit]
 table[row sep=crcr, y error plus index=2, y error minus index=3]{%
500	0.27081743572465	0.0151461832555148	0.0151461832555148\\
675	0.273999030998501	0.0103555780380488	0.0103555780380488\\
850	0.267766828506565	0.00887198169067153	0.00887198169067153\\
1025	0.268899286621294	0.00888997378994678	0.00888997378994678\\
1200	0.26917334326382	0.00721266081748251	0.00721266081748251\\
1375	0.265277731112254	0.00645206638318307	0.00645206638318307\\
1550	0.268079405949568	0.00647150883576467	0.00647150883576467\\
1725	0.266596932260199	0.00539928095397943	0.00539928095397943\\
1900	0.265034963809817	0.00527104574132127	0.00527104574132127\\
2075	0.266196732012886	0.00486213339751761	0.00486213339751761\\
2250	0.26550174724066	0.00434593852876568	0.00434593852876568\\
2425	0.264391175816755	0.00427722157997021	0.00427722157997021\\
2600	0.265099475164623	0.00407089008326109	0.00407089008326109\\
2775	0.264913678127443	0.00346116393472861	0.00346116393472861\\
2950	0.263827943588057	0.0033599045168743	0.0033599045168743\\
3125	0.265167419330251	0.00309725449758731	0.00309725449758731\\
3300	0.264431649478238	0.00285917361998007	0.00285917361998007\\
3475	0.263964096185375	0.00273610771410003	0.00273610771410003\\
3650	0.264890349206702	0.00237078416191444	0.00237078416191444\\
3825	0.264008792322133	0.0021015604000239	0.0021015604000239\\
4000	0.264250597488496	0.00214454781659997	0.00214454781659997\\
};
\addlegendentry{$n_f$ = 1};

\addplot [color=mycolor1,dashdotted,line width=1.5pt]
 plot [error bars/.cd, y dir = both, y explicit]
 table[row sep=crcr, y error plus index=2, y error minus index=3]{%
500	0.2055814983009	0.0148081504985152	0.0148081504985152\\
675	0.20915624237838	0.0108852187720009	0.0108852187720009\\
850	0.202576256044447	0.00917036661365273	0.00917036661365273\\
1025	0.204231162873852	0.00906519095311691	0.00906519095311691\\
1200	0.204358437302519	0.00756981401524884	0.00756981401524884\\
1375	0.200908328261235	0.00679599779702251	0.00679599779702251\\
1550	0.202752553477628	0.00675527119877517	0.00675527119877517\\
1725	0.201769721028638	0.00584085774099718	0.00584085774099718\\
1900	0.200355574623029	0.00546814234583103	0.00546814234583103\\
2075	0.201306638143905	0.00545929420708955	0.00545929420708955\\
2250	0.201127791024585	0.00464147037759537	0.00464147037759537\\
2425	0.199883844250013	0.00456620231443068	0.00456620231443068\\
2600	0.200402505838085	0.00427847409829389	0.00427847409829389\\
2775	0.200474788256506	0.00389422010838154	0.00389422010838154\\
2950	0.199451735314477	0.00384637756949918	0.00384637756949918\\
3125	0.200539527406684	0.00357915581338426	0.00357915581338426\\
3300	0.199966413064951	0.00345799304318188	0.00345799304318188\\
3475	0.199293128462804	0.00316108057149042	0.00316108057149042\\
3650	0.200265784041464	0.00297755549301315	0.00297755549301315\\
3825	0.199668272348928	0.00288198376778039	0.00288198376778039\\
4000	0.199443929068044	0.002737590040864	0.002737590040864\\
};
\addlegendentry{$n_f$ = 8};

\end{axis}
\end{tikzpicture}%
%

	\caption{Convergence of the Kendall coefficient of agreement. We determined mean $\pm$ standard deviation of the agreement in a Monte-Carlo simulation at different error levels $n_f$ for our sanity checks.}
	\label{fig:convergence}
\end{figure}
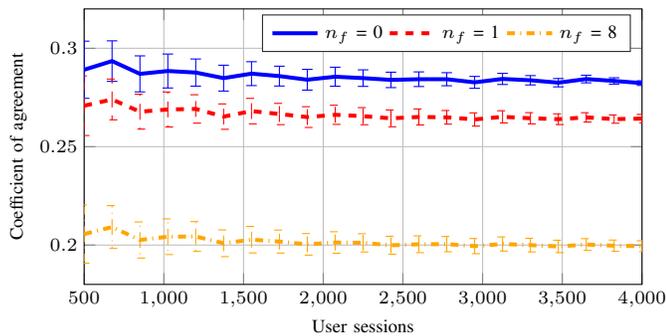

\Fref{fig:convergence} depicts Kendall's $u$ for different numbers of sessions included in the evaluation and different error levels $n_f \in \{0, 1, 8\}$ tolerated for the sanity checks. We performed a Monte-Carlo simulation similar to \cite{Lai2016} and randomly sampled the respective number of sessions. The mean and standard deviation of Kendall's $u$ was determined over 1,000 samples for each number of sessions. We found that Kendall's $u$ converges in terms of the standard deviation and becomes stable beyond 2,000 sessions with lower values for higher error levels $n_f$. The inter-observer variance is considerably higher if we omit the sanity checks (\ie $n_f = 8$). This result justifies the sanity checks that aim at removing outliers from the paired comparisons. We use $n_f = 1$ as a tradeoff between outlier removal and tolerating mistakenly entered votes of observers.

\subsection{Image Quality and Computation Time Tradeoff}

\begin{figure}[!t]
	\centering
	\scriptsize
	\setlength \figurewidth{0.46\textwidth}
	\setlength \figureheight{0.69\figurewidth}
	\subfloat[Mean B-T score vs. computation time ($2\times$ magnification)]{%
%
\begin{tikzpicture}

\begin{axis}[%
width=0.956\figurewidth,
height=\figureheight,
at={(0\figurewidth,0\figureheight)},
scale only axis,
xmode=log,
xmin=1.8,
xmax=2500,
xminorticks=true,
xlabel={Computation time [s]},
ymin=-2.5,
ymax=1.3,
ylabel={Mean B-T score},
axis background/.style={fill=white},
xmajorgrids,
xminorgrids,
ymajorgrids,
xlabel near ticks,ylabel near ticks,scaled y ticks=false,yticklabel style={/pgf/number format/fixed, /pgf/number format/precision=2},
]
\addplot [color=blue,only marks,mark=square*,mark options={solid,fill=blue},forget plot]
  table[row sep=crcr]{%
98.8508036	-0.103476315049352\\
};
\node[below right, align=left, text=black]
at (axis cs:98.851,-0.103) {EBSR};
\addplot [color=blue,only marks,mark=square*,mark options={solid,fill=blue},forget plot]
  table[row sep=crcr]{%
1112.6335734	0.892754519400259\\
};
\node[above right, align=left, text=black]
at (axis cs:1112.634,0.893) {ScSR};
\addplot [color=blue,only marks,mark=square*,mark options={solid,fill=blue},forget plot]
  table[row sep=crcr]{%
2.1614548	0.102528481712978\\
};
\node[below right, align=left, text=black]
at (axis cs:2.161,0.103) {NBSRF};
\addplot [color=red,only marks,mark=*,mark options={solid,fill=red},forget plot]
  table[row sep=crcr]{%
116.6395862	0.242740344311205\\
};
\node[above left, align=left, text=black]
at (axis cs:116.64,0.243) {VSRnet};
\addplot [color=red,only marks,mark=*,mark options={solid,fill=red},forget plot]
  table[row sep=crcr]{%
88.9483898	-0.348977287671961\\
};
\node[below left, align=left, text=black]
at (axis cs:88.948,-0.349) {NUISR};
\addplot [color=red,only marks,mark=*,mark options={solid,fill=red},forget plot]
  table[row sep=crcr]{%
84.1725538	-0.0172780302942174\\
};
\node[above right, align=left, text=black]
at (axis cs:84.173,-0.017) {WNUISR};
\addplot [color=red,only marks,mark=*,mark options={solid,fill=red},forget plot]
  table[row sep=crcr]{%
186.19319	-0.0286782111979787\\
};
\node[below right, align=left, text=black]
at (axis cs:186.193,-0.029) {HYSR};
\addplot [color=red,only marks,mark=*,mark options={solid,fill=red},forget plot]
  table[row sep=crcr]{%
368.2274878	0.0210286550398795\\
};
\node[below right, align=left, text=black]
at (axis cs:368.227,0.021) {DBRSR};
\addplot [color=red,only marks,mark=*,mark options={solid,fill=red},forget plot]
  table[row sep=crcr]{%
640.649143923301	1.05900690981381\\
};
\node[above right, align=left, text=black]
at (axis cs:640.649,1.059) {SRB};
\addplot [color=red,only marks,mark=*,mark options={solid,fill=red},forget plot]
  table[row sep=crcr]{%
104.8184712	0.773819451911175\\
};
\node[above left, align=left, text=black]
at (axis cs:104.818,0.774) {L1BTV};
\addplot [color=red,only marks,mark=*,mark options={solid,fill=red},forget plot]
  table[row sep=crcr]{%
399.8067354	0.795389807371986\\
};
\node[above right, align=left, text=black]
at (axis cs:399.807,0.795) {IRWSR};
\addplot [color=red,only marks,mark=*,mark options={solid,fill=red},forget plot]
  table[row sep=crcr]{%
248.56564	-2.41700166117318\\
};
\node[above right, align=left, text=black]
at (axis cs:248.566,-2.417) {BVSR};
\addplot [color=blue,only marks,mark=square*,mark options={solid,fill=blue},forget plot]
  table[row sep=crcr]{%
26.7126866	0.124776463253059\\
};
\node[below left, align=left, text=black]
at (axis cs:26.713,0.125) {SRCNN};
\addplot [color=red,only marks,mark=*,mark options={solid,fill=red},forget plot]
  table[row sep=crcr]{%
150.1940462	0.916848019373532\\
};
\node[above right, align=left, text=black]
at (axis cs:150.194,0.917) {BEPSR};
\addplot [color=blue,only marks,mark=square*,mark options={solid,fill=blue},forget plot]
  table[row sep=crcr]{%
684.088228244955	0.159983272746521\\
};
\node[above right, align=left, text=black]
at (axis cs:684.088,0.16) {SESR};
\addplot [color=blue,only marks,mark=square*,mark options={solid,fill=blue},forget plot]
  table[row sep=crcr]{%
426.1368944	0.100565569432627\\
};
\node[above left, align=left, text=black]
at (axis cs:426.137,0.101) {DRCN};
\addplot [color=blue,only marks,mark=square*,mark options={solid,fill=blue},forget plot]
  table[row sep=crcr]{%
32.0812772	0.1520108990873\\
};
\node[above left, align=left, text=black]
at (axis cs:32.081,0.152) {VDSR};
\addplot [color=blue,only marks,mark=square*,mark options={solid,fill=blue},forget plot]
  table[row sep=crcr]{%
11.637235	0.128846205063714\\
};
\node[above left, align=left, text=black]
at (axis cs:11.637,0.129) {A+};
\end{axis}
\end{tikzpicture}%
%
}\\
	\subfloat[Mean B-T score vs. computation time ($4\times$ magnification)]{%
%
\begin{tikzpicture}

\begin{axis}[%
width=0.956\figurewidth,
height=\figureheight,
at={(0\figurewidth,0\figureheight)},
scale only axis,
xmode=log,
xmin=1.8,
xmax=2500,
xminorticks=true,
xlabel={Computation time [s]},
ymin=-3.5,
ymax=1.9,
ylabel={Mean B-T score},
axis background/.style={fill=white},
xmajorgrids,
xminorgrids,
ymajorgrids,
xlabel near ticks,ylabel near ticks,scaled y ticks=false,yticklabel style={/pgf/number format/fixed, /pgf/number format/precision=2},
]
\addplot [color=blue,only marks,mark=square*,mark options={solid,fill=blue},forget plot]
  table[row sep=crcr]{%
171.736519	-0.071723092182609\\
};
\node[above right, align=left, text=black]
at (axis cs:171.737,-0.072) {EBSR};
\addplot [color=blue,only marks,mark=square*,mark options={solid,fill=blue},forget plot]
  table[row sep=crcr]{%
1002.8634222	0.393778724686659\\
};
\node[above right, align=left, text=black]
at (axis cs:1002.863,0.394) {ScSR};
\addplot [color=blue,only marks,mark=square*,mark options={solid,fill=blue},forget plot]
  table[row sep=crcr]{%
21.3740906	0.480386966968694\\
};
\node[above left, align=left, text=black]
at (axis cs:21.374,0.48) {NBSRF};
\addplot [color=red,only marks,mark=*,mark options={solid,fill=red},forget plot]
  table[row sep=crcr]{%
119.3518032	0.714504349029375\\
};
\node[below left, align=left, text=black]
at (axis cs:119.352,0.715) {VSRnet};
\addplot [color=red,only marks,mark=*,mark options={solid,fill=red},forget plot]
  table[row sep=crcr]{%
87.8762854	-1.73368006797886\\
};
\node[above right, align=left, text=black]
at (axis cs:87.876,-1.734) {NUISR};
\addplot [color=red,only marks,mark=*,mark options={solid,fill=red},forget plot]
  table[row sep=crcr]{%
85.1500134	-0.179135698503267\\
};
\node[above left, align=left, text=black]
at (axis cs:85.15,-0.179) {WNUISR};
\addplot [color=red,only marks,mark=*,mark options={solid,fill=red},forget plot]
  table[row sep=crcr]{%
258.3917378	-0.897644764735103\\
};
\node[above right, align=left, text=black]
at (axis cs:258.392,-0.898) {HYSR};
\addplot [color=red,only marks,mark=*,mark options={solid,fill=red},forget plot]
  table[row sep=crcr]{%
451.8890986	-0.241941212518542\\
};
\node[above right, align=left, text=black]
at (axis cs:451.889,-0.242) {DBRSR};
\addplot [color=red,only marks,mark=*,mark options={solid,fill=red},forget plot]
  table[row sep=crcr]{%
1491.40392944781	1.10681662829315\\
};
\node[above right, align=left, text=black]
at (axis cs:1491.404,1.107) {SRB};
\addplot [color=red,only marks,mark=*,mark options={solid,fill=red},forget plot]
  table[row sep=crcr]{%
258.2068604	1.10187279705925\\
};
\node[above left, align=left, text=black]
at (axis cs:258.207,1.102) {L1BTV};
\addplot [color=red,only marks,mark=*,mark options={solid,fill=red},forget plot]
  table[row sep=crcr]{%
410.0523792	1.18886967075448\\
};
\node[below right, align=left, text=black]
at (axis cs:410.052,1.189) {IRWSR};
\addplot [color=red,only marks,mark=*,mark options={solid,fill=red},forget plot]
  table[row sep=crcr]{%
1507.8252374	-3.30053953785429\\
};
\node[above left, align=left, text=black]
at (axis cs:1507.825,-3.301) {BVSR};
\addplot [color=blue,only marks,mark=square*,mark options={solid,fill=blue},forget plot]
  table[row sep=crcr]{%
29.5376378	0.756942318474413\\
};
\node[above right, align=left, text=black]
at (axis cs:29.538,0.757) {SRCNN};
\addplot [color=red,only marks,mark=*,mark options={solid,fill=red},forget plot]
  table[row sep=crcr]{%
897.0671582	1.53003828324768\\
};
\node[above right, align=left, text=black]
at (axis cs:897.067,1.53) {BEPSR};
\addplot [color=blue,only marks,mark=square*,mark options={solid,fill=blue},forget plot]
  table[row sep=crcr]{%
416.150394986427	0.724587135885472\\
};
\node[below right, align=left, text=black]
at (axis cs:416.15,0.725) {SESR};
\addplot [color=blue,only marks,mark=square*,mark options={solid,fill=blue},forget plot]
  table[row sep=crcr]{%
472.461677	1.4039710693179\\
};
\node[above left, align=left, text=black]
at (axis cs:472.462,1.404) {DRCN};
\addplot [color=blue,only marks,mark=square*,mark options={solid,fill=blue},forget plot]
  table[row sep=crcr]{%
45.3557678	1.3434512508086\\
};
\node[above right, align=left, text=black]
at (axis cs:45.356,1.343) {VDSR};
\addplot [color=blue,only marks,mark=square*,mark options={solid,fill=blue},forget plot]
  table[row sep=crcr]{%
9.4125812	0.357052409592026\\
};
\node[below left, align=left, text=black]
at (axis cs:9.413,0.357) {A+};
\end{axis}
\end{tikzpicture}%
%
}
	\caption{Tradeoff between image quality in terms of B-T scores and the computation times for $2\times$ and $4\times$ magnification. We compare SISR (shown in blue) and MFSR methods (shown in red).}
	\label{fig:qualityTimeTradeoff}
\end{figure}
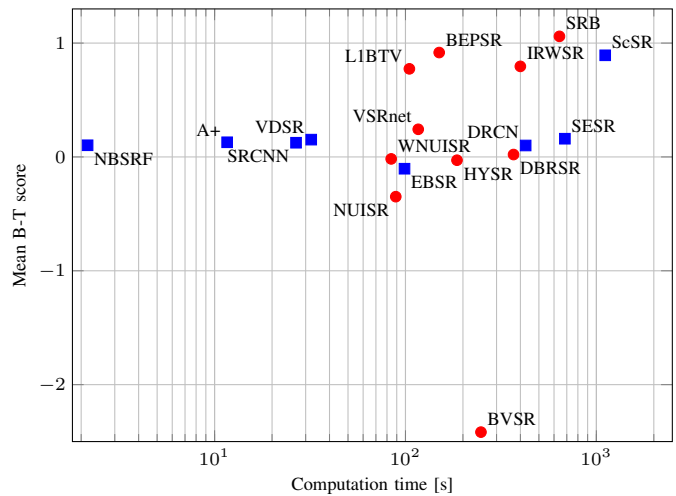
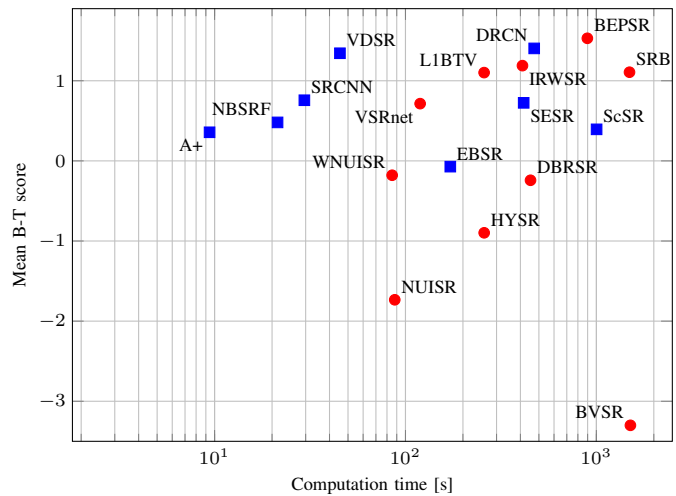

Computation time is relevant to many practical SR applications.
For all methods, MATLAB sources on the CPU are used, where some modules are accelerated by C++. SR is computed to the 1200$\times$960 pixels resolution of the ground truth images and computation times for MFSR include optical flow estimation. \Fref{fig:qualityTimeTradeoff} shows the tradeoff between image quality in the B-T model and computation time for $2\times$ and $4\times$ magnification. We identify three classes of algorithms: 
\begin{itemize}
	\item Five SISR (EBSR, NBSRF, SRCNN, DRCN, VDSR) and six MFSR methods (HYSR, DBSR, SRB, L1BTV, BEPSR, BVSR) have higher complexities for larger magnifications. The computation time of MFSR methods additionally depends on the number of input images.
	\item The computation time of three SISR methods (ScSR, A+, SESR) are mainly influenced by the input image resolution, but not by magnification.
	\item The computation time of four MFSR methods (VSRnet, NUISR, WNUISR, IRWSR) is unaffected by these factors.
\end{itemize}

Overall, VDSR and SRCNN -- two recent deep learning methods -- have excellent quality/time tradeoffs. For small magnifications, ScSR yields higher quality but is much slower. NBSRF and A+ are faster but achieve lower quality. In terms of MFSR, non-blind reconstruction (L1BTV, IRWSR, BEPSR) as well as VSRnet show good quality/time tradeoffs. The interpolation-based NUISR and WNUISR are slightly faster but achieve lower image quality. However, interpolation-based SR doubles the complexity of motion estimation as optical flow is computed in forward and backward direction \cite{Batz2016}.

\subsection{Correlation to Quantitative Study}
\label{sec:CorrelationToQuantitativeStudy}

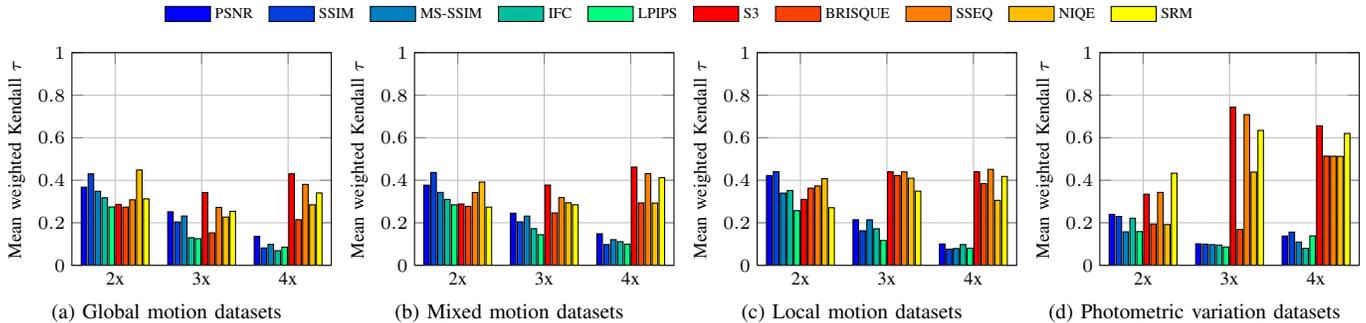
\begin{figure*}[!t]
	\scriptsize 
	\centering
	\setlength \figurewidth{1.0\textwidth}
	\subfloat{%
%
\definecolor{mycolor1}{rgb}{0.00000,0.25000,0.87500}%
\definecolor{mycolor2}{rgb}{0.00000,0.50000,0.75000}%
\definecolor{mycolor3}{rgb}{0.00000,0.75000,0.62500}%
\definecolor{mycolor4}{rgb}{0.00000,1.00000,0.50000}%
\definecolor{mycolor5}{rgb}{1.00000,0.75000,0.00000}%
\definecolor{mycolor6}{rgb}{1.00000,1.00000,0.00000}%
\begin{tikzpicture}
\scriptsize
\begin{axis}[%
    const plot,
		hide axis,
    xmin=10,
    xmax=50,
    ymin=0,
    ymax=0,
    legend style={
			legend image code/.code={2mm
				\draw [#1] (0cm,-0.065cm) rectangle (0.25cm,0.065cm);
			},
			anchor=north,
			inner ysep=0pt,
			draw=none, 
			legend cell align=left,
			legend columns=-1, 
			/tikz/every even column/.append style={column sep=0.15cm},
			nodes={scale=0.815, transform shape}
		}
	]
	\addlegendimage{black, mark=none, fill=blue, area legend}
	\addlegendentry{PSNR}
	\addlegendimage{black, mark=none, fill=mycolor1, area legend}
	\addlegendentry{SSIM}
	\addlegendimage{black, mark=none, fill=mycolor2, area legend}
	\addlegendentry{MS-SSIM}
	\addlegendimage{black, mark=none, fill=mycolor3, area legend}
	\addlegendentry{IFC}
	\addlegendimage{black, mark=none, fill=mycolor4, area legend}
	\addlegendentry{LPIPS}
	\addlegendimage{black, mark=none, fill=red, area legend}
	\addlegendentry{S3}
	\addlegendimage{black, mark=none, fill=red!50!orange, area legend}
	\addlegendentry{BRISQUE}
	\addlegendimage{black, mark=none, fill=orange, area legend}
	\addlegendentry{SSEQ}
	\addlegendimage{black, mark=none, fill=mycolor5, area legend}
	\addlegendentry{NIQE}
	\addlegendimage{black, mark=none, fill=mycolor6, area legend}
	\addlegendentry{SRM}
\end{axis}

\end{tikzpicture}%
%
}\\[-1.25ex]
	\setlength \figurewidth{0.20\textwidth}
	\setlength \figureheight{0.78\figurewidth}
	\setcounter{subfigure}{0}
	\subfloat[Global motion datasets]{%
%
\definecolor{mycolor1}{rgb}{0.00000,0.25000,0.87500}%
\definecolor{mycolor2}{rgb}{0.00000,0.50000,0.75000}%
\definecolor{mycolor3}{rgb}{0.00000,0.75000,0.62500}%
\definecolor{mycolor4}{rgb}{0.00000,1.00000,0.50000}%
\definecolor{mycolor5}{rgb}{1.00000,0.75000,0.00000}%
\definecolor{mycolor6}{rgb}{1.00000,1.00000,0.00000}%
\begin{tikzpicture}

\begin{axis}[%
width=0.951\figurewidth,
height=\figureheight,
at={(0\figurewidth,0\figureheight)},
scale only axis,
xmin=1.5,
xmax=4.5,
xtick={2,3,4},
xticklabels={{2x},{3x},{4x}},
ymin=0,
ymax=1,
ylabel={Mean weighted Kendall $\tau$},
axis background/.style={fill=white},
xmajorgrids,
ymajorgrids,
xlabel near ticks,ylabel near ticks,scaled y ticks=false,yticklabel style={/pgf/number format/fixed, /pgf/number format/precision=2},
]
\addplot[ybar,bar width=0.064,bar shift=-0.36,fill=blue,draw=black,area legend] plot table[row sep=crcr] {%
2	0.36746836669201\\
3	0.251520056239384\\
4	0.136536927299562\\
};
\addplot[forget plot,color=white!15!black] table[row sep=crcr] {%
1.5	0\\
4.5	0\\
};
\addplot[ybar,bar width=0.064,bar shift=-0.28,fill=mycolor1,draw=black,area legend] plot table[row sep=crcr] {%
2	0.429981920640484\\
3	0.203905071330161\\
4	0.0818139266229345\\
};
\addplot[forget plot,color=white!15!black] table[row sep=crcr] {%
1.5	0\\
4.5	0\\
};
\addplot[ybar,bar width=0.064,bar shift=-0.2,fill=mycolor2,draw=black,area legend] plot table[row sep=crcr] {%
2	0.348267014144361\\
3	0.231749597843123\\
4	0.0985343721643267\\
};
\addplot[forget plot,color=white!15!black] table[row sep=crcr] {%
1.5	0\\
4.5	0\\
};
\addplot[ybar,bar width=0.064,bar shift=-0.12,fill=mycolor3,draw=black,area legend] plot table[row sep=crcr] {%
2	0.317542492032205\\
3	0.129980218339622\\
4	0.0687895756989917\\
};
\addplot[forget plot,color=white!15!black] table[row sep=crcr] {%
1.5	0\\
4.5	0\\
};
\addplot[ybar,bar width=0.064,bar shift=-0.04,fill=mycolor4,draw=black,area legend] plot table[row sep=crcr] {%
2	0.274356242302701\\
3	0.125367038232185\\
4	0.085662007870499\\
};
\addplot[forget plot,color=white!15!black] table[row sep=crcr] {%
1.5	0\\
4.5	0\\
};
\addplot[ybar,bar width=0.064,bar shift=0.04,fill=red,draw=black,area legend] plot table[row sep=crcr] {%
2	0.286330563400485\\
3	0.342378451466896\\
4	0.430277729115444\\
};
\addplot[forget plot,color=white!15!black] table[row sep=crcr] {%
1.5	0\\
4.5	0\\
};
\addplot[ybar,bar width=0.064,bar shift=0.12,fill=red!50!orange,draw=black,area legend] plot table[row sep=crcr] {%
2	0.273653142579752\\
3	0.152632614171159\\
4	0.214479874473277\\
};
\addplot[forget plot,color=white!15!black] table[row sep=crcr] {%
1.5	0\\
4.5	0\\
};
\addplot[ybar,bar width=0.064,bar shift=0.2,fill=orange,draw=black,area legend] plot table[row sep=crcr] {%
2	0.307952254748294\\
3	0.272234961291022\\
4	0.380524377764076\\
};
\addplot[forget plot,color=white!15!black] table[row sep=crcr] {%
1.5	0\\
4.5	0\\
};
\addplot[ybar,bar width=0.064,bar shift=0.28,fill=mycolor5,draw=black,area legend] plot table[row sep=crcr] {%
2	0.448254332819215\\
3	0.227006481966336\\
4	0.284654011220297\\
};
\addplot[forget plot,color=white!15!black] table[row sep=crcr] {%
1.5	0\\
4.5	0\\
};
\addplot[ybar,bar width=0.064,bar shift=0.36,fill=mycolor6,draw=black,area legend] plot table[row sep=crcr] {%
2	0.312439650614806\\
3	0.253917932203989\\
4	0.340328267620976\\
};
\addplot[forget plot,color=white!15!black] table[row sep=crcr] {%
1.5	0\\
4.5	0\\
};
\end{axis}
\end{tikzpicture}%
%
}~
	\subfloat[Mixed motion datasets]{%
%
\definecolor{mycolor1}{rgb}{0.00000,0.25000,0.87500}%
\definecolor{mycolor2}{rgb}{0.00000,0.50000,0.75000}%
\definecolor{mycolor3}{rgb}{0.00000,0.75000,0.62500}%
\definecolor{mycolor4}{rgb}{0.00000,1.00000,0.50000}%
\definecolor{mycolor5}{rgb}{1.00000,0.75000,0.00000}%
\definecolor{mycolor6}{rgb}{1.00000,1.00000,0.00000}%
\begin{tikzpicture}

\begin{axis}[%
width=0.951\figurewidth,
height=\figureheight,
at={(0\figurewidth,0\figureheight)},
scale only axis,
xmin=1.5,
xmax=4.5,
xtick={2,3,4},
xticklabels={{2x},{3x},{4x}},
ymin=0,
ymax=1,
ylabel={Mean weighted Kendall $\tau$},
axis background/.style={fill=white},
xmajorgrids,
ymajorgrids,
xlabel near ticks,ylabel near ticks,scaled y ticks=false,yticklabel style={/pgf/number format/fixed, /pgf/number format/precision=2},
]
\addplot[ybar,bar width=0.064,bar shift=-0.36,fill=blue,draw=black,area legend] plot table[row sep=crcr] {%
2	0.376859042650676\\
3	0.24524983226207\\
4	0.148239637392321\\
};
\addplot[forget plot,color=white!15!black] table[row sep=crcr] {%
1.5	0\\
4.5	0\\
};
\addplot[ybar,bar width=0.064,bar shift=-0.28,fill=mycolor1,draw=black,area legend] plot table[row sep=crcr] {%
2	0.436655492434561\\
3	0.20431951775563\\
4	0.0976053184722493\\
};
\addplot[forget plot,color=white!15!black] table[row sep=crcr] {%
1.5	0\\
4.5	0\\
};
\addplot[ybar,bar width=0.064,bar shift=-0.2,fill=mycolor2,draw=black,area legend] plot table[row sep=crcr] {%
2	0.342928460420536\\
3	0.231435638569709\\
4	0.120489371401992\\
};
\addplot[forget plot,color=white!15!black] table[row sep=crcr] {%
1.5	0\\
4.5	0\\
};
\addplot[ybar,bar width=0.064,bar shift=-0.12,fill=mycolor3,draw=black,area legend] plot table[row sep=crcr] {%
2	0.310049155843\\
3	0.172346392572945\\
4	0.111050782293942\\
};
\addplot[forget plot,color=white!15!black] table[row sep=crcr] {%
1.5	0\\
4.5	0\\
};
\addplot[ybar,bar width=0.064,bar shift=-0.04,fill=mycolor4,draw=black,area legend] plot table[row sep=crcr] {%
2	0.284464734833673\\
3	0.143902890079779\\
4	0.0995010471952417\\
};
\addplot[forget plot,color=white!15!black] table[row sep=crcr] {%
1.5	0\\
4.5	0\\
};
\addplot[ybar,bar width=0.064,bar shift=0.04,fill=red,draw=black,area legend] plot table[row sep=crcr] {%
2	0.288395715852859\\
3	0.377318202698963\\
4	0.462019283610693\\
};
\addplot[forget plot,color=white!15!black] table[row sep=crcr] {%
1.5	0\\
4.5	0\\
};
\addplot[ybar,bar width=0.064,bar shift=0.12,fill=red!50!orange,draw=black,area legend] plot table[row sep=crcr] {%
2	0.277279680488078\\
3	0.246230454686107\\
4	0.293449150011553\\
};
\addplot[forget plot,color=white!15!black] table[row sep=crcr] {%
1.5	0\\
4.5	0\\
};
\addplot[ybar,bar width=0.064,bar shift=0.2,fill=orange,draw=black,area legend] plot table[row sep=crcr] {%
2	0.342428369551655\\
3	0.318908528301888\\
4	0.431179057398485\\
};
\addplot[forget plot,color=white!15!black] table[row sep=crcr] {%
1.5	0\\
4.5	0\\
};
\addplot[ybar,bar width=0.064,bar shift=0.28,fill=mycolor5,draw=black,area legend] plot table[row sep=crcr] {%
2	0.391464544851686\\
3	0.29407076356111\\
4	0.292837107355435\\
};
\addplot[forget plot,color=white!15!black] table[row sep=crcr] {%
1.5	0\\
4.5	0\\
};
\addplot[ybar,bar width=0.064,bar shift=0.36,fill=mycolor6,draw=black,area legend] plot table[row sep=crcr] {%
2	0.273742378036893\\
3	0.284840334482795\\
4	0.412318229712398\\
};
\addplot[forget plot,color=white!15!black] table[row sep=crcr] {%
1.5	0\\
4.5	0\\
};
\end{axis}
\end{tikzpicture}%
%
}~
	\subfloat[Local motion datasets]{%
%
\definecolor{mycolor1}{rgb}{0.00000,0.25000,0.87500}%
\definecolor{mycolor2}{rgb}{0.00000,0.50000,0.75000}%
\definecolor{mycolor3}{rgb}{0.00000,0.75000,0.62500}%
\definecolor{mycolor4}{rgb}{0.00000,1.00000,0.50000}%
\definecolor{mycolor5}{rgb}{1.00000,0.75000,0.00000}%
\definecolor{mycolor6}{rgb}{1.00000,1.00000,0.00000}%
\begin{tikzpicture}

\begin{axis}[%
width=0.951\figurewidth,
height=\figureheight,
at={(0\figurewidth,0\figureheight)},
scale only axis,
xmin=1.5,
xmax=4.5,
xtick={2,3,4},
xticklabels={{2x},{3x},{4x}},
ymin=0,
ymax=1,
ylabel={Mean weighted Kendall $\tau$},
axis background/.style={fill=white},
xmajorgrids,
ymajorgrids,
xlabel near ticks,ylabel near ticks,scaled y ticks=false,yticklabel style={/pgf/number format/fixed, /pgf/number format/precision=2},
]
\addplot[ybar,bar width=0.064,bar shift=-0.36,fill=blue,draw=black,area legend] plot table[row sep=crcr] {%
2	0.421597653638181\\
3	0.214224851174427\\
4	0.100432863713421\\
};
\addplot[forget plot,color=white!15!black] table[row sep=crcr] {%
1.5	0\\
4.5	0\\
};
\addplot[ybar,bar width=0.064,bar shift=-0.28,fill=mycolor1,draw=black,area legend] plot table[row sep=crcr] {%
2	0.440361364127441\\
3	0.162579369093161\\
4	0.0762592157922487\\
};
\addplot[forget plot,color=white!15!black] table[row sep=crcr] {%
1.5	0\\
4.5	0\\
};
\addplot[ybar,bar width=0.064,bar shift=-0.2,fill=mycolor2,draw=black,area legend] plot table[row sep=crcr] {%
2	0.339591562368267\\
3	0.214033296787845\\
4	0.079100803042113\\
};
\addplot[forget plot,color=white!15!black] table[row sep=crcr] {%
1.5	0\\
4.5	0\\
};
\addplot[ybar,bar width=0.064,bar shift=-0.12,fill=mycolor3,draw=black,area legend] plot table[row sep=crcr] {%
2	0.3519067976151\\
3	0.171921780586395\\
4	0.0976402623439305\\
};
\addplot[forget plot,color=white!15!black] table[row sep=crcr] {%
1.5	0\\
4.5	0\\
};
\addplot[ybar,bar width=0.064,bar shift=-0.04,fill=mycolor4,draw=black,area legend] plot table[row sep=crcr] {%
2	0.25693445767305\\
3	0.116941026522658\\
4	0.080732674589808\\
};
\addplot[forget plot,color=white!15!black] table[row sep=crcr] {%
1.5	0\\
4.5	0\\
};
\addplot[ybar,bar width=0.064,bar shift=0.04,fill=red,draw=black,area legend] plot table[row sep=crcr] {%
2	0.310124502077691\\
3	0.439684511950834\\
4	0.439769613898251\\
};
\addplot[forget plot,color=white!15!black] table[row sep=crcr] {%
1.5	0\\
4.5	0\\
};
\addplot[ybar,bar width=0.064,bar shift=0.12,fill=red!50!orange,draw=black,area legend] plot table[row sep=crcr] {%
2	0.362808428965034\\
3	0.42243974575704\\
4	0.384150132250472\\
};
\addplot[forget plot,color=white!15!black] table[row sep=crcr] {%
1.5	0\\
4.5	0\\
};
\addplot[ybar,bar width=0.064,bar shift=0.2,fill=orange,draw=black,area legend] plot table[row sep=crcr] {%
2	0.373213738605943\\
3	0.439197601073492\\
4	0.451341876500328\\
};
\addplot[forget plot,color=white!15!black] table[row sep=crcr] {%
1.5	0\\
4.5	0\\
};
\addplot[ybar,bar width=0.064,bar shift=0.28,fill=mycolor5,draw=black,area legend] plot table[row sep=crcr] {%
2	0.407362293722427\\
3	0.409522801485605\\
4	0.30555860389295\\
};
\addplot[forget plot,color=white!15!black] table[row sep=crcr] {%
1.5	0\\
4.5	0\\
};
\addplot[ybar,bar width=0.064,bar shift=0.36,fill=mycolor6,draw=black,area legend] plot table[row sep=crcr] {%
2	0.270896052202025\\
3	0.348931527101646\\
4	0.417992708521052\\
};
\addplot[forget plot,color=white!15!black] table[row sep=crcr] {%
1.5	0\\
4.5	0\\
};
\end{axis}
\end{tikzpicture}%
%
}~
	\subfloat[Photometric variation datasets]{%
%
\definecolor{mycolor1}{rgb}{0.00000,0.25000,0.87500}%
\definecolor{mycolor2}{rgb}{0.00000,0.50000,0.75000}%
\definecolor{mycolor3}{rgb}{0.00000,0.75000,0.62500}%
\definecolor{mycolor4}{rgb}{0.00000,1.00000,0.50000}%
\definecolor{mycolor5}{rgb}{1.00000,0.75000,0.00000}%
\definecolor{mycolor6}{rgb}{1.00000,1.00000,0.00000}%
\begin{tikzpicture}

\begin{axis}[%
width=0.951\figurewidth,
height=\figureheight,
at={(0\figurewidth,0\figureheight)},
scale only axis,
xmin=1.5,
xmax=4.5,
xtick={2,3,4},
xticklabels={{2x},{3x},{4x}},
ymin=0,
ymax=1,
ylabel={Mean weighted Kendall $\tau$},
axis background/.style={fill=white},
xmajorgrids,
ymajorgrids,
xlabel near ticks,ylabel near ticks,scaled y ticks=false,yticklabel style={/pgf/number format/fixed, /pgf/number format/precision=2},
]
\addplot[ybar,bar width=0.064,bar shift=-0.36,fill=blue,draw=black,area legend] plot table[row sep=crcr] {%
2	0.239743900371621\\
3	0.101092018660946\\
4	0.137826664483441\\
};
\addplot[forget plot,color=white!15!black] table[row sep=crcr] {%
1.5	0\\
4.5	0\\
};
\addplot[ybar,bar width=0.064,bar shift=-0.28,fill=mycolor1,draw=black,area legend] plot table[row sep=crcr] {%
2	0.22995498669273\\
3	0.099781096977257\\
4	0.156207738522494\\
};
\addplot[forget plot,color=white!15!black] table[row sep=crcr] {%
1.5	0\\
4.5	0\\
};
\addplot[ybar,bar width=0.064,bar shift=-0.2,fill=mycolor2,draw=black,area legend] plot table[row sep=crcr] {%
2	0.157512889575794\\
3	0.0971157596564312\\
4	0.109477985663053\\
};
\addplot[forget plot,color=white!15!black] table[row sep=crcr] {%
1.5	0\\
4.5	0\\
};
\addplot[ybar,bar width=0.064,bar shift=-0.12,fill=mycolor3,draw=black,area legend] plot table[row sep=crcr] {%
2	0.221452386546988\\
3	0.0952715390287701\\
4	0.0801152421973712\\
};
\addplot[forget plot,color=white!15!black] table[row sep=crcr] {%
1.5	0\\
4.5	0\\
};
\addplot[ybar,bar width=0.064,bar shift=-0.04,fill=mycolor4,draw=black,area legend] plot table[row sep=crcr] {%
2	0.158611729714892\\
3	0.0863905846144335\\
4	0.138433687165689\\
};
\addplot[forget plot,color=white!15!black] table[row sep=crcr] {%
1.5	0\\
4.5	0\\
};
\addplot[ybar,bar width=0.064,bar shift=0.04,fill=red,draw=black,area legend] plot table[row sep=crcr] {%
2	0.334550858781297\\
3	0.743573195726989\\
4	0.656204593760037\\
};
\addplot[forget plot,color=white!15!black] table[row sep=crcr] {%
1.5	0\\
4.5	0\\
};
\addplot[ybar,bar width=0.064,bar shift=0.12,fill=red!50!orange,draw=black,area legend] plot table[row sep=crcr] {%
2	0.194210805294468\\
3	0.168588561848625\\
4	0.513525284913366\\
};
\addplot[forget plot,color=white!15!black] table[row sep=crcr] {%
1.5	0\\
4.5	0\\
};
\addplot[ybar,bar width=0.064,bar shift=0.2,fill=orange,draw=black,area legend] plot table[row sep=crcr] {%
2	0.342967060855975\\
3	0.709042075479038\\
4	0.513083603006455\\
};
\addplot[forget plot,color=white!15!black] table[row sep=crcr] {%
1.5	0\\
4.5	0\\
};
\addplot[ybar,bar width=0.064,bar shift=0.28,fill=mycolor5,draw=black,area legend] plot table[row sep=crcr] {%
2	0.191395143958418\\
3	0.438469678051852\\
4	0.512367861093996\\
};
\addplot[forget plot,color=white!15!black] table[row sep=crcr] {%
1.5	0\\
4.5	0\\
};
\addplot[ybar,bar width=0.064,bar shift=0.36,fill=mycolor6,draw=black,area legend] plot table[row sep=crcr] {%
2	0.433289641423124\\
3	0.63449639380107\\
4	0.620375016369228\\
};
\addplot[forget plot,color=white!15!black] table[row sep=crcr] {%
1.5	0\\
4.5	0\\
};
\end{axis}
\end{tikzpicture}%
%
}
	\caption{Mean weighted Kendall $\tau$ distance among the B-T scores and different quantitative quality measures. The lower the weighted Kendall $\tau$ the higher the correlation to human visual perception expressed by the B-T model. We compare five full-reference measures (shown with blue color map) as well as five no-reference measures (shown with red color map) for different datasets and magnification factors.}
	\label{fig:correlation}
\end{figure*}

To analyze the correlation between quantitative quality assessment and human visual perception expressed by the B-T model, we employ the weighted Kendall $\tau$ distance \cite{Liu2013}. This compares algorithm rankings obtained by the quantitative measures to rankings in the B-T model, where lower distances express stronger correlations. \Fref{fig:correlation} shows the mean weighted Kendall $\tau$ (normalized to $[0, 1]$) of ten measures for different datasets and magnification factors. We found that for most measures their respective weighted Kendall $\tau$ depends on the magnification factor. In case of full-reference measures, the higher the magnification the lower the weighted Kendall $\tau$. This is consistent with prior studies \cite{Blau2018, Ma2017} and suggests that quantitative evaluations should focus on higher magnifications if one desires fair correlations to human visual perception.

For no-reference quality assessment, BRISQUE shows the strongest correlations for most datasets and magnification factors. For the full-reference measures, LPIPS and IFC show high correlations, while the commonly used PSNR, SSIM, and MS-SSIM are often weaker indicators for human visual perception. Particularly, the deep features exploited by LPIPS yields a more suitable perceptual measure than the hand-crafted features. Most full-reference measures also outperform their no-reference counterparts, especially for larger magnifications. This is remarkable as prior work \cite{Blau2018} proposed to use no-reference measures to quantify perceptual quality. We explain this contradiction by the fact that current no-reference measures are either generic (\eg NIQE) or customized to SR but developed under simplified conditions (\eg SRM). For instance, SRM was trained from SISR results on simulated data, which explains the lower performance on real data with MFSR facets. The weak correlations are particularly noticeable under photometric variations. These observations are other occurrences of the simulated-to-real gap and also underline the importance of ground truth data for SR benchmarking.

\section{Discussion}
\label{sec:Discussion}

The benchmark reveals several interesting conclusions related to the simulated-to-real gap and provides guidelines for future research.

\subsection{Remarks on the Quantitative Study}

In SISR, the use of external training data outperforms self-exemplar approaches \cite{Huang2015a}. Surprisingly, depending on the quality measure, popular deep nets \cite{Dong2014,Kim2016,Kim2016a} do not clearly outperform classical methods \cite{Kim2010,Timofte2015,Salvador2015,Yang2010}. This contrasts benchmarks on simulated data, where deep nets perform best.

We also found that MFSR surpasses SISR in baseline experiments with pure global motion. Despite the success of learning-based methods, classical sparsity priors \cite{Farsiu2004a,Kohler2015c,Zeng2013} are still invaluable in this field. We consider the further development of such priors as well as their combination with learning-based architectures as a promising way.

SR quality heavily depends on environmental conditions.  Generally, MFSR is sensitive to failures in optical flow estimation. This explains the weaker performance on mixed or local motion as well as photometric variations, while SISR is unaffected by these factors. Within MFSR, reconstruction \cite{Farsiu2004a,Kohler2015c,Zeng2013} or deep learning \cite{Kappeler2016} methods are more robust than interpolation-based approaches \cite{Batz2016,Batz2016b,Park2003}. 

Video compression, \eg H.265/HEVC, challenges all methods and MFSR in particular. The proposed dataset can be useful to create compressed training data for future learning-based methods. Also for future work, hybrid approaches to combine strengths of SISR and MFSR appear to be highly relevant, like \cite{Batz2015} or architectures like VSRnet \cite{Kappeler2016}. 

\subsection{Remarks on the Human Observer Study}

Visual perception of SR image quality in our observer study shows reasonable agreements to the algorithm rankings in the quantitative study. Specifically, classical reconstruction-based algorithms like \cite{Farsiu2004a,Kohler2015c,Zeng2013} (in MFSR) and learning-based methods like \cite{Yang2010,Kim2016,Kim2016a} (in SISR) are ranked highest in a B-T model derived from pair-wise comparisons.

Inter-observer variances heavily depend on SR parameters. Most importantly, our analysis shows that larger magnification factors lead to a higher consensus between different observers. In view of this finding, reliable evaluations should conduct SR with large magnifications if agreements to human visual perception are desired.

If low computational complexities are important, non-blind reconstruction algorithms \cite{Farsiu2004a,Kohler2015c,Zeng2013} and VSRnet \cite{Kappeler2016} (in MFSR) as well as deep learning \cite{Dong2014,Kim2016} (in SISR) feature good tradeoffs between image quality and computation time.

The actual correlation between human perception and quantitative benchmarks depends on the quality measure, as also reported in~\cite{Yang2014a,Lai2016}. The popular PSNR shows weak correlations compared to the information-theoretic IFC and the data-driven LPIPS. For future work, we encourage the use of such elaborated measures to reliably benchmark SR. Beyond that full-reference measures have stronger correlations compared to no-reference measures on real images, which underlines the importance of ground truth data and the need to improve current no-reference methods.

\section{Conclusion}
\label{sec:Conclusion}

This paper presented a database to conduct the largest SR benchmark to date, across both SISR and MFSR algorithms. This is the first dataset to combine image sequences of real LR acquisitions with ground truth data. Additionally, the size of the database is a magnitude larger than currently existing benchmarks. We also demonstrated that evaluations on simulated data do not necessarily reflect the performance of SR on real data. Our data captured by means of hardware binning with challenging effects like non-Gaussian noise provides an important step to close this simulated-to-real gap.

We plan to extend our benchmark by more state-of-the-art approaches that focus on reconstruction accuracy \cite{Lim2017, Zhang2018b}, perceptual image quality \cite{Ledig2017, Sajjadi2017, Wang2018}, or applications with limited computational resources \cite{Romano2016, Li2018}. We also encourage other authors to validate their future algorithms on our data and to provide their results to enable fair cross-comparisons among SR methods. Future works can also use the data to move model training from simulated to real images, and to perform thorough quantitative evaluations. Using the results of our human observer study, the benchmark can also serve as a testbed for image quality assessment methods.

\noindent
\textit{Limitations.}
Our database considers various types of environmental aspects (local motion and photometric variation), technological aspects (hardware binning) as well as software aspects (video coding) without involving simulations. We note that several facets of real data are, however, left for future work. For instance, sensor-specific artifacts like the color-filter array (CFA) or rolling shutter \cite{Punnappurath2015} are not acquired. Such artifacts are crucial when dealing with commercially available low-cost CMOS sensors. Also, the level of realism of outdoor scenes is sacrificed by our lab scenes and the acquisition of an accurate ground truth. One typical artifact in outdoor scenes is motion blur due to a freely moving camera. This could be addressed by \textit{recording-and-playback} of motion similar to \cite{Kohler2012}. In our future work, we plan to extend our database by such aspects towards benchmarking SR in the wild.

\section*{Acknowledgments}

We would like to thank all observers of our human observer study for their participation. We also thank the authors of prior works for providing the source codes for our benchmark. We are also grateful to the anonymous reviewers for their constructive comments that helped to improve this work.

\bibliographystyle{IEEEtran} 
\bibliography{egbib}      

%




\begin{IEEEbiography}[{\includegraphics[width=1in,height=1.25in,clip,keepaspectratio]{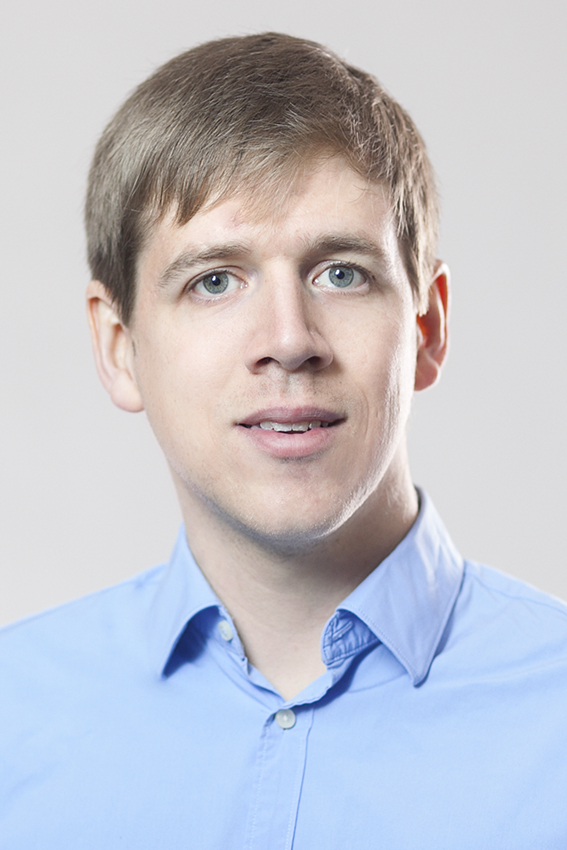}}]{Thomas K\"ohler}
received the M.Sc. degree in computer science at the University of Applied Sciences Munich, Munich, Germany in 2010 and the Ph.D. degree in computer science from the Friedrich-Alexander-Universtit\"at (FAU) Erlangen-N\"urnberg, Erlangen, Germany, in 2017.

From 2011 to 2016, he was a researcher at the Pattern Recognition Lab at FAU. From 2016 to 2017, he joined the Department of Neuroradiology at the University Clinics Erlangen, Germany. Since 2017, he is a Software Engineer for computer vision applications in the Cognitive Services and Augmented Reality Group at e.solutions GmbH, Erlangen, Germany. His research interests include machine learning, computer vision and image processing, especially for inverse imaging problems and computational photography.
\end{IEEEbiography}

\begin{IEEEbiography}[{\includegraphics[width=1in,height=1.25in,clip,keepaspectratio]{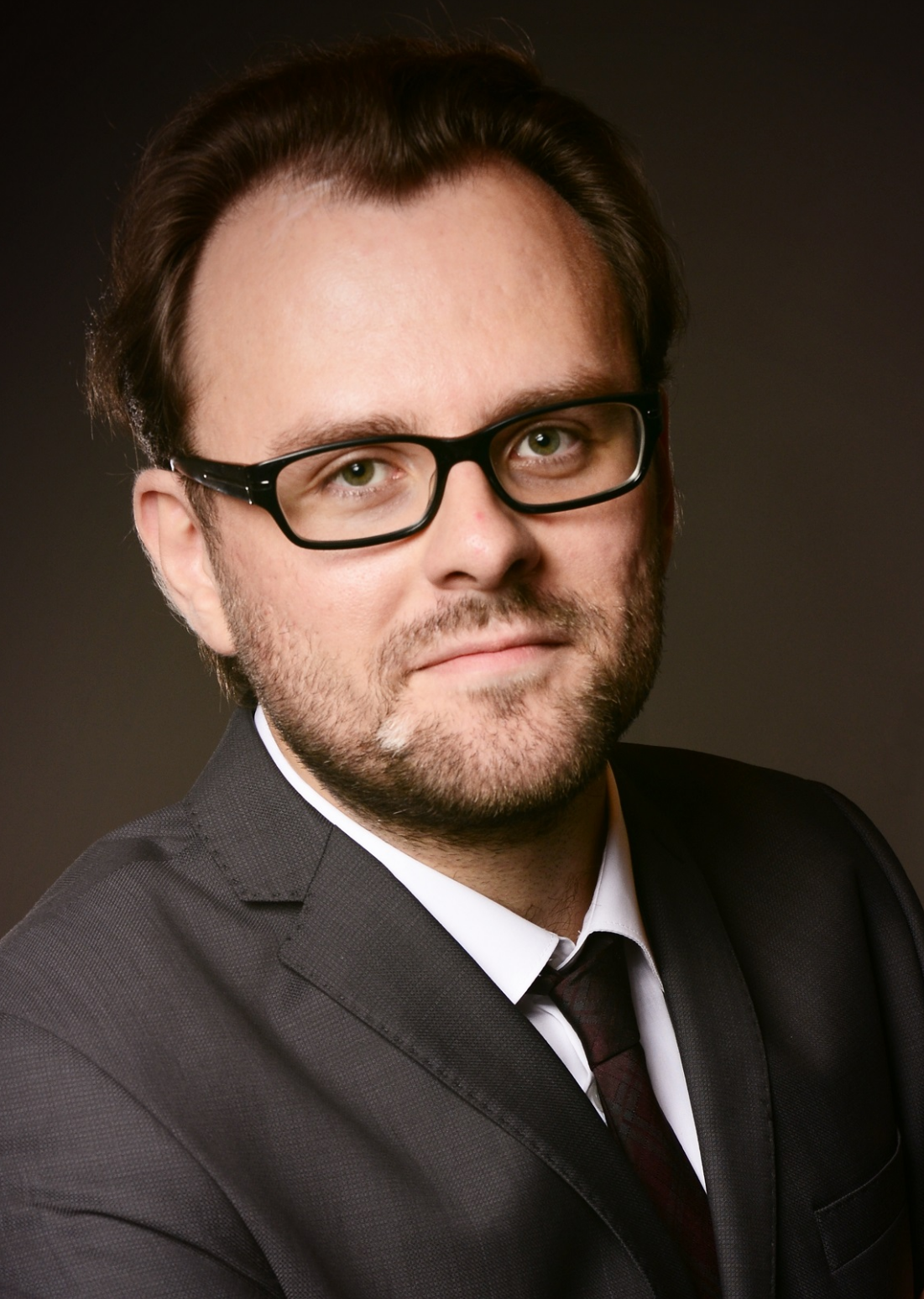}}]{Michel
Bätz}
received the Dipl.-Ing. degree in systems of information and multimedia technology and the Dr.-Ing. degree in electrical engineering from Friedrich-Alexander-Universtit\"at (FAU) Erlangen-N\"urnberg, Erlangen, Germany, in 2012 and 2018, respectively.
 From 2012 to 2017, he was with the Chair of Multimedia Communications and Signal Processing at FAU, where he conducted his research on spatial, temporal, and radiometric resolution enhancement techniques for images and videos.
Currently, he is with the Fraunhofer Institute for Integrated Circuits, Erlangen, Germany.
His research interests include super-resolution as well as computational imaging with a particular focus on light field processing.
\end{IEEEbiography}

\begin{IEEEbiography}[{\includegraphics[width=1in,height=1.25in,clip,keepaspectratio]{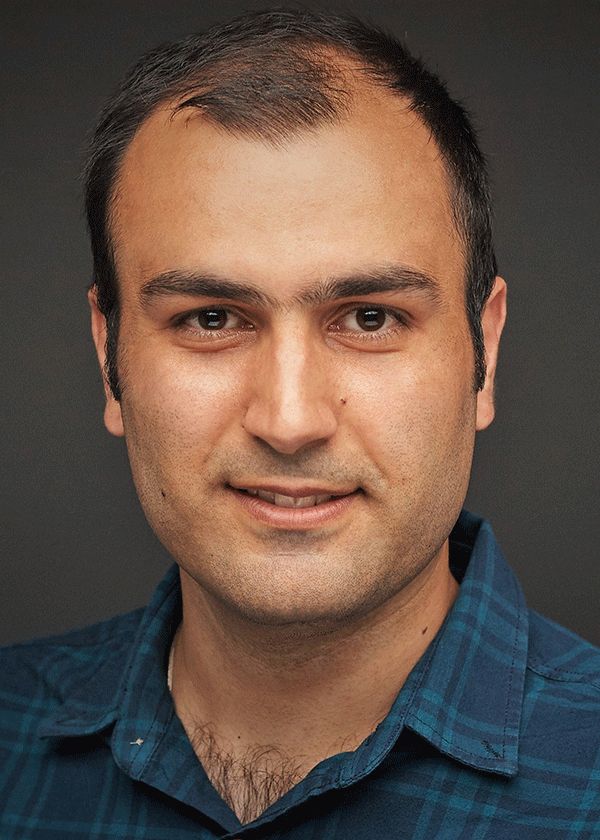}}]{Farzad Naderi}
received the M.Sc. degree in the field of communications and multimedia engineering from the Friedrich-Alexander-Universtit\"at (FAU) Erlangen-N\"urnberg, Erlangen, Germany. He is currently a Research Associate at the Fraunhofer Institute for Integrated Circuits. His research interests are deep learning and natural language processing. His contribution to this paper was at the time he worked at the Pattern Recognition Lab at FAU.
\end{IEEEbiography}

\begin{IEEEbiography}[{\includegraphics[width=1in,height=1.25in,clip,keepaspectratio]{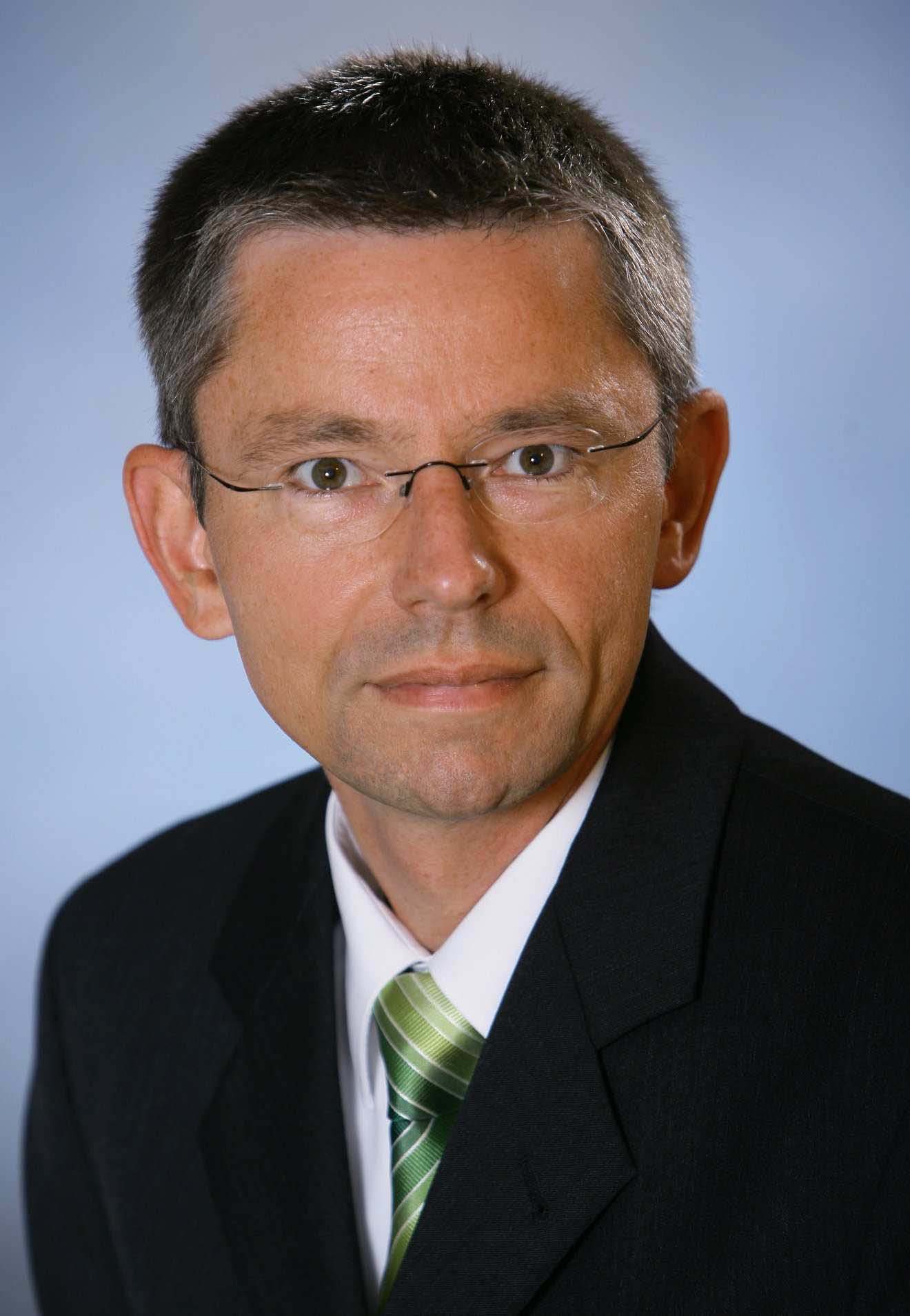}}]{André Kaup}
(M’96–SM’99–F’13) received the Dipl.-Ing. and Dr.-Ing. degrees in electrical engineering from Rheinisch-Westfälische Technische Hochschule (RWTH) Aachen University, Aachen, Germany, in 1989 and 1995, respectively.

He was with the Institute for Communication Engineering, RWTH Aachen University, from 1989 to 1995. He joined the Networks and Multimedia Communications Department, Siemens Corporate Technology, Munich, Germany, in 1995 and became Head of the Mobile Applications and Services Group in 1999. Since 2001 he has been a Full Professor and the Head of the Chair of Multimedia Communications and Signal Processing, Friedrich-Alexander-Universtit\"at (FAU) Erlangen-N\"urnberg, Erlangen, Germany. From 1997 to 2001 he was the Head of the German MPEG delegation. From 2005 to 2007 he was a Vice Speaker of the DFG Collaborative Research Center 603. From 2015 to 2017 he served as Head of the Department of Electrical Engineering and Vice Dean of the Faculty of Engineering. He has authored around 350 journal and conference papers and has over 100 patents granted or pending. His research interests include image and video signal processing and coding, and multimedia communication.

André Kaup is a member of the IEEE Multimedia Signal Processing Technical Committee, a member of the scientific advisory board of the German VDE/ITG, and a Fellow of the IEEE. He served as an Associate Editor for IEEE Transactions on Circuits and Systems for Video Technology and was a Guest Editor for IEEE Journal of Selected Topics in Signal Processing. From 1998 to 2001 he served as an Adjunct Professor with the Technical University of Munich, Munich. He was a Siemens Inventor of the Year 1998 and obtained the 1999 ITG Award. He received several best paper awards, and his group won the Grand Video Compression Challenge at the Picture Coding Symposium 2013. He received the teaching award of the Faculty of Engineering in 2015. In 2018 he was elected full member of the Bavarian Academy of Sciences.
\end{IEEEbiography}

\begin{IEEEbiography}[{\includegraphics[width=1in,height=1.25in,clip,keepaspectratio]{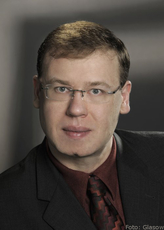}}]{Andreas Maier}
(M'05) graduated in computer science in 2005 and received the Ph.D. degree in computer science at the Friedrich-Alexander-Universit\"{a}t (FAU) Erlangen-N\"{u}rnberg, Erlangen, Germany, in 2009.

From 2005 to 2009, he was working at the Pattern Recognition Lab at FAU. His major research subject was medical signal processing in speech data. In this period, he developed the first online speech intelligibility assessment tool - PEAKS - that has been used to analyze over 4.000 patient and control subjects so far. From 2009 to 2010, he was Postdoctoral Fellow at the Radiological Sciences Laboratory of the Stanford University, Stanford, CA. From 2011 to 2012, he joined Siemens Healthcare as Innovation Project Manager and was responsible for reconstruction topics in the angiography and X-ray business unit. In 2012, he returned to FAU as Head of the Medical Image Reconstruction Group at the Pattern Recognition Lab. Since 2015, he is the Head of the Pattern Recognition Lab. In 2018, he was awarded an ERC Synergy Grant 4D nanoscope.
\end{IEEEbiography}

\begin{IEEEbiography}[{\includegraphics[width=1in,height=1.25in,clip,keepaspectratio]{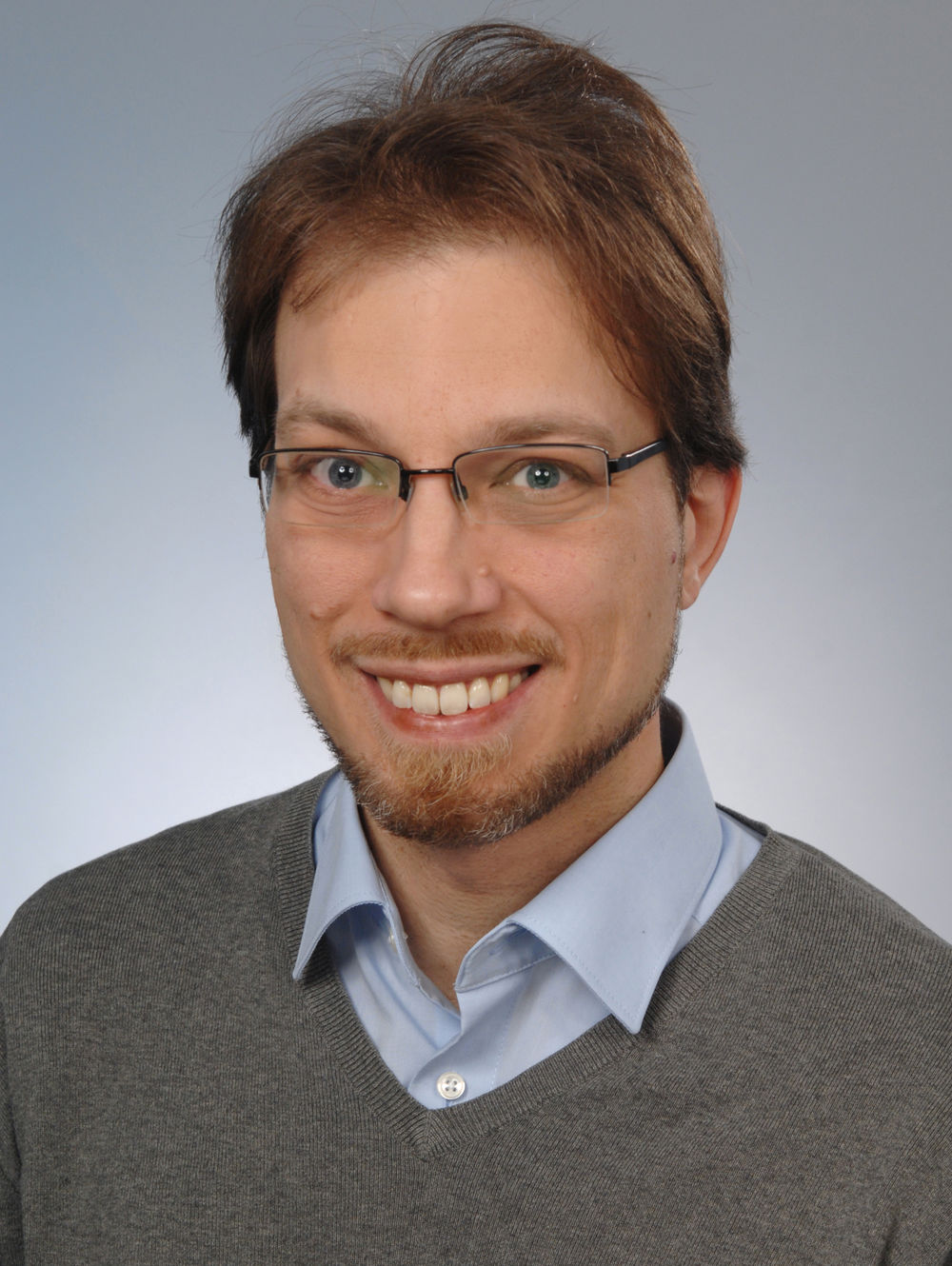}}]{Christian Riess}
received the Ph.D. degree in computer science from the Friedrich-Alexander
University Erlangen-N{\"u}rnberg (FAU), Erlangen, Germany, in 2012.  From 2013
to 2015, he was a Postdoc at the Radiological Sciences Laboratory, Stanford
University, Stanford, CA, USA. Since 2015, he is the head of the Phase-Contrast X-ray Group at the Pattern
Recognition Laboratory at FAU. Since 2016, he is senior researcher and head
of the Multimedia Security Group at the IT Infrastructures Lab at FAU.
He is currently a member of the IEEE Information Forensics and
Security Technical Committee. His research interests include all aspects of
image processing and imaging, particularly with applications in image and video
forensics, X-ray phase contrast, color image processing, and computer vision.
\end{IEEEbiography}

\vfill

\end{document}